\documentclass[runningheads]{llncs}
\newif\ifsupplementary
\supplementaryfalse

\usepackage{eccv}

\usepackage{eccvabbrv}

\usepackage{graphicx}
\usepackage{booktabs}

\usepackage[accsupp]{axessibility}  

\usepackage{hyperref}

\usepackage{orcidlink}

\usepackage[export]{adjustbox}
\usepackage{multirow}
\usepackage{tablefootnote}

\newif\ifdraft
\draftfalse

\definecolor{orange}{rgb}{1,0.5,0}
\definecolor{violet}{RGB}{70,0,170}
\definecolor{magenta}{RGB}{170,0,170}
\definecolor{dgreen}{RGB}{0,150,0}

\ifdraft
 \newcommand{\PF}[1]{{\color{red}{\bf PF: #1}}}
 
 \newcommand{\FS}[1]{{\color{blue}{\bf FS: #1}}}

 \newcommand{\BG}[1]{{\color{olive}{\bf BG: #1}}}

 \newcommand{\TODO}[1]{\textbf{\color{red}[TODO: #1]}}
 \newcommand{\NT}[1]{{\color{violet}{\bf NT: #1}}}
 
\else
 \newcommand{\PF}[1]{}
 
 \newcommand{\FS}[1]{}

 \newcommand{\BG}[1]{}
 
 \newcommand{\ME}[1]{}
  
  \newcommand{\TODO}[1]{}
  
  \newcommand{\NT}[1]{{\color{violet}{}}}
  
\fi

\newcommand{\comment}[1]{}
\newcommand{\parag}[1]{\vspace{-3mm}\paragraph{#1}}

\newcommand{\bx}{\mathbf{x}}

\newlength{\mytabcolsep}
\begin{document}

\makeatletter
\newcommand{\definetrim}[2]{%
  \define@key{Gin}{#1}[]{\setkeys{Gin}{trim=#2,clip}}%
}
\makeatother

\ifsupplementary
  \title{Supplementary Material for Neural Surface Detection for Unsigned Distance Fields}
  \renewcommand\thefigure{A.\arabic{figure}} 
  \renewcommand\thetable{A.\arabic{table}} 
  \renewcommand\thesection{A.\arabic{section}}
\else
  \title{Neural Surface Detection for Unsigned Distance Fields}
\fi

\ifsupplementary
  \titlerunning{Supplementary Material for Neural Surface Detection for UDFs}
\fi

\author{%
  Federico Stella\orcidlink{0009-0001-7490-7356} \and
  Nicolas Talabot\orcidlink{0000-0003-0952-5906} \and
  Hieu Le\orcidlink{0000-0001-7855-2778} \and
  Pascal Fua\orcidlink{0000-0002-6702-9970}
}

\authorrunning{F.~Stella et al.}

\institute{CVLab, École Polytechnique Fédérale de Lausanne (EPFL), \\ \texttt{\{federico.stella, nicolas.talabot, minh.le, pascal.fua\}@epfl.ch} }

\maketitle

\ifsupplementary

\section{Input normalization}

\begin{table}[]
    \renewcommand{\arraystretch}{1.0}
    \caption{\small \textbf{Effect of normalization and training with multiple resolutions.} Median L2 Chamfer Distance $\times 10^{-5}$ with 200k sample points (CD) and Image Consistency (IC) are reported at varying grid resolutions. Triangulation is performed using Marching Cubes. Method in bold is the one presented in the main paper.
    }
    \vspace{-5mm}
    \label{tab:supp_norm}
    \begin{scriptsize}
        \begin{center}
            \setlength{\tabcolsep}{3pt}
            \begin{tabular}{c|cc|cc|cc|cc} 
                 & \multicolumn{2}{c}{ABC~\cite{Koch19a} 32} & \multicolumn{2}{c}{ABC 64} & \multicolumn{2}{c}{ABC 128} & \multicolumn{2}{c}{ABC 256} \\ 
                 Method & CD $\downarrow$ & IC $\uparrow$ & CD $\downarrow$ & IC $\uparrow$ & CD $\downarrow$ & IC $\uparrow$ & CD $\downarrow$ & IC $\uparrow$ \\ 
                \midrule
                    Not normalized          & 28.1 & 90.1           & 4.62 & 95.4           & 2.54 & 97.3           & 2.33 & 98.2 \\
                    Multiresolution         & 24.0 & 90.6           & 4.61 & 95.4           & 2.54 & 97.4           & 2.33 & 98.2 \\
                    \textbf{Normalized}     & 19.0 & 91.8           & 4.46 & 95.4           & 2.52 & 97.4           & 2.33 & 98.2 \\
                    \midrule 
            \end{tabular}
        \end{center}
    \end{scriptsize}
\end{table}

As mentioned in the main paper - Sec. 3.2 - we normalized the input UDF values by the size of the grid cells to make the input invariant to the shape resolution from which we sample UDFs. This is important because the UDF value decides if there is a surface nearby, this depends on the size of the grid cell. For example, a UDF value of $0.1$ would mean that there probably is a surface in a cell with a grid size of $0.2$, but would not mean the same with a grid size of $0.05$. Our normalization step ensures a consistent relation between the input normalized UDF values and the presence of the surface, regardless of the field resolution.

To validate the approach, we train a network without this normalization step, denoted as \textit{Not normalized}. This network is trained only with UDFs sampled from 128-resolution shapes, just like the main approach presented in the main paper. We also train a network without normalization and with multiple resolutions (32, 64, 128, 256, as opposed to only 128), denoted as \textit{Multiresolution}. The results are shown in Table \ref{tab:supp_norm}, where the model presented in the main paper is in bold. The results show that training with multiple resolutions increases the accuracy of the network in localizing surfaces at resolutions different from 128. Our normalization step, however, can achieve even better results, and only needs to be trained on a single resolution (128).

\section{Noise types}

\begin{table}[t]
    \renewcommand{\arraystretch}{1.0}
    \caption{\small \textbf{Effects of different noise types on the model.} Median L2 Chamfer Distance $\times 10^{-5}$ with 200k sample points (CD) and Image Consistency (IC) are reported at varying grid resolutions. Triangulation is performed using Marching Cubes. Method in bold refers to what presented in the main paper.
    }
    \vspace{-5mm}
    \label{tab:supp_noise}
    \begin{scriptsize}
        \begin{center}
            \setlength{\tabcolsep}{3pt}
            \begin{tabular}{c|cc|cc} 
                 & \multicolumn{2}{c}{ABC~\cite{Koch19a} 128} & \multicolumn{2}{c}{MGN~\cite{Bhatnagar19} autodec. 256} \\ 
                 Method & CD $\downarrow$ & IC $\uparrow$ & CD $\downarrow$ & IC $\uparrow$ \\ 
                \midrule
                    No noise                & 2.55 & 97.4                   & 2.04 & 94.5 \\
                    Add                     & 5.23 & 92.8                   & 3.86 & 81.0 \\
                    Add-Exp                 & 2.54 & 97.3                   & 2.01 & 94.9 \\
                    \textbf{Scale}          & 2.52 & 97.4                   & 2.02 & 94.8 \\
                    Scale (re-trained)      & 2.54 & 97.4                   & 2.02 & 94.9 \\
                    Scale-Exp               & 2.55 & 97.4                   & 2.01 & 94.9 \\
                    Scale+GradSwap          & 3.89 & 96.7                   & 2.30 & 93.2 \\
                    \midrule 
            \end{tabular}
        \end{center}
    \end{scriptsize}
\end{table}

When training our proposed model to predict sign configurations, we add noise to the input UDF and gradient values in order to make it more robust to imperfect UDFs, such as those output from a neural network. In Table~\ref{tab:supp_noise}, we perform an ablation study on the noise and noise type. 
Given an input vector $c\in\mathbb{R}^{8\times4}$ consisting of the UDF values and gradient vectors on its 8 corners extracted at a grid cell, we augment $c$ with different types of noise to generate an augmented input vector $\tilde{c}$ for training the network. We experiment with the following types of noise:
%
\begin{itemize}
	\item \textit{No noise}: Noiseless baseline with $\tilde{c}=c$, also reported in the ablation study in the main paper.
	\item\textit{Add}: We add i.i.d.~Gaussian noise to both the UDF and the gradient values. 
	$$
	\tilde{c} = c + \epsilon, \;\text{with}\; \epsilon\sim\mathcal{N}(0,\sigma^2)^{8\times4}
	$$
	We report results with $\sigma=1.0$, which we empirically found to be best.
	\item \textit{Add-Exp}: We add i.i.d.~Gaussian noise, rescaled by the input value. For each element $c_i$ of $c$, we add a scaled Gaussian noise based on its value:
	$$
	\tilde{c}_i = c_i +  \epsilon_i \cdot \exp^{-\left(\frac{\bar{c}_i}{\sigma}\right)^2}, \;\text{with}\; \epsilon_i\sim\mathcal{N}(0, \sigma^2)
	$$ 
	and $\bar{\cdot}$ is the \textit{stop-gradient} operation. The rationale is that smaller values of $c_i$ corresponds to points closer to the surface, which should be augmented more aggresively with higher noise in the case of neural UDFs. We use $\sigma=0.1$, which we empirically found to be best.
	\item \textit{Scale}: We rescale the grid cell by i.i.d.~Gaussian noise:
	$$
	\tilde{c} = c \cdot (1 + \epsilon), \;\text{with}\; \epsilon\sim\mathcal{N}(0,\sigma^2)^{8\times4}.
	$$
	We use $\sigma=1.0$, which we empirically found to be best. This is the method used in the main paper.
	\item \textit{Scale-Exp}: We rescale the grid cell by i.i.d.~Gaussian noise, which is itself rescaled by the cell value, with a similar motivation as \textit{Add-Exp}: 
	$$
	\tilde{c}_i = c_i \cdot \left(1 + \epsilon_i \cdot \exp^{-\left(\frac{\bar{c}_i}{\sigma}\right)^2}\right), \;\text{with}\; \epsilon_i\sim\mathcal{N}(0, \sigma^2),
	$$
	and $\sigma=0.1$.
	\item \textit{GradSwap}: We randomly swap the gradient direction, with higher probability where the UDF is small:
	$$
	\mathbf{g}_i = \begin{cases}
		- \mathbf{g}_i & \text{if } p_i < \frac{1}{2}\exp^{-\left(\frac{u_i}{\sigma}\right)^2} \\
		\mathbf{g}_i & \text{else}
	\end{cases}
	$$
	where $u_i$ is the UDF value at corner $i$ and $\mathbf{g}_i$ its gradient, and $p_i\sim\mathcal{U}(0, 1)$. As above, we reason that close to the surface, neural UDFs have a greater risk to have a wrong gradient. We empirically found $\sigma=0.02$ to be best.
\end{itemize}
We also trained a second model with the \textit{Scale} strategy, denoted as \textit{Scale (re-trained)} to show that there is a small amount of variance, but the results are stable. As can be seen in Table~\ref{tab:supp_noise}, many noise configurations lead to the same performances. In particular, having a noise model that scales differently based on the distance from the surface seems to be beneficial. This happens in \textit{Add-Exp}, \textit{Scale}, \textit{Scale-Exp}, and they all share a very similar performance. Therefore, we decided to use the simple formulation of the \textit{Scale} noise to train our proposed model. We also include the \textit{No noise} model as a baseline, which performs slightly worse on the MGN autodecoder and is on par on the ABC ground-truth. As shown in the main paper, this baseline can lead to misreconstructions in case of thin surfaces at lower resolutions, which is solved by the \textit{Scale} noise.

\section{Autodecoder training details}
To train the autodecoder models used in the main paper, we follow a similar approach as DeepSDF~\cite{Park2019deepSDFLC}. The input meshes are rescaled and centered to a $[-1,1]^3$ volume and, through a data preparation step, we sample the points used for training. For each mesh, we sample 20k points uniformly in the volume, and 400k points near the surface. The latter points are determined by first uniformly sampling 200k points on the surface, and then adding a small amount of Gaussian noise. We apply a Gaussian noise with mean 0 and standard deviation $\sqrt{0.005}$ to determine 200k points near the surface, and a Gaussian noise with mean 0 and standard deviation $\sqrt{0.0005}$ to determine the remaining 200k points. The autodecoder network consists in 12 layers, each with 1024 hidden nodes and ReLU activations, and latent codes of size 512. It is trained with L1 loss without regularization and without Fourier encoding for 10k epochs with a batch size of 16. The concentrate the network capacity around the surface, the UDF is clamped to 0.1. The Adam optimizer is used with learning rates of 0.0005 for the model and 0.001 for the latent codes. The learning rates are decayed at epoch 1600 and 3500 by a factor of 0.35.


\section{Post-processing on UNDC}

\begin{table}[t]
    \renewcommand{\arraystretch}{1.0}
    \caption{\small \textbf{Post-processing on UNDC.} Median L2 Chamfer Distance $\times 10^{-5}$ with 2M sample points (CD) and Image Consistency (IC) are reported at varying grid resolutions. The post-processed algorithm is denoted as \textit{UNDC pp}. *Resolution is doubled for experiments with ShapeNet-Cars due to the higher complexity of the shapes.
    }
    \vspace{-5mm}
    \label{tab:supp_undc}
    \begin{scriptsize}
        \begin{center}
            \setlength{\tabcolsep}{3pt}
            \begin{tabular}{cc|cc|cc|cc|cc|cc} 
                \multicolumn{2}{c}{} & \multicolumn{2}{c}{ABC~\cite{Koch19a}} & \multicolumn{2}{c}{MGN~\cite{Bhatnagar19}} & \multicolumn{2}{c}{Cars~\cite{Chang15}}  & \multicolumn{2}{c}{MGN autodec.} & \multicolumn{2}{c}{Cars autodec.} \\ 
                Res. & Method & CD $\downarrow$ & IC $\uparrow$ & CD $\downarrow$ & IC $\uparrow$ & CD $\downarrow$ & IC $\uparrow$ & CD $\downarrow$ & IC $\uparrow$ & CD $\downarrow$ & IC $\uparrow$ \\ 
                \midrule
                \multirow{2}{*}{32*}
                    & UNDC~\cite{Chen22b}         & 11.8 & 92.4     & 6.81 & 89.7       & 7.60 & 88.2       & 10.8 & 86.2       & 14.9 & 86.4 \\
                    & UNDC pp                     & 13.3 & 93.5     & 7.80 & 91.5       & 9.41 & 88.8       & 14.7 & 88.9       & 20.8 & 87.4 \\
                    \midrule 
                \multirow{2}{*}{64*}
                    & UNDC~\cite{Chen22b}         & 0.783 & 95.8    & 0.926 & 0.958     & 1.35 & 91.0       & 1.83 & 90.7       & 16.8 & 85.9 \\
                    & UNDC pp                     & 0.806 & 96.4    & 92.9 & 94.7       & 1.55 & 91.9       & 1.81 & 93.4       & 24.7 & 86.8 \\
                    \midrule 
                \multirow{2}{*}{128*}
                    & UNDC~\cite{Chen22b}         & 0.0877 & 97.2   & 0.140 & 94.7      & 0.195 & 94.0      & 1.06 & 88.7       & 57.7 & 74.8 \\
                    & UNDC pp                     & 0.0781 & 97.6   & 0.126 & 96.4      & 0.214 & 94.7      & 1.25 & 92.6       & 77.6 & 74.5 \\
                    \midrule 
                \multirow{2}{*}{256*}
                    & UNDC~\cite{Chen22b}         & 0.0146 & 98.0   & 0.0251 & 96.5     & - & -             & 1.68 & 82.3       & - & - \\
                    & UNDC pp                     & 0.0126 & 98.3   & 0.0206 & 97.5     & - & -             & 3.86 & 84.4       & - & - \\
                    \midrule 
            \end{tabular}
        \end{center}
    \end{scriptsize}
\end{table}

As mentioned in the main paper, Sec. 4.2, Unsigned Neural Dual Contouring (UNDC)~\cite{Chen22b} also proposes a post-processing step to reduce holes in the meshes. For fairness to the other algorithms, UNDC has been evaluated without such post-processing, but we report here the results of the post-processed algorithm as well. In Table \ref{tab:supp_undc} we show that the post-processed algorithm improves consistently the Image Consistency scores, while often worsening the Chamfer Distance, especially at lower resolutions or in the complex ShapeNet-Cars~\cite{Chang15} dataset. Nevertheless, comparing these results to the ones in Table 2 of the main paper, the post-processed UNDC is still outperformed by both DualMesh-UDF~\cite{Zhang23b} and our method.

\section{Mean value tables}

\begin{table}[t]
    \renewcommand{\arraystretch}{1.0}
    \caption{\small \textbf{Marching Cubes-based triangulation.} Mean L2 Chamfer Distance $\times 10^{-5}$ with 200k sample points (CD) and Image Consistency (IC) are reported at varying grid resolutions. Results meshed with Marching Cubes on the ground-truth SDF are reported for reference on ABC, where extracting an SDF is possible. *Resolution is doubled for experiments with ShapeNet-Cars due to the higher complexity of the shapes. $\dagger$Method failed to reconstruct some shapes, leading to unbound metrics: the shown mean values do not include these cases, hence they should not be directly compared to the others. Comparing them nonetheless: the best results are in bold, the second-best in italics. 
    }
    \vspace{-5mm}
    \label{tab:supp_mc}
    \begin{scriptsize}
        \begin{center}
            \setlength{\tabcolsep}{3pt}
            \begin{tabular}{cc|cc|cc|cc|cc|cc} 
                \multicolumn{2}{c}{} & \multicolumn{2}{c}{ABC~\cite{Koch19a}} & \multicolumn{2}{c}{MGN~\cite{Bhatnagar19}} & \multicolumn{2}{c}{Cars~\cite{Chang15}}  & \multicolumn{2}{c}{MGN autodec.} & \multicolumn{2}{c}{Cars autodec.} \\ 
                Res. & Method & CD $\downarrow$ & IC $\uparrow$ & CD $\downarrow$ & IC $\uparrow$ & CD $\downarrow$ & IC $\uparrow$ & CD $\downarrow$ & IC $\uparrow$ & CD $\downarrow$ & IC $\uparrow$ \\ 
                \midrule
                \multirow{4}{*}{32*} 
                    & CAP-UDF~\cite{Zhou22}             & 3440 & 50.4 $\dagger$                         & 149 & 64.5                            & 98.6 & 78.6                               & 248 & 56.4                            & 225 & 69.6 \\
                    & MeshUDF~\cite{Guillard22b}        & \textit{134} & \textit{85.0} $\dagger$        & \textit{20.1} & \textit{89.3}         & \textit{22.4} & \textit{85.4}             & \textit{20.4} & \textit{89.3}         & \textit{27.4} & \textit{86.3} \\
                    & DCUDF~\cite{Hou2023DCUDF}         & 1640 & 75.1 $\dagger$                         & 667 & 74.6                            & 887 & 70.4                                & 558 & 75.7                            & 427 & 78.1 \\
                    & Ours + MC~\cite{Lewiner03}        & \textbf{35.3} & \textbf{87.9}                 & \textbf{8.94} & \textbf{90.9}         & \textbf{10.6} & \textbf{87.9}             & \textbf{9.10} & \textbf{90.8}         & \textbf{15.8} & \textbf{87.3} \\
                    \midrule 
                    & GT + MC & 331 & 86.8 $\dagger$ & - & - & - & - & - & - & - & - \\ \toprule
                \multirow{4}{*}{64*} 
                    & CAP-UDF~\cite{Zhou22}             & 2380 & 65.2 $\dagger$                         & 12.6 & 85.2                           & 28.1 & 88.1                               & 22.7 & 78.6                           & 78.7 & 81.8 \\
                    & MeshUDF~\cite{Guillard22b}        & \textit{49.5} & \textit{91.5}                 & \textit{3.79} & \textit{93.9}         & \textit{7.17} & \textit{89.7}             & \textit{4.25} & \textit{93.3}         & \textit{15.2} & \textbf{88.7} \\
                    & DCUDF~\cite{Hou2023DCUDF}         & 845 & 82.6 $\dagger$                          & 172 & 84.4                            & 228 & 80.5                                & 117 & 86.6                            & 71.2 & 86.8 \\
                    & Ours + MC~\cite{Lewiner03}        & \textbf{9.62} & \textbf{93.2}                 & \textbf{2.25} & \textbf{94.9}         & \textbf{4.59} & \textbf{91.4}             & \textbf{2.95} & \textbf{94.1}         & \textbf{13.1} & \textit{88.5} \\
                    \midrule 
                    & GT + MC & 55.5 & 93.7 $\dagger$ & - & - & - & - & - & - & - & - \\ 
                    \toprule
                \multirow{4}{*}{128*} 
                    & CAP-UDF~\cite{Zhou22}             & 898 & 86.2 $\dagger$                          & 2.14 & 95.4                           & 4.57 & \textit{92.8}                      & 4.99 & 91.2                           & 54.5 & 86.2 \\
                    & MeshUDF~\cite{Guillard22b}        & \textit{4.76} & \textit{95.4}                 & \textit{1.71} & \textit{96.3}         & \textit{3.91} & \textit{92.8}             & \textit{2.55} & 94.5                  & \textit{21.4} & \textbf{88.2} \\
                    & DCUDF~\cite{Hou2023DCUDF}         & 263 & 89.5 $\dagger$                          & 39.8 & 91.8                           & 135 & 85.7                                & 11.8 & \textbf{94.8}                  & 1100 & 76.0 \\
                    & Ours + MC~\cite{Lewiner03}        & \textbf{3.86} & \textbf{95.9}                 & \textbf{1.48} & \textbf{96.9}         & \textbf{3.44} & \textbf{94.1}             & \textbf{2.36} & \textbf{94.8}         & \textbf{19.4} & \textit{87.6} \\
                    \midrule 
                    & GT + MC & 53.6 & 96.0 $\dagger$ & - & - & - & - & - & - & - & - \\ 
                    \toprule
                \multirow{4}{*}{256*} 
                    & CAP-UDF~\cite{Zhou22}             & \textbf{3.11} & 96.7                          & \textit{1.39} & \textit{97.6}         & 3.79 & 95.0                               & 3.71 & 94.1                           & \textbf{55.2} & \textbf{86.4} \\
                    & MeshUDF~\cite{Guillard22b}        & 5.22 & \textit{96.9}                          & 1.42 & 97.4                           & \textit{3.35} & \textit{95.1}             & \textit{2.61} & \textbf{94.4}         & 144 & 78.8 \\
                    & DCUDF~\cite{Hou2023DCUDF}         & 251 & 92.8 $\dagger$                          & 15.0 & 94.5                           & 152 & 87.5                                & 39.1 & 88.9 $\dagger$                 & 4640 & 34.8 \\
                    & Ours + MC~\cite{Lewiner03}        & \textit{5.15} & \textbf{97.1}                 & \textbf{1.38} & \textbf{97.8}         & \textbf{3.25} & \textbf{95.8}             & \textbf{2.51} & \textbf{94.4}         & \textit{89.0} & \textit{82.0} \\ 
                    \midrule
                    & GT + MC & 52.2 & 97.1 $\dagger$ & - & - & - & - & - & - & - & - \\ 
                    \toprule
            \end{tabular}
        \end{center}
    \end{scriptsize}
\end{table}


\begin{table}[t]
    \renewcommand{\arraystretch}{1.0}
    \caption{\small \textbf{Dual Contouring-based triangulation.} Mean L2 Chamfer Distance $\times 10^{-5}$ with 2M sample points (CD) and Image Consistency (IC) are reported at varying grid resolutions. *Resolution is doubled for experiments with ShapeNet-Cars due to the higher complexity of the shapes. The best results are in bold, the second-best in italics. 
    }
    \vspace{-5mm}
    \label{tab:supp_dc}
    \begin{scriptsize}
        \begin{center}
            \setlength{\tabcolsep}{3pt}
            \resizebox{1.0\linewidth}{!}{
            \begin{tabular}{cc|cc|cc|cc|cc|cc} 
                \multicolumn{2}{c}{} & \multicolumn{2}{c}{ABC~\cite{Koch19a}} & \multicolumn{2}{c}{MGN~\cite{Bhatnagar19}} & \multicolumn{2}{c}{Cars~\cite{Chang15}}  & \multicolumn{2}{c}{MGN autodec.} & \multicolumn{2}{c}{Cars autodec.} \\ 
                Res. & Method & CD $\downarrow$ & IC $\uparrow$ & CD $\downarrow$ & IC $\uparrow$ & CD $\downarrow$ & IC $\uparrow$ & CD $\downarrow$ & IC $\uparrow$ & CD $\downarrow$ & IC $\uparrow$ \\  
                \midrule
                \multirow{4}{*}{32*} 
                    & UNDC~\cite{Chen22b}           & 61.8 & 86.6                           & 7.41 & 89.0                           & 7.56 & 88.1                           & 11.7 & 85.2                           & 20.7 & 86.3 \\
                    & DMUDF~\cite{Zhang23b}         & \textit{15.6} & \textbf{95.4}         & \textbf{3.15} & \textbf{93.9}         & \textit{4.25} & \textit{92.3}         & 1640 & 61.4                           & 1110 & 43.3 \\
                    & DMUDF-T                       & - & -                                 & - & -                                 & - & -                                 & \textbf{3.24} & \textbf{93.5}         & \textit{7.63} & \textbf{89.9} \\
                    & Ours+DMUDF                    & \textbf{15.4} & \textbf{95.4}         & \textit{3.35} & \textbf{93.9}         & \textbf{4.02} & \textbf{92.4}         & \textit{3.42} & \textbf{93.5}         & \textbf{7.38} & \textbf{89.9} \\
                    \midrule 
                \multirow{4}{*}{64*} 
                    & UNDC~\cite{Chen22b}           & \textbf{2.21} & 93.8                  & 1.03 & 92.5                           & 1.33 & 91.3                           & 1.98 & 90.2                           & 27.8 & 84.6 \\
                    & DMUDF~\cite{Zhang23b}         & \textit{2.78} & \textbf{97.5}         & \textbf{0.262} & \textbf{96.5}        & \textit{0.824} & \textbf{94.2}        & 1120 & 60.9                           & 1600 & 42.8 \\
                    & DMUDF-T                       & - & -                                 & - & -                                 & - & -                                 & \textbf{0.937} & \textbf{94.9}        & \textit{7.45} & \textit{89.4} \\
                    & Ours+DMUDF                    & 2.80 & \textbf{97.5}                  & \textit{0.265} & \textbf{96.5}        & \textbf{0.771} & \textbf{94.2}        & \textit{0.943} & \textbf{94.9}        & \textbf{7.24} & \textbf{89.7} \\
                    \midrule 
                \multirow{4}{*}{128*} 
                    & UNDC~\cite{Chen22b}           & 18.6 & 95.8                           & 0.162 & 94.5                          & \textit{0.200} & 94.0                 & 1.46 & 88.3                           & 95.1 & 73.4 \\
                    & DMUDF~\cite{Zhang23b}         & \textbf{0.293} & \textbf{98.2}        & \textit{0.0311} & \textbf{97.8}       & \textit{0.200} & \textbf{95.7}        & 993 & 59.0                            & 1440 & 42.6 \\
                    & DMUDF-T                       & - & -                                 & - & -                                 & - & -                                 & \textit{0.900} & \textit{94.7}        & \textit{14.3} & \textit{86.7} \\
                    & Ours+DMUDF                    & \textit{0.295} & \textbf{98.2}        & \textbf{0.0302} & \textbf{97.8}       & \textbf{0.174} & \textbf{95.7}        & \textbf{0.886} & \textbf{94.8}        & \textbf{13.7} & \textbf{88.1} \\
                    \midrule 
                \multirow{4}{*}{256*} 
                    & UNDC~\cite{Chen22b}           & \textbf{0.0191} & 97.1                & 0.0284 & 96.5                         & - & -                                 & 3.84 & 81.2                           & - & - \\
                    & DMUDF~\cite{Zhang23b}         & \textit{0.0541} & \textbf{98.3}       & \textit{0.00360} & \textbf{98.3}      & \textit{0.430} & \textit{96.3}        & 948 & 57.1                            & 1560 & 41.1 \\
                    & DMUDF-T                       & - & -                                 & - & -                                 & - & -                                 & \textit{1.25} & \textit{92.5}         & \textit{51.8} & \textit{78.1} \\
                    & Ours+DMUDF                    & 0.0546 & \textbf{98.3}                & \textbf{0.00345} & \textbf{98.3}      & \textbf{0.0386} & \textbf{96.6}       & \textbf{1.18} & \textbf{93.2}         & \textbf{51.5} & \textbf{81.1} \\ 
                    \midrule
            \end{tabular}
            }
        \end{center}
    \end{scriptsize}
\end{table}

We report in Tab~\ref{tab:supp_mc} and Tab~\ref{tab:supp_dc} the mean values of the Chamfer Distance and Image Consistency metrics for the experiments presented in the main paper. The results are consistent with the median values, and the same conclusions can be drawn. Notice, however, that the mean values are more sensitive to outliers, especially the Chamfer Distance, which is unbounded. This is why we chose to report the median values in the main paper, as they reflect the general performance of the models better. Notice also that some methods completely fail to reconstruct some shapes, they are indicated with a $\dagger$ in Tab~\ref{tab:supp_mc}. When this happens the mean values cannot be computed, hence we exclude the unreconstructed shapes from the computation. As a result, the mean values cannot be directly compared to the others, and should be interpreted with caution. Nevertheless, our method - as well as the methods in Tab~\ref{tab:supp_dc} - never fails to reconstruct a shape and the metrics are consistent with the median values reported in the main paper. One notable exception is represented by the Chamfer Distance on ABC~\cite{Koch19a} at resolution 256, where CAP-UDF~\cite{Zhou22} inverts the trend and scores the best mean value. We notice that this is due to three shapes, namely \#9622, \#9626 and \#9766, that miss one large surface each in the reconstruction with our method and with MeshUDF~\cite{Guillard22b}. Both methods share the same Marching Cubes implementation, namely the scikit-image one~\cite{scikit-image}, so we suspect that the missing surface is due to a miss in the MC look-up table. CAP-UDF uses a different MC implementation, PyMCubes~\cite{PyMCubes}, and does not suffer from this problem. To corroborate this hypothesis, we notice that when our method is paired with DualMesh-UDF~\cite{Zhang23b} this problem does not arise. We will investigate this issue further, and test our method with different MC implementations.

\section{Execution times}
Execution time has not been a core objective of this work. However, we report that our method is slightly faster than MeshUDF - the currently most popular UDF meshing algorithm. For 20 cars from the Cars autodecoder at resolution $128^3$ with an NVIDIA A100 GPU, our method requires 8.14 seconds per instance while MeshUDF requires 8.64 seconds. The grid value querying time is the same but our inference and triangulation is faster: 0.45 for our method, compared to 0.94 for MeshUDF. Most of the time is spent querying the autodecoder, which is GPU-bound.

\section{Additional experiments with DCUDF}

\begin{table}[t]
    \renewcommand{\arraystretch}{1.0}
    \caption{\small \textbf{Additional experiments on DCUDF.} Median L2 Chamfer Distance $\times 10^{-5}$ with 200k sample points (CD) and Image Consistency (IC) are reported at varying grid resolutions. The best results are in bold. "DCUDF-ls" is executed with a lower smoothing parameter (500 instead of 2000). "DCUDF-nc" is executed without the cutting step. *Resolution is doubled for experiments with ShapeNet-Cars due to the higher complexity of the shapes.
    }
    \label{tab:supp_dcudf}
    \begin{scriptsize}
        \begin{center}
            \setlength{\tabcolsep}{3pt}
            \begin{tabular}{cc|cc|cc|cc|cc|cc} 
                \multicolumn{2}{c}{} & \multicolumn{2}{c}{ABC~\cite{Koch19a}} & \multicolumn{2}{c}{MGN~\cite{Bhatnagar19}} & \multicolumn{2}{c}{Cars~\cite{Chang15}}  & \multicolumn{2}{c}{MGN autodec.} & \multicolumn{2}{c}{Cars autodec.} \\ 
                Res. & Method & CD $\downarrow$ & IC $\uparrow$ & CD $\downarrow$ & IC $\uparrow$ & CD $\downarrow$ & IC $\uparrow$ & CD $\downarrow$ & IC $\uparrow$ & CD $\downarrow$ & IC $\uparrow$ \\ 
                \midrule
                \multirow{4}{*}{32*} 
                    & DCUDF~\cite{Hou2023DCUDF} & 1200 & 78.0                           & 603 & 76.0                        & 559 & 69.7                        & 503 & 77.0                        & 318 & 83.7 \\
                    & DCUDF-ls & 1200 & 77.8 & 596 & 76.3 & 506 & 79.0 & 419 & 78.6 & 307 & 83.8 \\
                    & DCUDF-nc & 572 & 82.5 & 484 & 77.4 & 194 & 81.9 & 413 & 78.3 & 97.1 & 84.9 \\
                    & Ours + MC~\cite{Lewiner03} & \textbf{19.0} & \textbf{91.8}     & \textbf{8.09} & \textbf{92.0}     & \textbf{9.89} & \textbf{87.4}     & \textbf{8.37} & \textbf{91.9}     & \textbf{13.2} & \textbf{87.3} \\
                    \midrule
                \multirow{4}{*}{64*} 
                    & DCUDF~\cite{Hou2023DCUDF} & 291 & 86.5                            & 155 & 85.6                        & 169 & 81.3                        & 92.4 & 87.3                       & 55.0 & 87.3 \\
                    & DCUDF-ls & 292 & 86.6 & 153 & 85.6 & 117 & 85.9 & 36.9 & 92.1 & 82.2 & 86.1 \\
                    & DCUDF-nc & 138 & 88.8 & 119 & 85.7 & 29.6 & 88.1 & 61.6 & 87.4 & 14.7 & \textbf{89.4} \\
                    & Ours + MC~\cite{Lewiner03} & \textbf{4.46} & \textbf{95.4}     & \textbf{2.10} & \textbf{95.5}     & \textbf{4.47} & \textbf{91.2}     & \textbf{2.64} & \textbf{94.7}     & \textbf{10.0} & 88.5 \\
                    \midrule
                \multirow{4}{*}{128*} 
                    & DCUDF~\cite{Hou2023DCUDF} & 44.6 & 92.5                           & 26.1 & 92.3                       & 113 & 87.7                        & 6.34 & \textbf{95.4}              & 355 & 78.2 \\
                    & DCUDF-ls & 42.0 & 92.2 & 26.5 & 92.3 & 113 & 87.6 & 5.47 & 95.3 & 330 & 77.1 \\
                    & DCUDF-nc & 26.1 & 93.8 & 20.0 & 92.3 & 9.48 & 91.2 & 2.29 & \textbf{95.4} & 69.6 & 82.3 \\
                    & Ours + MC~\cite{Lewiner03} & \textbf{2.52} & \textbf{97.4}     & \textbf{1.40} & \textbf{97.1}     & \textbf{3.13} & \textbf{94.2}     & \textbf{2.06} & 95.2     & \textbf{14.5} & \textbf{87.9} \\
                    \midrule
                \multirow{4}{*}{256*} 
                    & DCUDF~\cite{Hou2023DCUDF} & 11.3 & 95.0                           & 8.09 & 94.8                       & 104 & 89.5                        & 3.40 & 91.1                       & 1930 & 36.0 \\
                    & DCUDF-ls & 11.5 & 95.0 & 8.11 & 92.3 & 127 & 89.5 & 7.44 & 90.6 & 2250 & 34.0 \\
                    & DCUDF-nc & 8.50 & 96.0 & 6.14 & 94.9 & 4.58 & 93.3 & 4.13 & 91.1 & 626 & 43.8 \\
                    & Ours + MC~\cite{Lewiner03} & \textbf{2.33} & \textbf{98.2}     & \textbf{1.33} & \textbf{97.9}     & \textbf{3.03} & \textbf{95.8}     & \textbf{2.02} & \textbf{94.8}     & \textbf{62.2} & \textbf{83.8} \\ 
                    \midrule
            \end{tabular}
        \end{center}
    \end{scriptsize}
\end{table}

We report in Tab~\ref{tab:supp_dcudf} additional experiments with the DCUDF\cite{Hou2023DCUDF} pipeline. In particular, as suggested by the authors, we experiment with a lower smoothing aggressiveness which benefits lower resolutions, denoted as "DCUDF-ls", and with skipping the cutting step of the algorithm, denoted as "DCUDF-nc". The cutting step ensures that only one of the two collapsed surfaces is kept, however it can fail to cut the surface correctly, leading to missing areas of the object as seen in the main paper. Removing such step, however, makes the surface double layered, which may not be suitable for downstream tasks. The increased accuracy compared to the experiment in the main paper comes at the cost of additional parameter tuning. Overall, similar conclusions as in the main paper apply.

\section{Additional qualitative results}

\newlength{\mcfigwidthsupp}
\setlength{\mcfigwidthsupp}{0.18\linewidth}
\definetrim{trimabc}{2cm 1cm 1cm 1cm}
\definetrim{trimmgn}{5cm 2cm 5cm 1cm}
\definetrim{trimsn}{0cm 12cm 9cm 12cm}
\definetrim{trimmgnad}{4cm 30cm 4cm 0cm}
\definetrim{trimsnad}{0cm 15cm 0cm 20cm}
\setlength\mytabcolsep{\tabcolsep}
\setlength\tabcolsep{1pt}

\begin{figure}[ht!]
	\begin{center}
	{\scriptsize
		\begin{tabular}{llccccc}
			\rotatebox[origin=c]{90}{ABC~\cite{Koch19a}} & \rotatebox[origin=c]{90}{32} &
			\includegraphics[valign=m,width=\mcfigwidthsupp,trimabc]{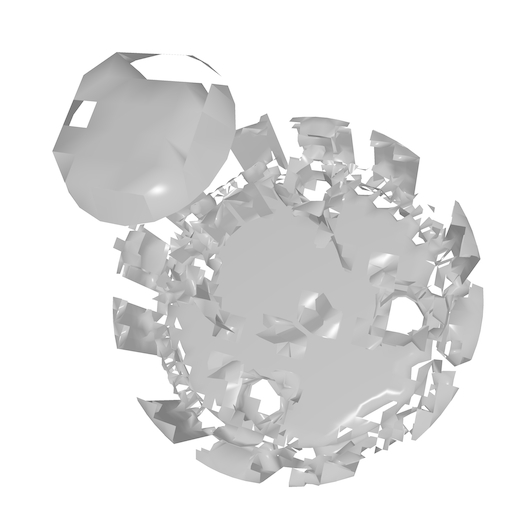}
			& \includegraphics[valign=m,width=\mcfigwidthsupp,trimabc]{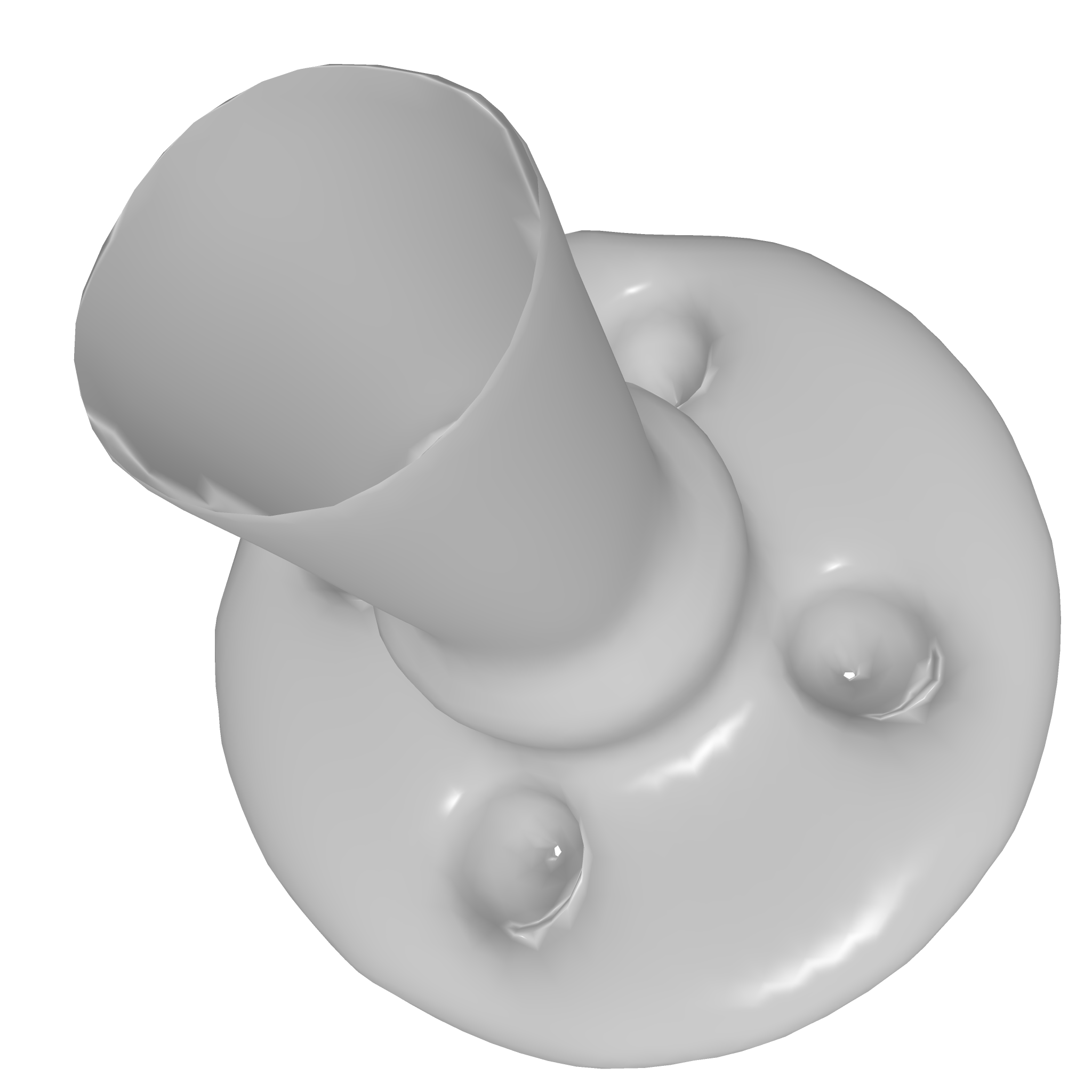}
			& \includegraphics[valign=m,width=\mcfigwidthsupp,trimabc]{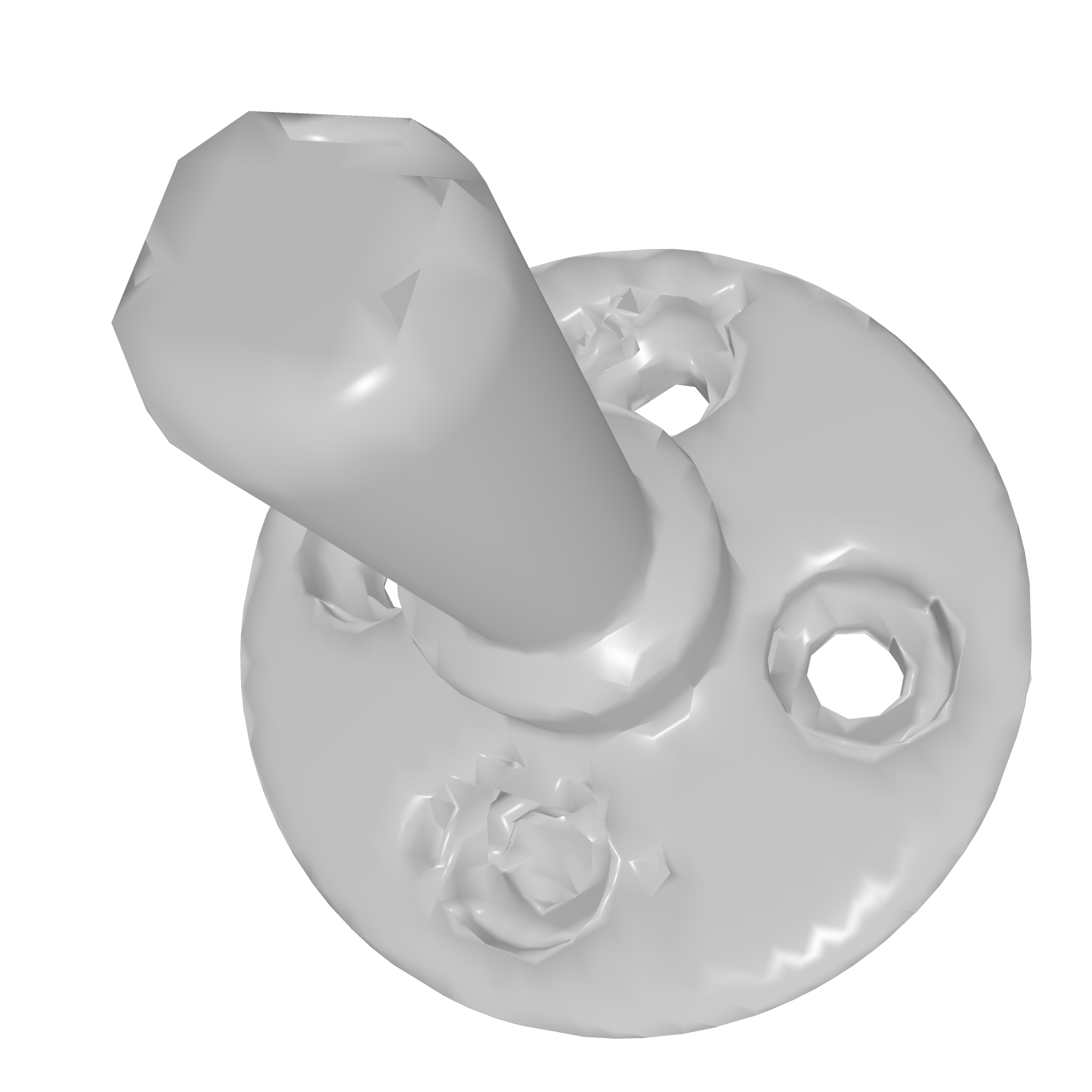}
			& \includegraphics[valign=m,width=\mcfigwidthsupp,trimabc]{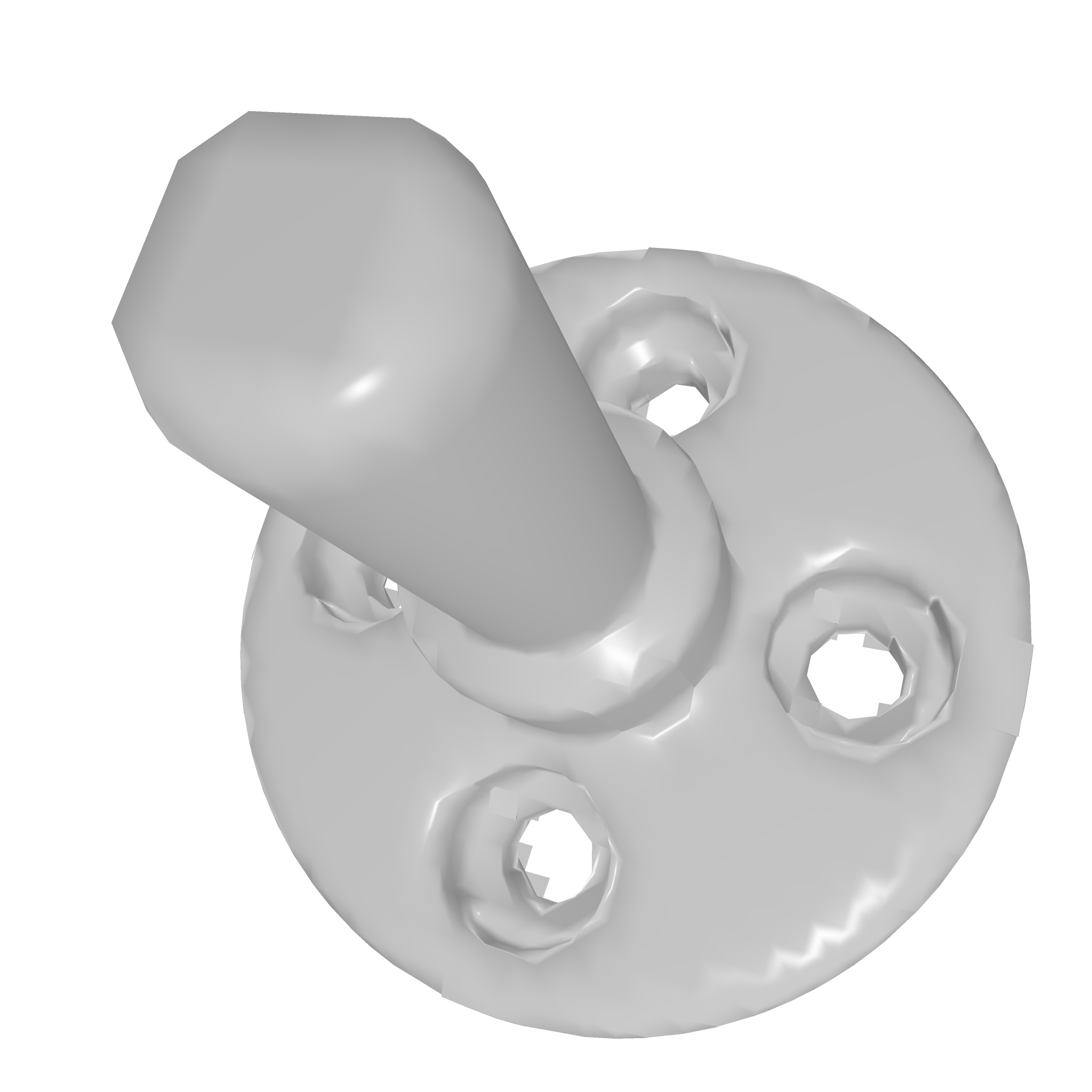} 
			& \includegraphics[valign=m,width=\mcfigwidthsupp,trimabc]{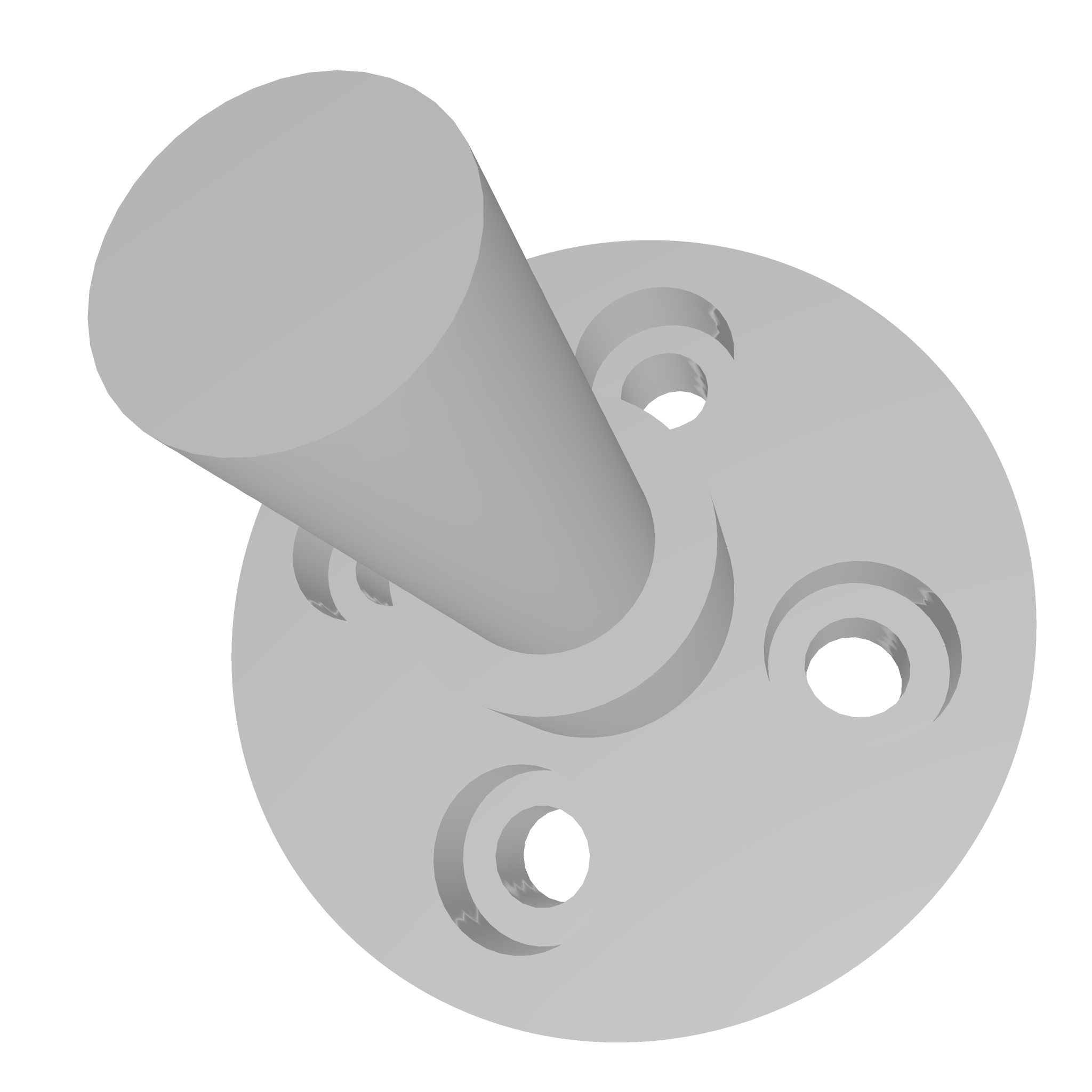}\\ \addlinespace[2\tabcolsep]
			\rotatebox[origin=c]{90}{ABC~\cite{Koch19a}} & \rotatebox[origin=c]{90}{64} &
			\includegraphics[valign=m,width=\mcfigwidthsupp,trimabc]{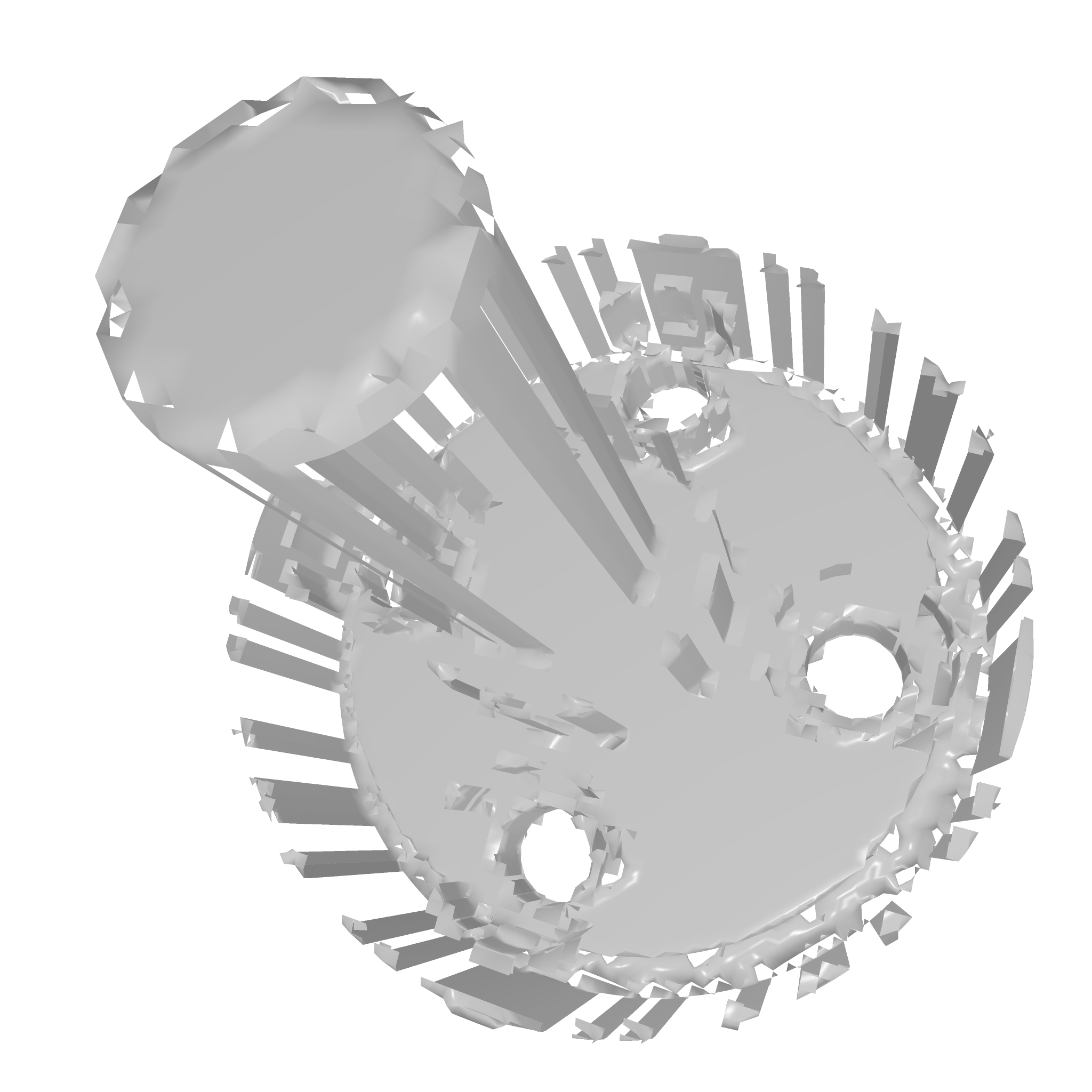}
			& \includegraphics[valign=m,width=\mcfigwidthsupp,trimabc]{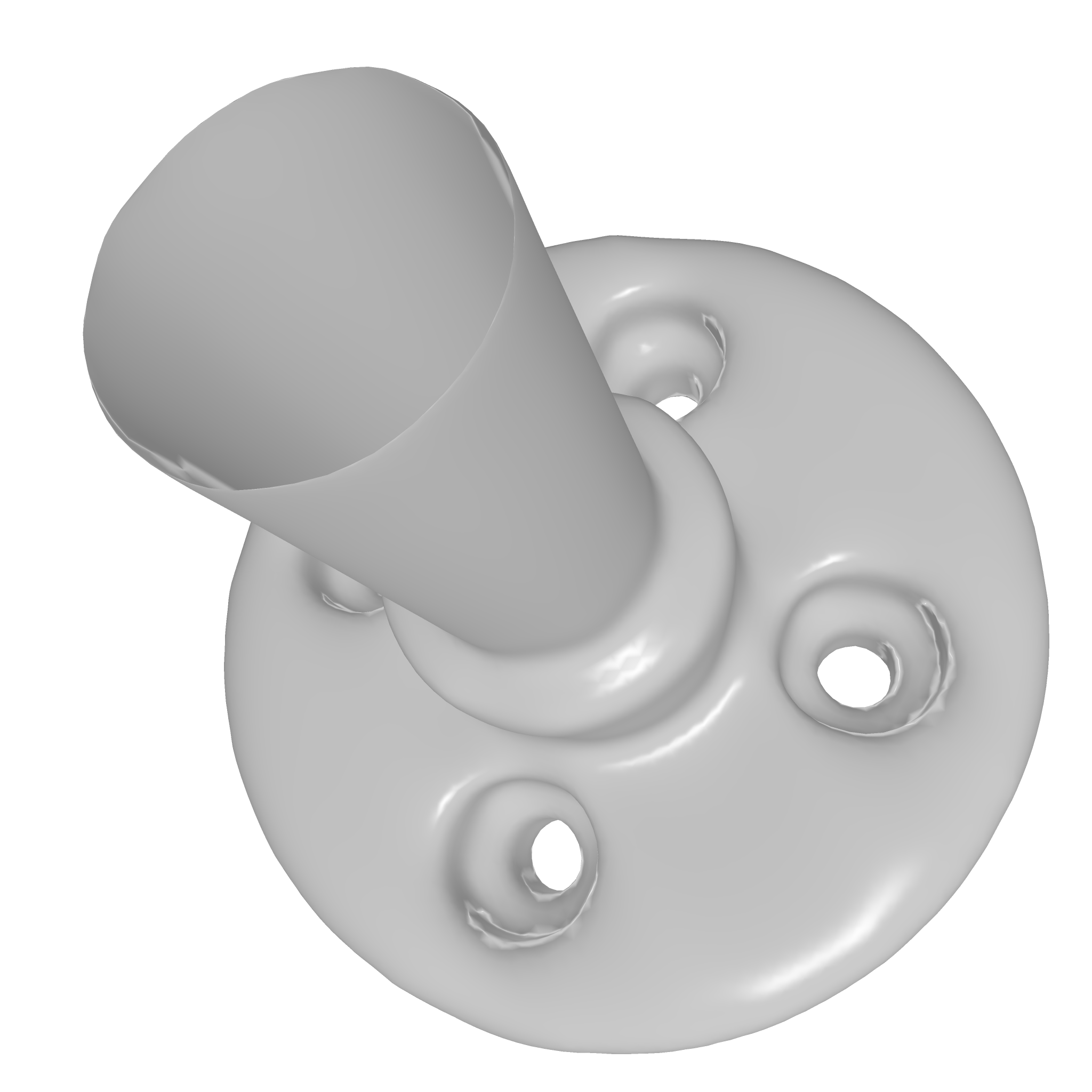}
			& \includegraphics[valign=m,width=\mcfigwidthsupp,trimabc]{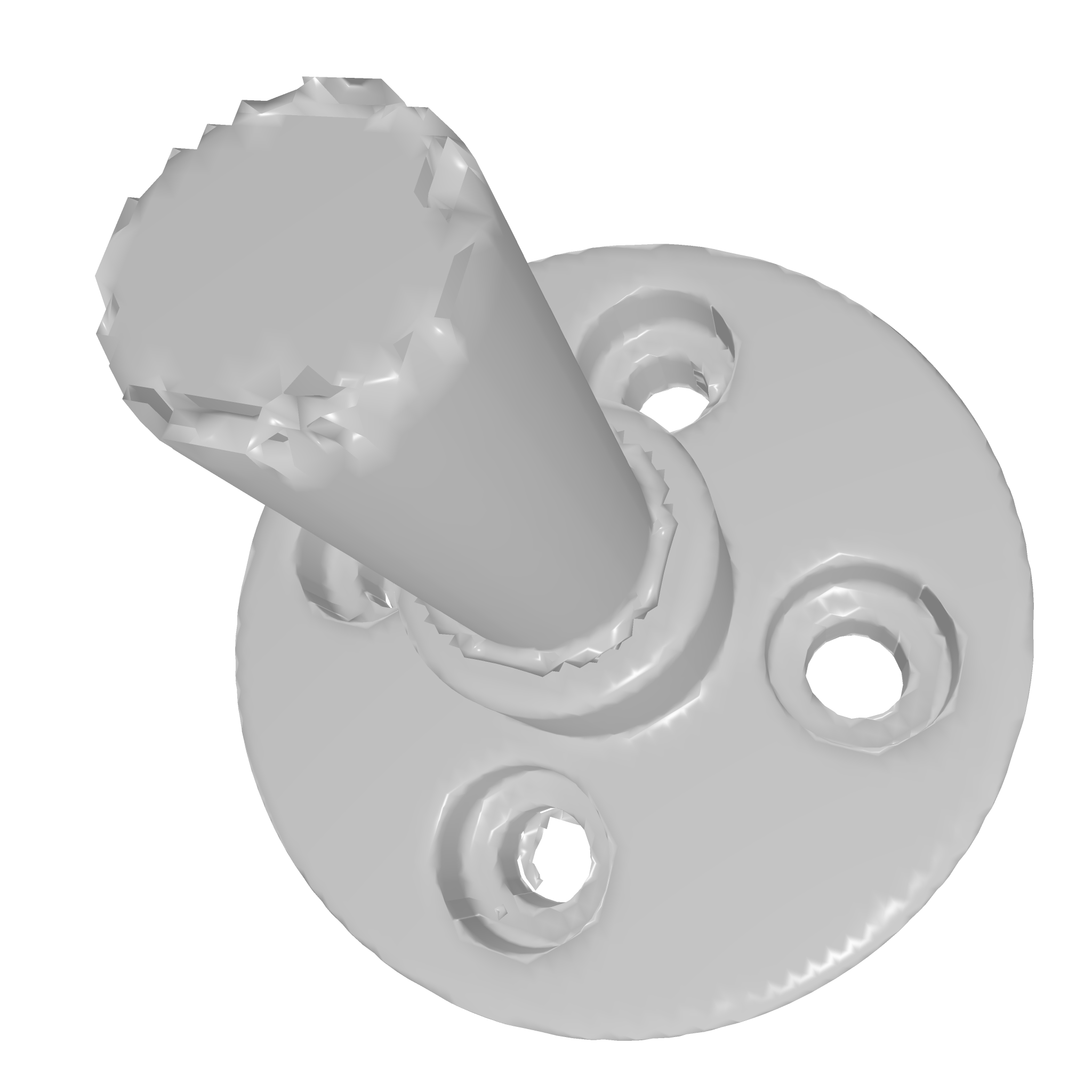}
			& \includegraphics[valign=m,width=\mcfigwidthsupp,trimabc]{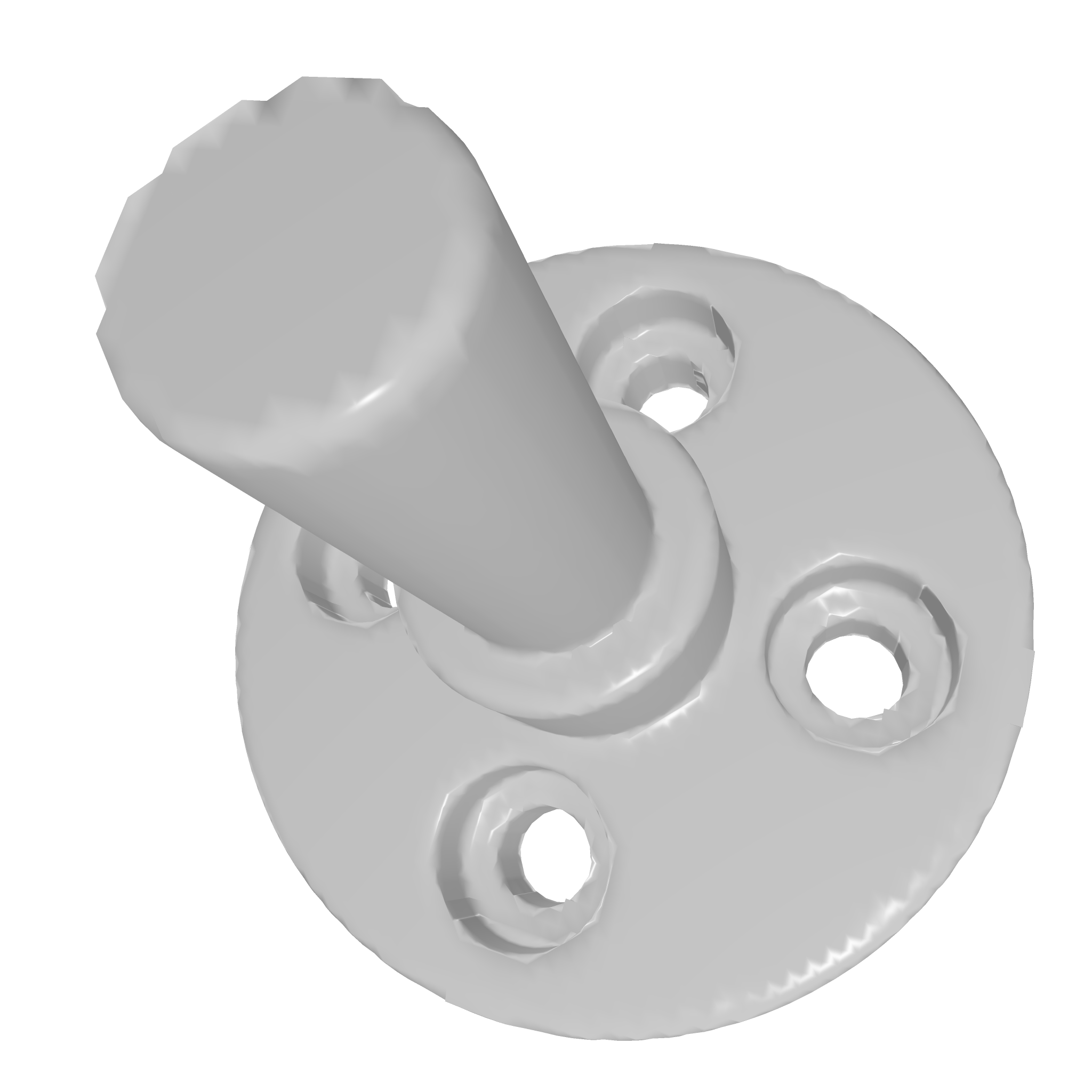} 
			& \includegraphics[valign=m,width=\mcfigwidthsupp,trimabc]{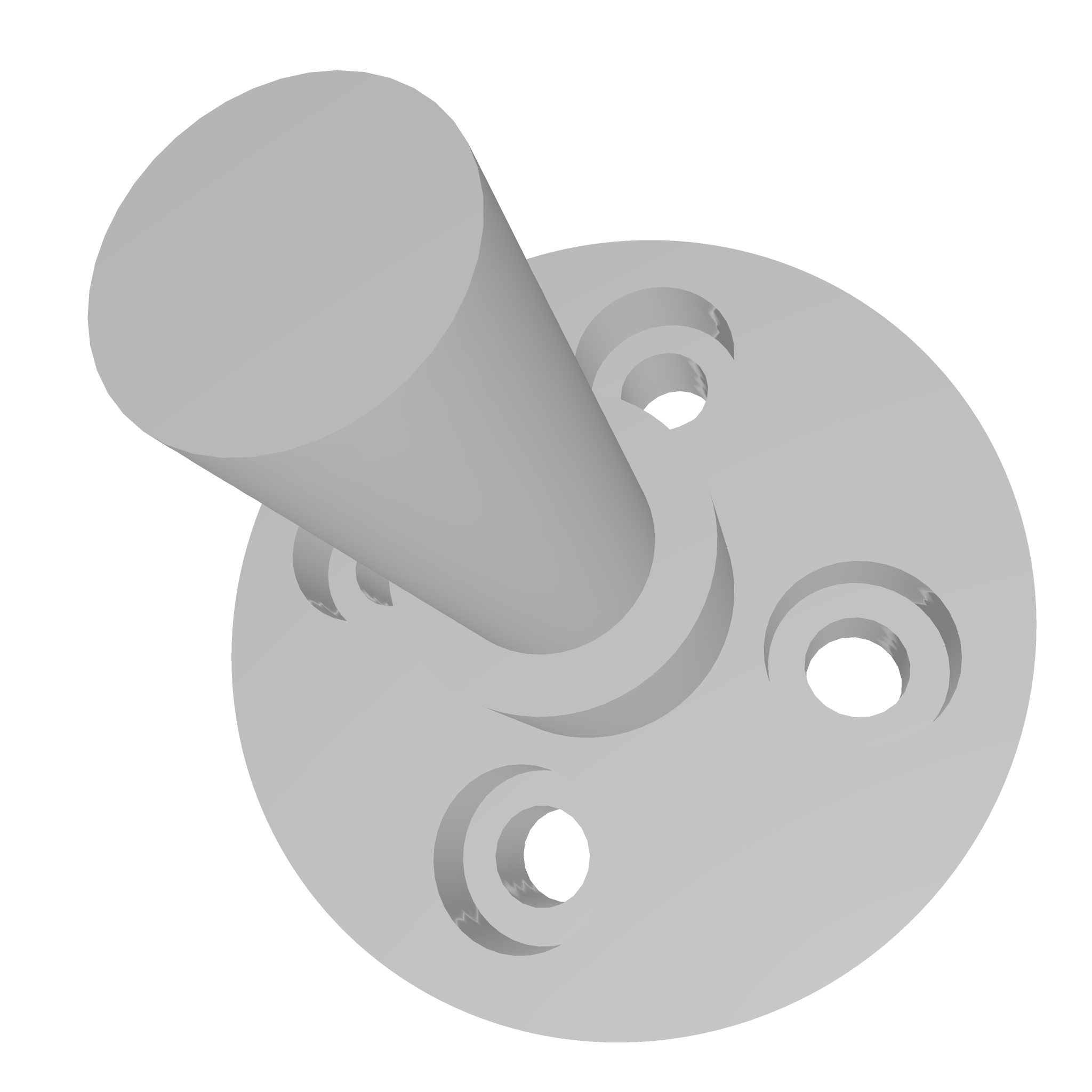}\\ \addlinespace[2\tabcolsep]
			\rotatebox[origin=c]{90}{ABC~\cite{Koch19a}} & \rotatebox[origin=c]{90}{128} &
			\includegraphics[valign=m,width=\mcfigwidthsupp,trimabc]{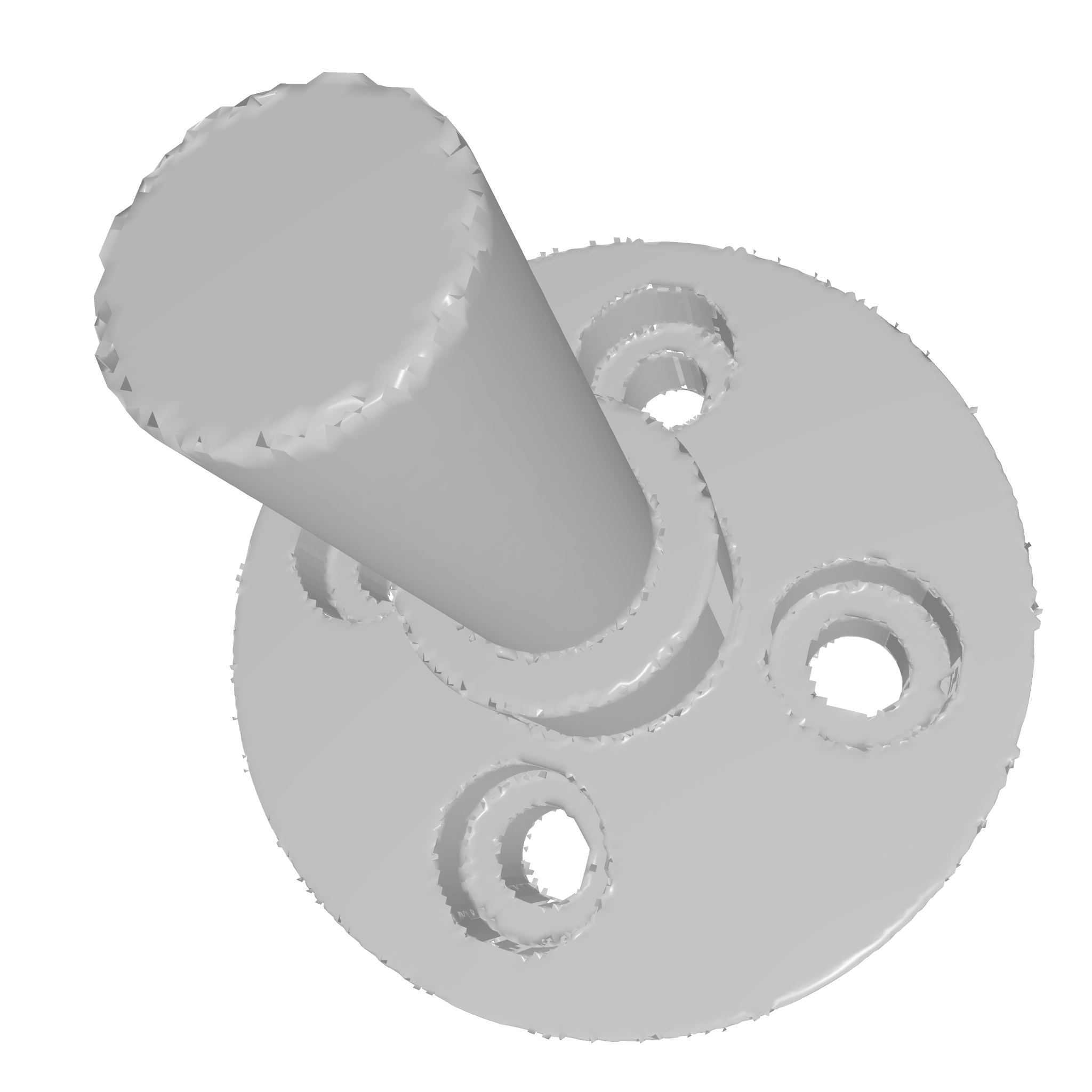}
			& \includegraphics[valign=m,width=\mcfigwidthsupp,trimabc]{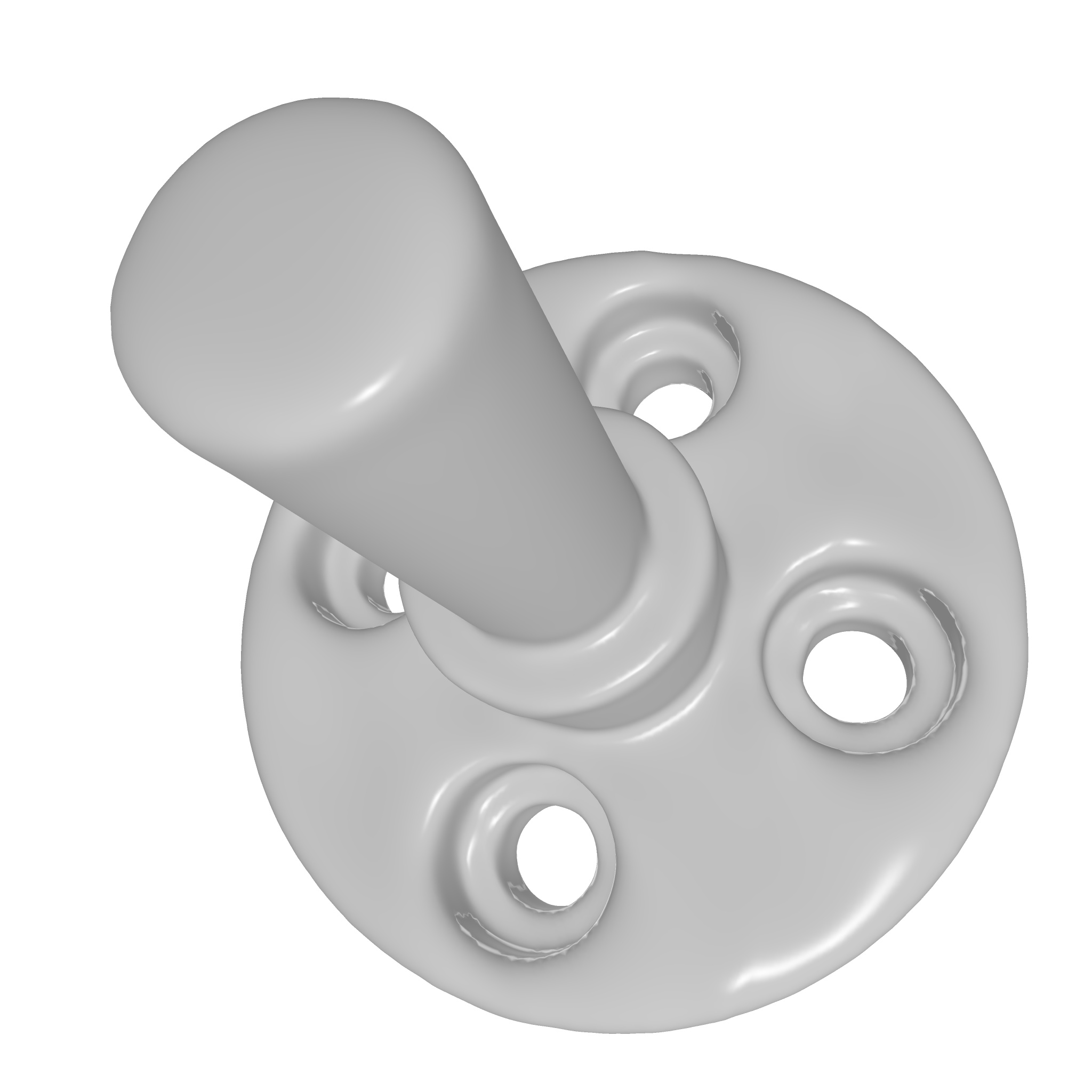}
			& \includegraphics[valign=m,width=\mcfigwidthsupp,trimabc]{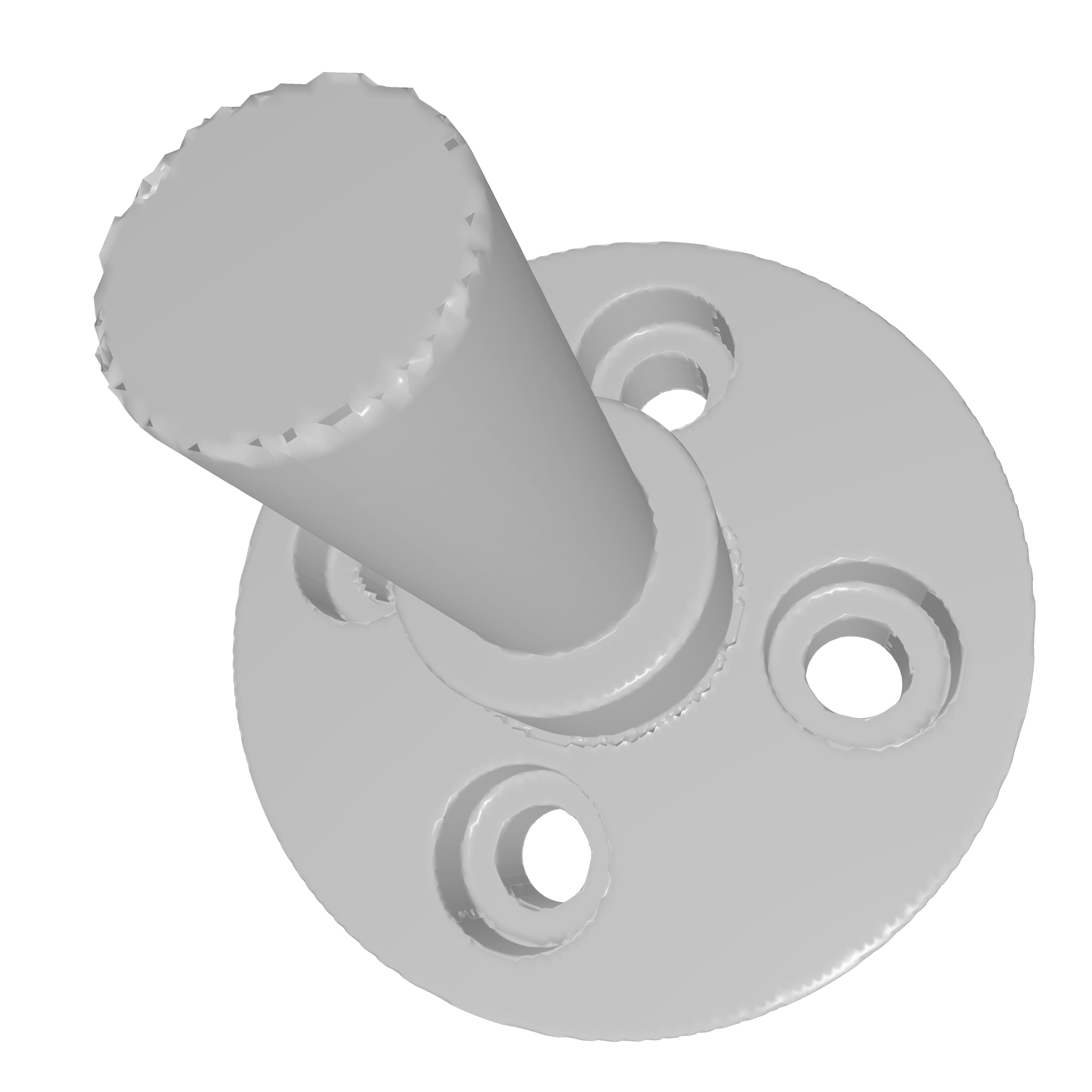}
			& \includegraphics[valign=m,width=\mcfigwidthsupp,trimabc]{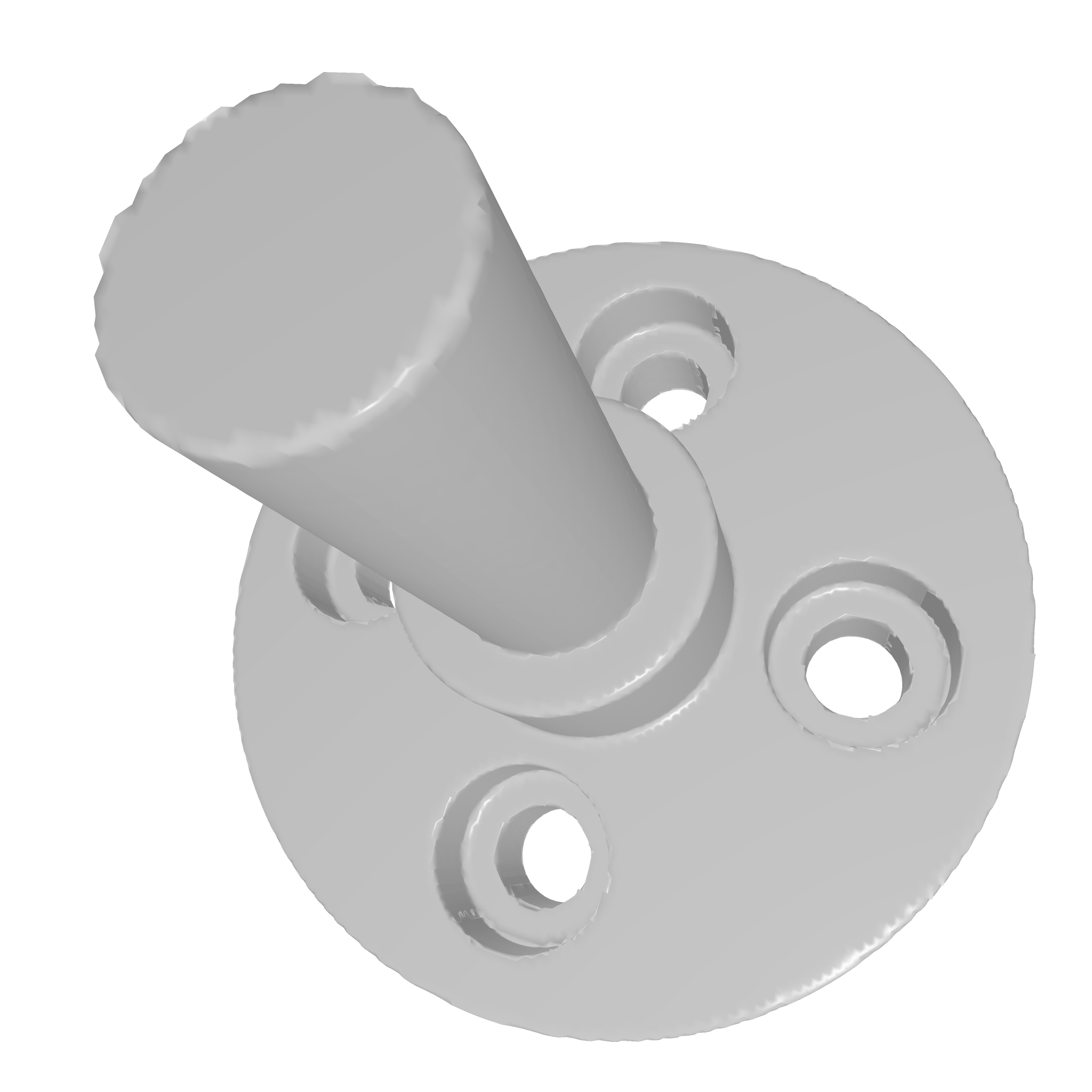} 
			& \includegraphics[valign=m,width=\mcfigwidthsupp,trimabc]{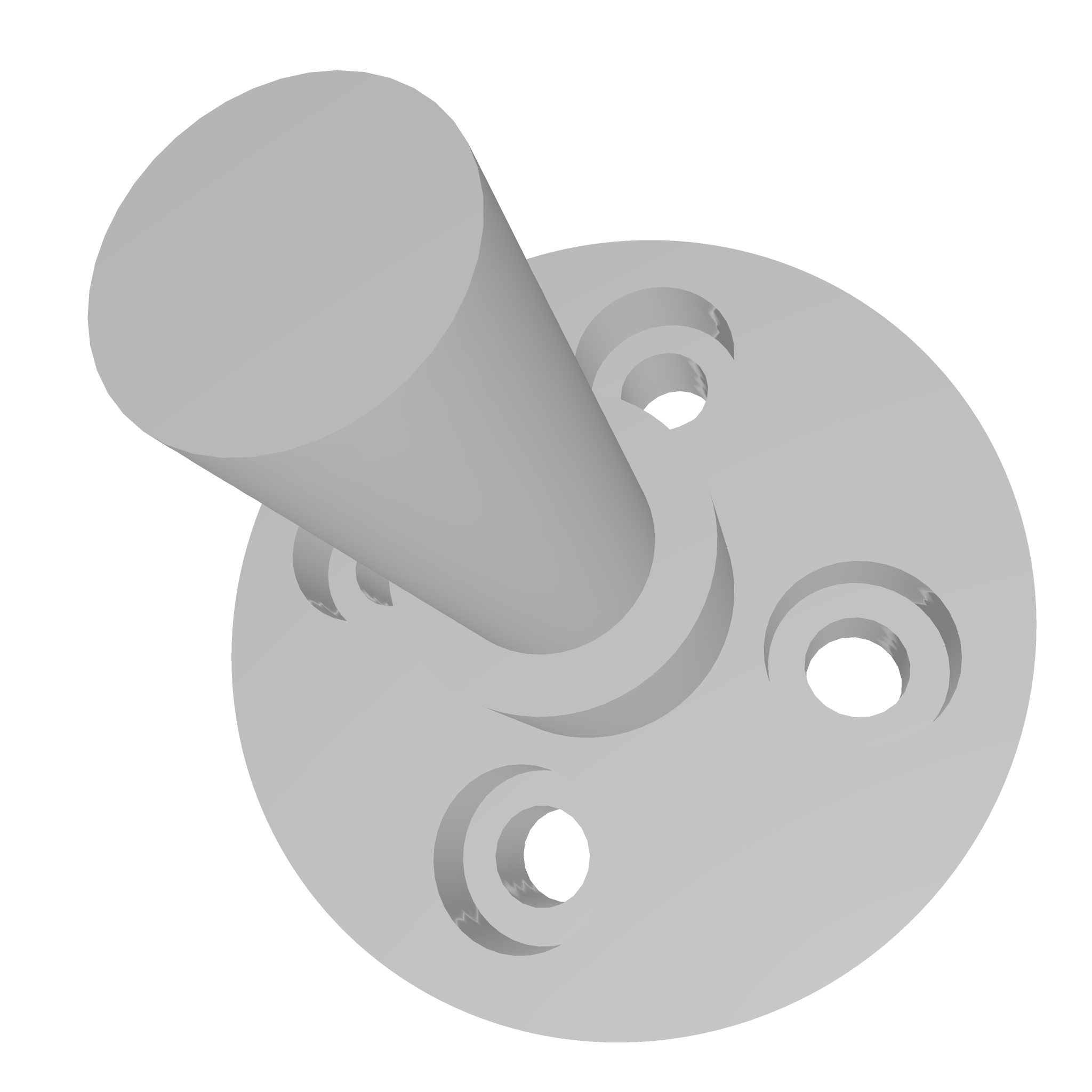}\\ \addlinespace[2\tabcolsep]
			\rotatebox[origin=c]{90}{ABC~\cite{Koch19a}} & \rotatebox[origin=c]{90}{256} &
			\includegraphics[valign=m,width=\mcfigwidthsupp,trimabc]{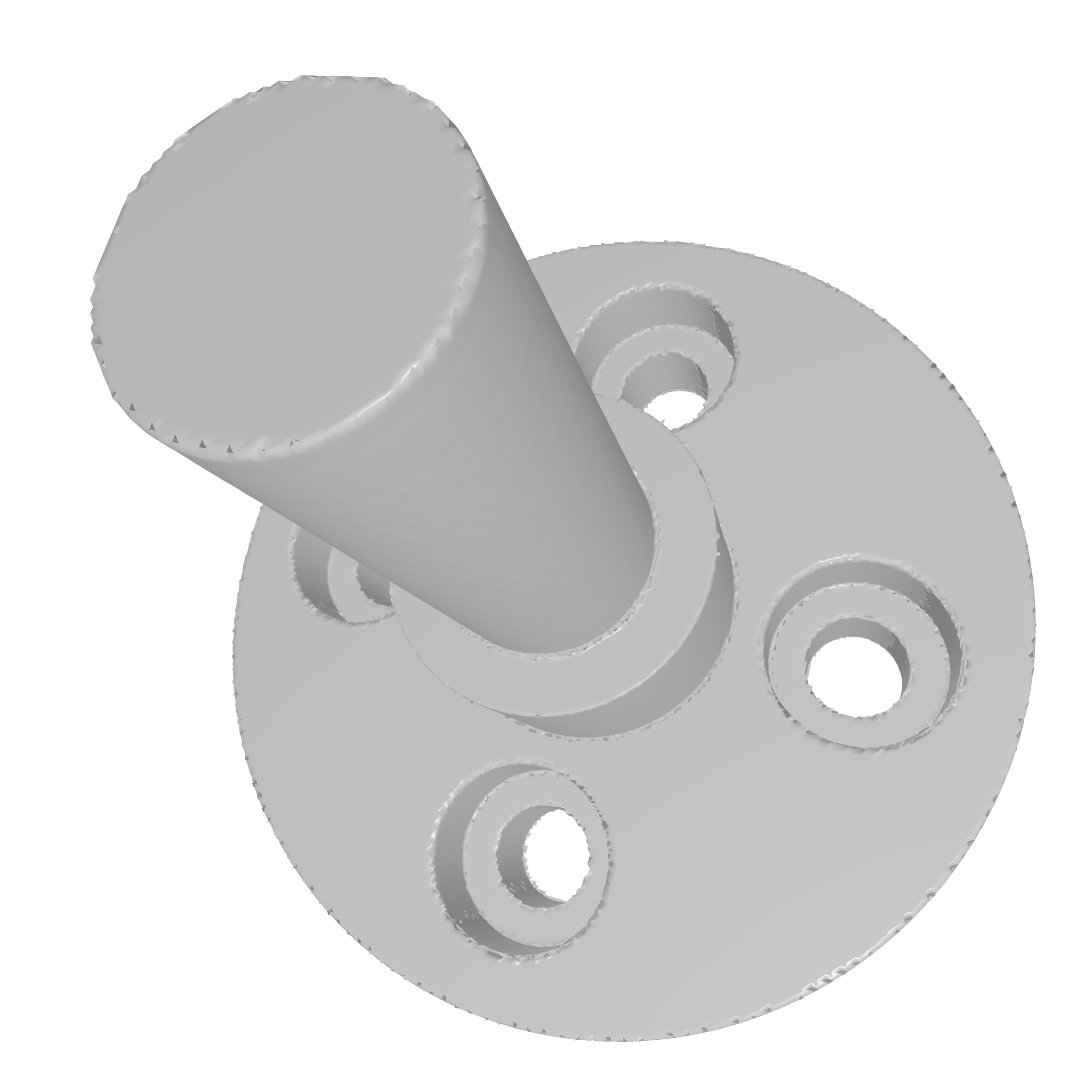}
			& \includegraphics[valign=m,width=\mcfigwidthsupp,trimabc]{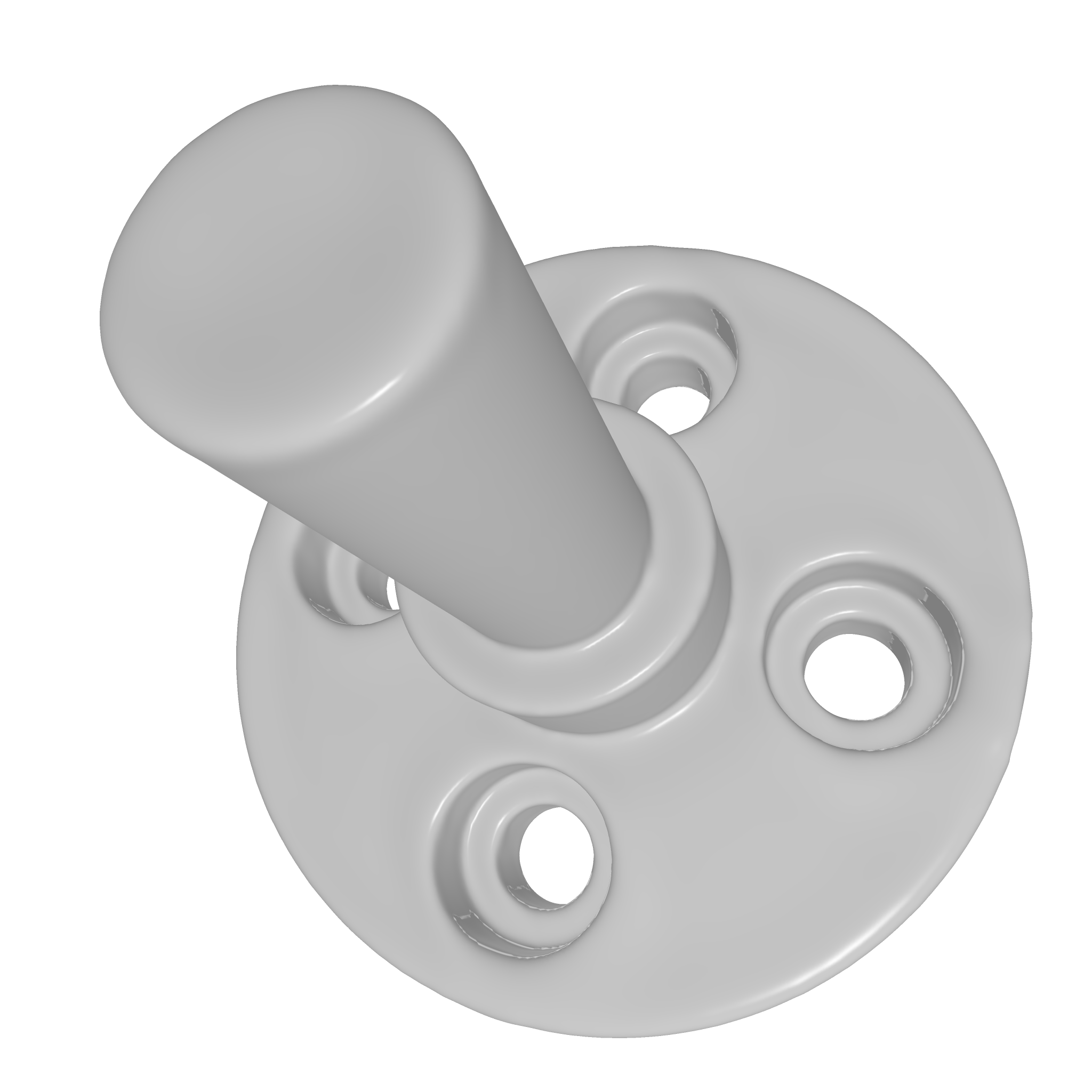}
			& \includegraphics[valign=m,width=\mcfigwidthsupp,trimabc]{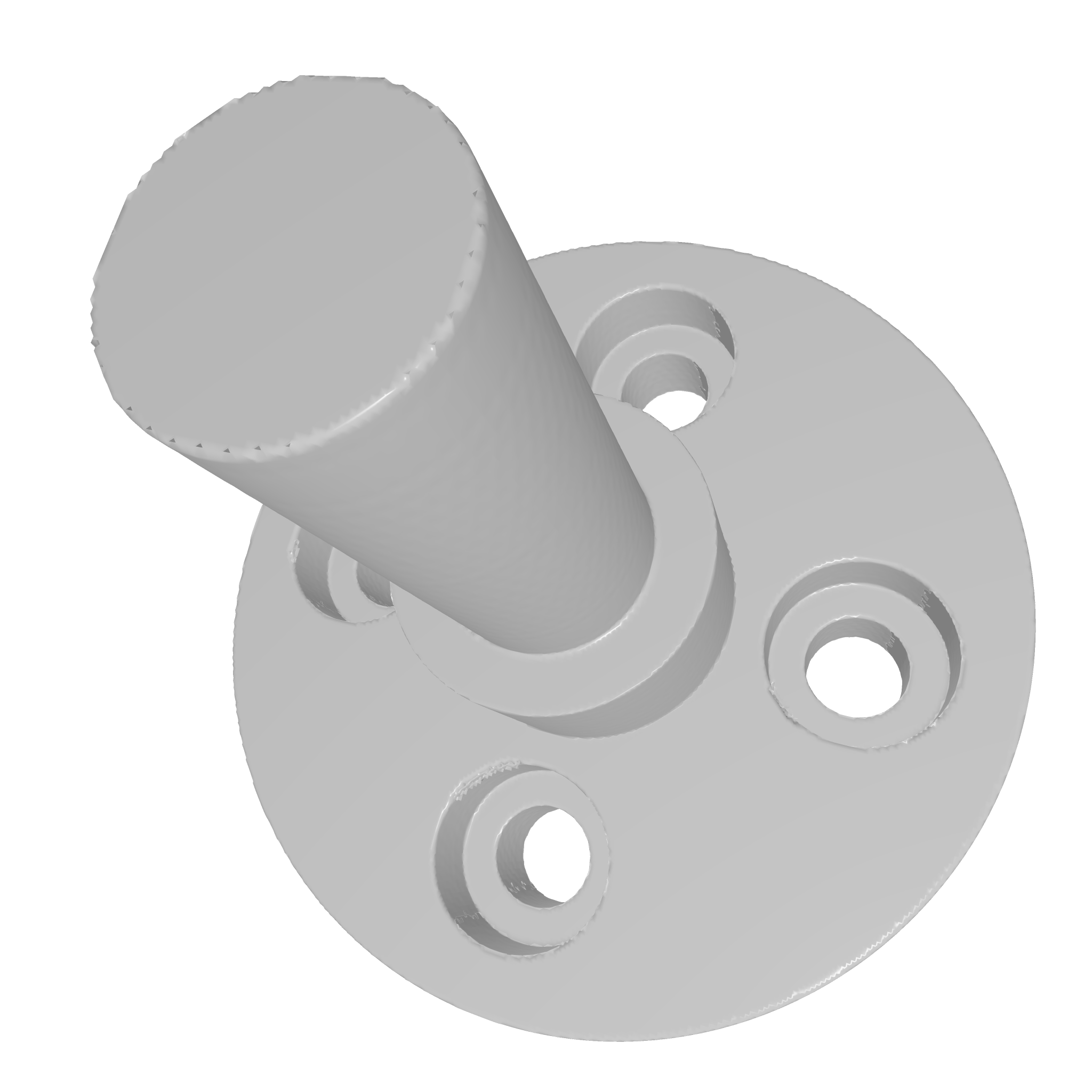}
			& \includegraphics[valign=m,width=\mcfigwidthsupp,trimabc]{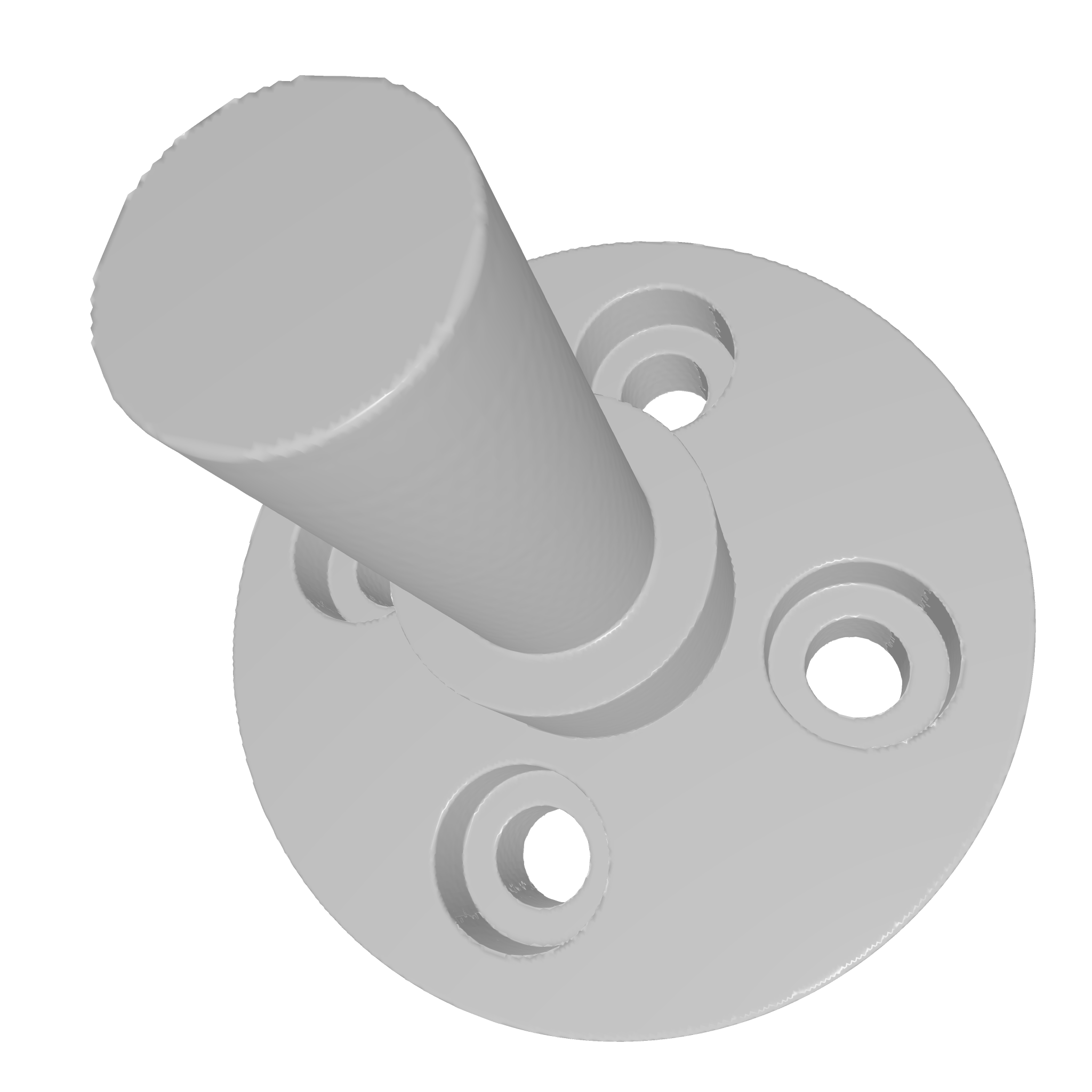} 
			& \includegraphics[valign=m,width=\mcfigwidthsupp,trimabc]{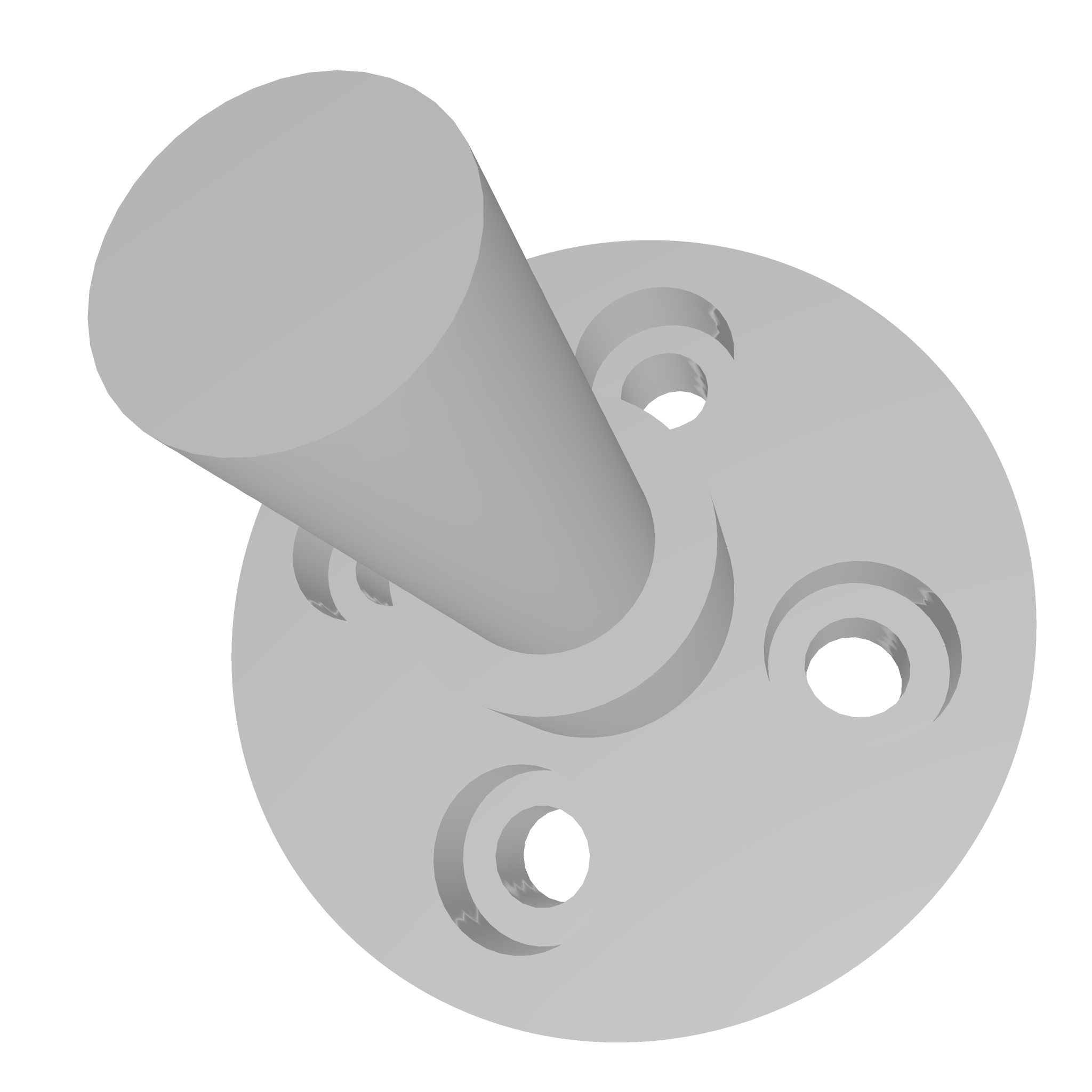}\\ \addlinespace[2\tabcolsep]
			&& CAP-UDF~\cite{Zhou22} & DCUDF~\cite{Hou2023DCUDF} & MeshUDF~\cite{Guillard22b} & Ours & GT
		\end{tabular}}
		\caption{\textbf{Reconstruction methods based on Marching Cubes on ABC~\cite{Koch19a} at varying resolutions.}}
		\label{fig:supp_mc_abc}
	\end{center}
\end{figure}

\begin{figure}[ht!]
	\begin{center}
	{\scriptsize
		\begin{tabular}{llccccc}
			\rotatebox[origin=c]{90}{MGN~\cite{Bhatnagar19}} & \rotatebox[origin=c]{90}{32} &
			\includegraphics[valign=m,width=\mcfigwidthsupp,trimmgn]{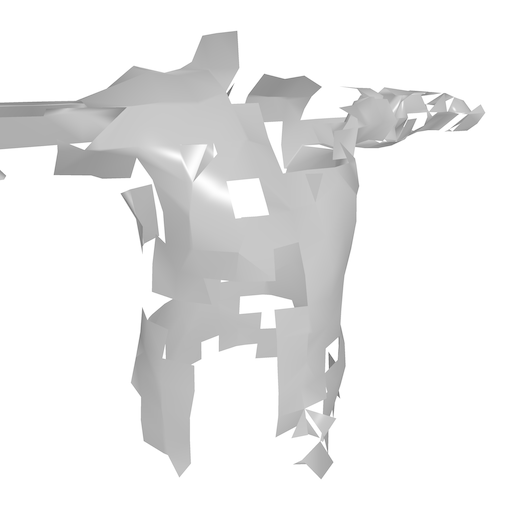}
			& \includegraphics[valign=m,width=\mcfigwidthsupp,trimmgn]{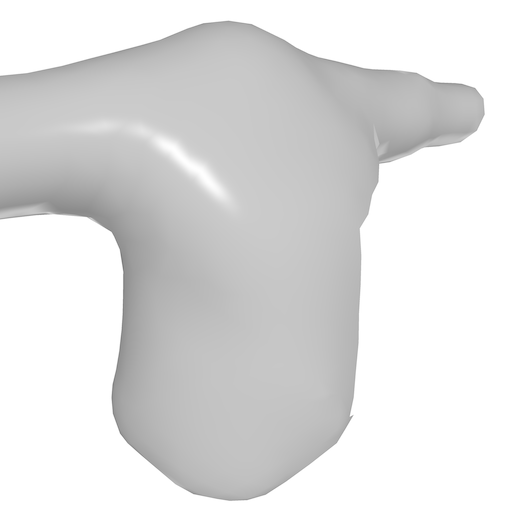}
			& \includegraphics[valign=m,width=\mcfigwidthsupp,trimmgn]{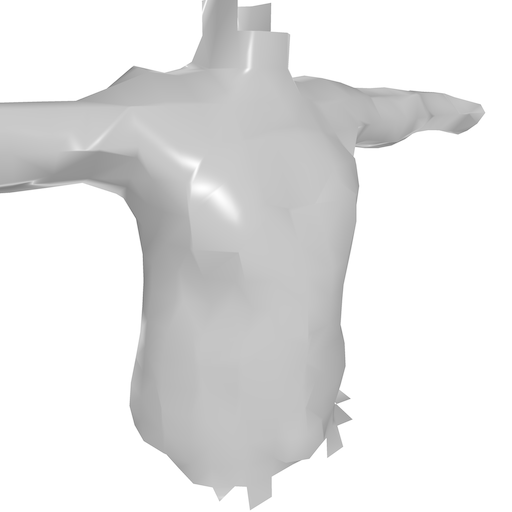}
			& \includegraphics[valign=m,width=\mcfigwidthsupp,trimmgn]{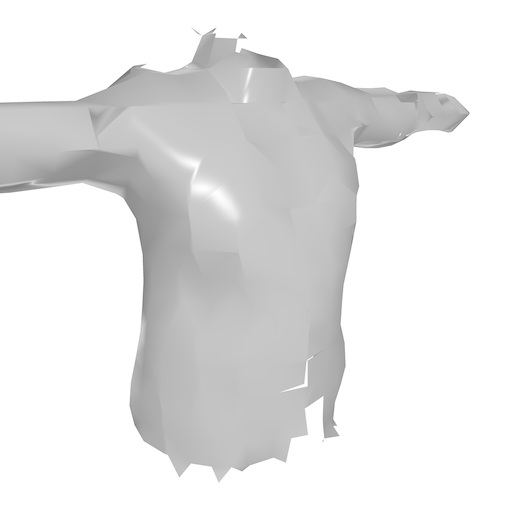}
			& \includegraphics[valign=m,width=\mcfigwidthsupp,trimmgn]{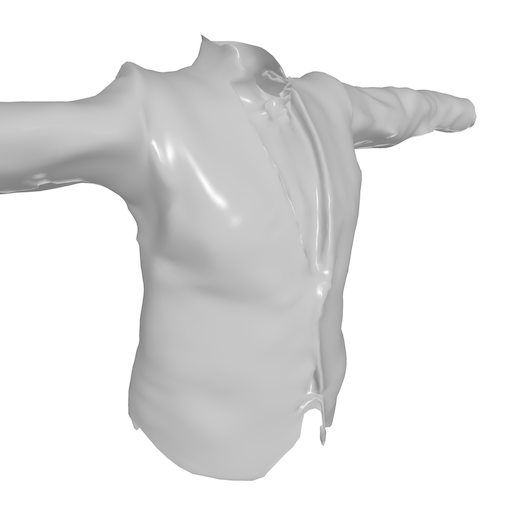}\\ \addlinespace[2\tabcolsep]
			\rotatebox[origin=c]{90}{MGN~\cite{Bhatnagar19}} & \rotatebox[origin=c]{90}{64} &
			\includegraphics[valign=m,width=\mcfigwidthsupp,trimmgn]{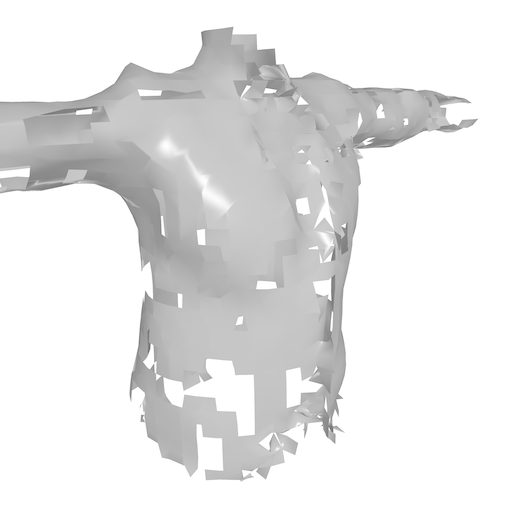}
			& \includegraphics[valign=m,width=\mcfigwidthsupp,trimmgn]{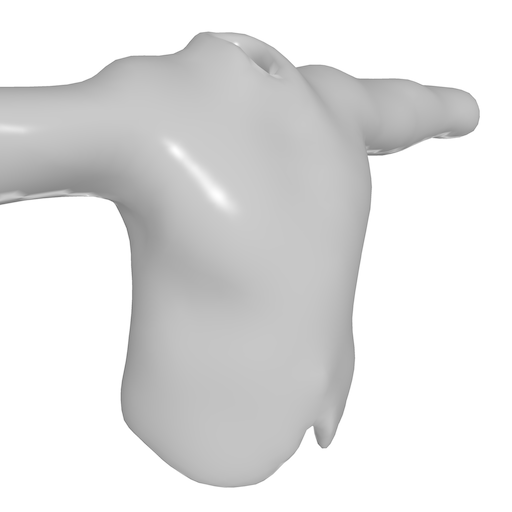}
			& \includegraphics[valign=m,width=\mcfigwidthsupp,trimmgn]{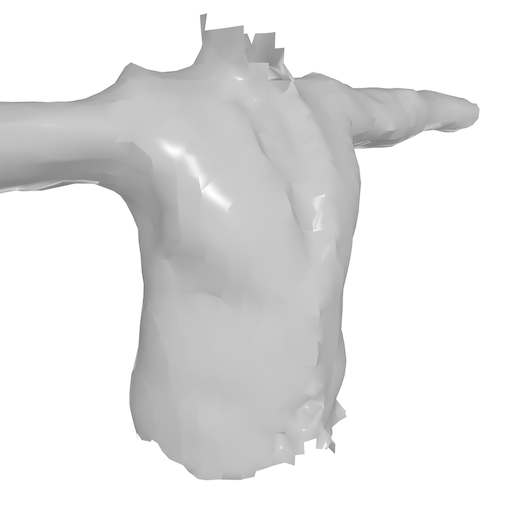}
			& \includegraphics[valign=m,width=\mcfigwidthsupp,trimmgn]{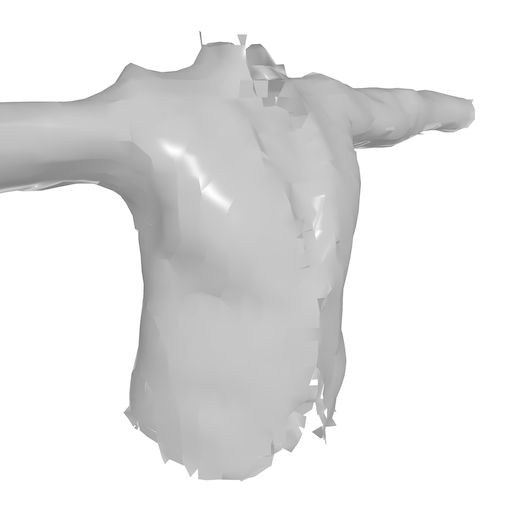}
			& \includegraphics[valign=m,width=\mcfigwidthsupp,trimmgn]{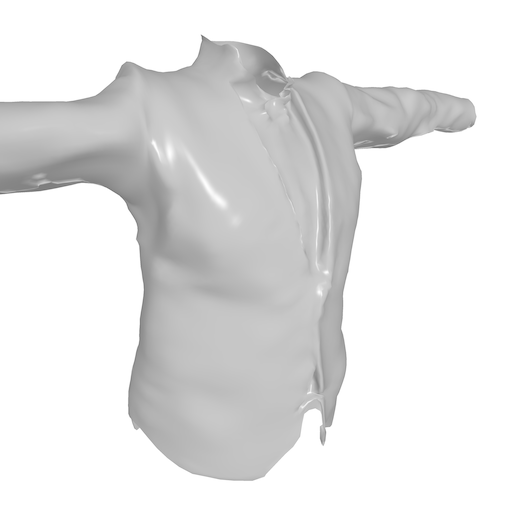}\\ \addlinespace[2\tabcolsep]
			\rotatebox[origin=c]{90}{MGN~\cite{Bhatnagar19}} & \rotatebox[origin=c]{90}{128} &
			\includegraphics[valign=m,width=\mcfigwidthsupp,trimmgn]{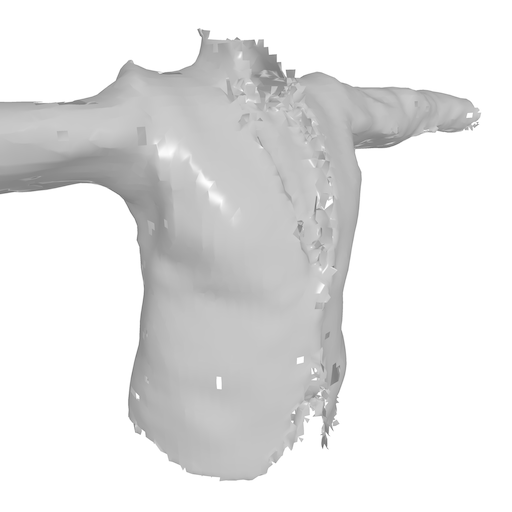}
			& \includegraphics[valign=m,width=\mcfigwidthsupp,trimmgn]{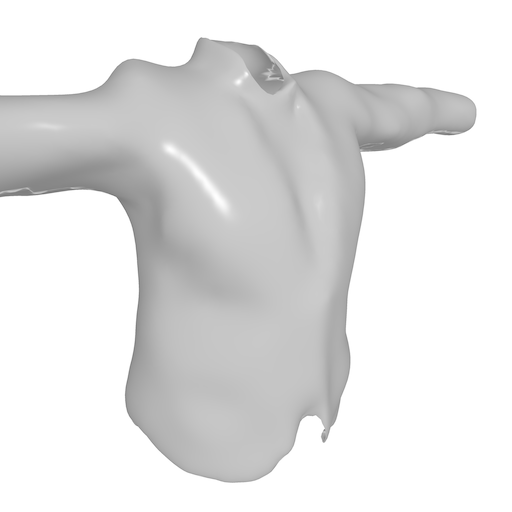}
			& \includegraphics[valign=m,width=\mcfigwidthsupp,trimmgn]{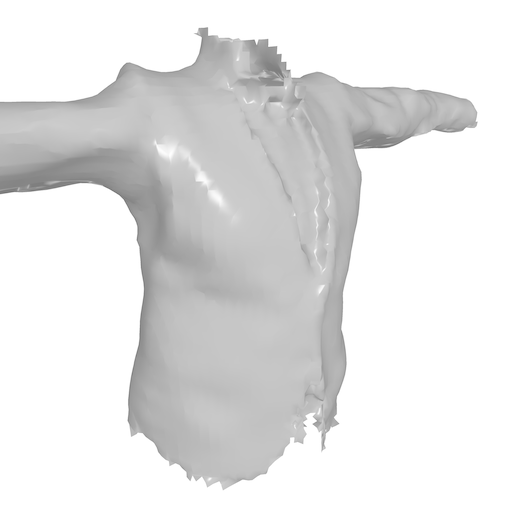}
			& \includegraphics[valign=m,width=\mcfigwidthsupp,trimmgn]{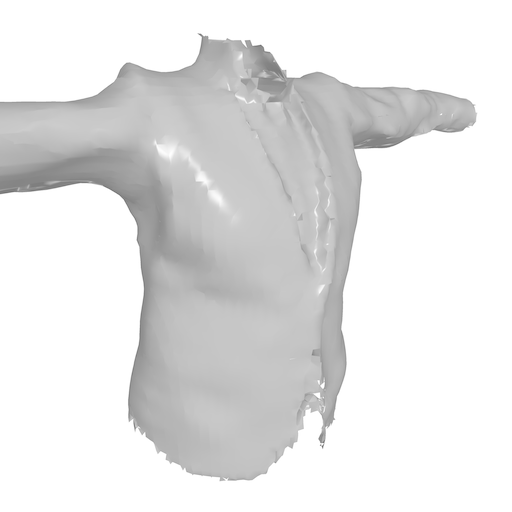}
			& \includegraphics[valign=m,width=\mcfigwidthsupp,trimmgn]{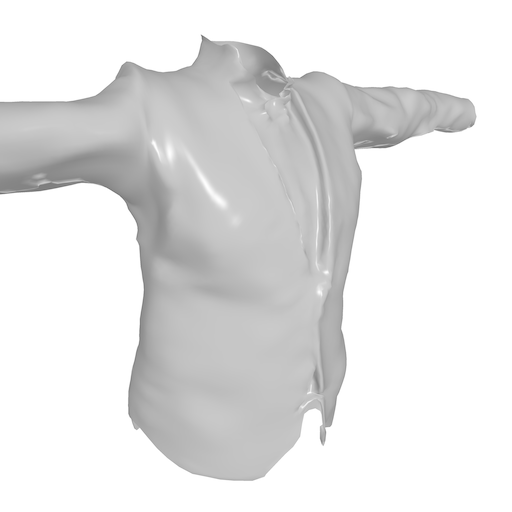}\\ \addlinespace[2\tabcolsep]
			\rotatebox[origin=c]{90}{MGN~\cite{Bhatnagar19}} & \rotatebox[origin=c]{90}{256} &
			\includegraphics[valign=m,width=\mcfigwidthsupp,trimmgn]{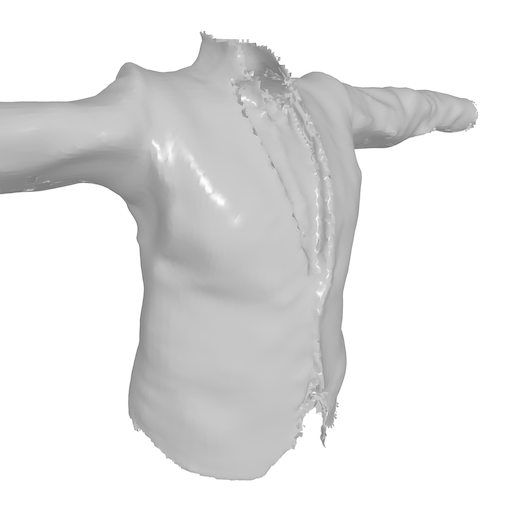}
			& \includegraphics[valign=m,width=\mcfigwidthsupp,trimmgn]{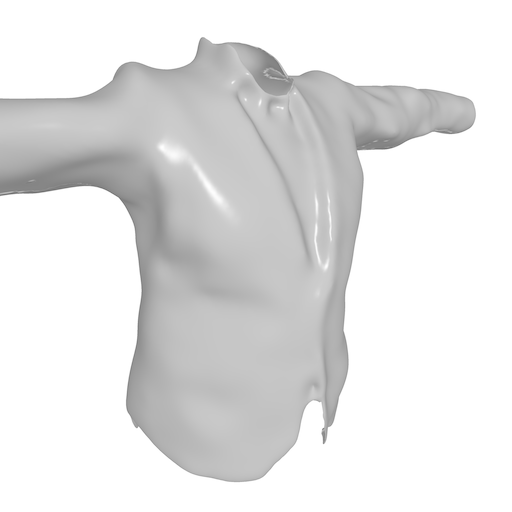}
			& \includegraphics[valign=m,width=\mcfigwidthsupp,trimmgn]{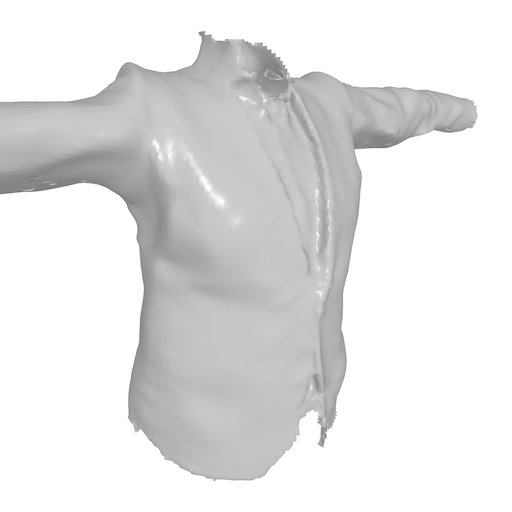}
			& \includegraphics[valign=m,width=\mcfigwidthsupp,trimmgn]{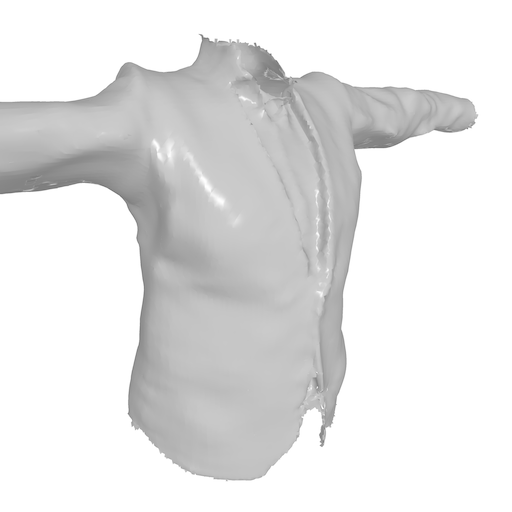}
			& \includegraphics[valign=m,width=\mcfigwidthsupp,trimmgn]{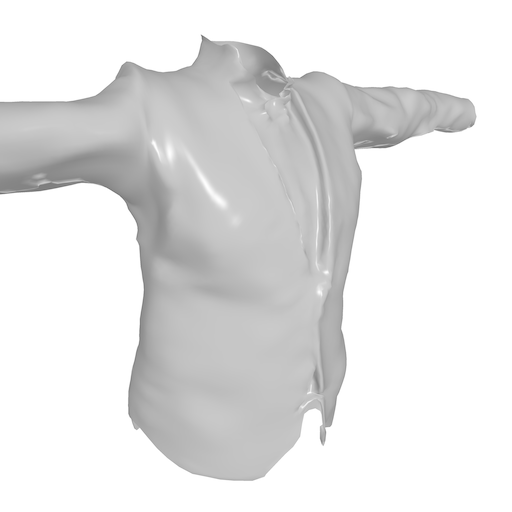}\\ \addlinespace[2\tabcolsep]
			&& CAP-UDF~\cite{Zhou22} & DCUDF~\cite{Hou2023DCUDF} & MeshUDF~\cite{Guillard22b} & Ours & GT
		\end{tabular}}
		\caption{\textbf{Reconstruction methods based on Marching Cubes on MGN~\cite{Bhatnagar19} at varying resolutions.}}
		\label{fig:supp_mc_mgn}
	\end{center}
\end{figure}

\begin{figure}[ht!]
		\begin{center}
		{\scriptsize
			\begin{tabular}{llccccc}
				\rotatebox[origin=c]{90}{Cars~\cite{Chang15}} & \rotatebox[origin=c]{90}{64} &
				\includegraphics[valign=m,width=\mcfigwidthsupp,trimsn]{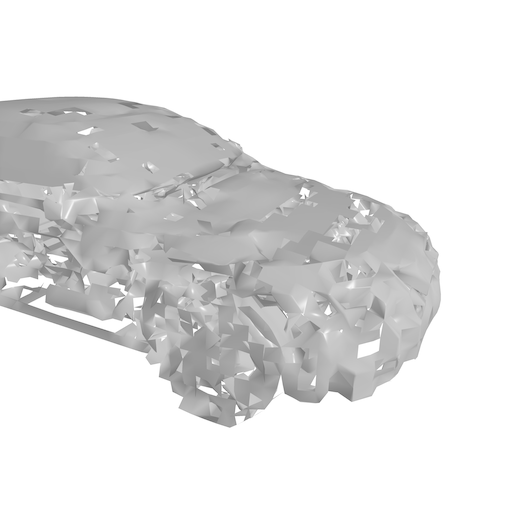}
				& \includegraphics[valign=m,width=\mcfigwidthsupp,trimsn]{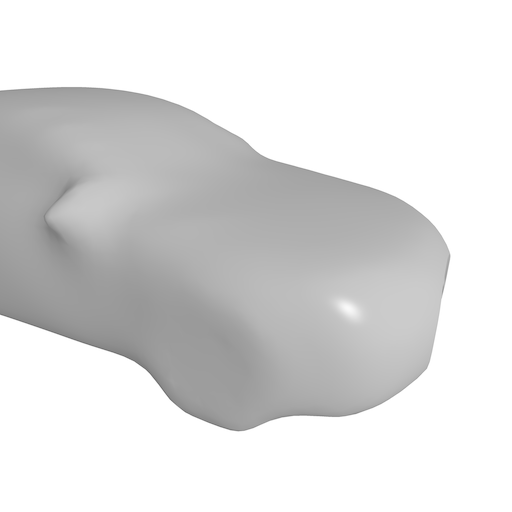}
				& \includegraphics[valign=m,width=\mcfigwidthsupp,trimsn]{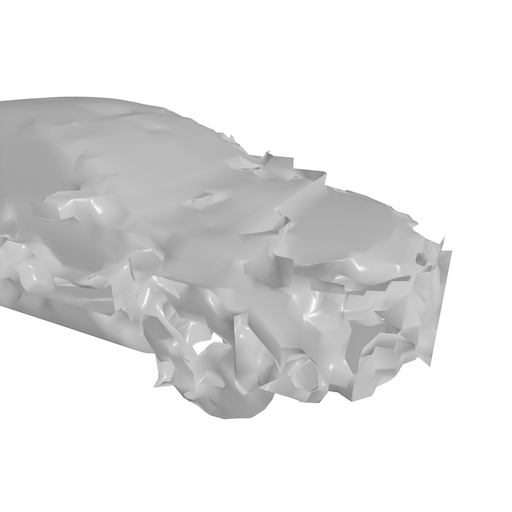}
				& \includegraphics[valign=m,width=\mcfigwidthsupp,trimsn]{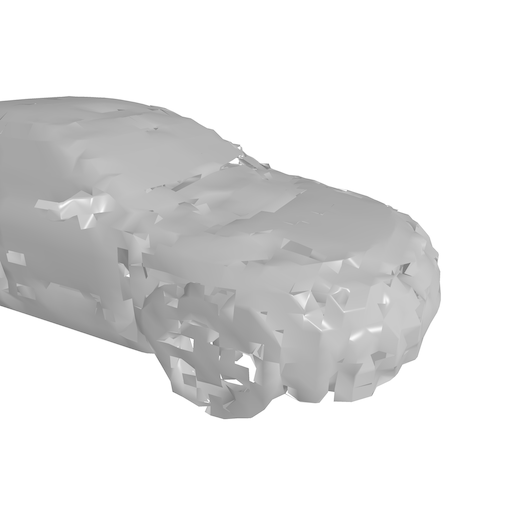}
				& \includegraphics[valign=m,width=\mcfigwidthsupp,trimsn]{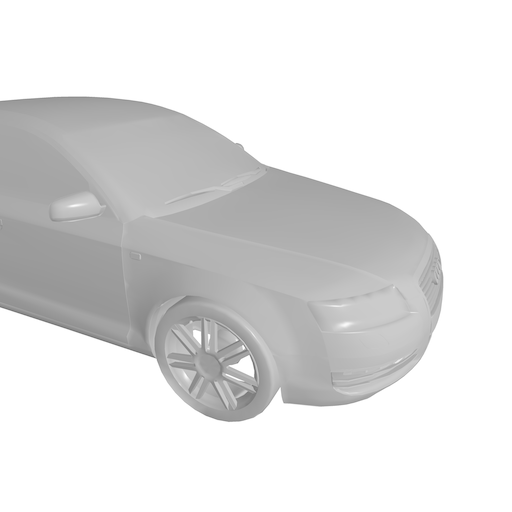}\\ \addlinespace[2\tabcolsep]
				\rotatebox[origin=c]{90}{Cars~\cite{Chang15}} & \rotatebox[origin=c]{90}{128} &
				\includegraphics[valign=m,width=\mcfigwidthsupp,trimsn]{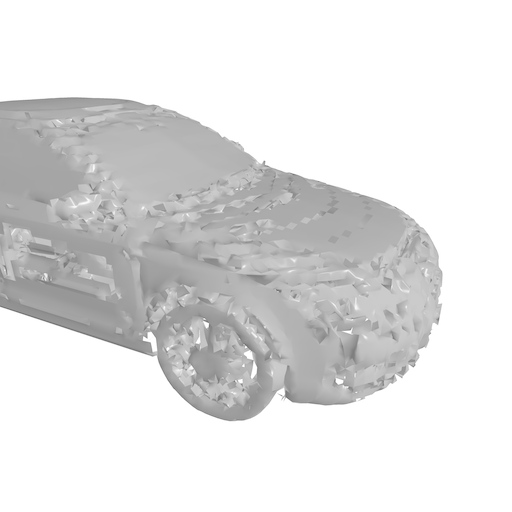}
				& \includegraphics[valign=m,width=\mcfigwidthsupp,trimsn]{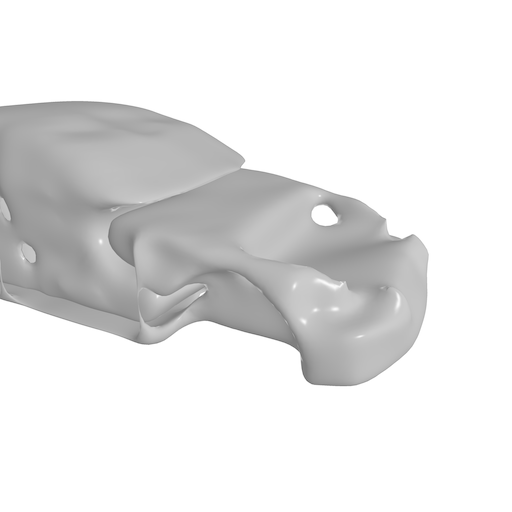}
				& \includegraphics[valign=m,width=\mcfigwidthsupp,trimsn]{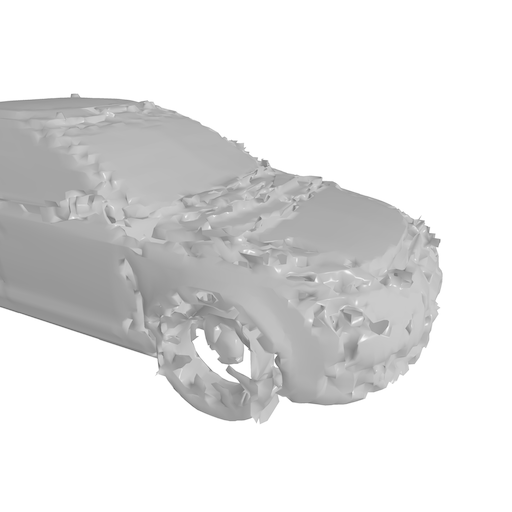}
				& \includegraphics[valign=m,width=\mcfigwidthsupp,trimsn]{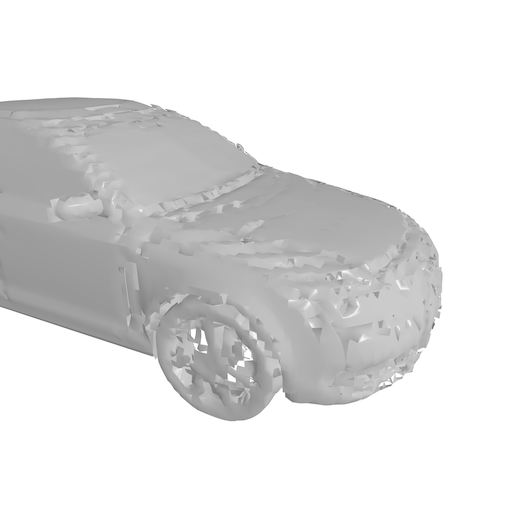}
				& \includegraphics[valign=m,width=\mcfigwidthsupp,trimsn]{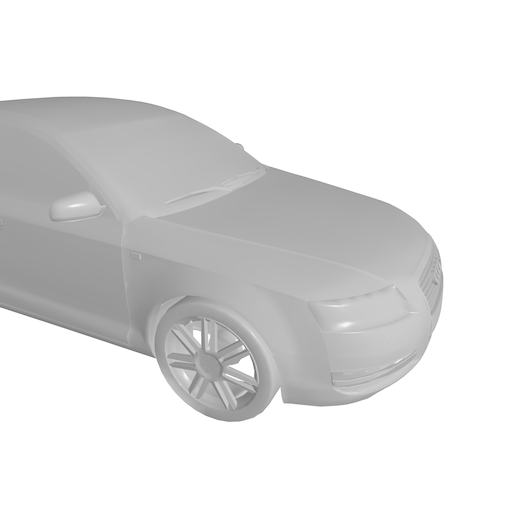}\\ \addlinespace[2\tabcolsep]
				\rotatebox[origin=c]{90}{Cars~\cite{Chang15}} & \rotatebox[origin=c]{90}{256} &
				\includegraphics[valign=m,width=\mcfigwidthsupp,trimsn]{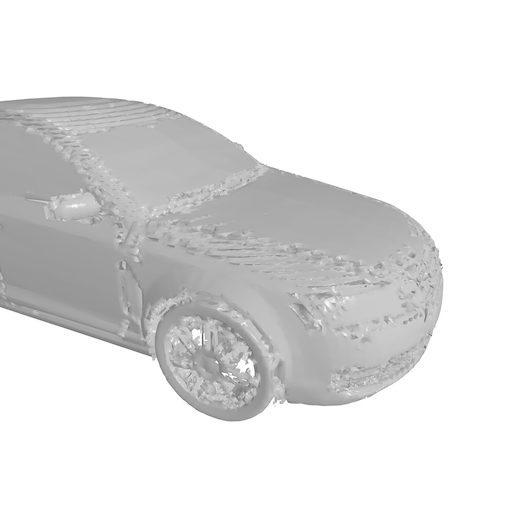}
				& \includegraphics[valign=m,width=\mcfigwidthsupp,trimsn]{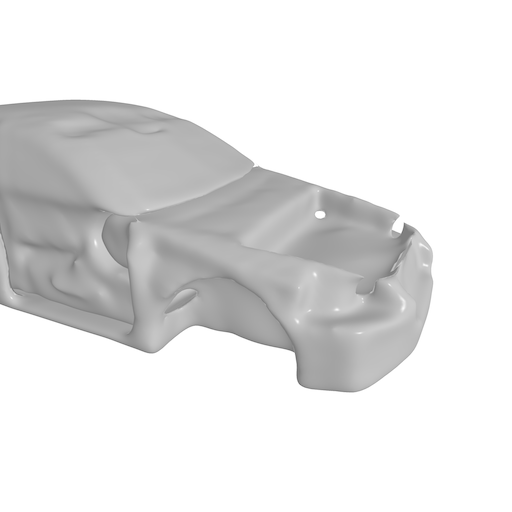}
				& \includegraphics[valign=m,width=\mcfigwidthsupp,trimsn]{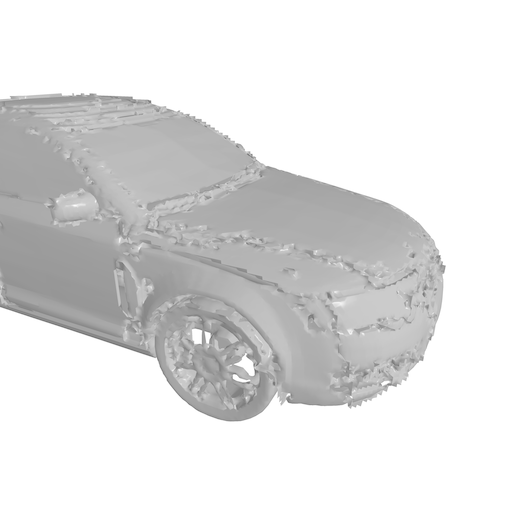}
				& \includegraphics[valign=m,width=\mcfigwidthsupp,trimsn]{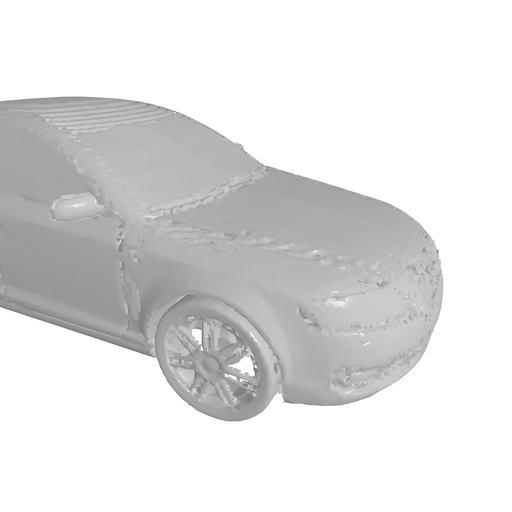}
				& \includegraphics[valign=m,width=\mcfigwidthsupp,trimsn]{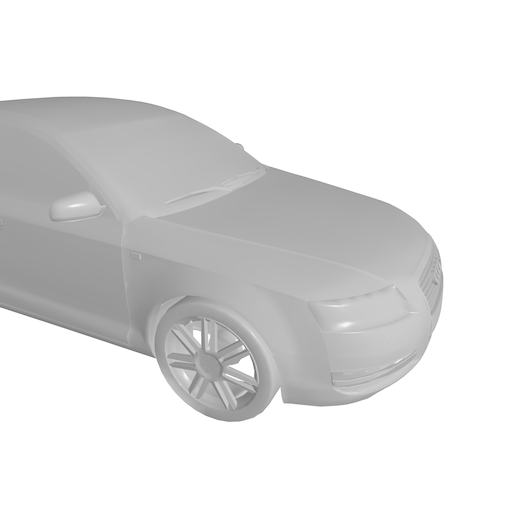}\\ \addlinespace[2\tabcolsep]
				\rotatebox[origin=c]{90}{Cars~\cite{Chang15}} & \rotatebox[origin=c]{90}{512} &
				\includegraphics[valign=m,width=\mcfigwidthsupp,trimsn]{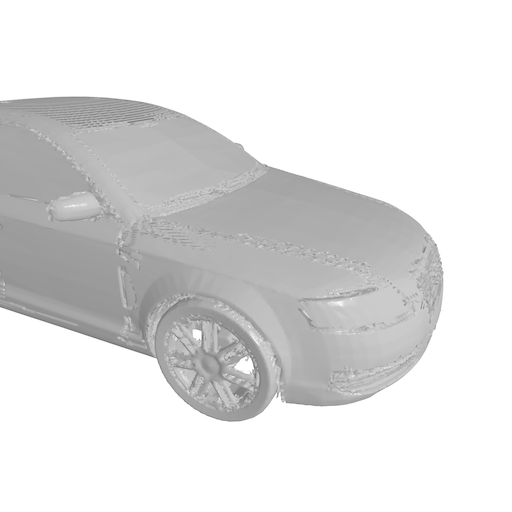}
				& \includegraphics[valign=m,width=\mcfigwidthsupp,trimsn]{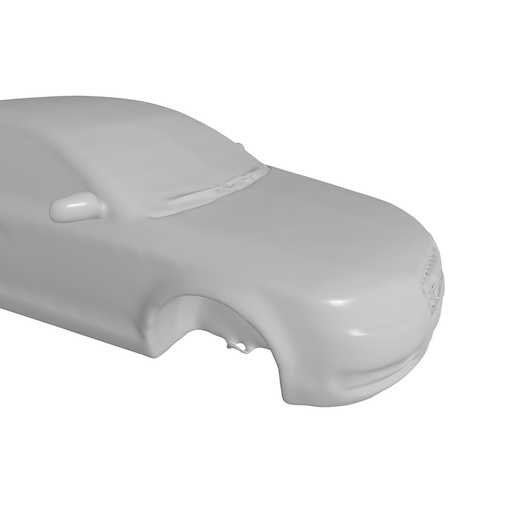}
				& \includegraphics[valign=m,width=\mcfigwidthsupp,trimsn]{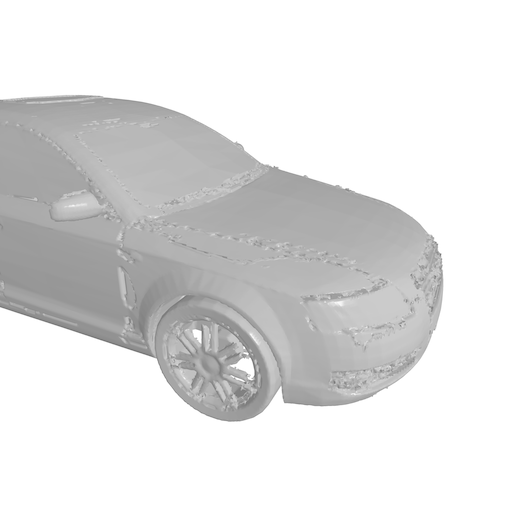}
				& \includegraphics[valign=m,width=\mcfigwidthsupp,trimsn]{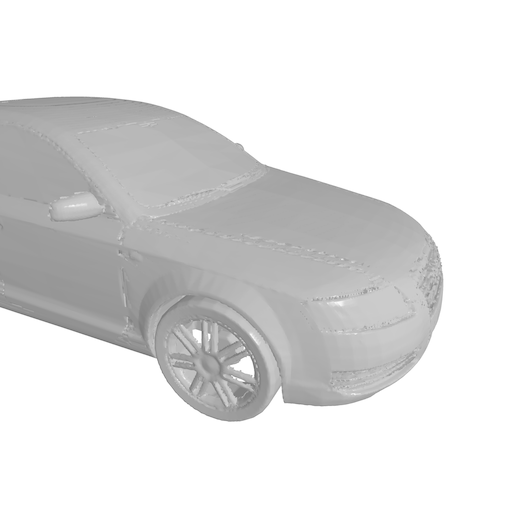}
				& \includegraphics[valign=m,width=\mcfigwidthsupp,trimsn]{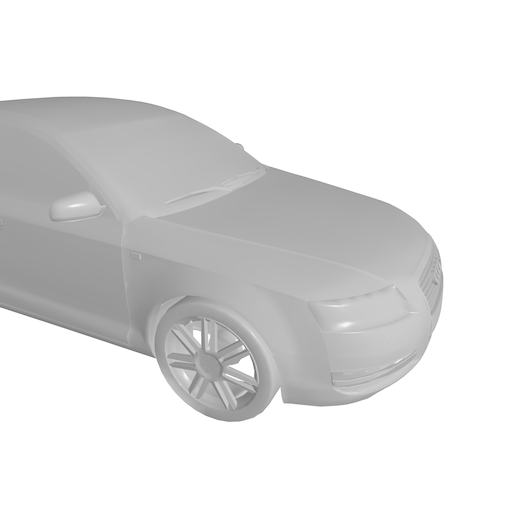}\\ \addlinespace[2\tabcolsep]
				&& CAP-UDF~\cite{Zhou22} & DCUDF~\cite{Hou2023DCUDF} & MeshUDF~\cite{Guillard22b} & Ours & GT
			\end{tabular}}
			\caption{\textbf{Reconstruction methods based on Marching Cubes on ShapeNet-Cars~\cite{Chang15} at varying resolutions.}}
			\label{fig:supp_mc_cars}
		\end{center}
	\end{figure}

	\begin{figure}[ht!]
			\begin{center}
			{\scriptsize
				\begin{tabular}{llccccc}
					\rotatebox[origin=c]{90}{MGN autodec.\phantom{AA}} & \rotatebox[origin=c]{90}{32} &
					\includegraphics[valign=m,width=\mcfigwidthsupp,trimmgnad]{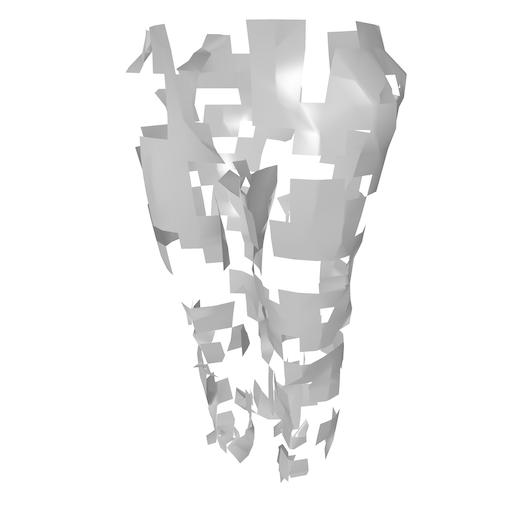}
					& \includegraphics[valign=m,width=\mcfigwidthsupp,trimmgnad]{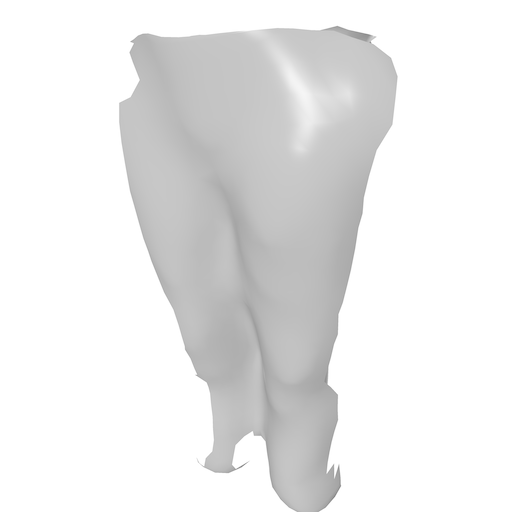}
					& \includegraphics[valign=m,width=\mcfigwidthsupp,trimmgnad]{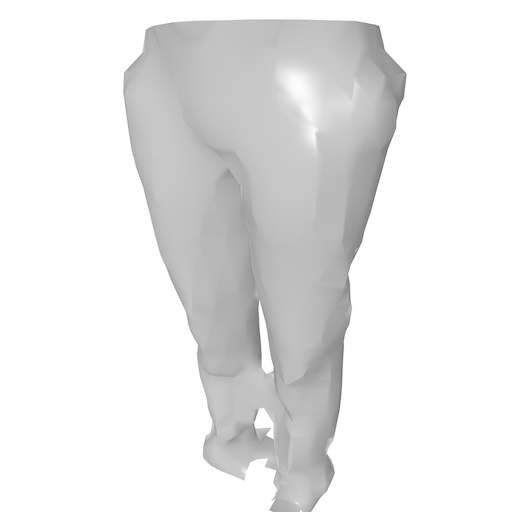}
					& \includegraphics[valign=m,width=\mcfigwidthsupp,trimmgnad]{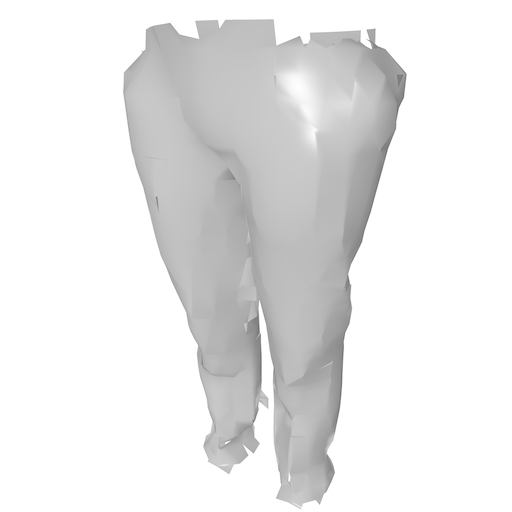}
					& \includegraphics[valign=m,width=\mcfigwidthsupp,trimmgnad]{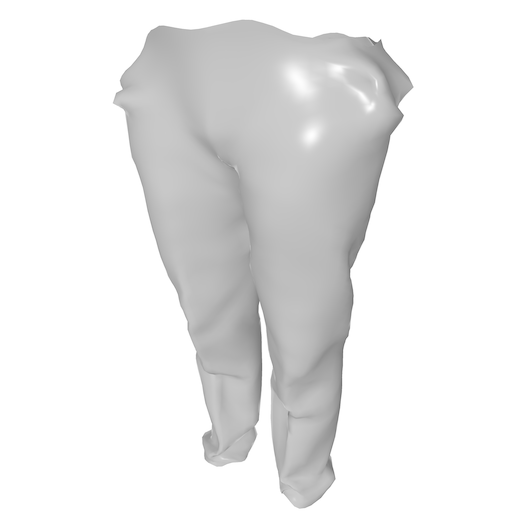}\\ \addlinespace[2\tabcolsep]
					\rotatebox[origin=c]{90}{MGN autodec.\phantom{AA}} & \rotatebox[origin=c]{90}{64} &
					\includegraphics[valign=m,width=\mcfigwidthsupp,trimmgnad]{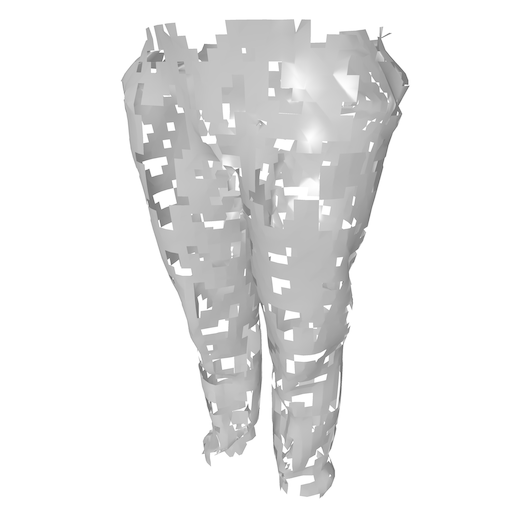}
					& \includegraphics[valign=m,width=\mcfigwidthsupp,trimmgnad]{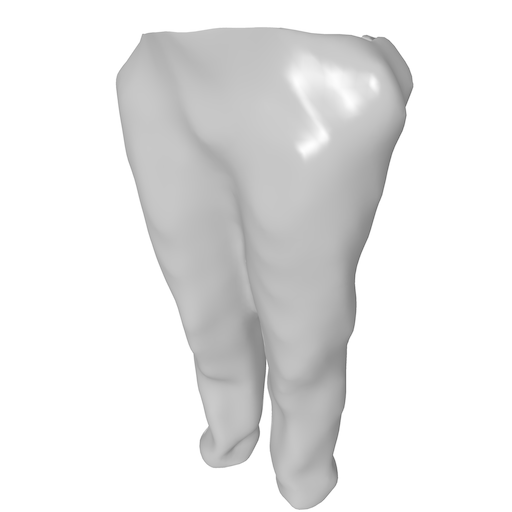}
					& \includegraphics[valign=m,width=\mcfigwidthsupp,trimmgnad]{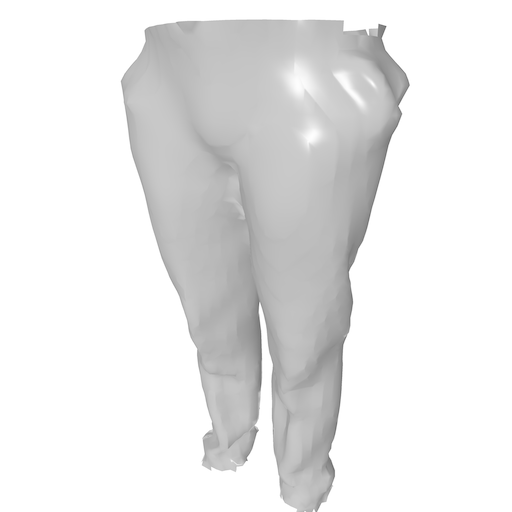}
					& \includegraphics[valign=m,width=\mcfigwidthsupp,trimmgnad]{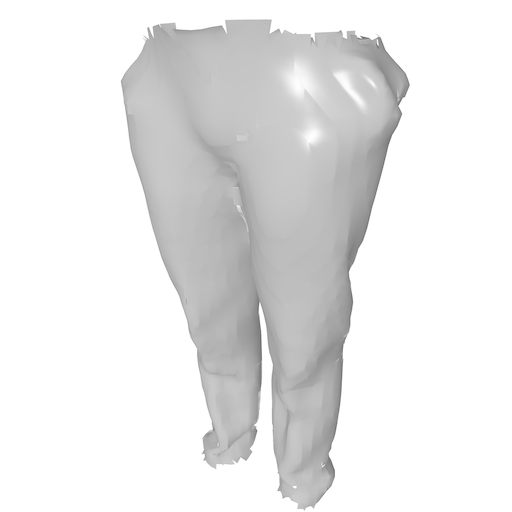}
					& \includegraphics[valign=m,width=\mcfigwidthsupp,trimmgnad]{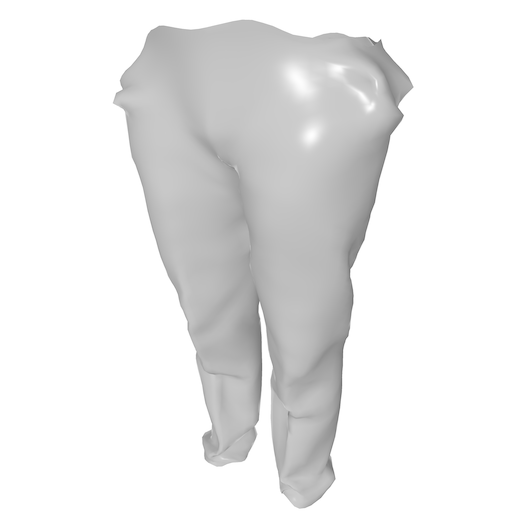}\\ \addlinespace[2\tabcolsep]
					\rotatebox[origin=c]{90}{MGN autodec.\phantom{AA}} & \rotatebox[origin=c]{90}{128} &
					\includegraphics[valign=m,width=\mcfigwidthsupp,trimmgnad]{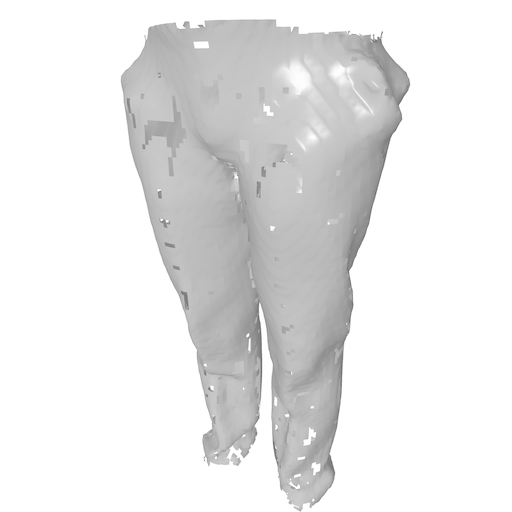}
					& \includegraphics[valign=m,width=\mcfigwidthsupp,trimmgnad]{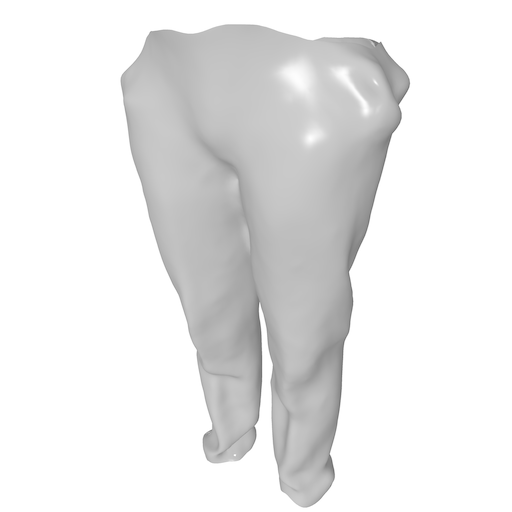}
					& \includegraphics[valign=m,width=\mcfigwidthsupp,trimmgnad]{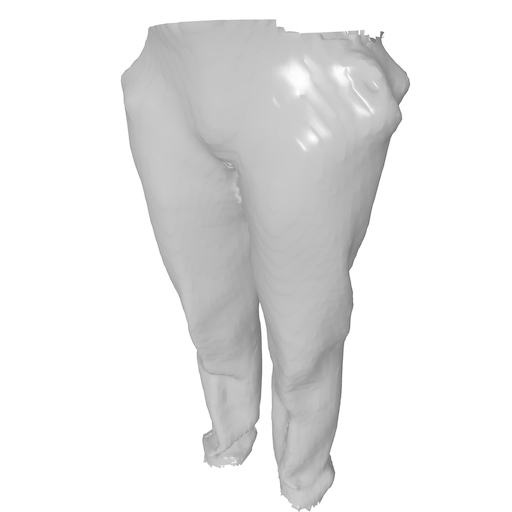}
					& \includegraphics[valign=m,width=\mcfigwidthsupp,trimmgnad]{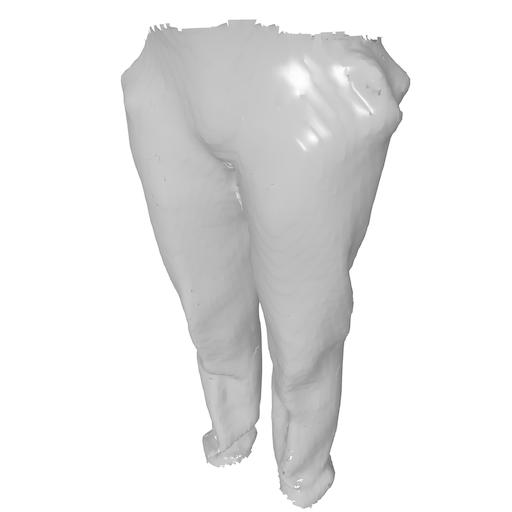}
					& \includegraphics[valign=m,width=\mcfigwidthsupp,trimmgnad]{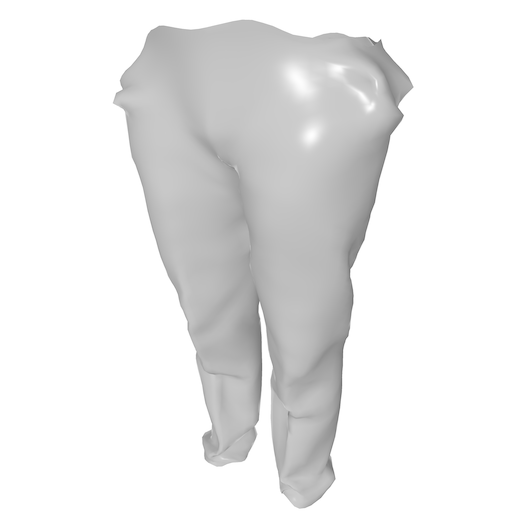}\\ \addlinespace[2\tabcolsep]
					\rotatebox[origin=c]{90}{MGN autodec.\phantom{AA}} & \rotatebox[origin=c]{90}{256} &
					\includegraphics[valign=m,width=\mcfigwidthsupp,trimmgnad]{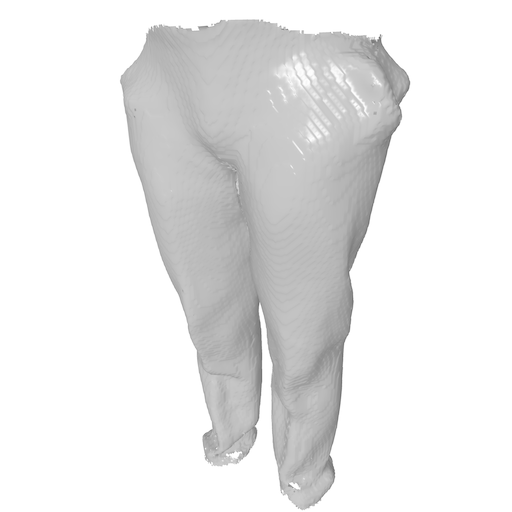}
					& \includegraphics[valign=m,width=\mcfigwidthsupp,trimmgnad]{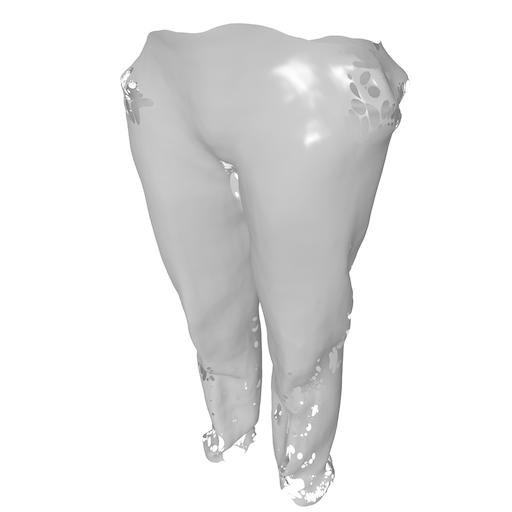}
					& \includegraphics[valign=m,width=\mcfigwidthsupp,trimmgnad]{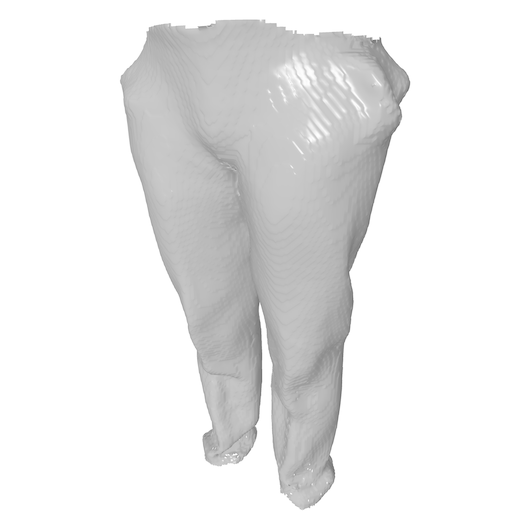}
					& \includegraphics[valign=m,width=\mcfigwidthsupp,trimmgnad]{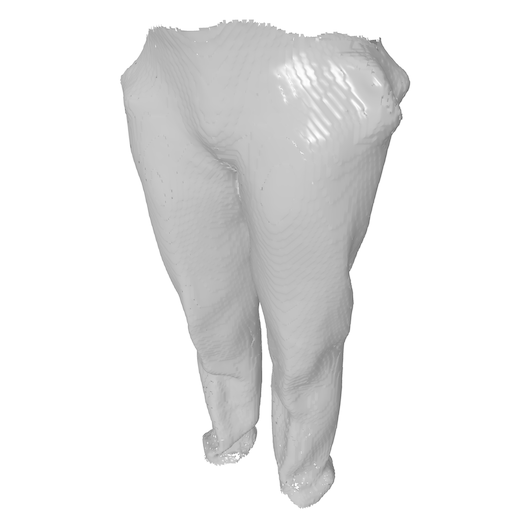}
					& \includegraphics[valign=m,width=\mcfigwidthsupp,trimmgnad]{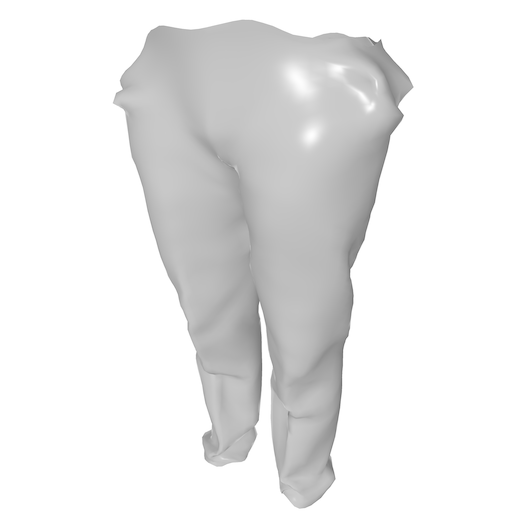}\\ \addlinespace[2\tabcolsep]
					&& CAP-UDF~\cite{Zhou22} & DCUDF~\cite{Hou2023DCUDF} & MeshUDF~\cite{Guillard22b} & Ours & GT
				\end{tabular}}
				\caption{\textbf{Reconstruction methods based on Marching Cubes on an autodecoder trained on MGN~\cite{Bhatnagar19} at varying resolutions.}}
				\label{fig:supp_mc_mgnautodec}
			\end{center}
		\end{figure}

		\begin{figure}[ht!]
				\begin{center}
				{\scriptsize
					\begin{tabular}{llccccc}
						\rotatebox[origin=c]{90}{Cars autodec.} & \rotatebox[origin=c]{90}{64} &
						\includegraphics[valign=m,width=\mcfigwidthsupp,trimsnad]{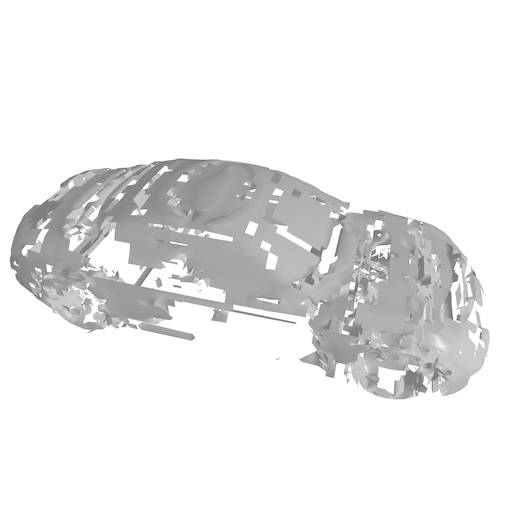}
						& \includegraphics[valign=m,width=\mcfigwidthsupp,trimsnad]{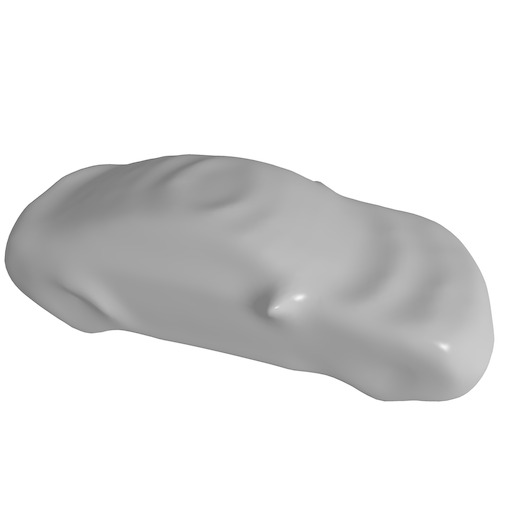}
						& \includegraphics[valign=m,width=\mcfigwidthsupp,trimsnad]{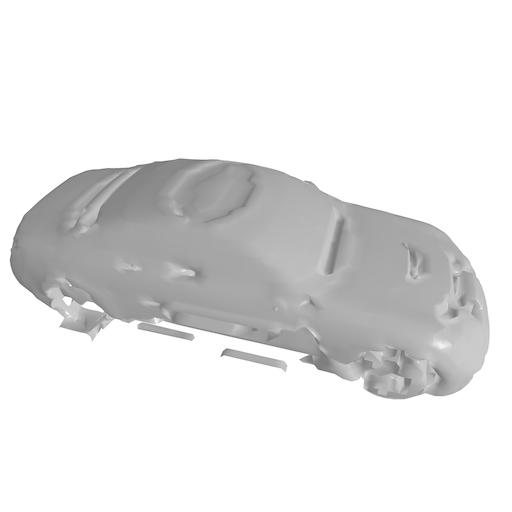}
						& \includegraphics[valign=m,width=\mcfigwidthsupp,trimsnad]{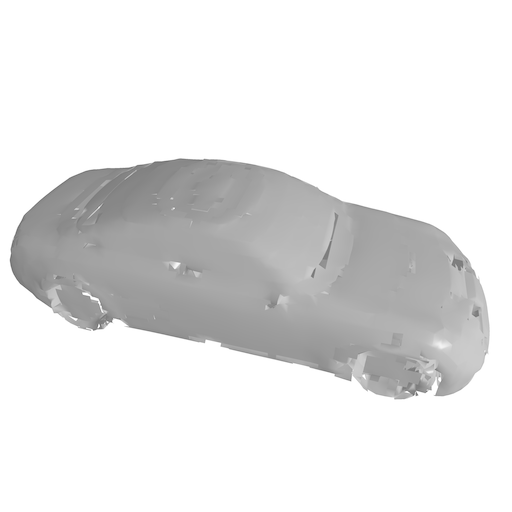}
						& \includegraphics[valign=m,width=\mcfigwidthsupp,trimsnad]{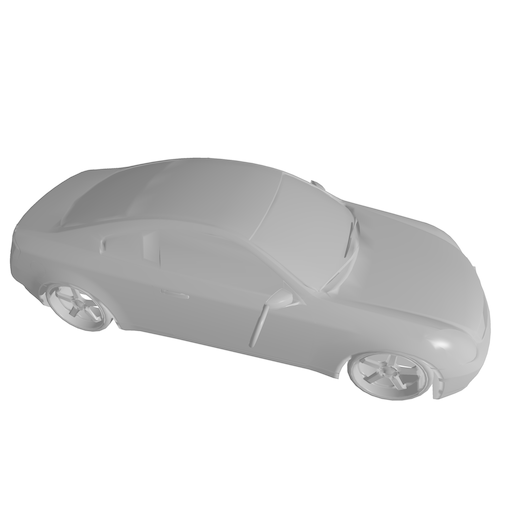}\\ \addlinespace[2\tabcolsep]
						\rotatebox[origin=c]{90}{Cars autodec.} & \rotatebox[origin=c]{90}{128} &
						\includegraphics[valign=m,width=\mcfigwidthsupp,trimsnad]{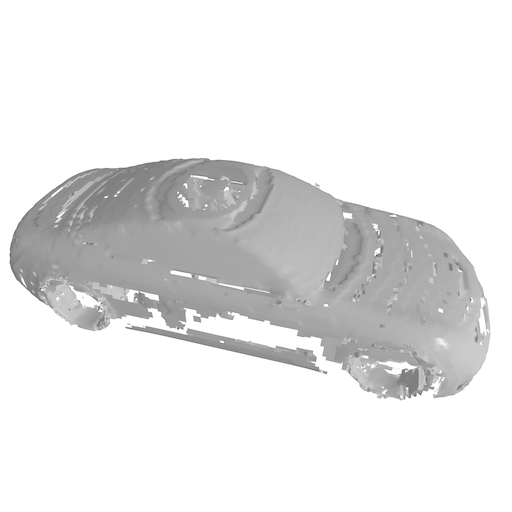}
						& \includegraphics[valign=m,width=\mcfigwidthsupp,trimsnad]{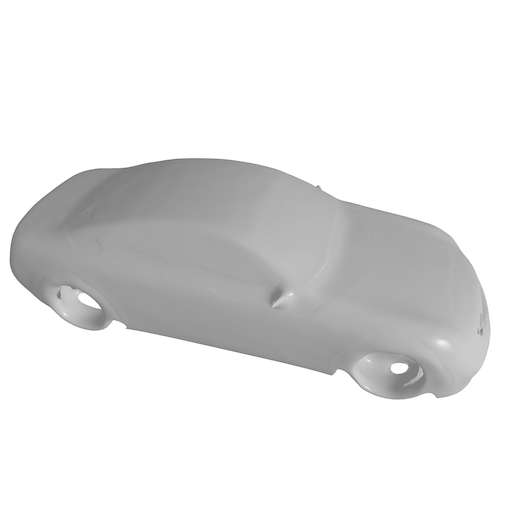}
						& \includegraphics[valign=m,width=\mcfigwidthsupp,trimsnad]{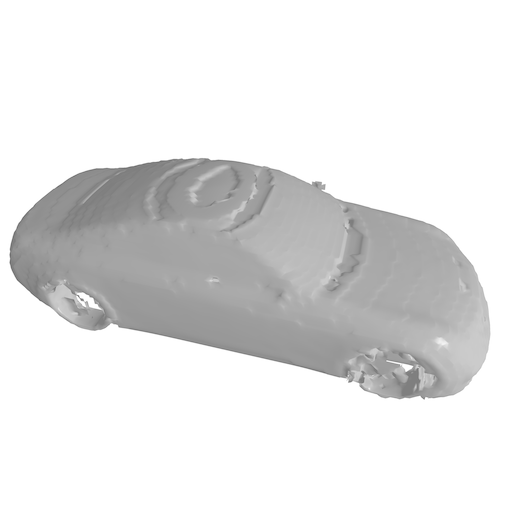}
						& \includegraphics[valign=m,width=\mcfigwidthsupp,trimsnad]{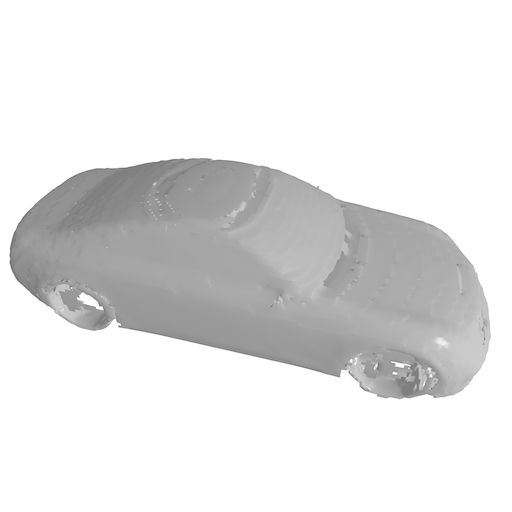}
						& \includegraphics[valign=m,width=\mcfigwidthsupp,trimsnad]{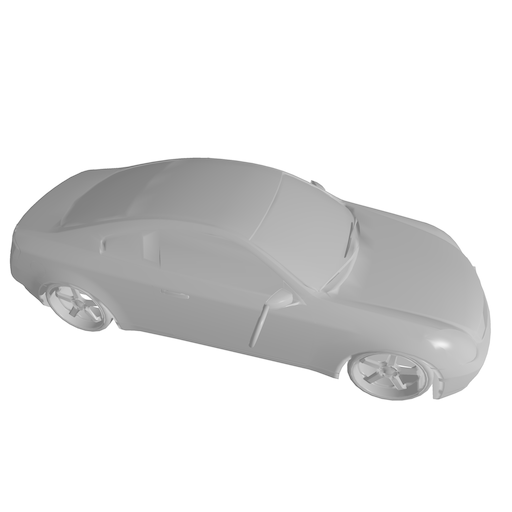}\\ \addlinespace[2\tabcolsep]
						\rotatebox[origin=c]{90}{Cars autodec.} & \rotatebox[origin=c]{90}{256} &
						\includegraphics[valign=m,width=\mcfigwidthsupp,trimsnad]{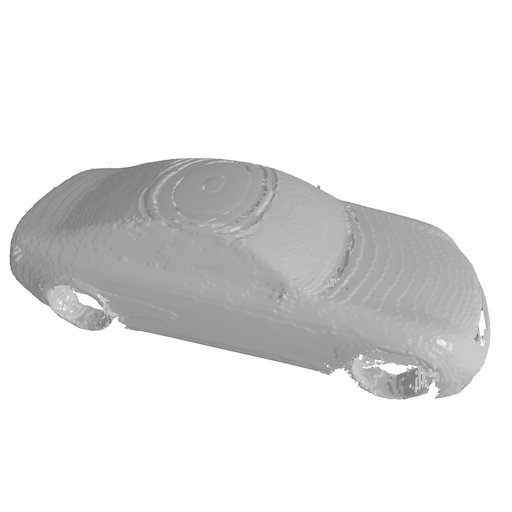}
						& \includegraphics[valign=m,width=\mcfigwidthsupp,trimsnad]{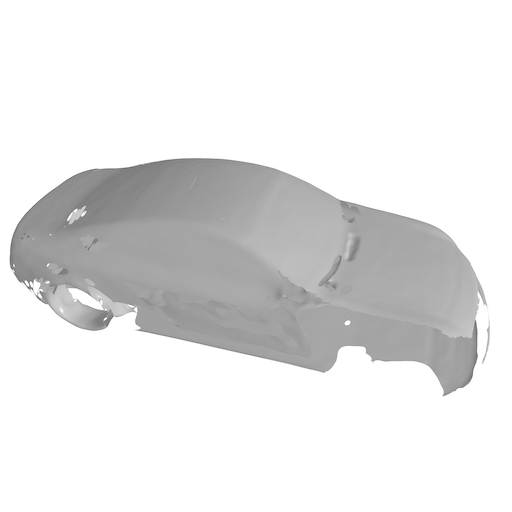}
						& \includegraphics[valign=m,width=\mcfigwidthsupp,trimsnad]{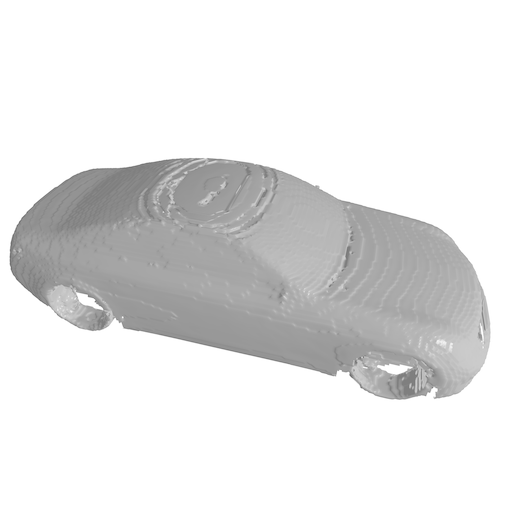}
						& \includegraphics[valign=m,width=\mcfigwidthsupp,trimsnad]{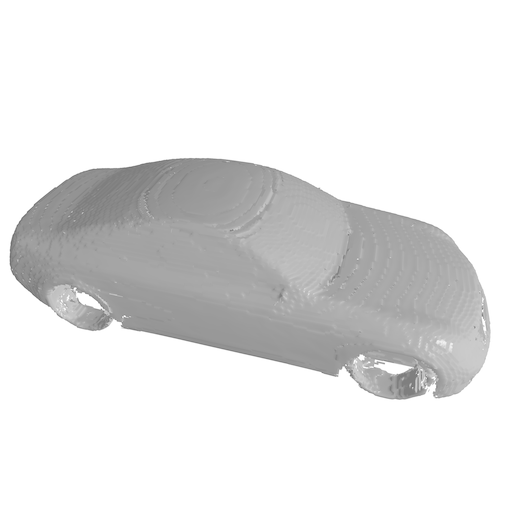}
						& \includegraphics[valign=m,width=\mcfigwidthsupp,trimsnad]{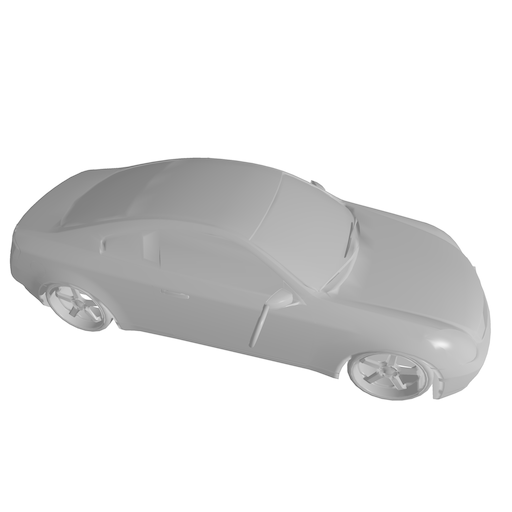}\\ \addlinespace[2\tabcolsep]
						\rotatebox[origin=c]{90}{Cars autodec.} & \rotatebox[origin=c]{90}{512} &
						\includegraphics[valign=m,width=\mcfigwidthsupp,trimsnad]{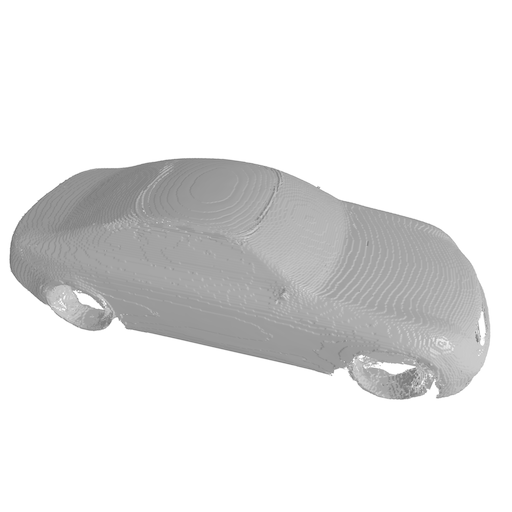}
						& \includegraphics[valign=m,width=\mcfigwidthsupp,trimsnad]{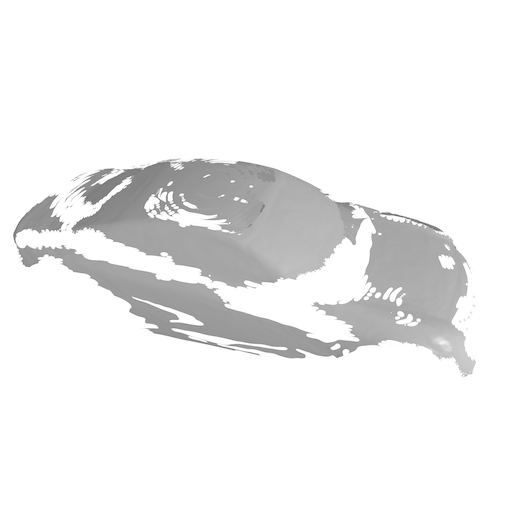}
						& \includegraphics[valign=m,width=\mcfigwidthsupp,trimsnad]{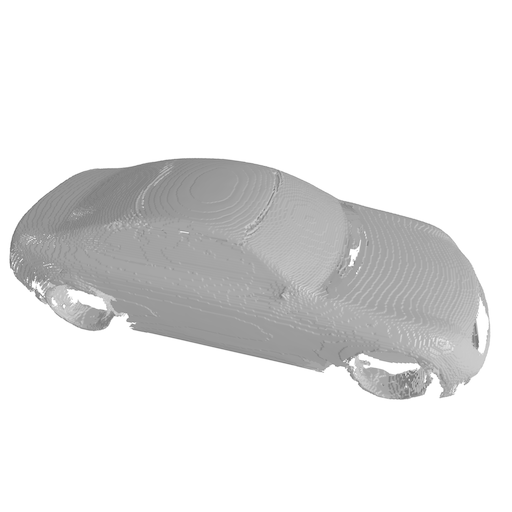}
						& \includegraphics[valign=m,width=\mcfigwidthsupp,trimsnad]{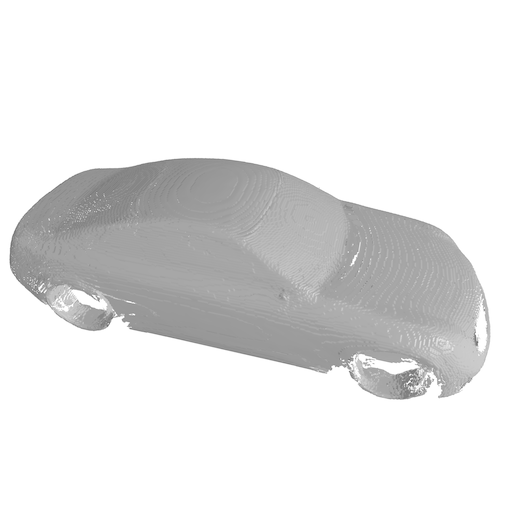}
						& \includegraphics[valign=m,width=\mcfigwidthsupp,trimsnad]{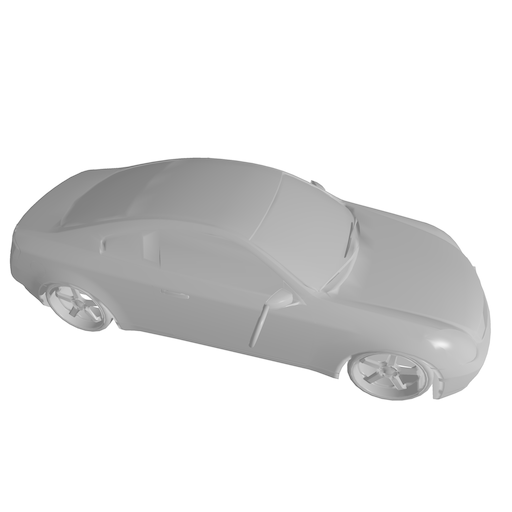}\\ \addlinespace[2\tabcolsep]
						&& CAP-UDF~\cite{Zhou22} & DCUDF~\cite{Hou2023DCUDF} & MeshUDF~\cite{Guillard22b} & Ours & GT
					\end{tabular}}
					\caption{\textbf{Reconstruction methods based on Marching Cubes on an autodecoder trained on ShapNet-Cars~\cite{Chang15} at varying resolutions.}}
					\label{fig:supp_mc_carsautodec}
				\end{center}
			\end{figure}

\setlength{\tabcolsep}{\mytabcolsep}

\newlength{\dcwidthsupp}
\setlength{\dcwidthsupp}{0.18\linewidth}
\definetrim{trimabc2}{2cm 1cm 1cm 1cm}
\definetrim{trimmgn2}{10cm 2cm 10cm 1cm}
\definetrim{trimsn2}{0cm 12cm 9cm 12cm}
\definetrim{trimmgnad2}{2cm 7cm 2cm 10cm}
\definetrim{trimsnad2}{0cm 12cm 0cm 25cm}
\setlength\mytabcolsep{\tabcolsep}
\setlength\tabcolsep{1pt}

\begin{figure}[ht!]
	\begin{center}
	{\scriptsize
		\begin{tabular}{llcccc}
			\rotatebox[origin=c]{90}{ABC~\cite{Koch19a}} & \rotatebox[origin=c]{90}{32} &
			\includegraphics[valign=m,width=\dcwidthsupp,trimabc2]{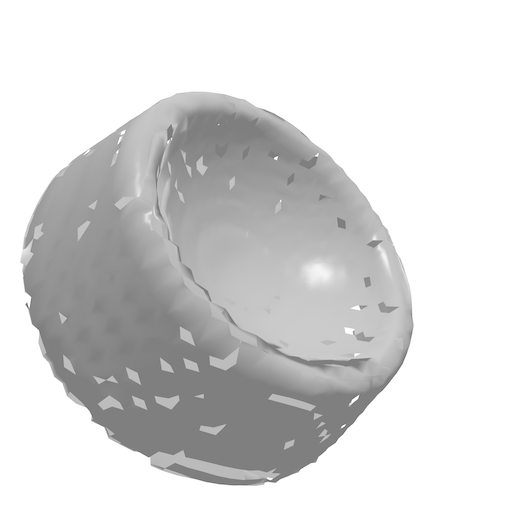}
			& \includegraphics[valign=m,width=\dcwidthsupp,trimabc2]{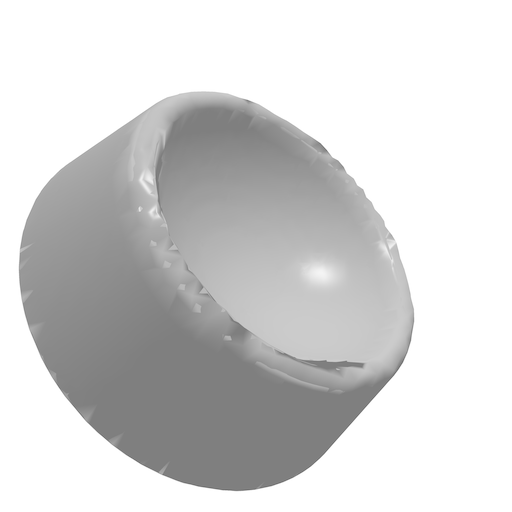}
			& \includegraphics[valign=m,width=\dcwidthsupp,trimabc2]{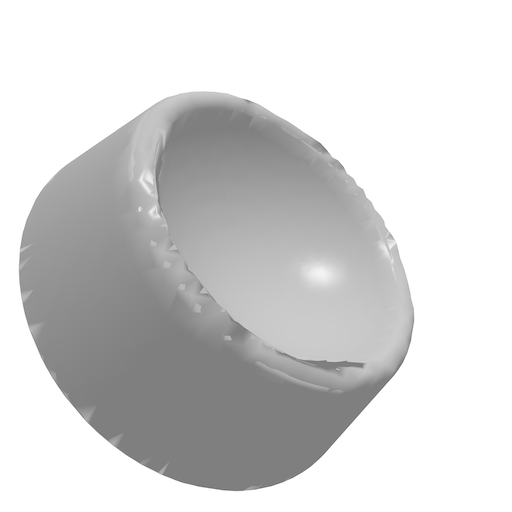} 
			& \includegraphics[valign=m,width=\dcwidthsupp,trimabc2]{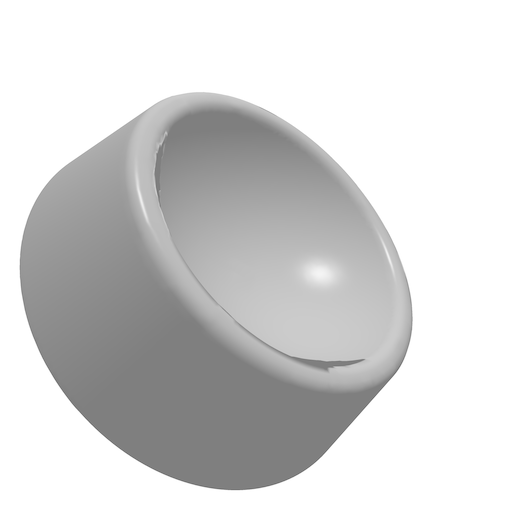}\\ \addlinespace[2\tabcolsep]
			\rotatebox[origin=c]{90}{ABC~\cite{Koch19a}} & \rotatebox[origin=c]{90}{64} &
			\includegraphics[valign=m,width=\dcwidthsupp,trimabc2]{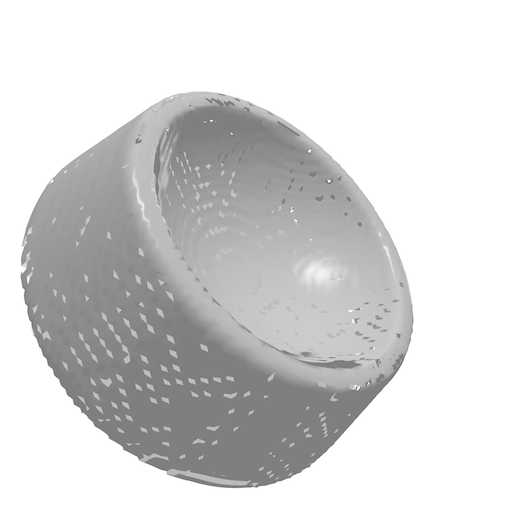}
			& \includegraphics[valign=m,width=\dcwidthsupp,trimabc2]{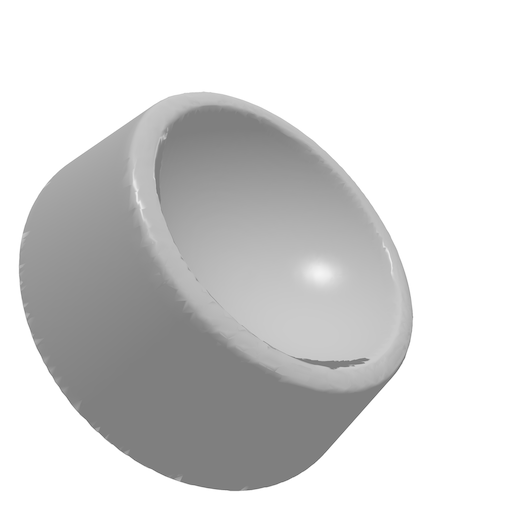}
			& \includegraphics[valign=m,width=\dcwidthsupp,trimabc2]{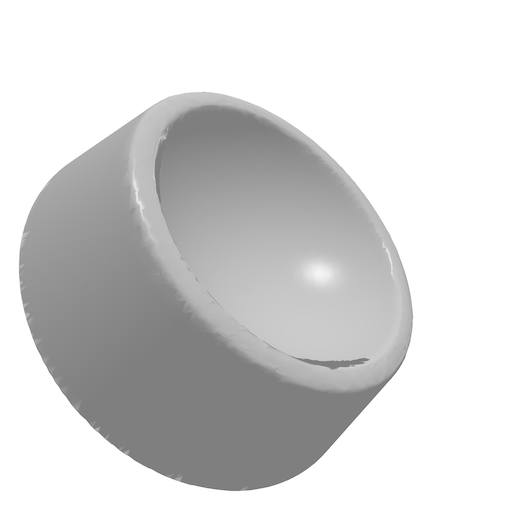} 
			& \includegraphics[valign=m,width=\dcwidthsupp,trimabc2]{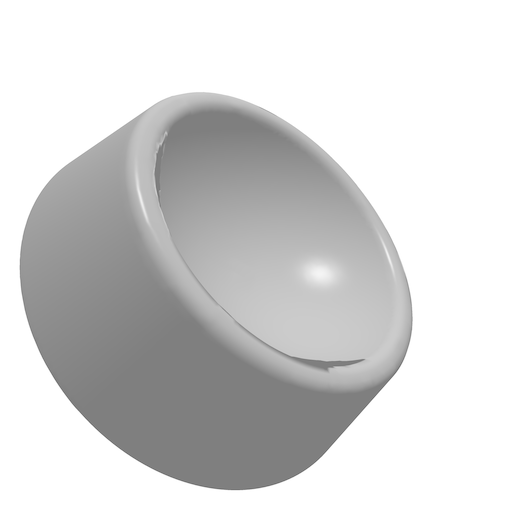}\\ \addlinespace[2\tabcolsep]
			\rotatebox[origin=c]{90}{ABC~\cite{Koch19a}} & \rotatebox[origin=c]{90}{128} &
			\includegraphics[valign=m,width=\dcwidthsupp,trimabc2]{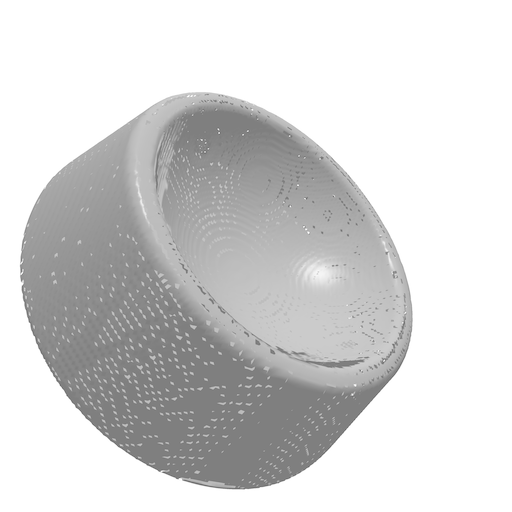}
			& \includegraphics[valign=m,width=\dcwidthsupp,trimabc2]{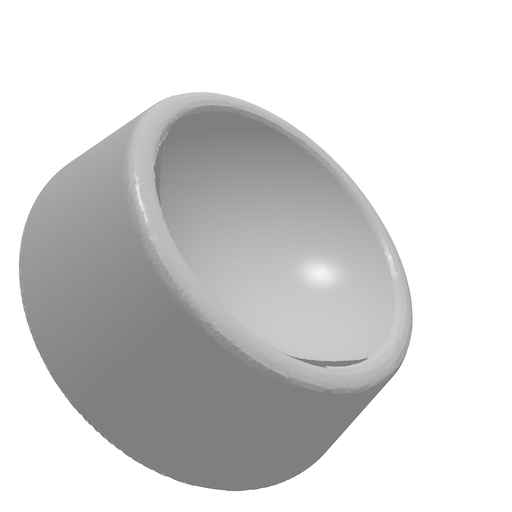}
			& \includegraphics[valign=m,width=\dcwidthsupp,trimabc2]{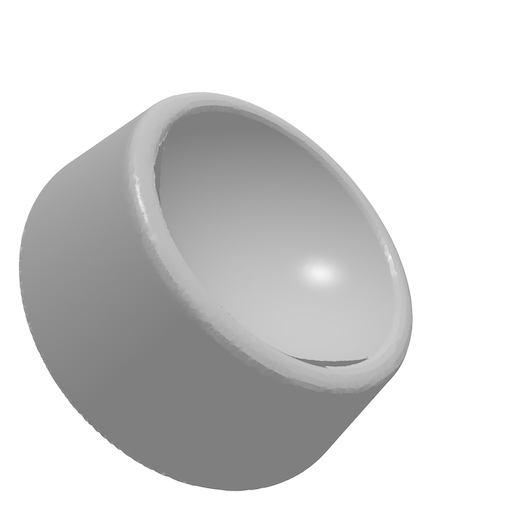} 
			& \includegraphics[valign=m,width=\dcwidthsupp,trimabc2]{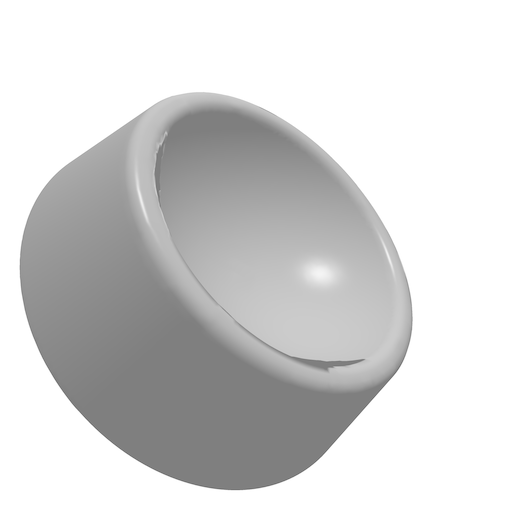}\\ \addlinespace[2\tabcolsep]
			\rotatebox[origin=c]{90}{ABC~\cite{Koch19a}} & \rotatebox[origin=c]{90}{256} &
			\includegraphics[valign=m,width=\dcwidthsupp,trimabc2]{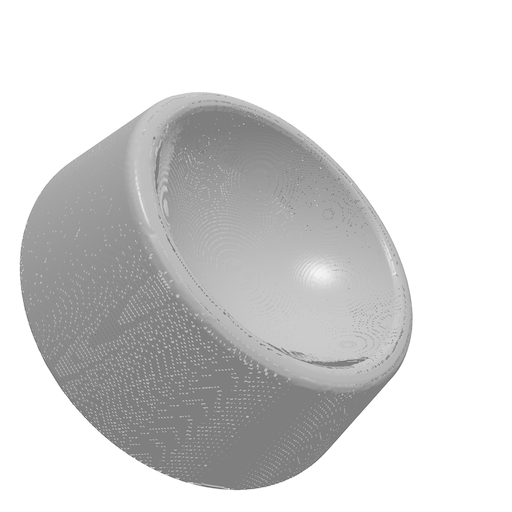}
			& \includegraphics[valign=m,width=\dcwidthsupp,trimabc2]{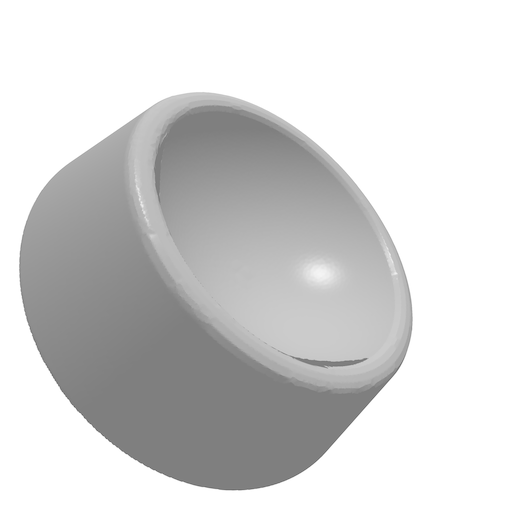}
			& \includegraphics[valign=m,width=\dcwidthsupp,trimabc2]{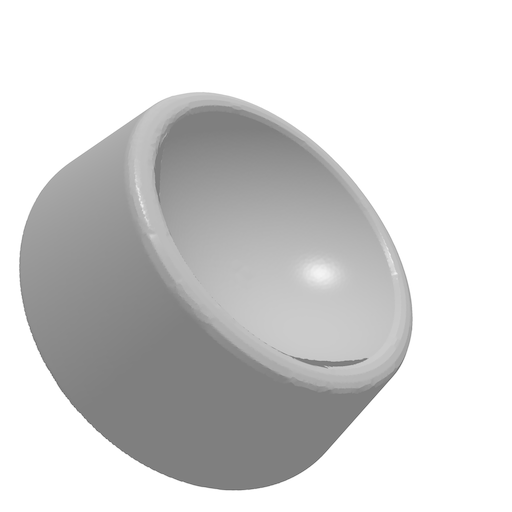} 
			& \includegraphics[valign=m,width=\dcwidthsupp,trimabc2]{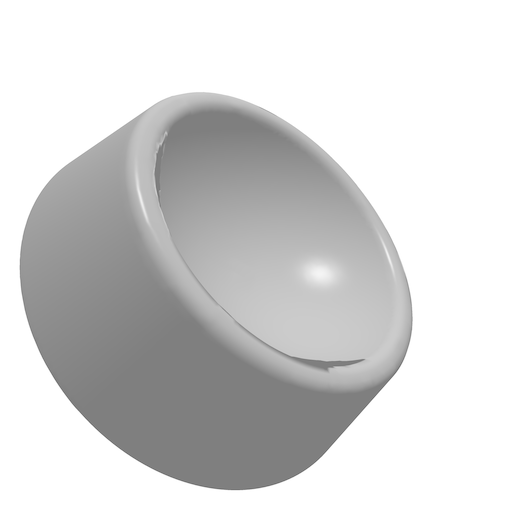}\\ \addlinespace[2\tabcolsep]
			&& UNDC~\cite{Chen22b} & DMUDF~\cite{Zhang23b} & Ours+DMUDF & GT
		\end{tabular}}
		\caption{\textbf{Reconstruction methods based on Dual Contouring on ABC~\cite{Koch19a} at varying resolutions.}}
		\label{fig:supp_dc_abc}
	\end{center}
\end{figure}

\begin{figure}[ht!]
		\begin{center}
		{\scriptsize
			\begin{tabular}{llcccc}
				\rotatebox[origin=c]{90}{MGN~\cite{Bhatnagar19}} & \rotatebox[origin=c]{90}{32} &
				\includegraphics[valign=m,width=\dcwidthsupp,trimmgn2]{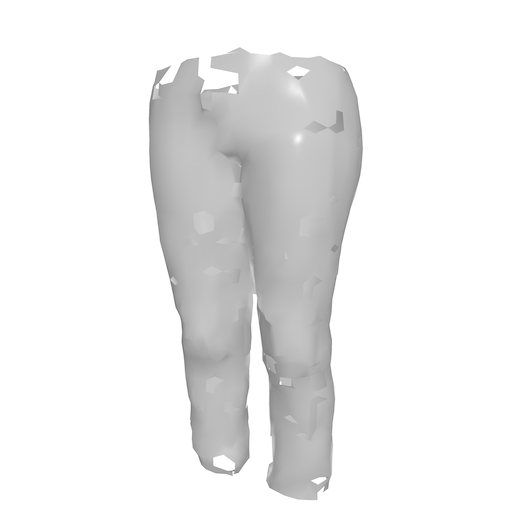}
				& \includegraphics[valign=m,width=\dcwidthsupp,trimmgn2]{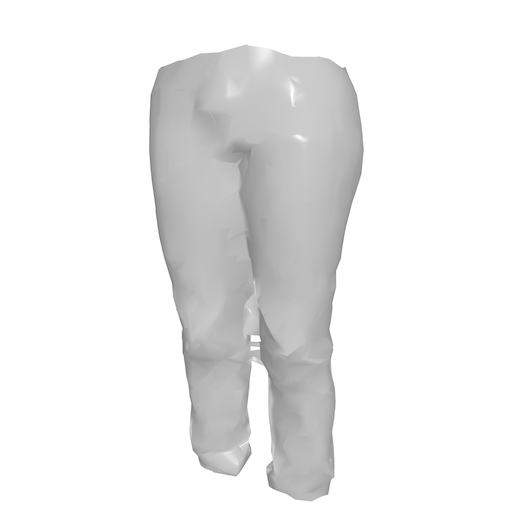}
				& \includegraphics[valign=m,width=\dcwidthsupp,trimmgn2]{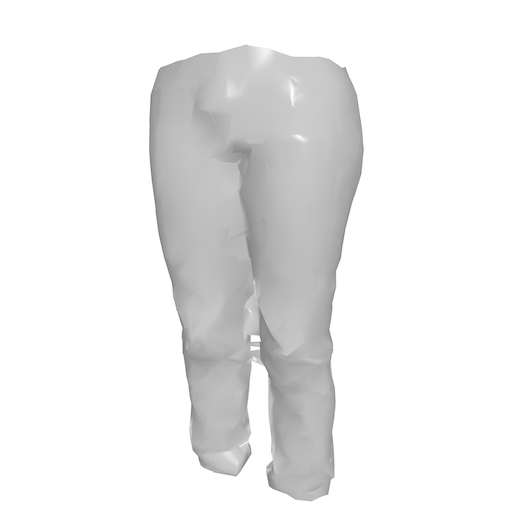}
				& \includegraphics[valign=m,width=\dcwidthsupp,trimmgn2]{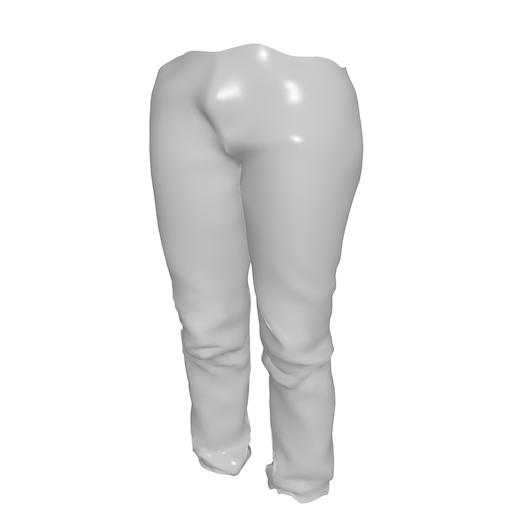}\\ \addlinespace[2\tabcolsep]
				\rotatebox[origin=c]{90}{MGN~\cite{Bhatnagar19}} & \rotatebox[origin=c]{90}{64} &
				\includegraphics[valign=m,width=\dcwidthsupp,trimmgn2]{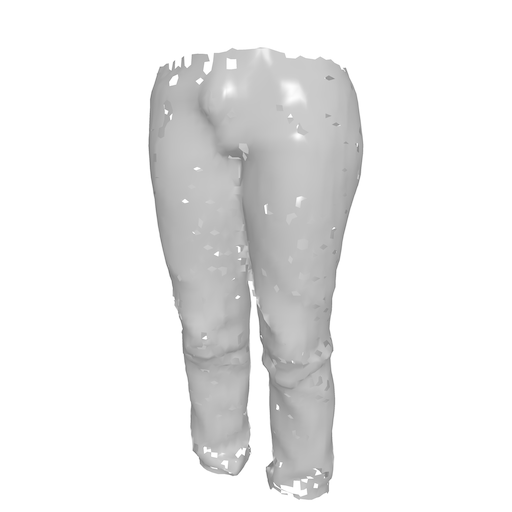}
				& \includegraphics[valign=m,width=\dcwidthsupp,trimmgn2]{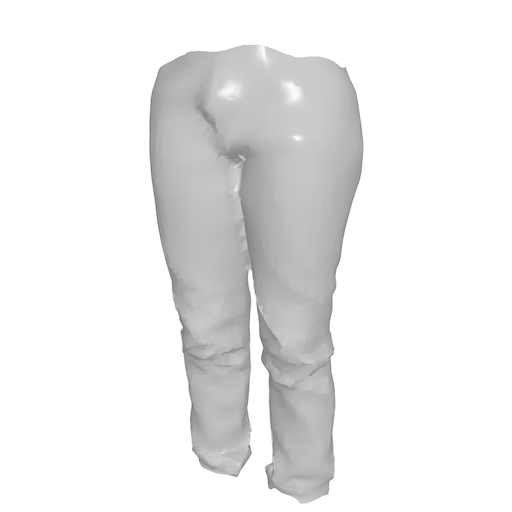}
				& \includegraphics[valign=m,width=\dcwidthsupp,trimmgn2]{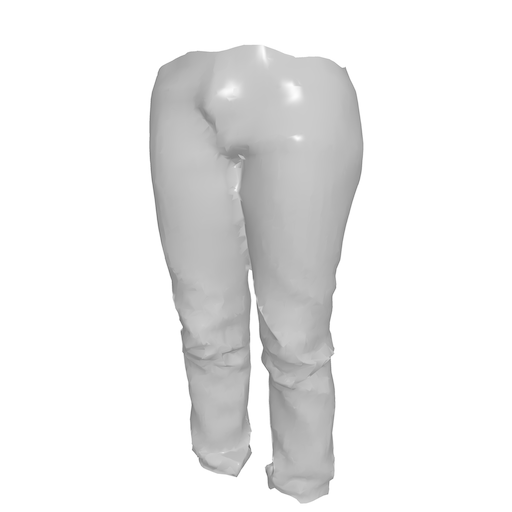}
				& \includegraphics[valign=m,width=\dcwidthsupp,trimmgn2]{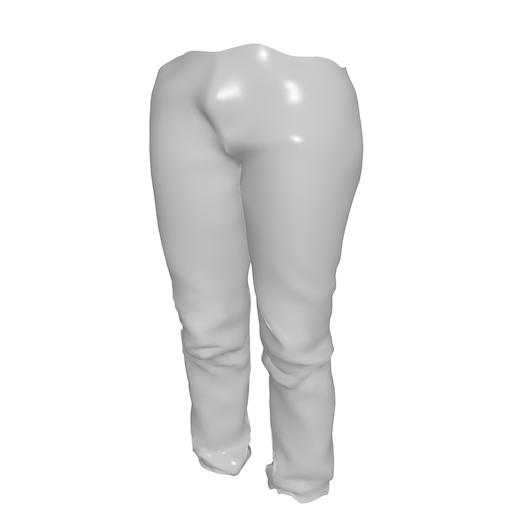}\\ \addlinespace[2\tabcolsep]
				\rotatebox[origin=c]{90}{MGN~\cite{Bhatnagar19}} & \rotatebox[origin=c]{90}{128} &
				\includegraphics[valign=m,width=\dcwidthsupp,trimmgn2]{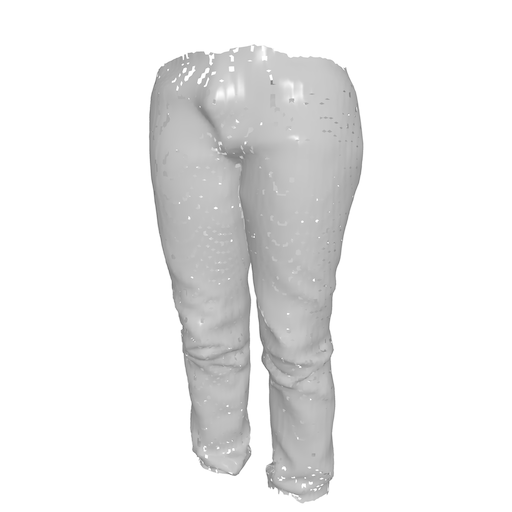}
				& \includegraphics[valign=m,width=\dcwidthsupp,trimmgn2]{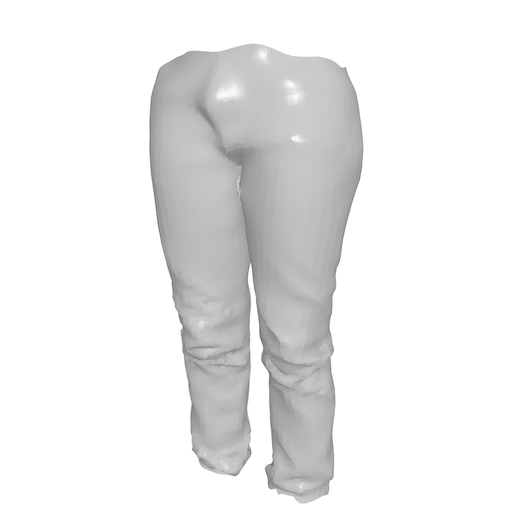}
				& \includegraphics[valign=m,width=\dcwidthsupp,trimmgn2]{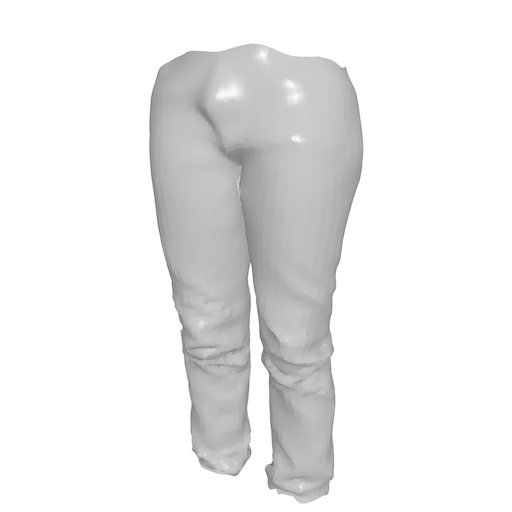}
				& \includegraphics[valign=m,width=\dcwidthsupp,trimmgn2]{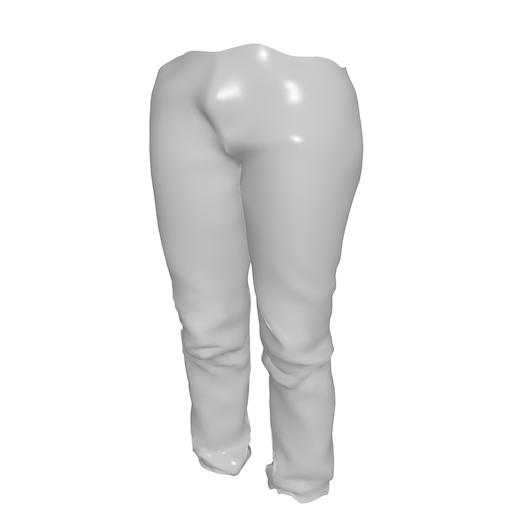}\\ \addlinespace[2\tabcolsep]
				\rotatebox[origin=c]{90}{MGN~\cite{Bhatnagar19}} & \rotatebox[origin=c]{90}{256} &
				\includegraphics[valign=m,width=\dcwidthsupp,trimmgn2]{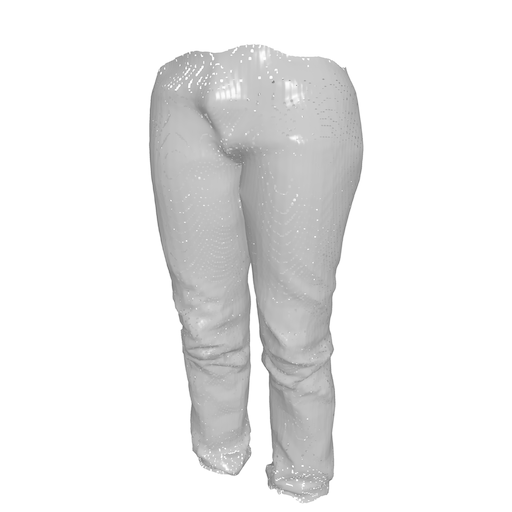}
				& \includegraphics[valign=m,width=\dcwidthsupp,trimmgn2]{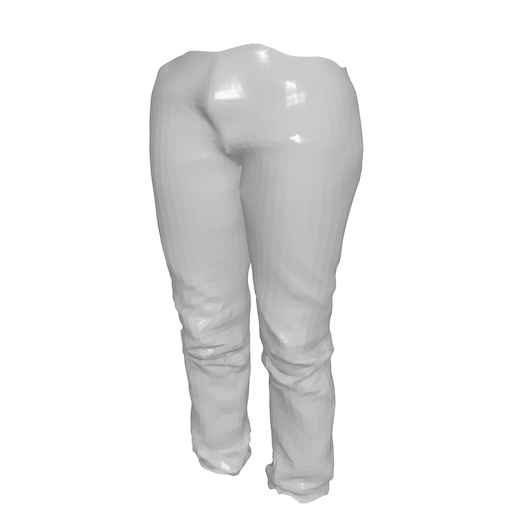}
				& \includegraphics[valign=m,width=\dcwidthsupp,trimmgn2]{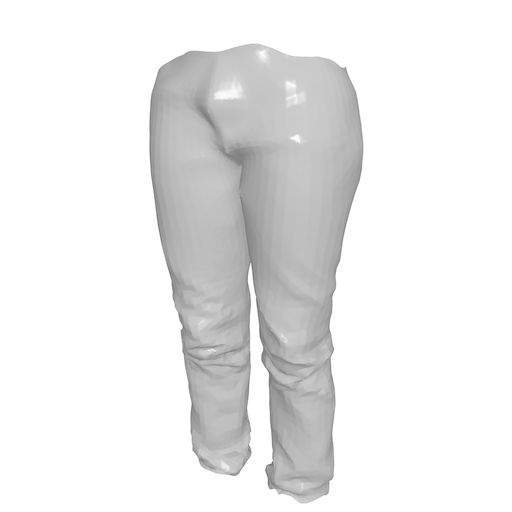}
				& \includegraphics[valign=m,width=\dcwidthsupp,trimmgn2]{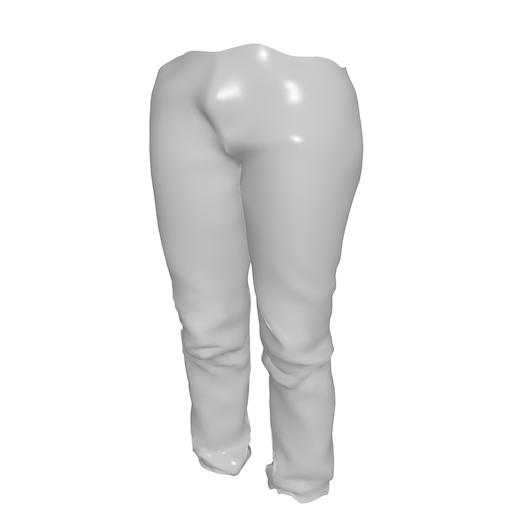}\\ \addlinespace[2\tabcolsep]
				&& UNDC~\cite{Chen22b} & DMUDF~\cite{Zhang23b} & Ours+DMUDF & GT
			\end{tabular}}
			\caption{\textbf{Reconstruction methods based on Dual Contouring on MGN~\cite{Bhatnagar19} at varying resolutions.}}
			\label{fig:supp_dc_mgn}
		\end{center}
	\end{figure}

	\begin{figure}[ht!]
			\begin{center}
			{\scriptsize
				\begin{tabular}{llcccc}
					\rotatebox[origin=c]{90}{Cars~\cite{Chang15}} & \rotatebox[origin=c]{90}{64} &
					\includegraphics[valign=m,width=\dcwidthsupp,trimsn2]{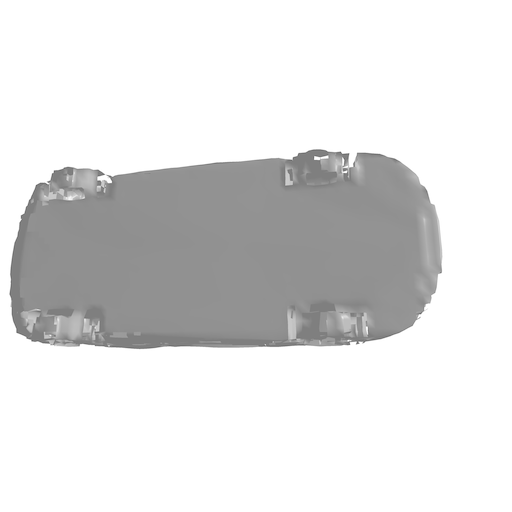}
					& \includegraphics[valign=m,width=\dcwidthsupp,trimsn2]{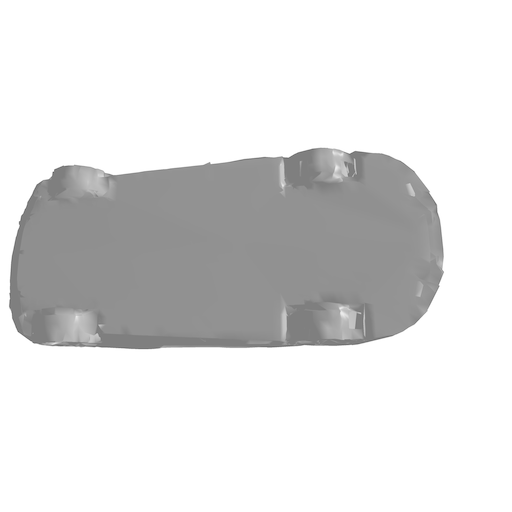}
					& \includegraphics[valign=m,width=\dcwidthsupp,trimsn2]{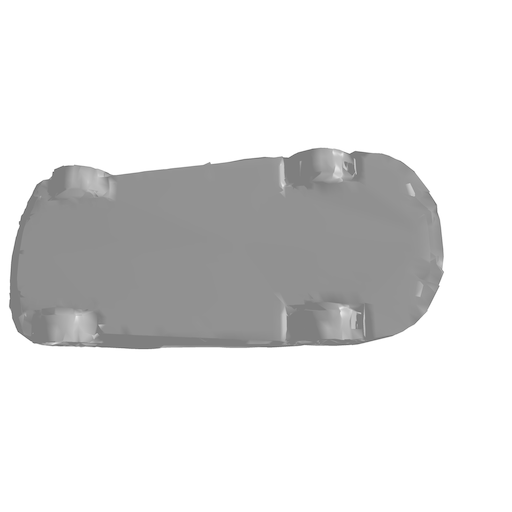}
					& \includegraphics[valign=m,width=\dcwidthsupp,trimsn2]{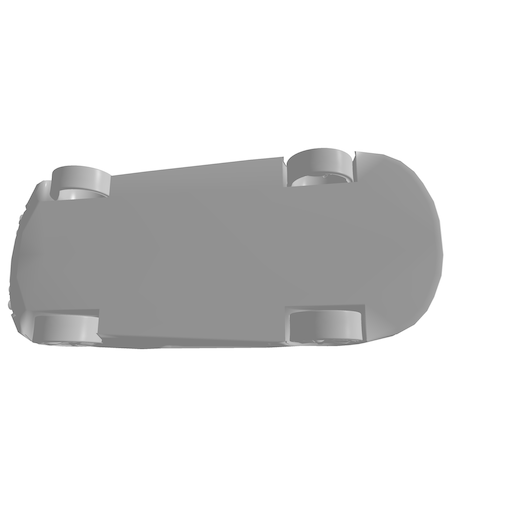}\\ \addlinespace[2\tabcolsep]
					\rotatebox[origin=c]{90}{Cars~\cite{Chang15}} & \rotatebox[origin=c]{90}{128} &
					\includegraphics[valign=m,width=\dcwidthsupp,trimsn2]{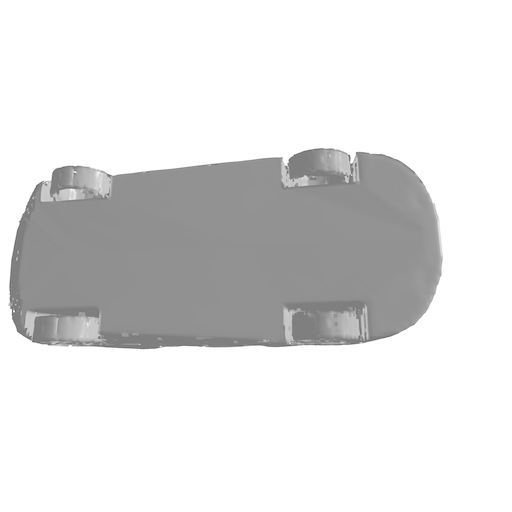}
					& \includegraphics[valign=m,width=\dcwidthsupp,trimsn2]{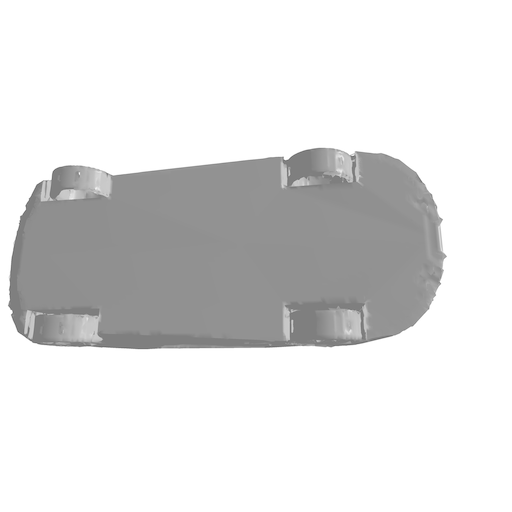}
					& \includegraphics[valign=m,width=\dcwidthsupp,trimsn2]{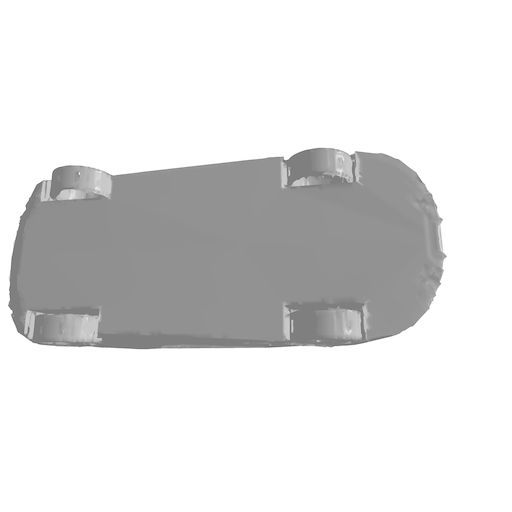}
					& \includegraphics[valign=m,width=\dcwidthsupp,trimsn2]{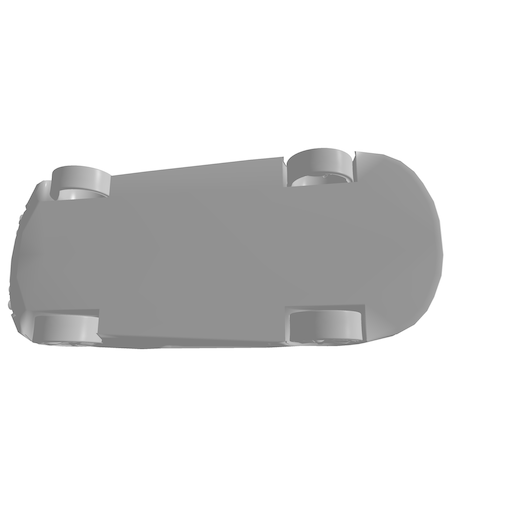}\\ \addlinespace[2\tabcolsep]
					\rotatebox[origin=c]{90}{Cars~\cite{Chang15}} & \rotatebox[origin=c]{90}{256} &
					\includegraphics[valign=m,width=\dcwidthsupp,trimsn2]{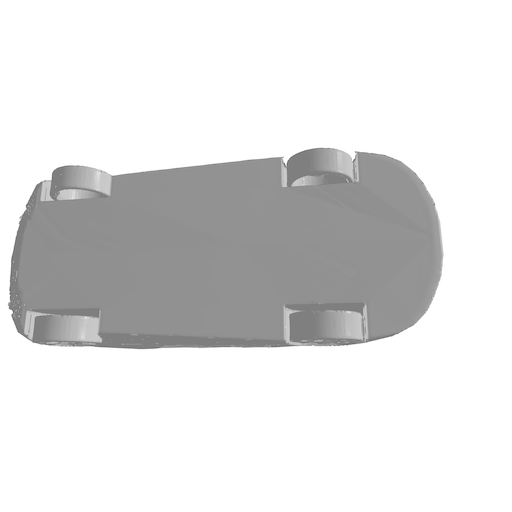}
					& \includegraphics[valign=m,width=\dcwidthsupp,trimsn2]{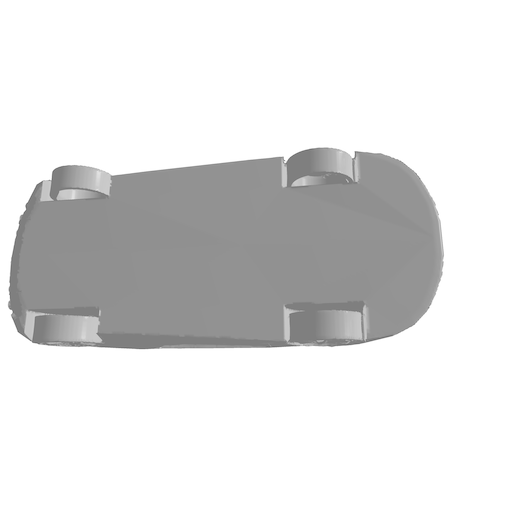}
					& \includegraphics[valign=m,width=\dcwidthsupp,trimsn2]{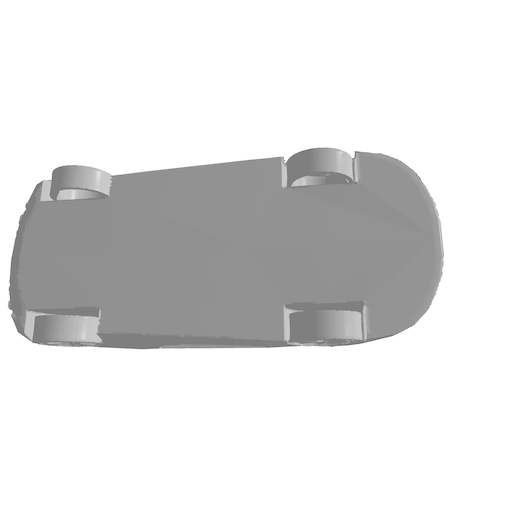}
					& \includegraphics[valign=m,width=\dcwidthsupp,trimsn2]{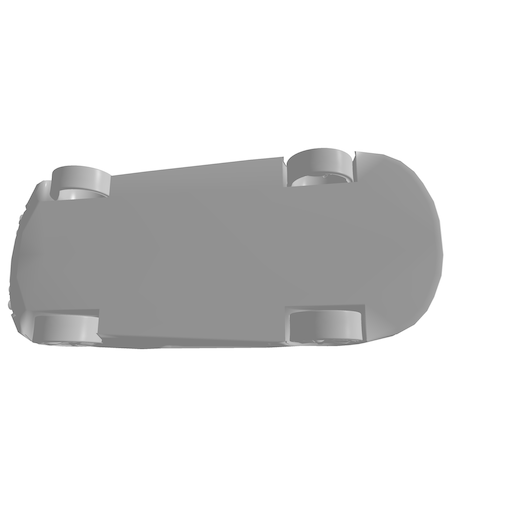}\\ \addlinespace[2\tabcolsep]
					\rotatebox[origin=c]{90}{Cars~\cite{Chang15}} & \rotatebox[origin=c]{90}{512} &
					\includegraphics[valign=m,width=\dcwidthsupp,trimsn2]{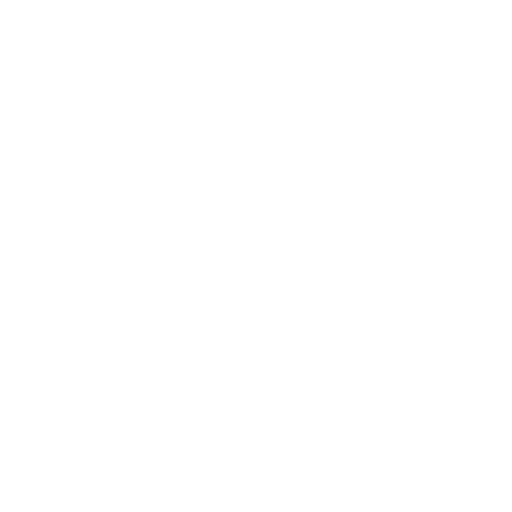}
					& \includegraphics[valign=m,width=\dcwidthsupp,trimsn2]{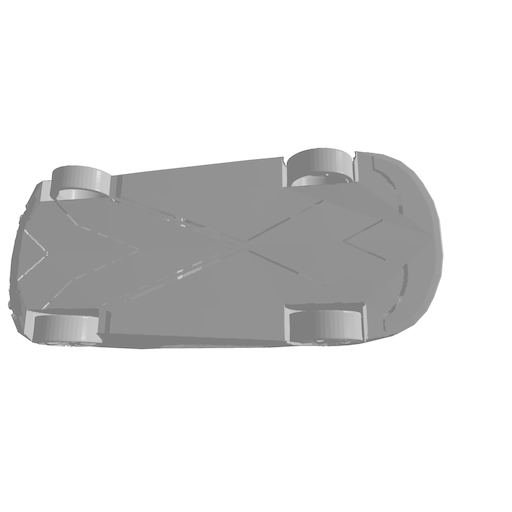}
					& \includegraphics[valign=m,width=\dcwidthsupp,trimsn2]{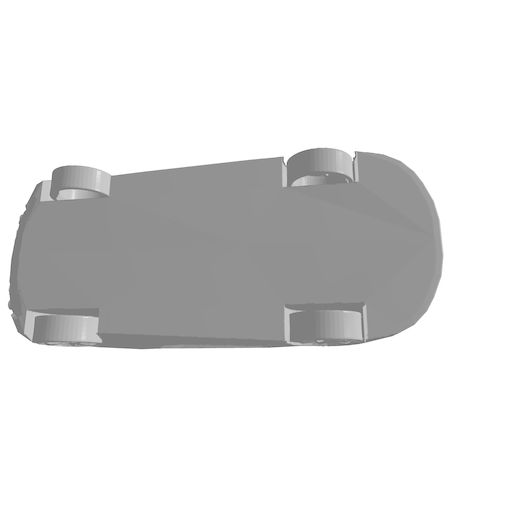}
					& \includegraphics[valign=m,width=\dcwidthsupp,trimsn2]{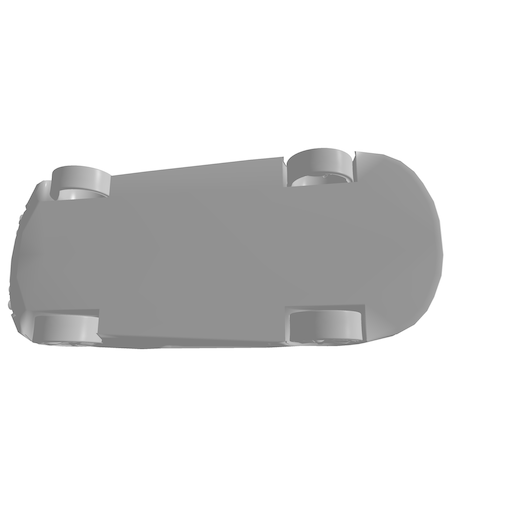}\\ \addlinespace[2\tabcolsep]
					&& UNDC~\cite{Chen22b} & DMUDF~\cite{Zhang23b} & Ours+DMUDF & GT
				\end{tabular}}
				\caption{\textbf{Reconstruction methods based on Dual Contouring on ShapeNet-Cars~\cite{Chang15} at varying resolutions.}}
				\label{fig:supp_dc_cars}
			\end{center}
		\end{figure}

		\begin{figure}[ht!]
				\begin{center}
				{\scriptsize
					\begin{tabular}{llccccc}
						\rotatebox[origin=c]{90}{MGN autodec.} & \rotatebox[origin=c]{90}{32} &
						\includegraphics[valign=m,width=\dcwidthsupp,trimmgnad2]{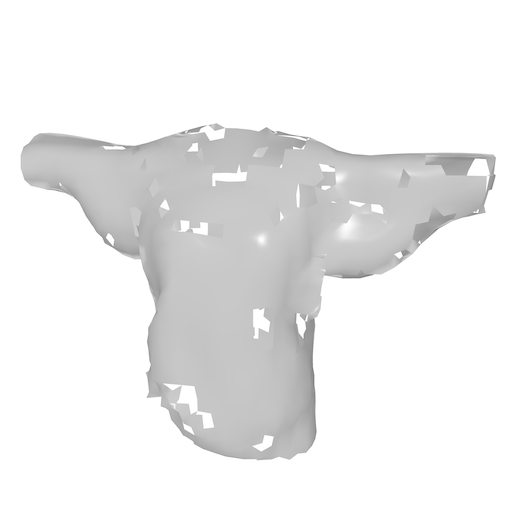}
						& \includegraphics[valign=m,width=\dcwidthsupp,trimmgnad2]{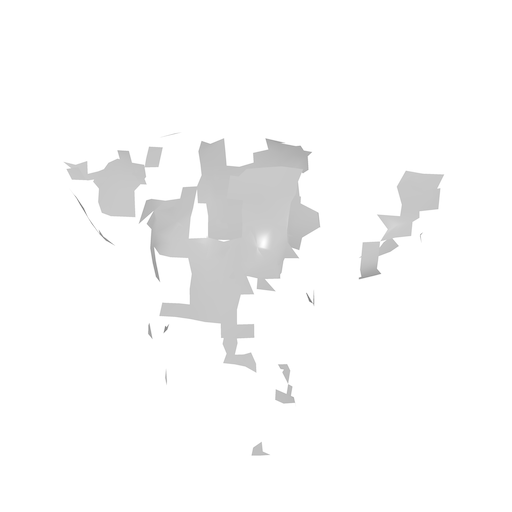}
						& \includegraphics[valign=m,width=\dcwidthsupp,trimmgnad2]{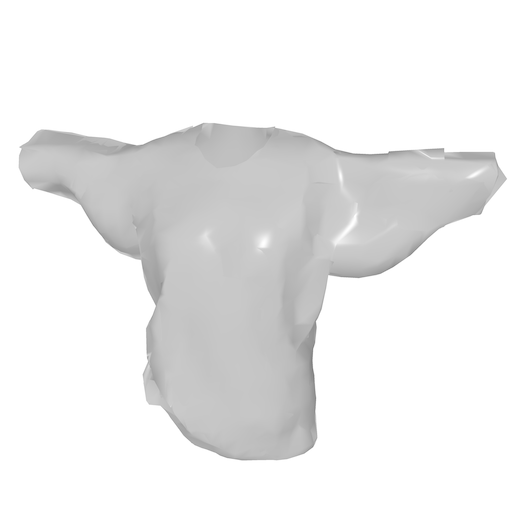}
						& \includegraphics[valign=m,width=\dcwidthsupp,trimmgnad2]{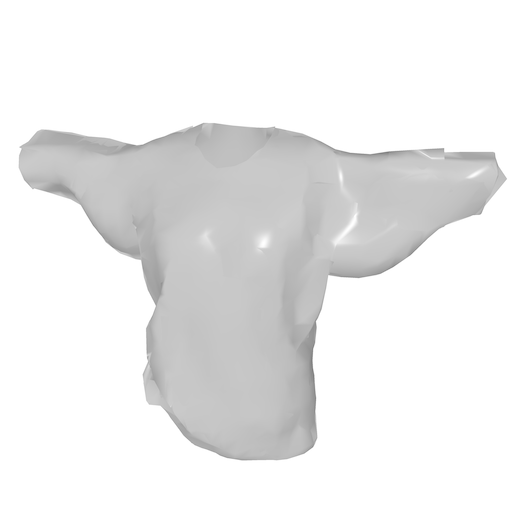}
						& \includegraphics[valign=m,width=\dcwidthsupp,trimmgnad2]{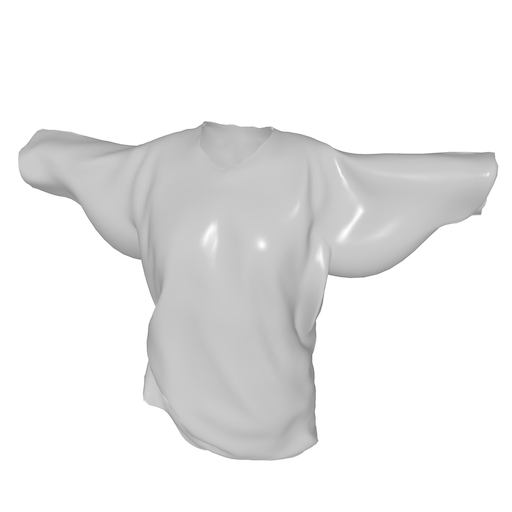}\\ \addlinespace[2\tabcolsep]
						\rotatebox[origin=c]{90}{MGN autodec.} & \rotatebox[origin=c]{90}{64} &
						\includegraphics[valign=m,width=\dcwidthsupp,trimmgnad2]{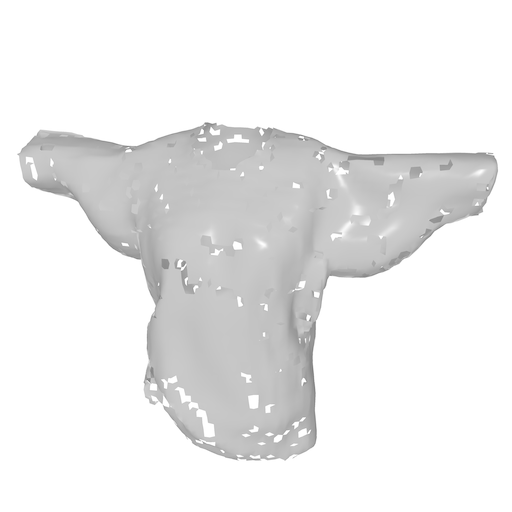}
						& \includegraphics[valign=m,width=\dcwidthsupp,trimmgnad2]{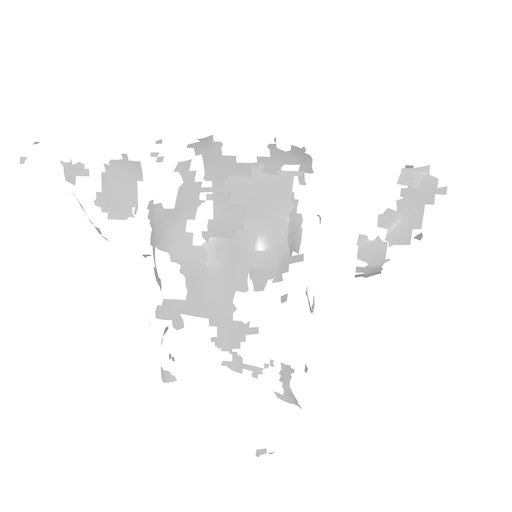}
						& \includegraphics[valign=m,width=\dcwidthsupp,trimmgnad2]{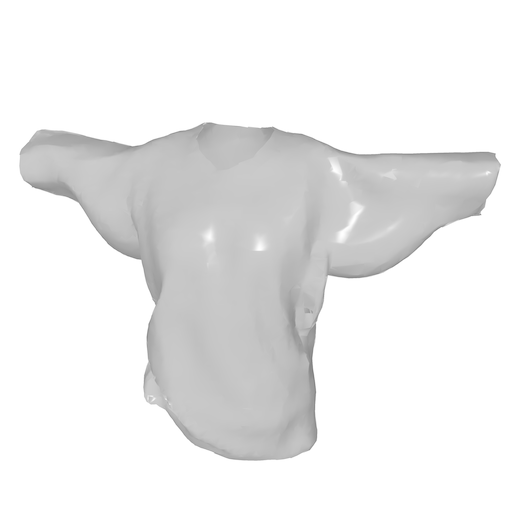}
						& \includegraphics[valign=m,width=\dcwidthsupp,trimmgnad2]{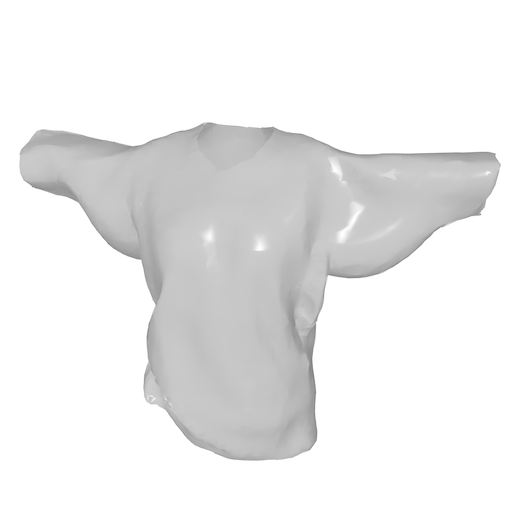}
						& \includegraphics[valign=m,width=\dcwidthsupp,trimmgnad2]{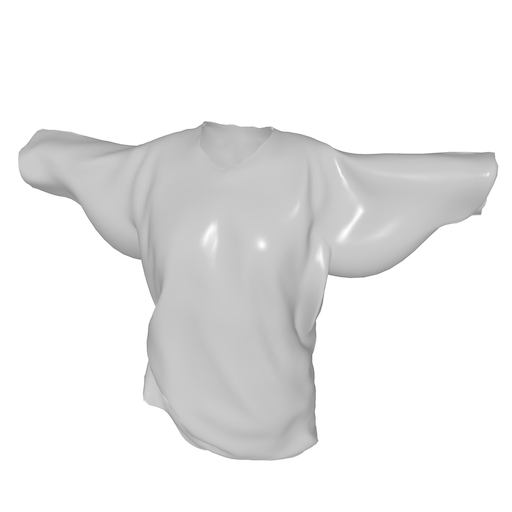}\\ \addlinespace[2\tabcolsep]
						\rotatebox[origin=c]{90}{MGN autodec.} & \rotatebox[origin=c]{90}{128} &
						\includegraphics[valign=m,width=\dcwidthsupp,trimmgnad2]{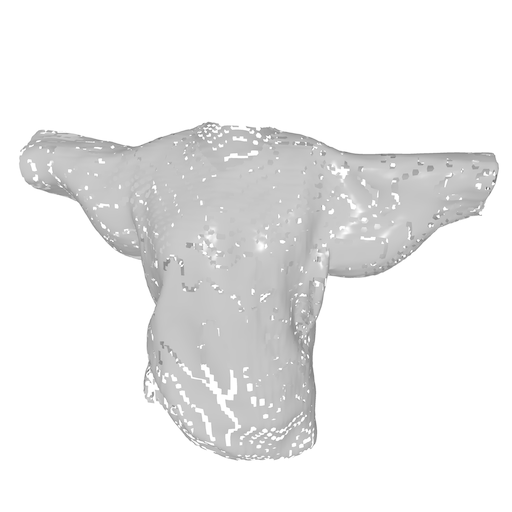}
						& \includegraphics[valign=m,width=\dcwidthsupp,trimmgnad2]{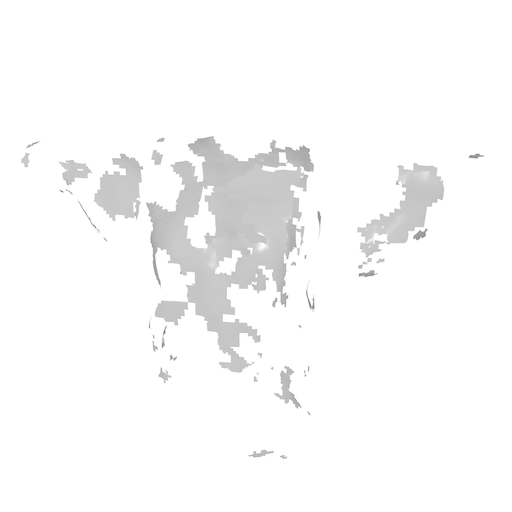}
						& \includegraphics[valign=m,width=\dcwidthsupp,trimmgnad2]{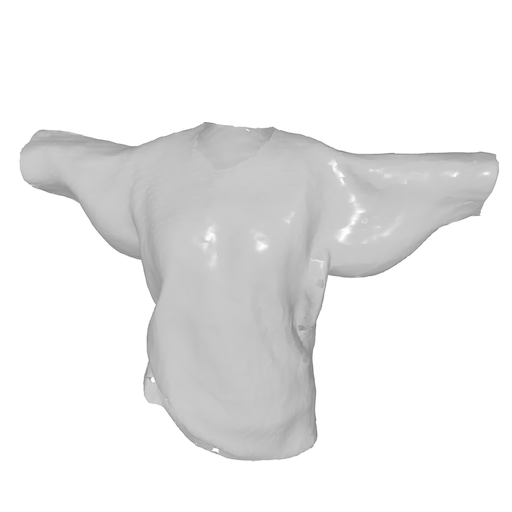}
						& \includegraphics[valign=m,width=\dcwidthsupp,trimmgnad2]{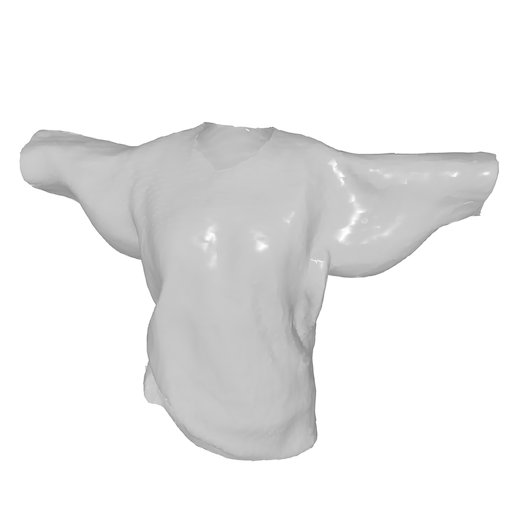}
						& \includegraphics[valign=m,width=\dcwidthsupp,trimmgnad2]{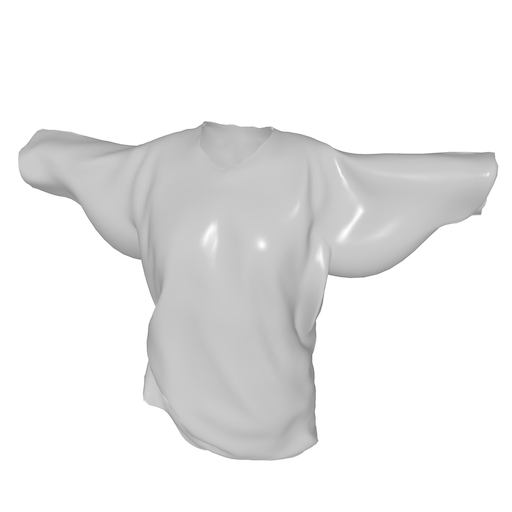}\\ \addlinespace[2\tabcolsep]
						\rotatebox[origin=c]{90}{MGN autodec.} & \rotatebox[origin=c]{90}{256} &
						\includegraphics[valign=m,width=\dcwidthsupp,trimmgnad2]{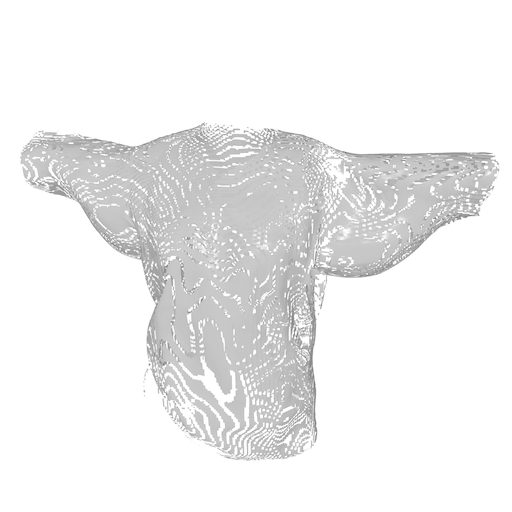}
						& \includegraphics[valign=m,width=\dcwidthsupp,trimmgnad2]{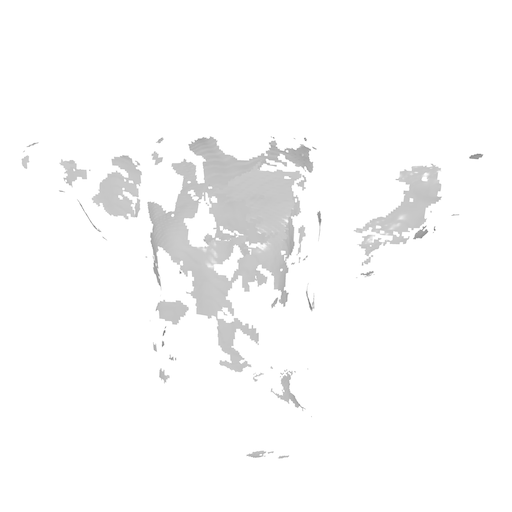}
						& \includegraphics[valign=m,width=\dcwidthsupp,trimmgnad2]{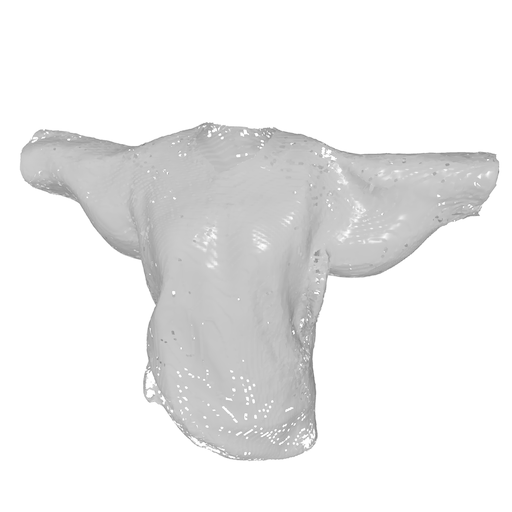}
						& \includegraphics[valign=m,width=\dcwidthsupp,trimmgnad2]{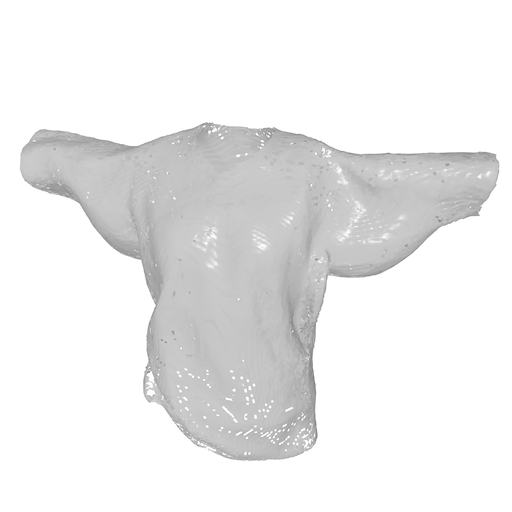}
						& \includegraphics[valign=m,width=\dcwidthsupp,trimmgnad2]{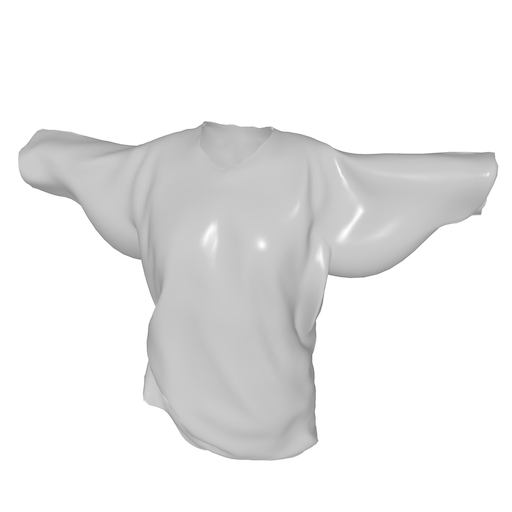}\\ \addlinespace[2\tabcolsep]
						&& UNDC~\cite{Chen22b} & DMUDF~\cite{Zhang23b} & DMUDF-T & Ours+DMUDF & GT
					\end{tabular}}
					\caption{\textbf{Reconstruction methods based on Dual Contouring on an autodecoder trained on MGN~\cite{Bhatnagar19} at varying resolutions.}}
					\label{fig:supp_dc_mgnautodec}
				\end{center}
			\end{figure}

			\begin{figure}[ht!]
					\begin{center}
					{\scriptsize
						\begin{tabular}{llccccc}
							\rotatebox[origin=c]{90}{Cars autodec.} & \rotatebox[origin=c]{90}{64} &
							\includegraphics[valign=m,width=\dcwidthsupp,trimsnad2]{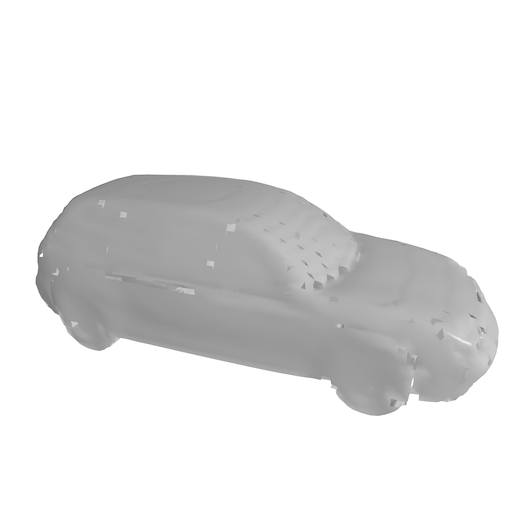}
							& \includegraphics[valign=m,width=\dcwidthsupp,trimsnad2]{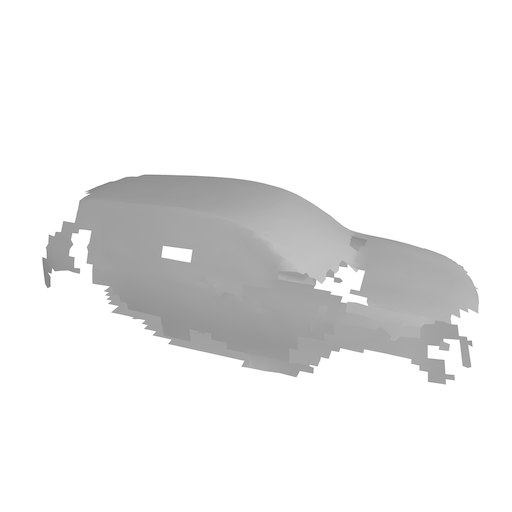}
							& \includegraphics[valign=m,width=\dcwidthsupp,trimsnad2]{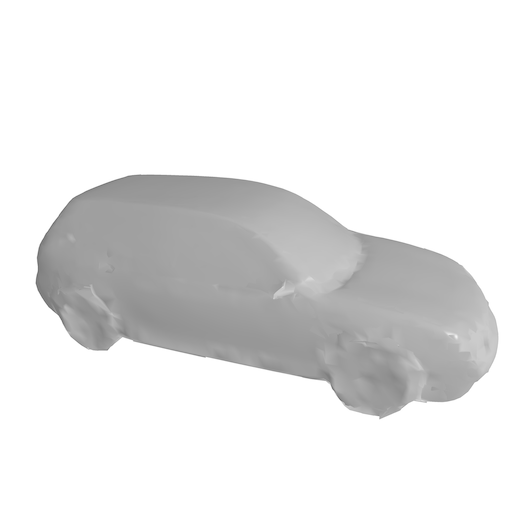}
							& \includegraphics[valign=m,width=\dcwidthsupp,trimsnad2]{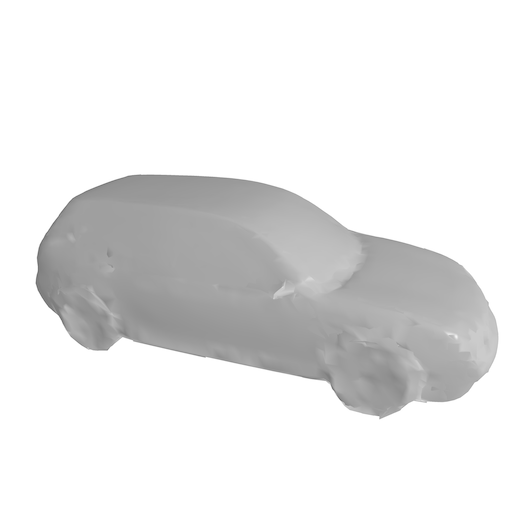}
							& \includegraphics[valign=m,width=\dcwidthsupp,trimsnad2]{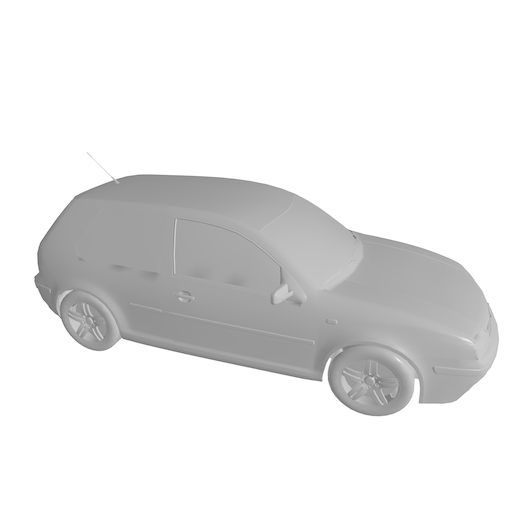}\\ \addlinespace[2\tabcolsep]
							\rotatebox[origin=c]{90}{Cars autodec.} & \rotatebox[origin=c]{90}{128} &
							\includegraphics[valign=m,width=\dcwidthsupp,trimsnad2]{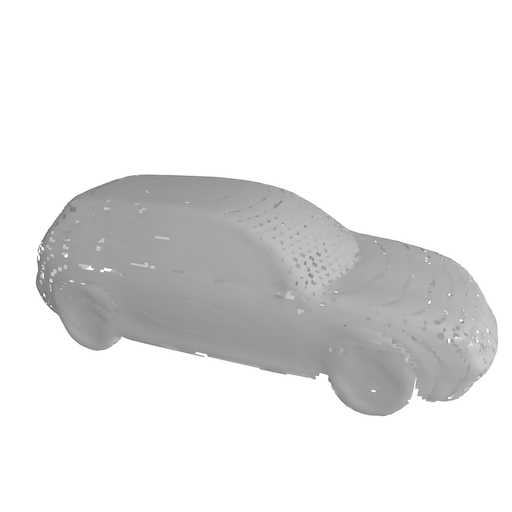}
							& \includegraphics[valign=m,width=\dcwidthsupp,trimsnad2]{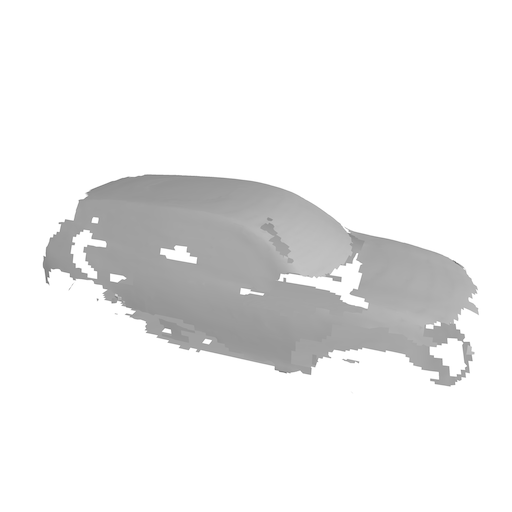}
							& \includegraphics[valign=m,width=\dcwidthsupp,trimsnad2]{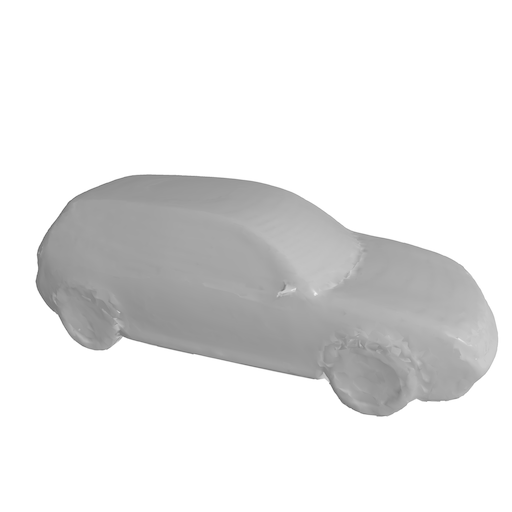}
							& \includegraphics[valign=m,width=\dcwidthsupp,trimsnad2]{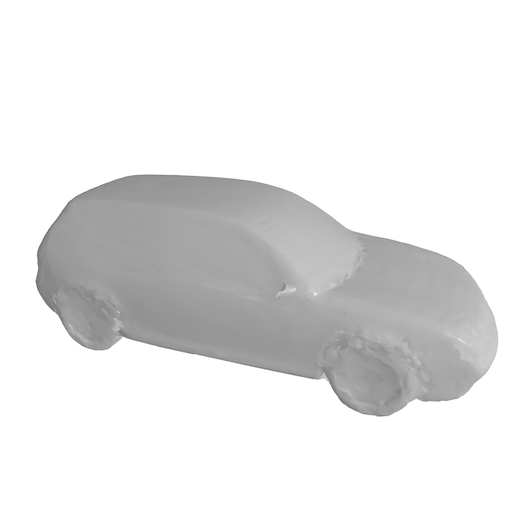}
							& \includegraphics[valign=m,width=\dcwidthsupp,trimsnad2]{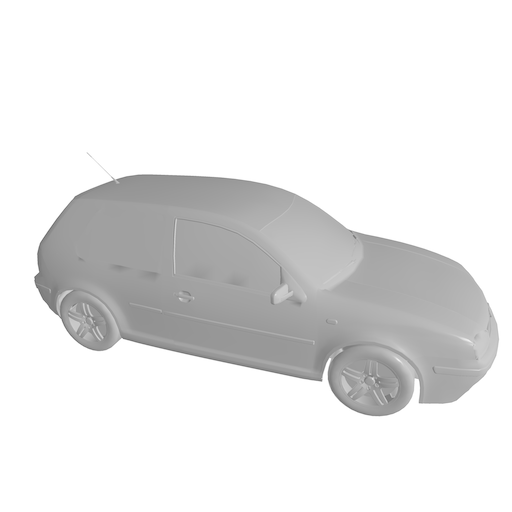}\\ \addlinespace[2\tabcolsep]
							\rotatebox[origin=c]{90}{Cars autodec.} & \rotatebox[origin=c]{90}{256} &
							\includegraphics[valign=m,width=\dcwidthsupp,trimsnad2]{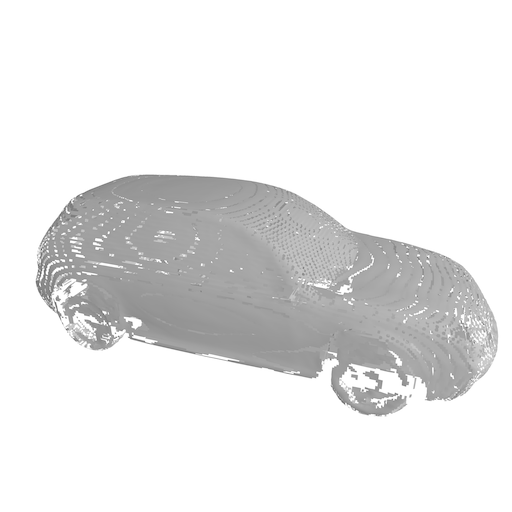}
							& \includegraphics[valign=m,width=\dcwidthsupp,trimsnad2]{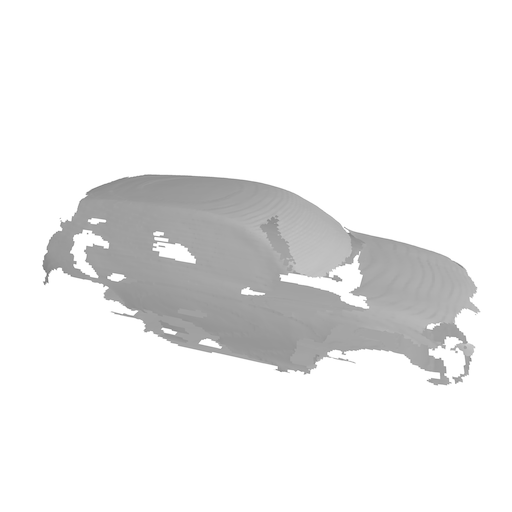}
							& \includegraphics[valign=m,width=\dcwidthsupp,trimsnad2]{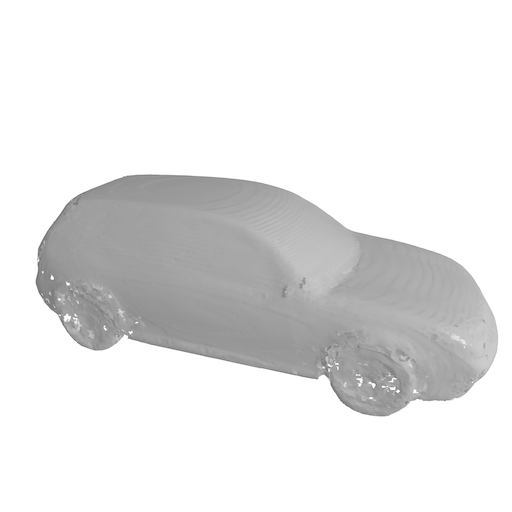}
							& \includegraphics[valign=m,width=\dcwidthsupp,trimsnad2]{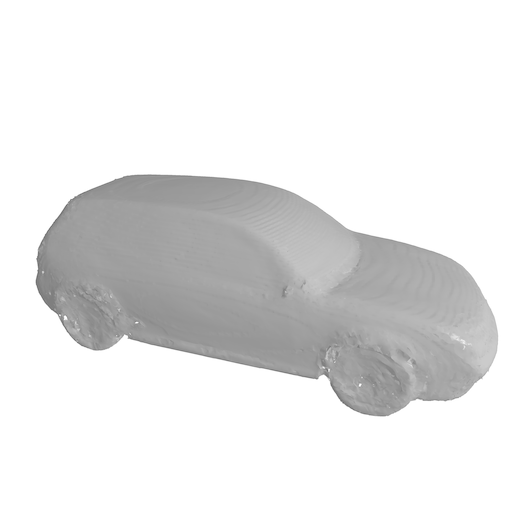}
							& \includegraphics[valign=m,width=\dcwidthsupp,trimsnad2]{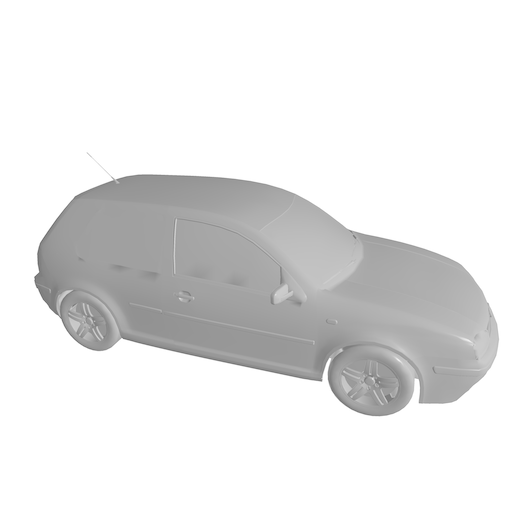}\\ \addlinespace[2\tabcolsep]
							\rotatebox[origin=c]{90}{Cars autodec.} & \rotatebox[origin=c]{90}{512} &
							\includegraphics[valign=m,width=\dcwidthsupp,trimsnad2]{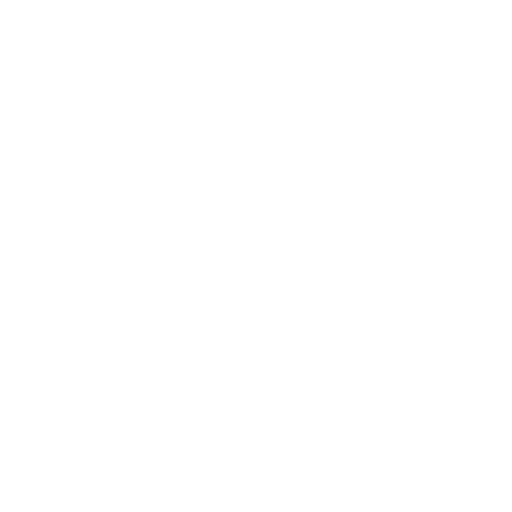}
							& \includegraphics[valign=m,width=\dcwidthsupp,trimsnad2]{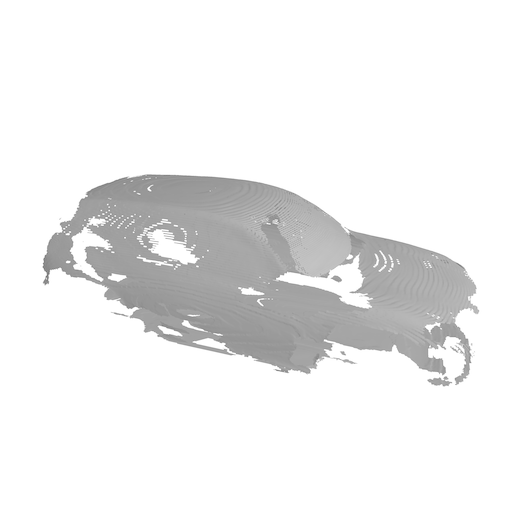}
							& \includegraphics[valign=m,width=\dcwidthsupp,trimsnad2]{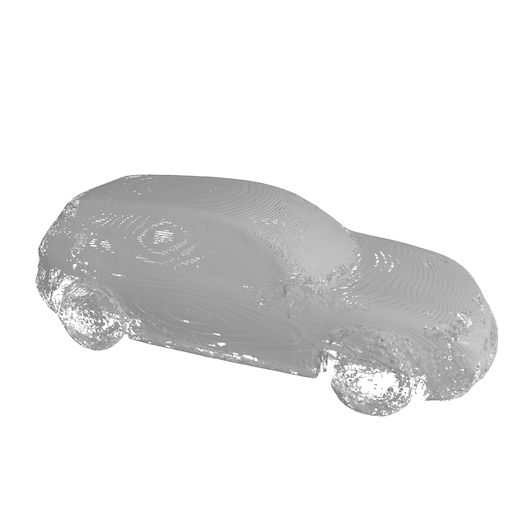}
							& \includegraphics[valign=m,width=\dcwidthsupp,trimsnad2]{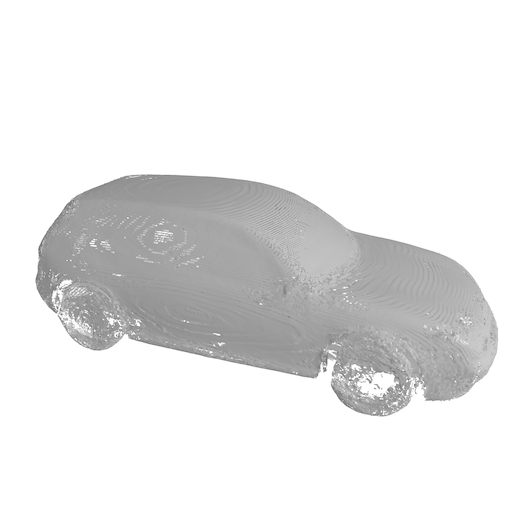}
							& \includegraphics[valign=m,width=\dcwidthsupp,trimsnad2]{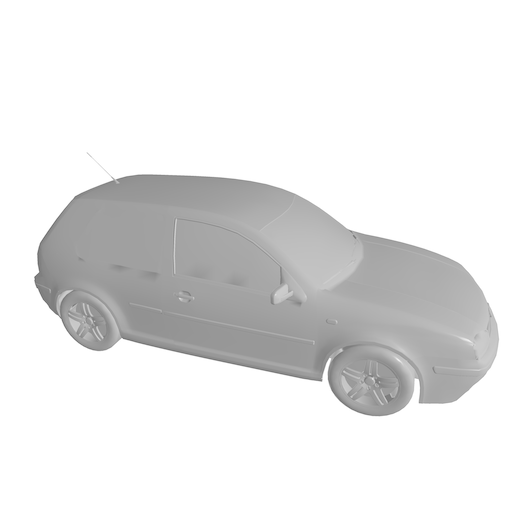}\\ \addlinespace[2\tabcolsep]
							&& UNDC~\cite{Chen22b} & DMUDF~\cite{Zhang23b} & DMUDF-T & Ours+DMUDF & GT
						\end{tabular}}
						\caption{\textbf{Reconstruction methods based on Dual Contouring on an autodecoder trained on ShapNet-Cars~\cite{Chang15} at varying resolutions.}}
						\label{fig:supp_dc_carsautodec}
					\end{center}
				\end{figure}

\setlength{\tabcolsep}{\mytabcolsep}


\newlength{\suppcompfigwidth}
\setlength{\suppcompfigwidth}{0.192\linewidth}
\definetrim{suppcomp1}{25cm 15cm 15cm 0cm}
\definetrim{suppcomp2}{30cm 5cm 10cm 20cm}
\setlength\mytabcolsep{\tabcolsep}
\setlength\tabcolsep{1pt}

\begin{figure}[ht!]
	\begin{center}
	{\scriptsize
		\begin{tabular}{llccccc}
		    \rotatebox[origin=c]{90}{DMUDF~\cite{Zhang23b}} & \rotatebox[origin=c]{90}{}
			& \includegraphics[valign=m,width=\suppcompfigwidth,suppcomp1]{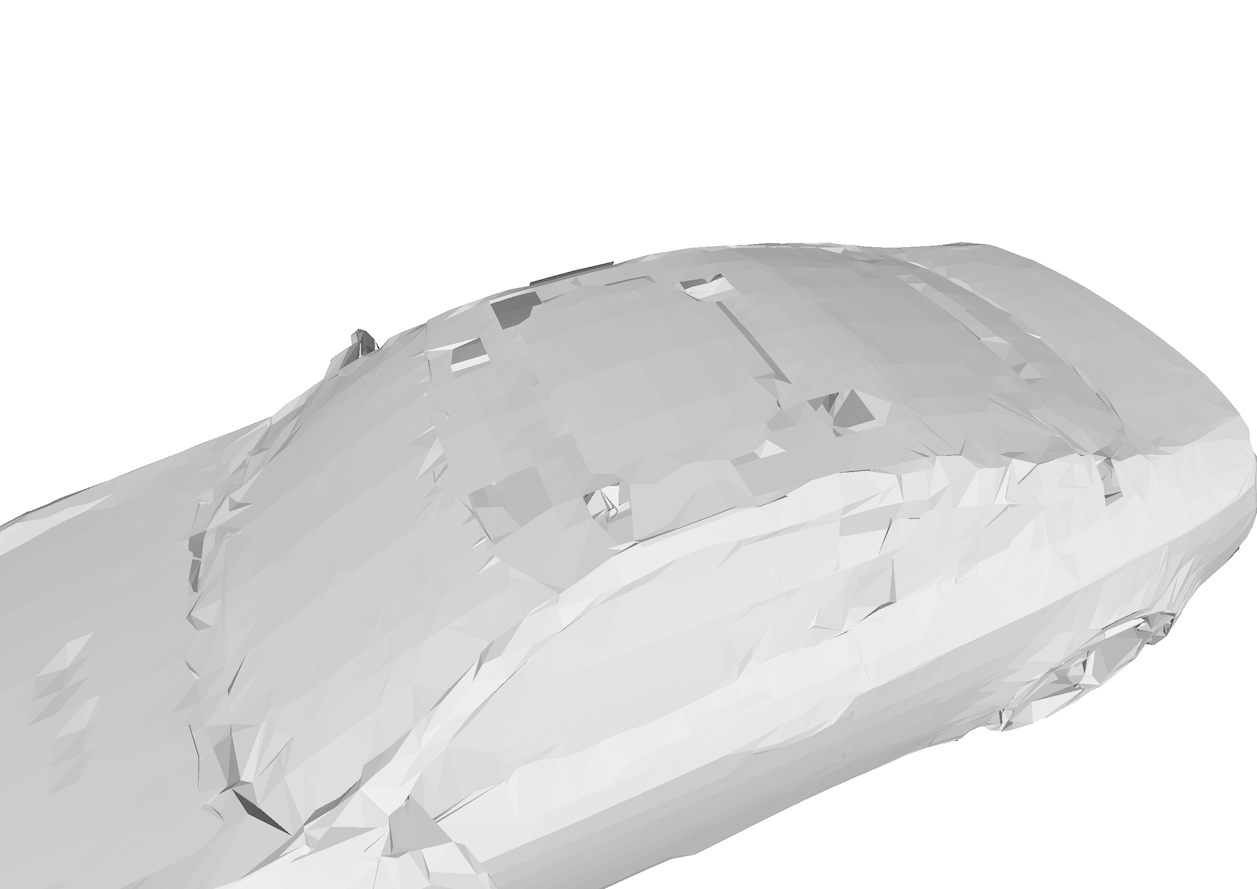}
			& \includegraphics[valign=m,width=\suppcompfigwidth,suppcomp1]{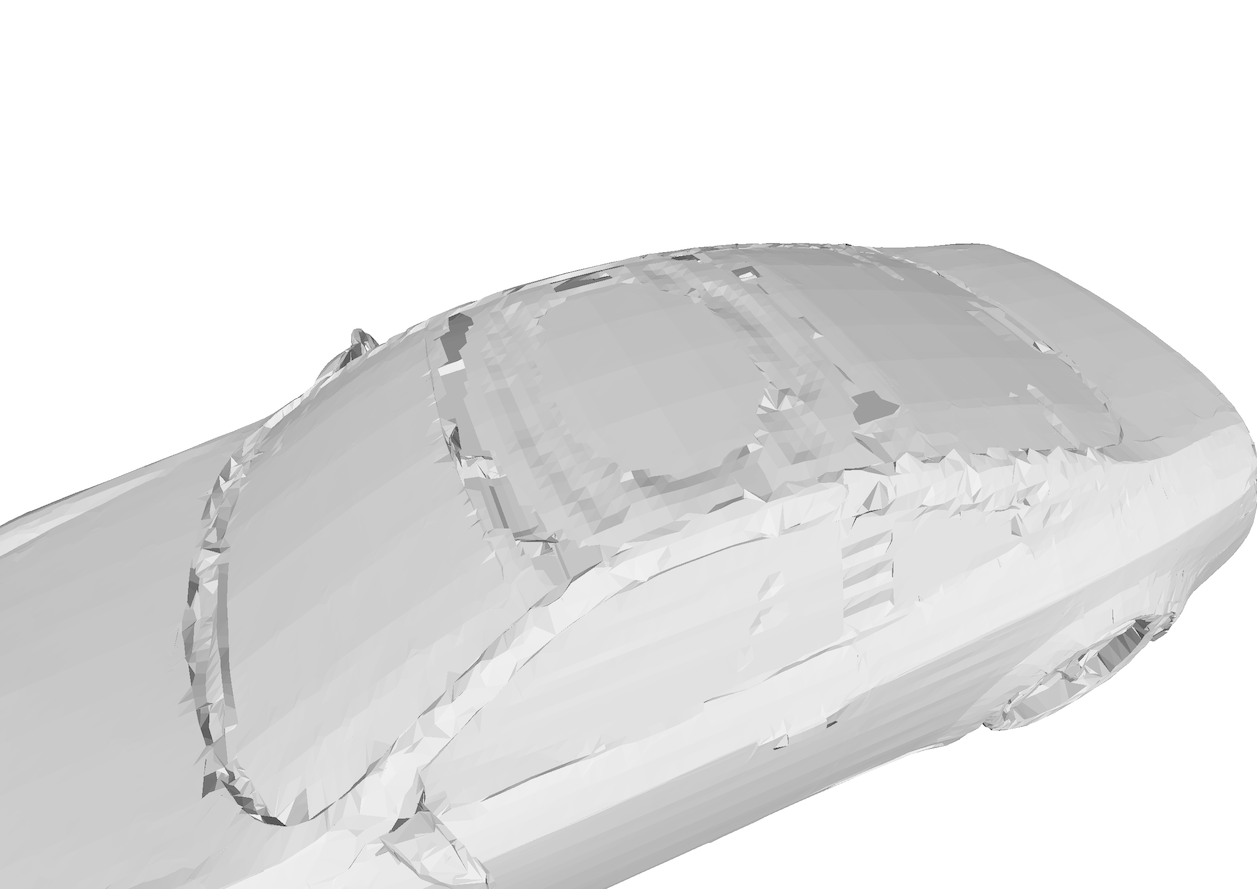}
			& \includegraphics[valign=m,width=\suppcompfigwidth,suppcomp1]{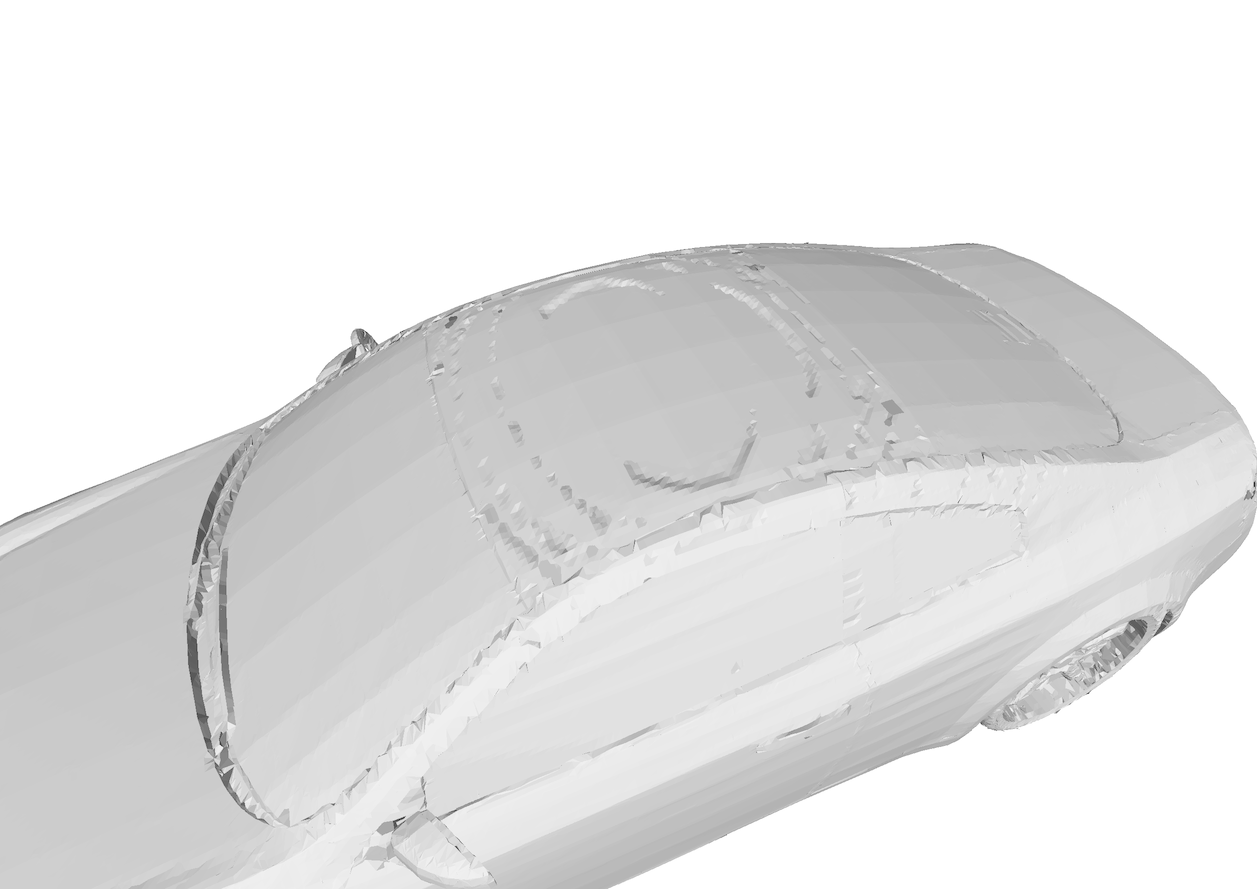}
			& \includegraphics[valign=m,width=\suppcompfigwidth,suppcomp1]{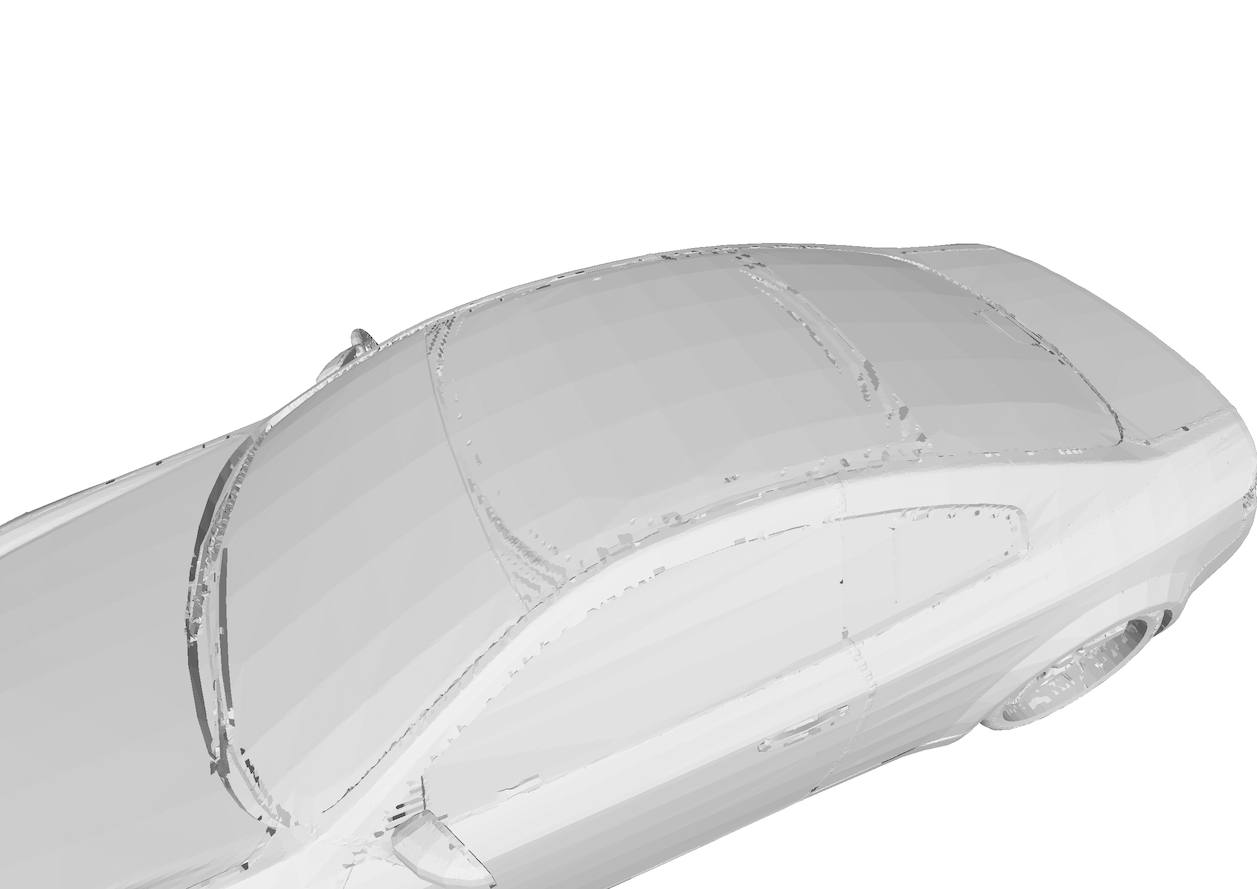}
			& \multirow{2}{*}{\includegraphics[valign=m,width=\suppcompfigwidth,suppcomp1]{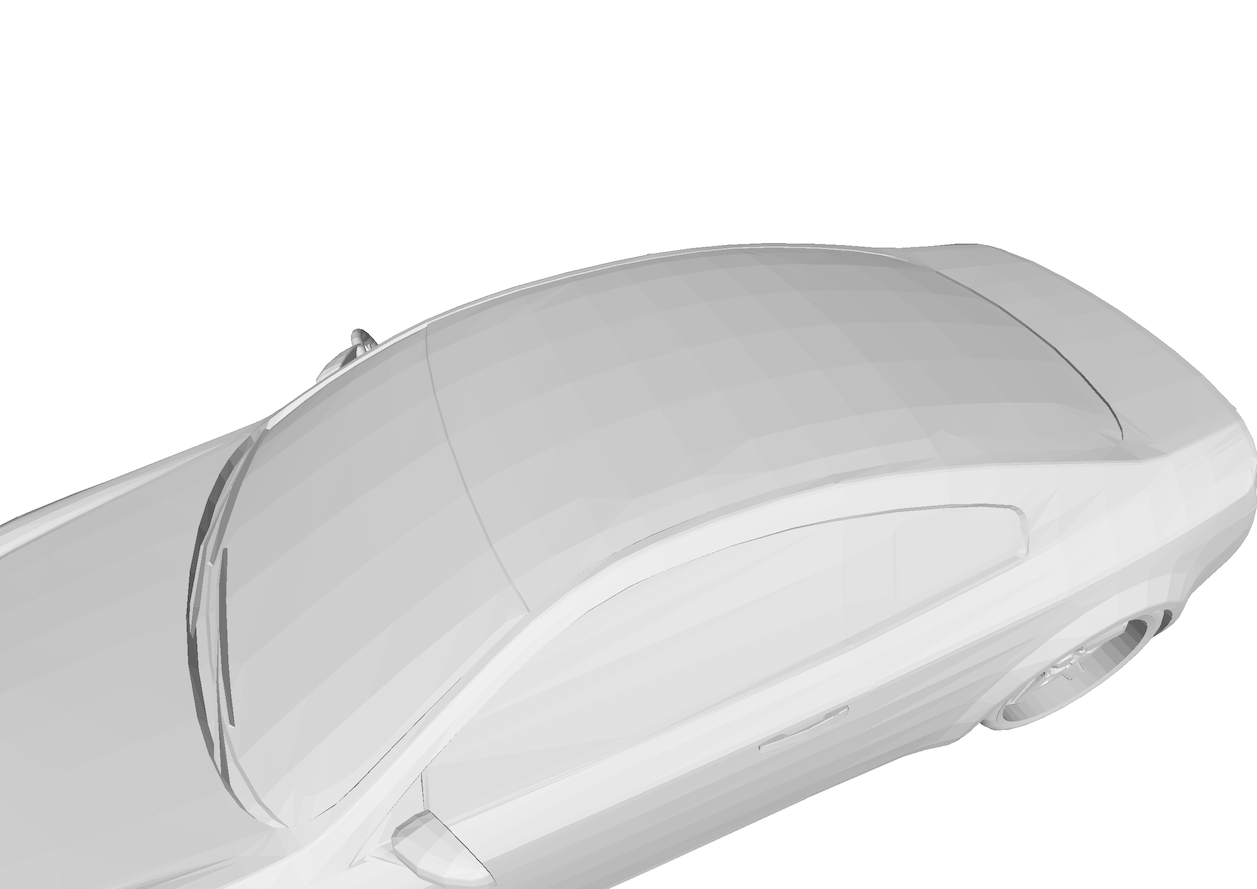}}\\ \addlinespace[2\tabcolsep]
			\rotatebox[origin=c]{90}{Ours+DMUDF} & \rotatebox[origin=c]{90}{}
			& \includegraphics[valign=m,width=\suppcompfigwidth,suppcomp1]{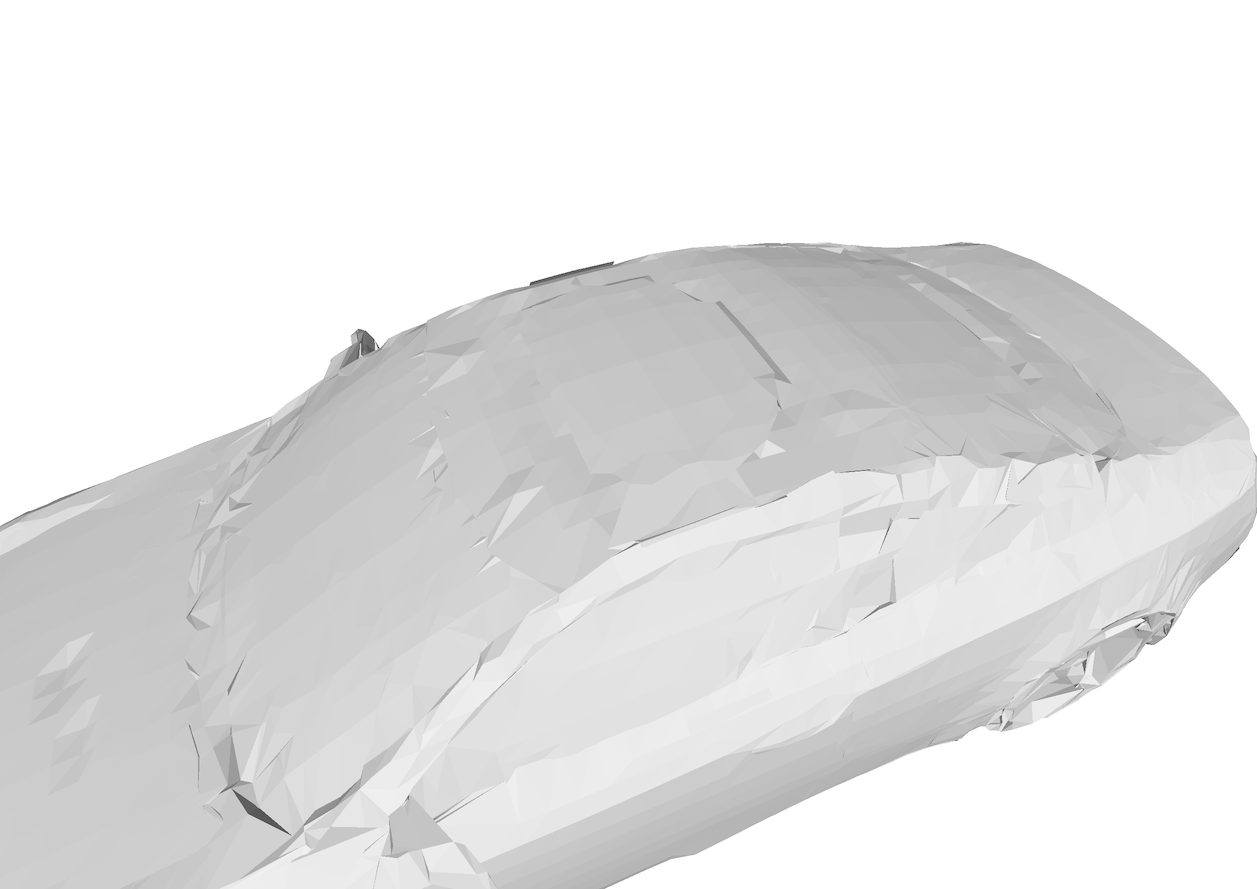} 
			& \includegraphics[valign=m,width=\suppcompfigwidth,suppcomp1]{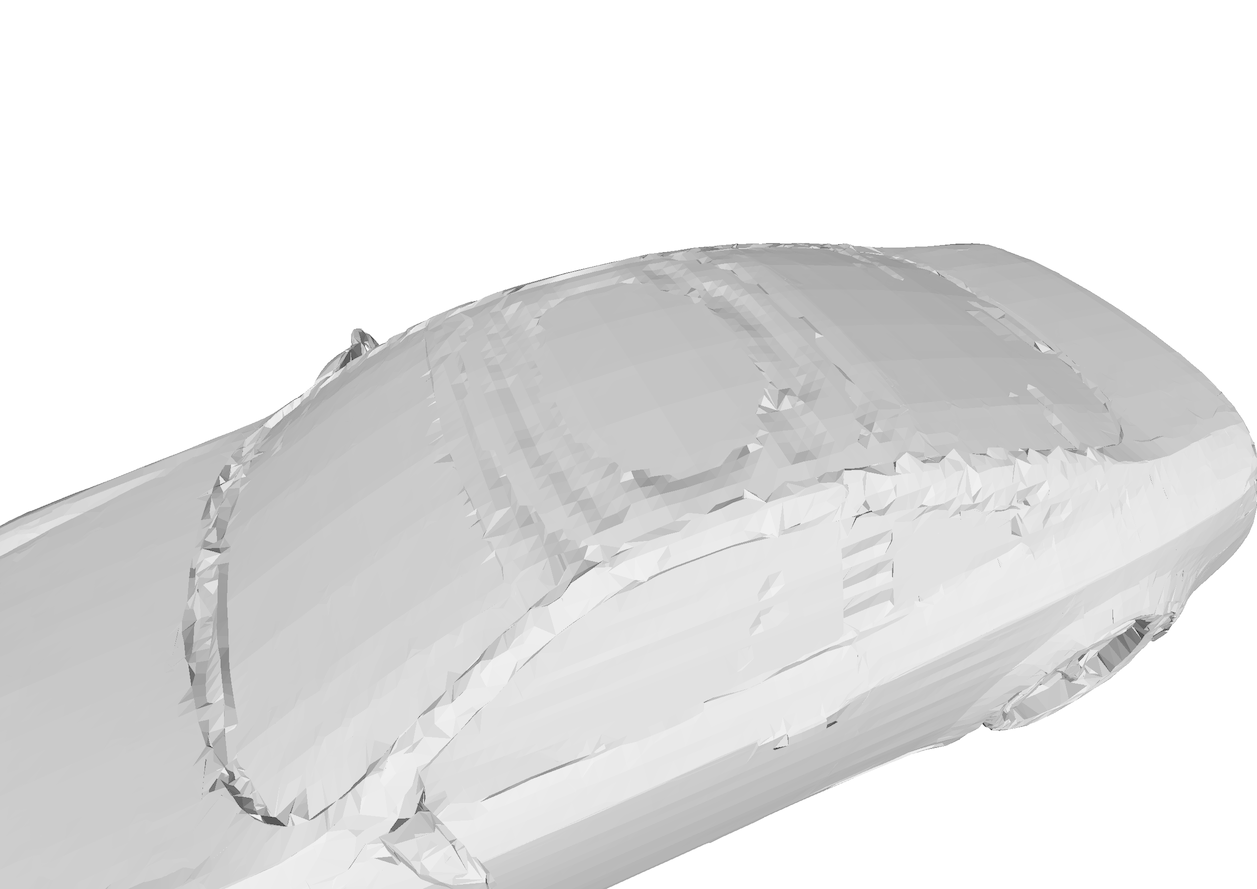} 
			& \includegraphics[valign=m,width=\suppcompfigwidth,suppcomp1]{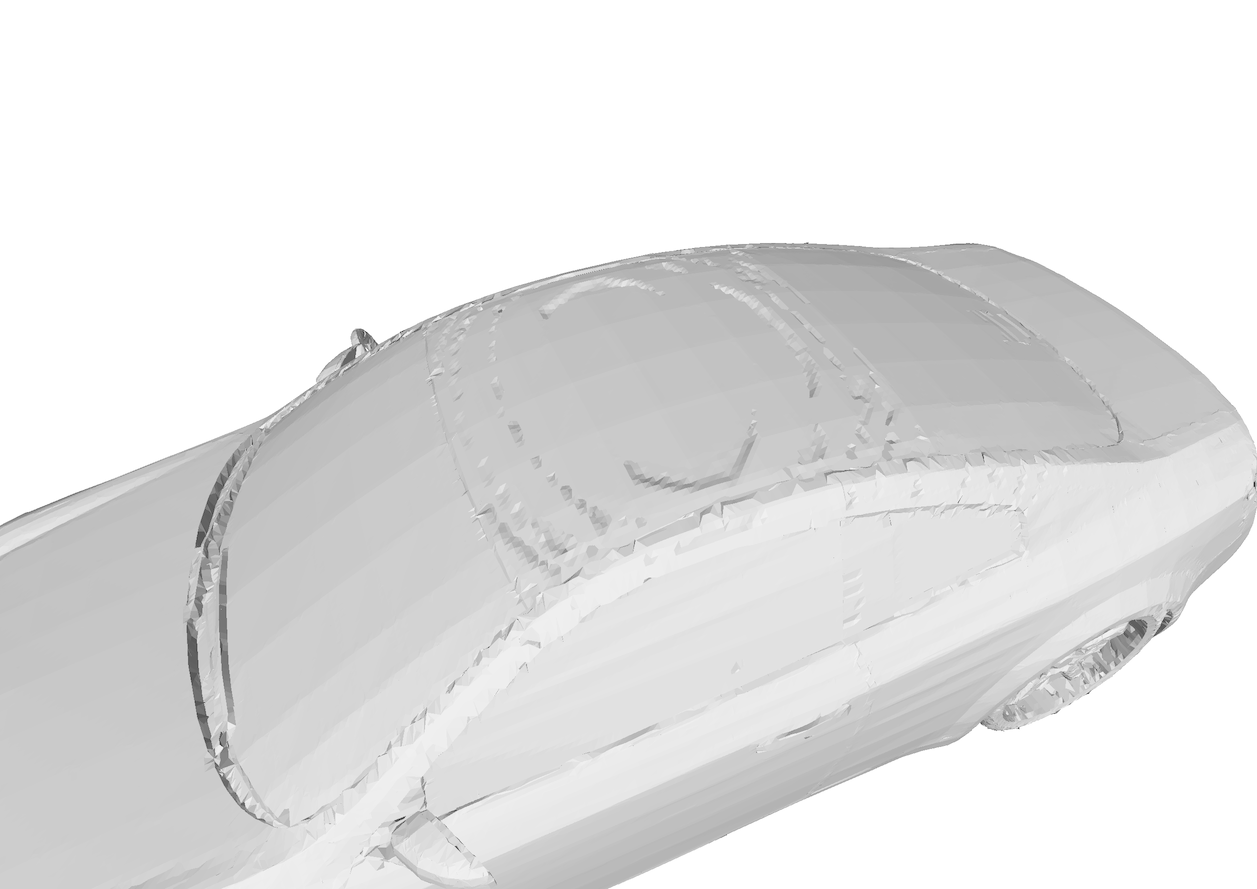} 
			& \includegraphics[valign=m,width=\suppcompfigwidth,suppcomp1]{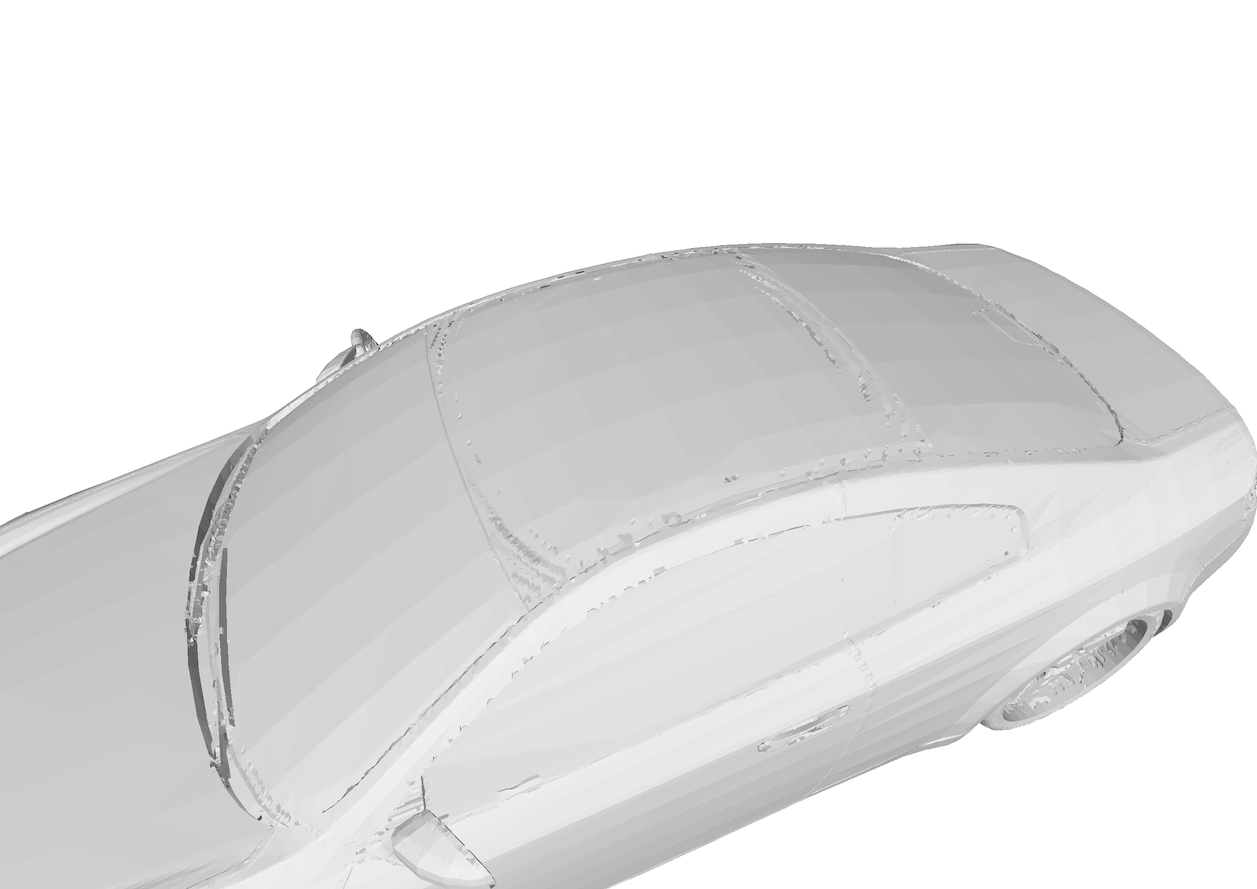} 
			\\ \addlinespace[2\tabcolsep] \hline \addlinespace[2\tabcolsep]

			\rotatebox[origin=c]{90}{DMUDF~\cite{Zhang23b}} & \rotatebox[origin=c]{90}{}
			& \includegraphics[valign=m,width=\suppcompfigwidth,suppcomp2]{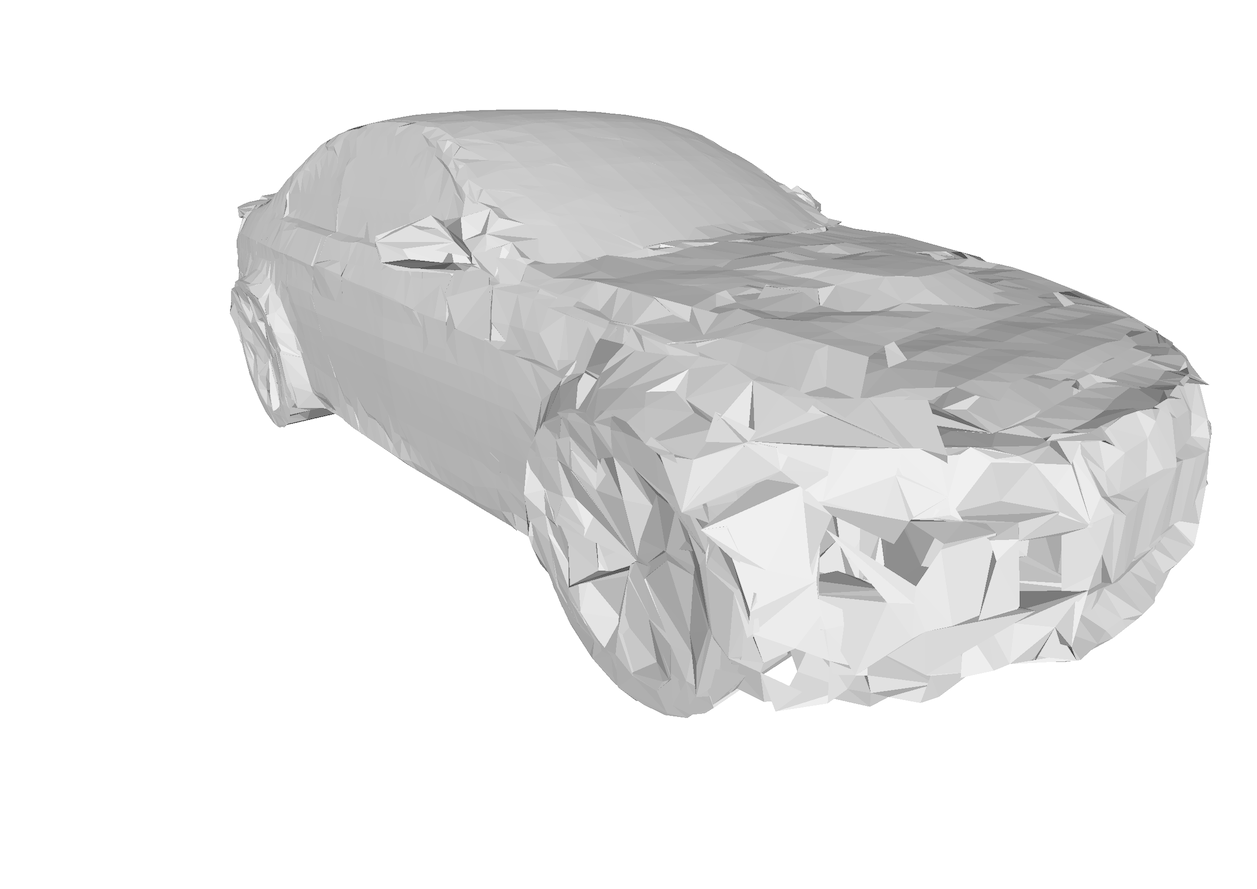}
			& \includegraphics[valign=m,width=\suppcompfigwidth,suppcomp2]{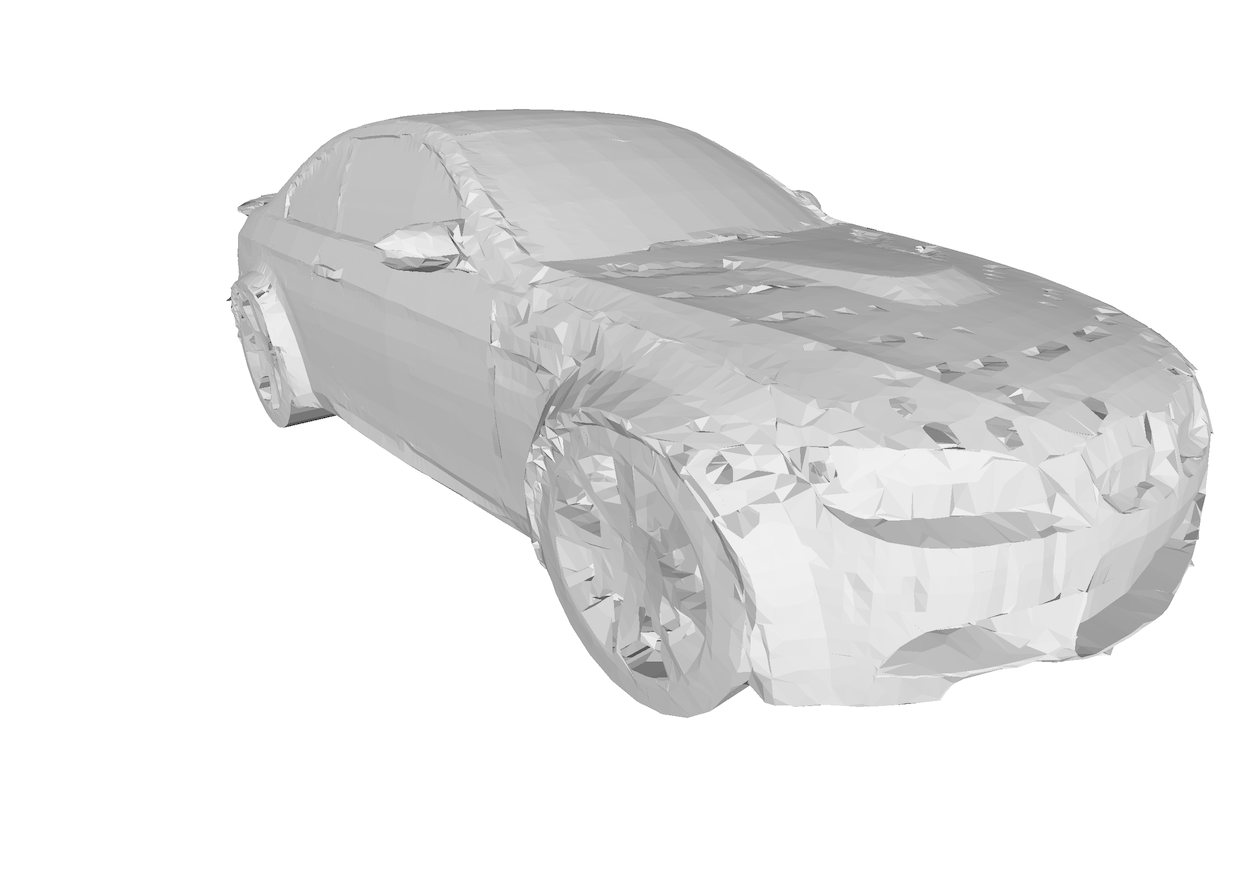}
			& \includegraphics[valign=m,width=\suppcompfigwidth,suppcomp2]{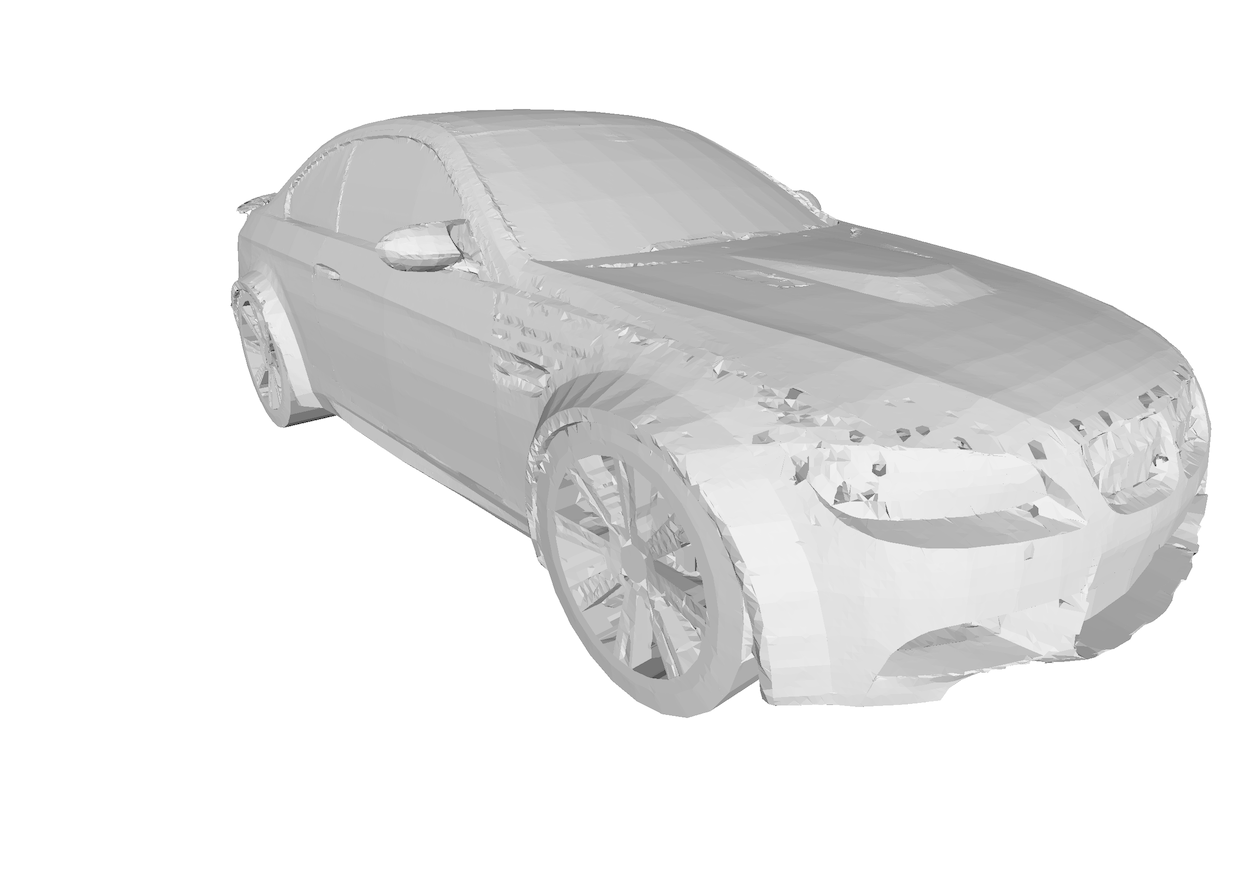}
			& \includegraphics[valign=m,width=\suppcompfigwidth,suppcomp2]{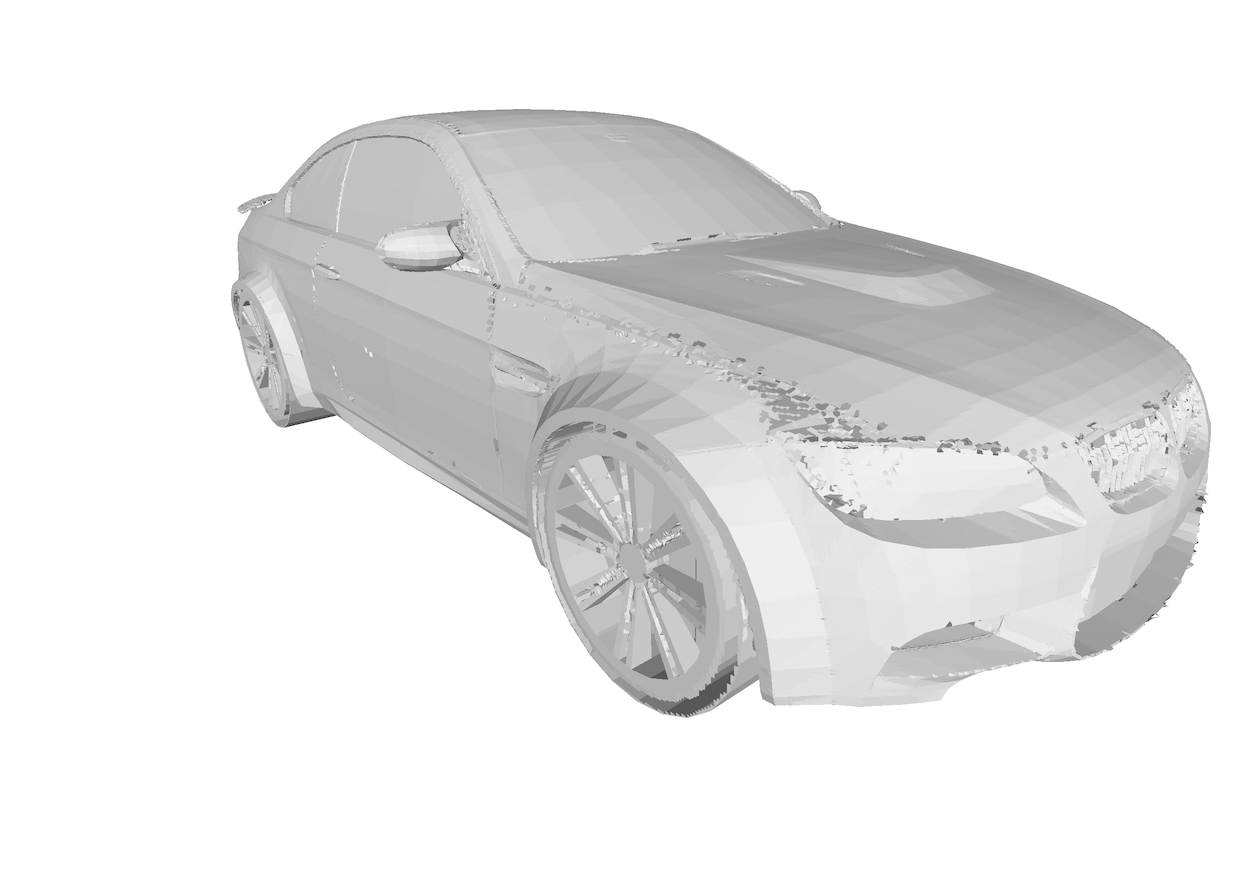}
			& \multirow{2}{*}{\includegraphics[valign=m,width=\suppcompfigwidth,suppcomp2]{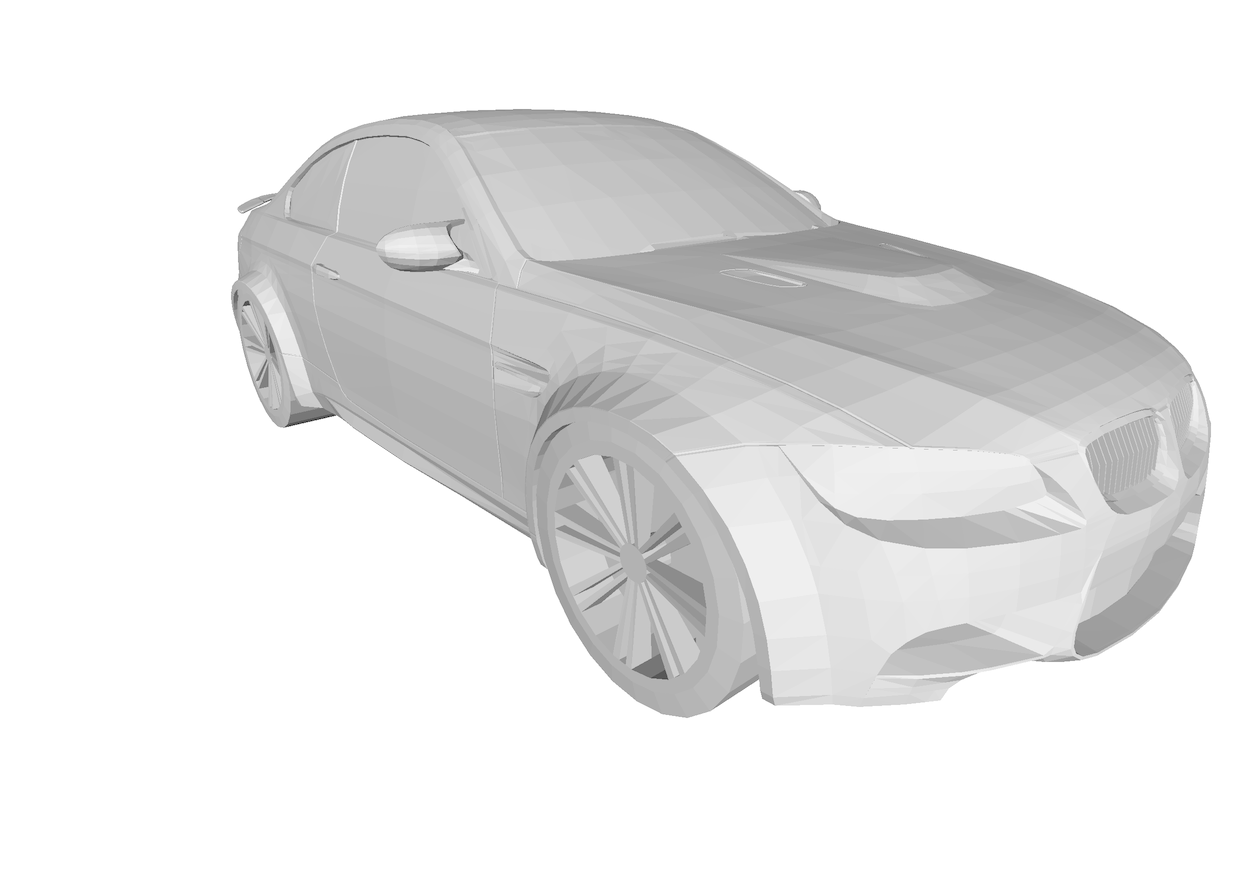}}\\ \addlinespace[2\tabcolsep]
			\rotatebox[origin=c]{90}{Ours+DMUDF} & \rotatebox[origin=c]{90}{}
			& \includegraphics[valign=m,width=\suppcompfigwidth,suppcomp2]{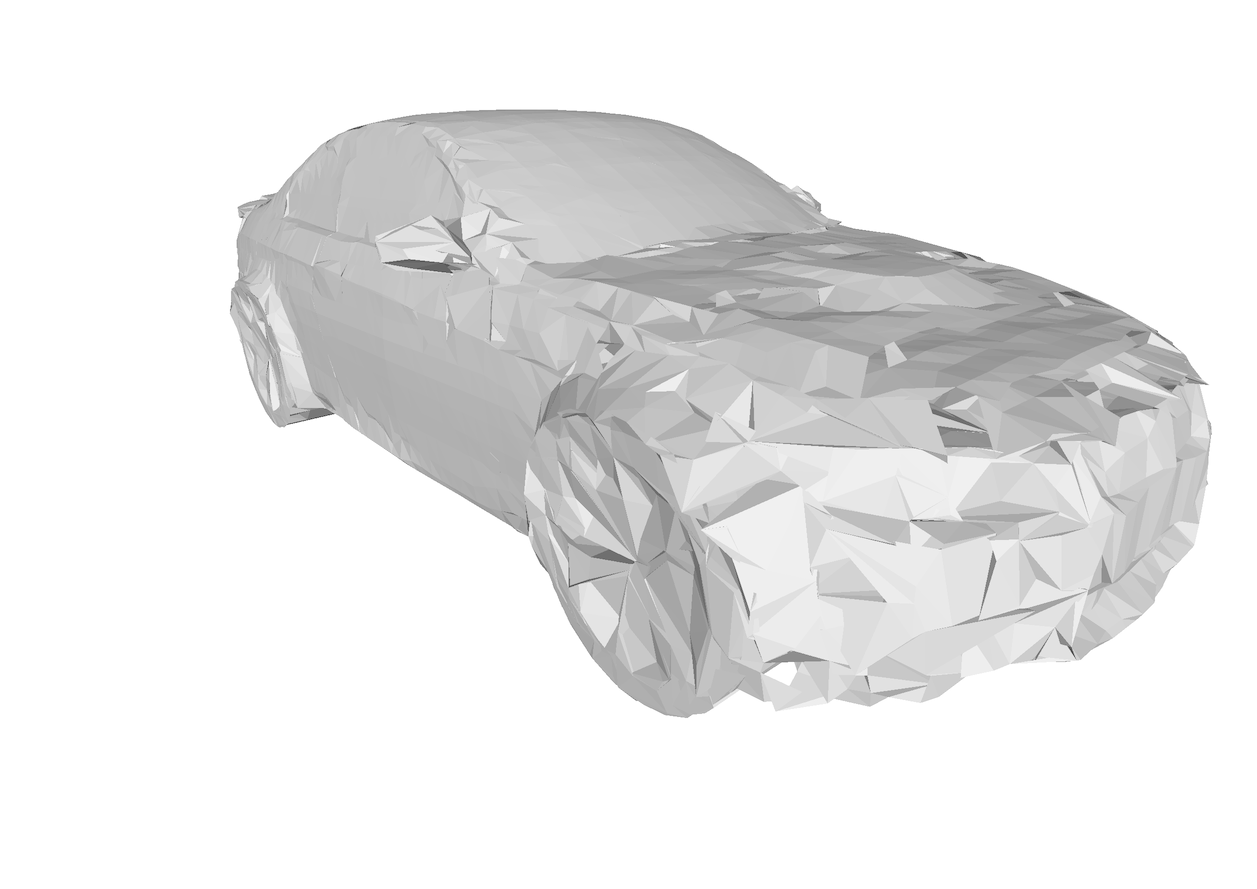} 
			& \includegraphics[valign=m,width=\suppcompfigwidth,suppcomp2]{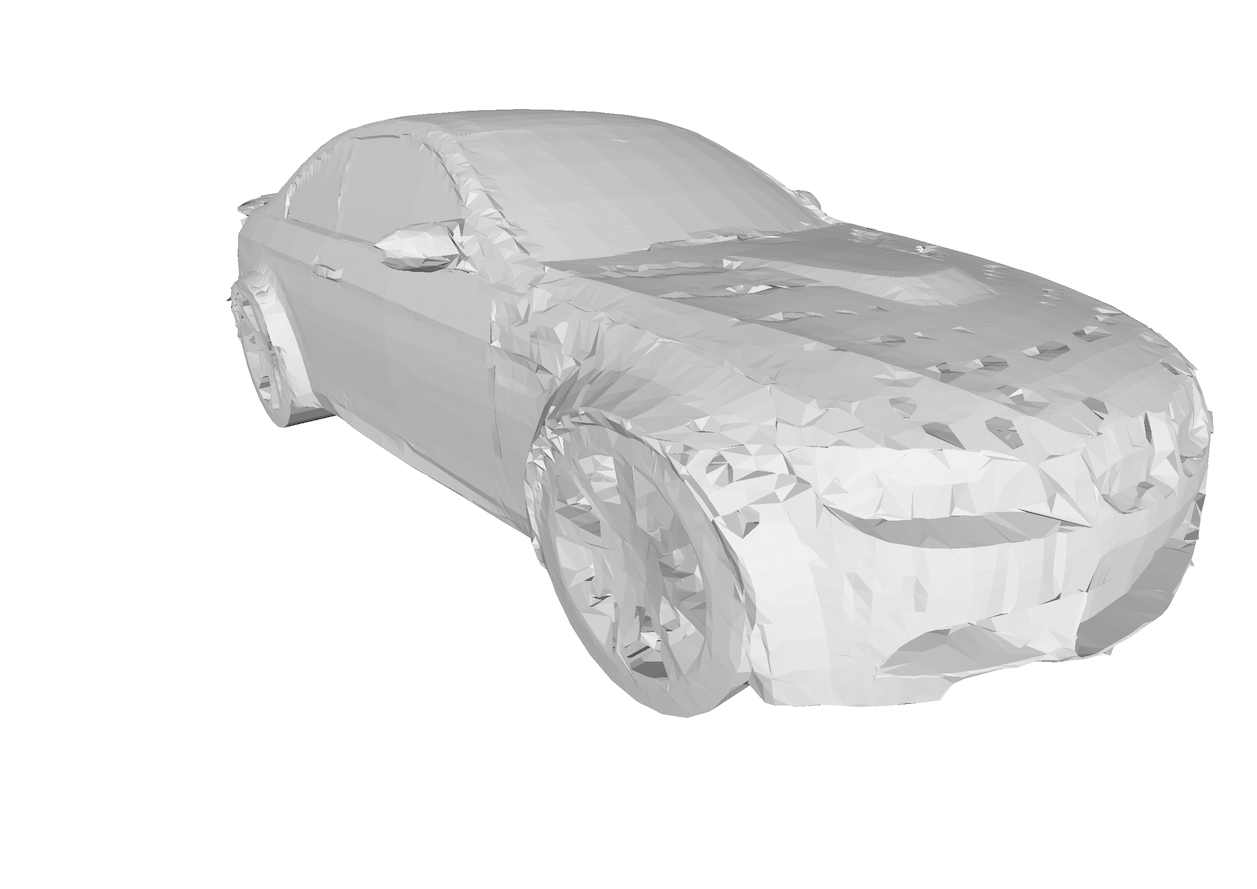} 
			& \includegraphics[valign=m,width=\suppcompfigwidth,suppcomp2]{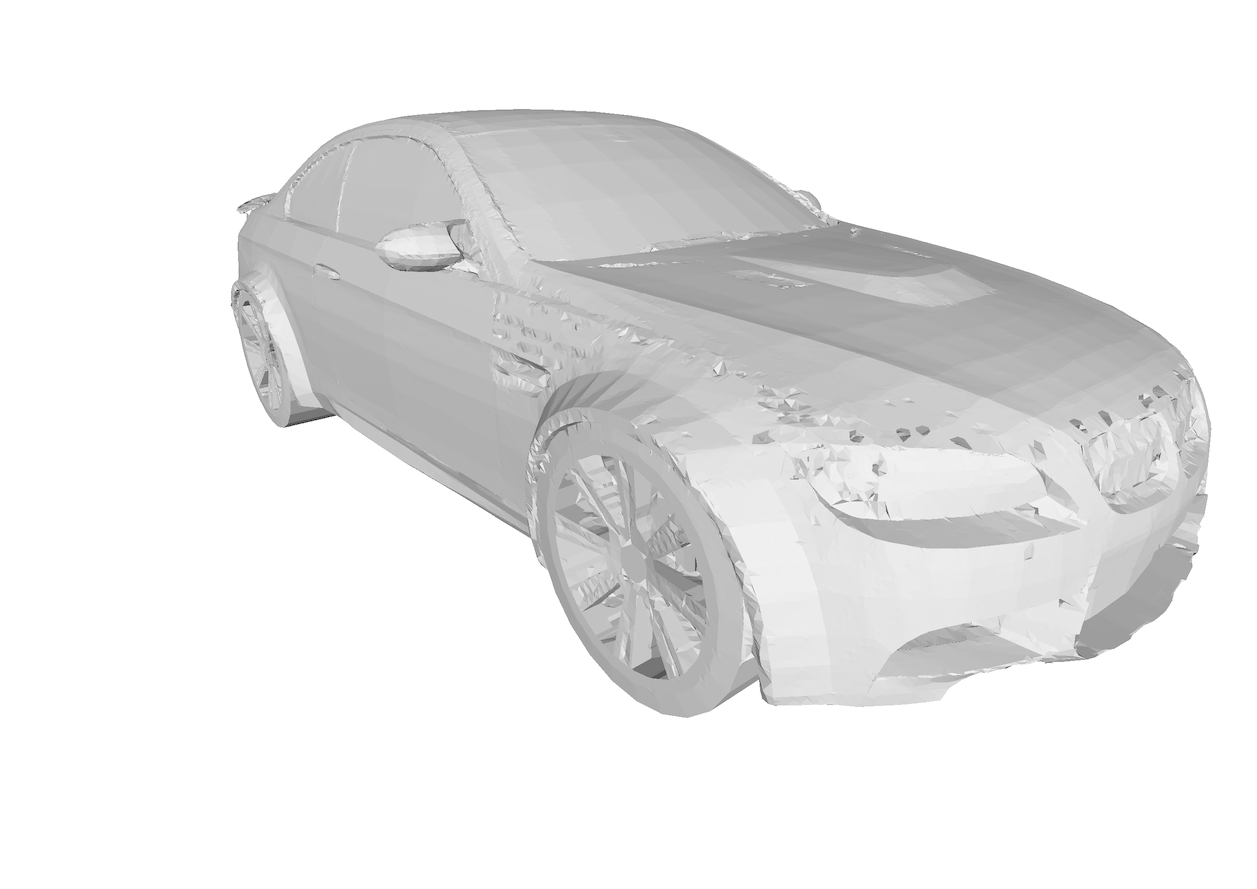} 
			& \includegraphics[valign=m,width=\suppcompfigwidth,suppcomp2]{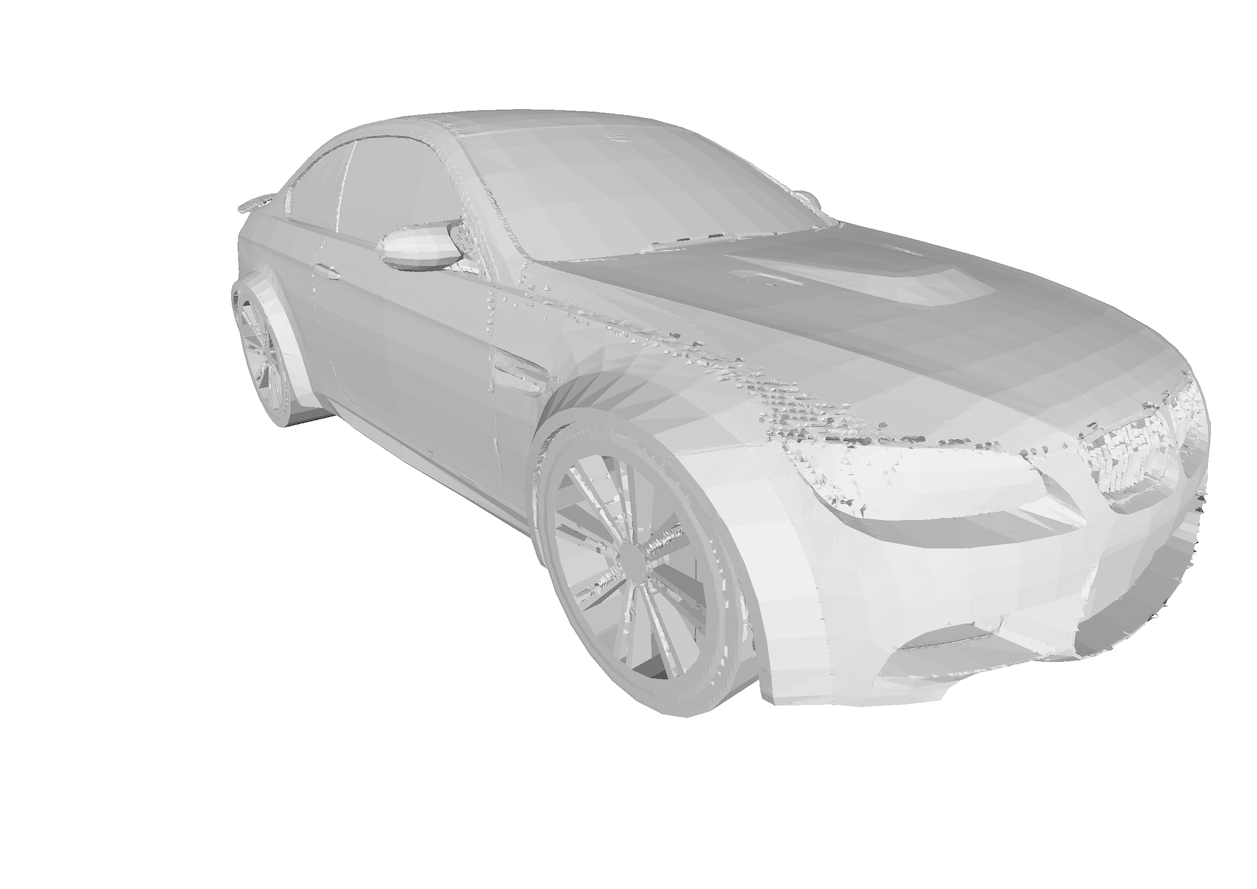} \\ \addlinespace[2\tabcolsep]
			&& 64 & 128 & 256 & 512 & GT
		\end{tabular}}
		\caption{\textbf{Improving on DualMesh-UDF~\cite{Zhang23b}.} Details of how our method can be used to improve the results of DualMesh-UDF by closing many of the holes.}
		\label{fig:supp_comparison}
	\end{center}
\end{figure}

We show in Figures~\ref{fig:supp_mc_abc}-\ref{fig:supp_dc_carsautodec} some additional reconstructions of the models presented in the main paper, across all tested resolutions.

We also show in Figure~\ref{fig:supp_comparison} a comparison between our method and DualMesh-UDF~\cite{Zhang23b} on the ShapeNet-Cars~\cite{Chang15} dataset. We can see that our method is able to fill many of the holes across all resolutions, for example on the roof of the car in the first two rows or in the front of the car in the second two rows. The front tire reconstructed by DualMesh-UDF at resolution 512 is also hollow, while our method is able to close it.

\clearpage

\else

\begin{abstract}

Extracting surfaces from Signed Distance Fields (SDFs) can be accomplished using traditional algorithms, such as Marching Cubes. However, since they rely on sign flips across the surface, these algorithms cannot be used directly on Unsigned Distance Fields (UDFs). In this work, we introduce a deep-learning approach to taking a UDF and turning it locally into an SDF, so that it can be effectively triangulated using existing algorithms. We show that it achieves better accuracy in surface detection than existing methods. Furthermore it generalizes well to unseen shapes and datasets, while being parallelizable. We also demonstrate the flexibily of the method by using it in conjunction with DualMeshUDF, a state of the art dual meshing method that can operate on UDFs, improving its results and removing the need to tune its parameters.


\end{abstract}




\section{Introduction}

By providing a continuous and differentiable representation of three-dimensional shapes whose topology can change, Signed Distance Fields (SDFs) have proved their worth. However, they are best suited to modeling watertight surfaces. Non-watertight ones, such as those garments are made of, can be handled by meshing a non-zero level set, which amounts to wrapping an SDF around them, but at the cost of an accuracy loss.

Thus, Unsigned Distance Fields (UDFs) have emerged as an effective alternative for representing open surfaces. Unfortunately, when it becomes necessary to convert an implicit surface into an explicit one, for example - rendering purposes, UDFs are at a disadvantage. While there are well-established approaches such as Marching Cubes \cite{Chernyaev95} and Dual Contouring \cite{Ju02} for triangulating SDF-based implicit surfaces, existing triangulation methods that can operate on UDFs are less reliable. Some rely on hand-crafted rules~\cite{Guillard22b, Zhou22, Zhang23b} while others use a neural network~\cite{Chen22b} trained to predict the vertex locations. As it is difficult to hand-craft an exhaustive set of rules, rule-based methods often produce artifacts, unwanted holes, and spurious surfaces. Furthermore, they can introduce  biases. The neural-based one~\cite{Chen22b} ignores important information, such as the gradient of the field. Furthermore, its accuracy is inherently bounded by that of the training data used to train the network, in this case the accuracy of the Dual Contouring~\cite{Ju02} algorithm used to produce it.



\newlength{\teaserfigwidth}
\setlength{\teaserfigwidth}{0.192\linewidth}
\definetrim{teaser}{31cm 15cm 11cm 20cm}
\setlength\mytabcolsep{\tabcolsep}
\setlength\tabcolsep{1pt}

\begin{figure}[t]
	\begin{center}
	{\scriptsize
		\begin{tabular}{llccccc}
		    \rotatebox[origin=c]{90}{} & \rotatebox[origin=c]{90}{} &
			\includegraphics[valign=m,width=\teaserfigwidth,teaser]{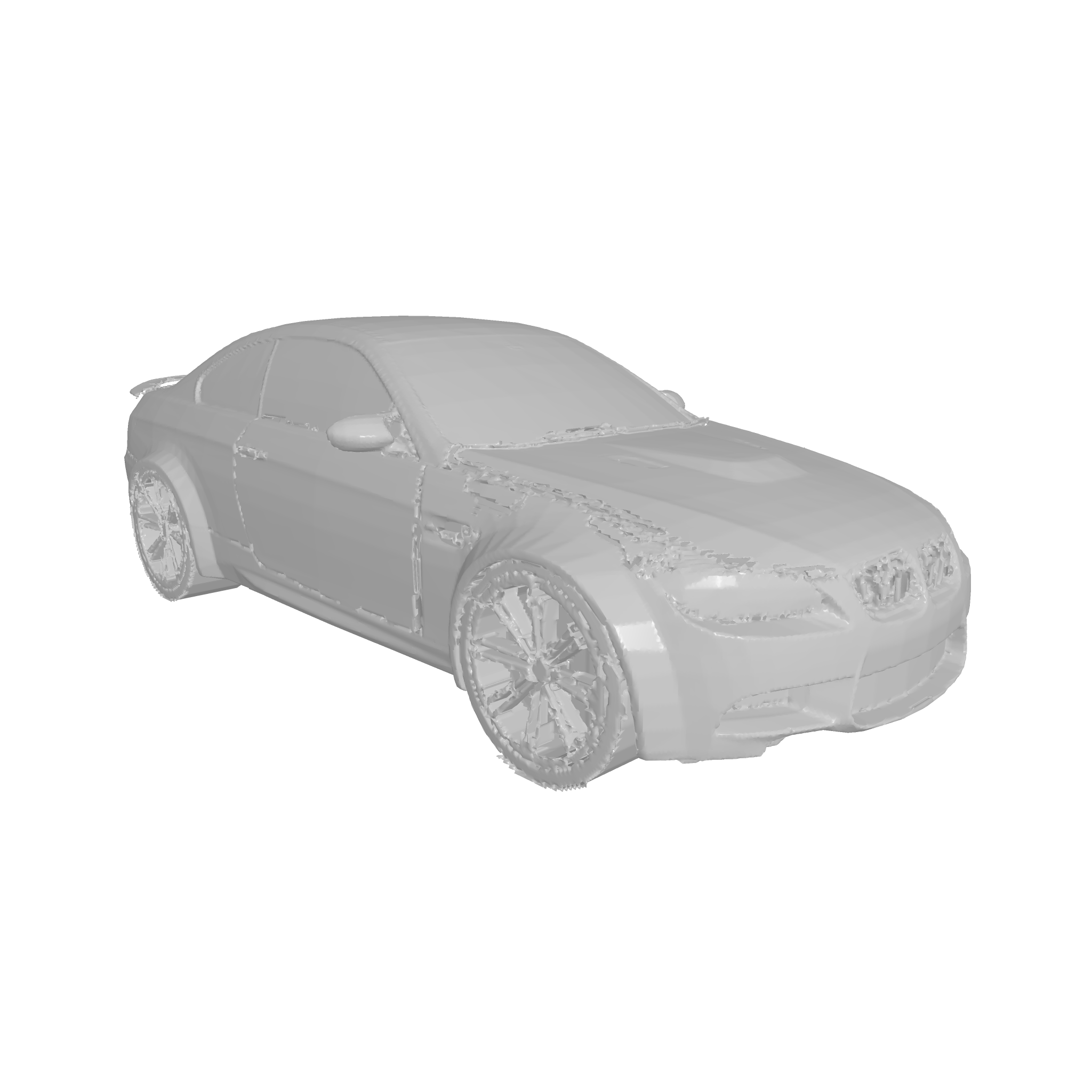}
			& \includegraphics[valign=m,width=\teaserfigwidth,teaser]{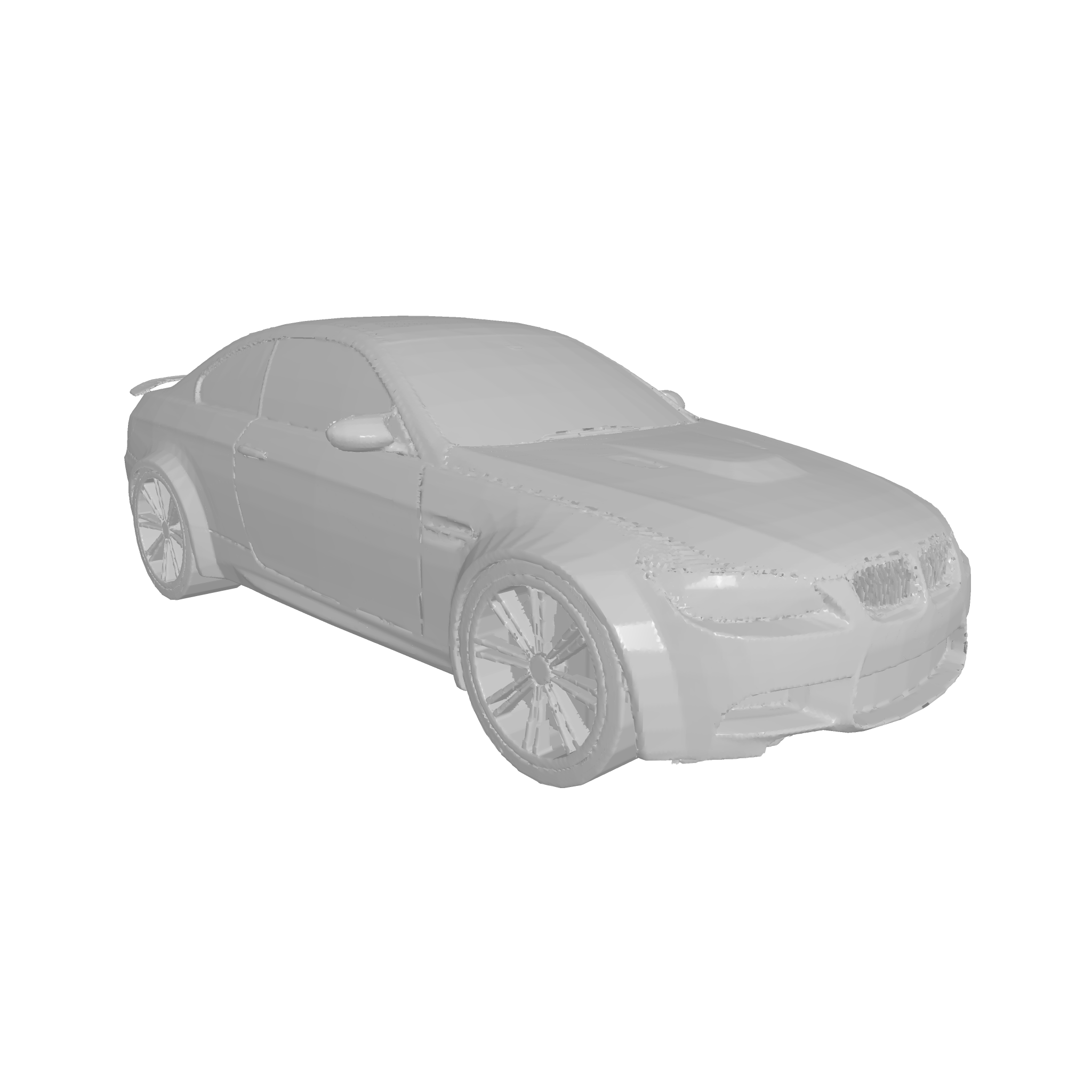}
			& \includegraphics[valign=m,width=\teaserfigwidth,teaser]{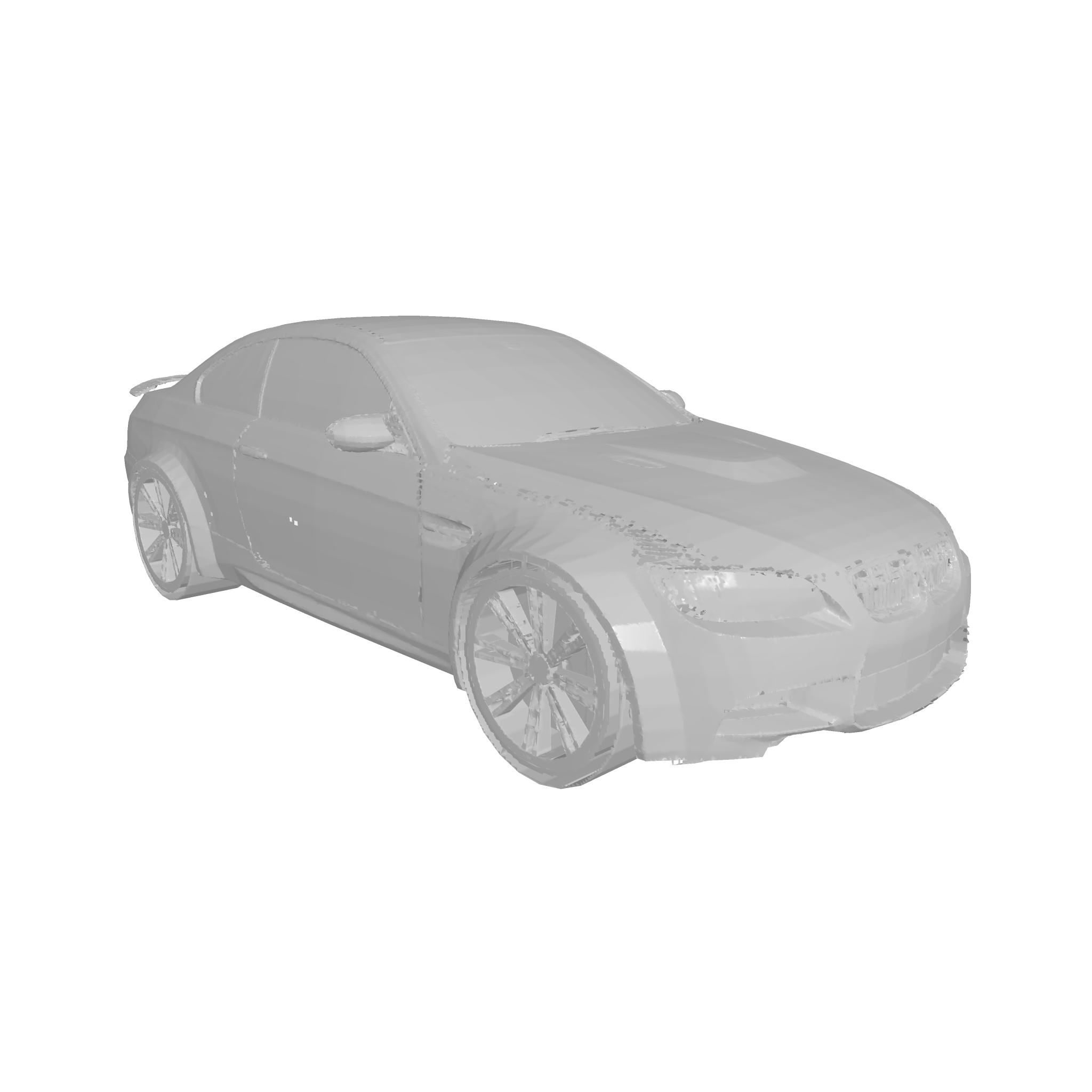}
			& \includegraphics[valign=m,width=\teaserfigwidth,teaser]{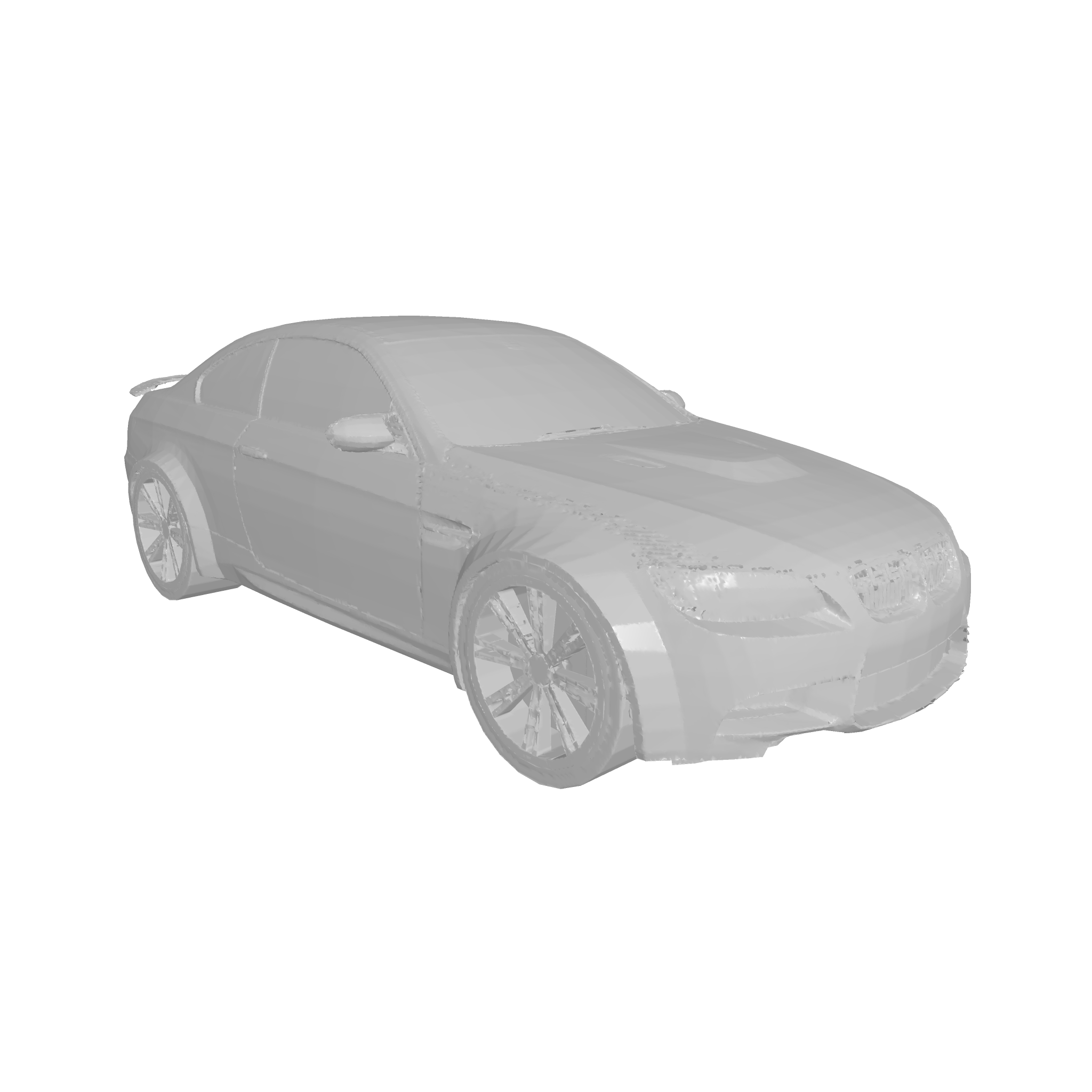}
			& \includegraphics[valign=m,width=\teaserfigwidth,teaser]{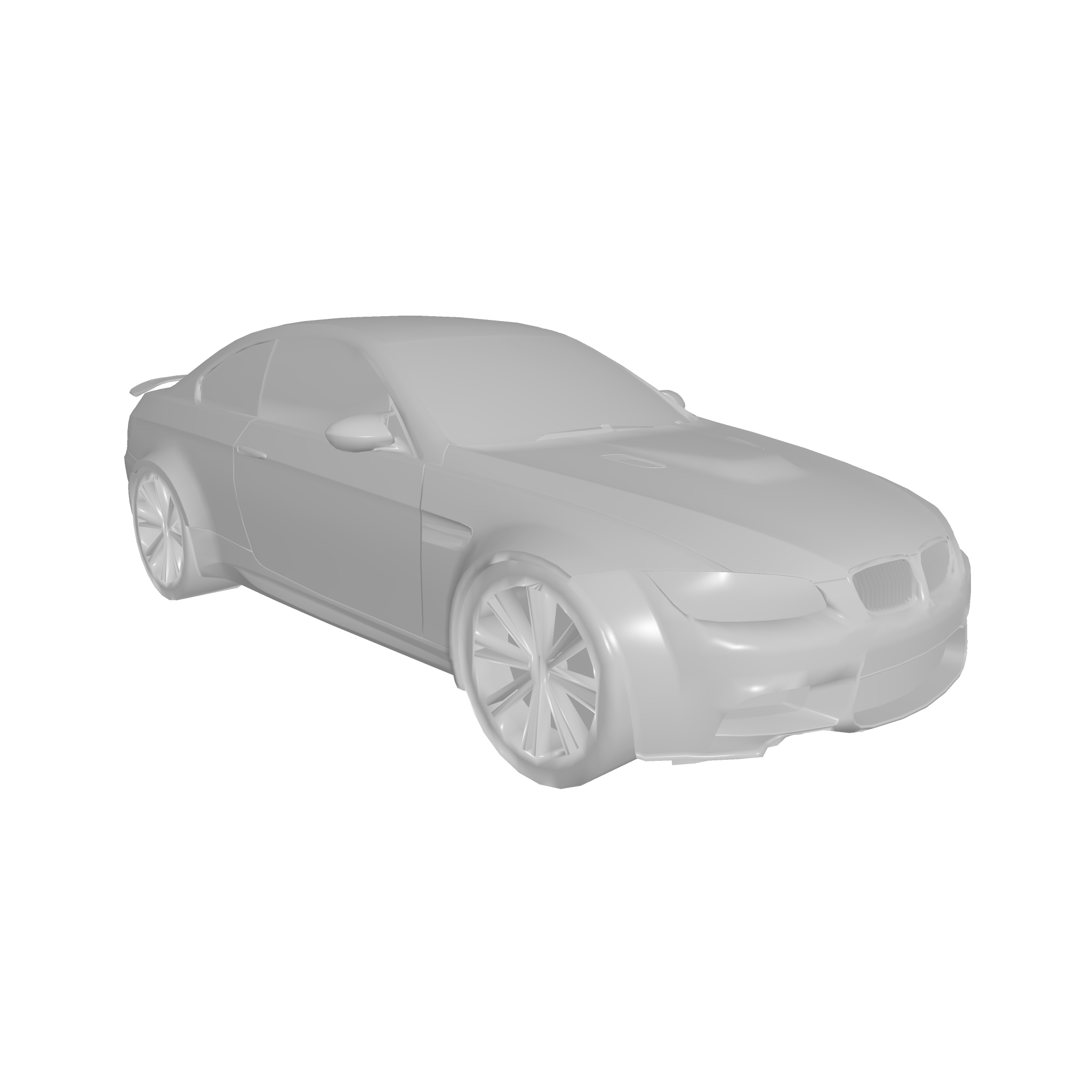}\\ \addlinespace[2\tabcolsep]
			&& MeshUDF~\cite{Guillard22b} & Ours+MC\cite{Lewiner03} & DMUDF\cite{Zhang23b} & Ours+DMUDF & GT
		\end{tabular}}
		\caption{\textbf{Neural Surface Detection.} We mesh a UDF by turning it into a pseudo-SDF using a neural network, and then meshing it using a triangulation algorithm. When coupled with Marching Cubes~\cite{Lewiner03}, our method yields better results than other MC-based methods. When coupled with DualMeshUDF~\cite{Zhang23b}, a dual method, it removes the requirement for parameter-tuning and closes most holes in the reconstructions.}
		\label{fig:teaser}
	\end{center}
\end{figure}

In this work, we introduce a deep-learning based approach to surface detection in unsigned distance fields that avoids these pitfalls. We formulate the surface detection problem as a local cell-wise classification one. More specifically, we start with an UDF whose values and gradients are given on a 3D grid and train a network to turn them into an SDF-like field in which signs flip across boundaries and that can be triangulated using the standard Marching Cubes algorithm~\cite{Lorensen87}. To train the network, we use watertight surfaces for which an SDF ground-truth can be computed precisely. However, once trained, the network can be used on non-watertight surfaces as well, because it makes local decisions. 

Our approach is entirely data-driven, without any hand-crafted rules that can result in harmful biases. This is unlike a method such as MeshUDF~\cite{Guillard22b} that incorporates heuristics favoring thin, continuous surfaces. These are suitable for meshing garments but not for more complicated objects such as cars with double surfaces or inner structures. This is also unlike DualMesh-UDF~\cite{Zhang23b}, which requires manually setting thresholds to asses whether the vertices in a given cell should be discarded or not, along with a simplistic heuristic to connect neighboring cells and form surfaces. Consequently, DualMesh-UDF is prone to creating holes, especially in flat or noisy regions, and to connecting nearby but different surfaces.

In short, our contribution is a novel data-driven approach for neural open surface detection. It comes without any annotation cost. The network is easy to train, adapt, and scalable to various scenarios. We demonstrate on the ABC~\cite{Koch19a}, ShapeNet-Cars~\cite{Chang15}, and MGN garments~\cite{Bhatnagar19} datasets that our approach outperforms the state-of-the-art MeshUDF and DualMesh-UDF~\cite{Zhang23b} algorithms, when used in conjunction with either Marching Cubes or Dual Contouring, respectively.

\section{Related Work}
\label{sec:related}

Our work focuses on converting an UDF into an explicit mesh, which is closely related to methods for triangulating signed and unsigned 3D fields.  Apart from UDFs, learning different representations for open surfaces is also an active research topic\cite{Wang22b,Chen22e}. However, it is beyond the scope of our paper.

\parag{Triangulating Signed Fields.}
Marching Cubes (MC)~\cite{Lorensen87} and Dual Contouring (DC)~\cite{Ju02} are two of the most popular meshing algorithms for signed fields. They both start with a 3D grid of positive and negative values and look for grid cells for which the vertex values are of different signs. While MC creates mesh vertices on the edges of such cells using an interpolation mechanism, DC creates them inside the cells and solves quadratic optimization problems to determine their exact position. More recently, deep-learning based methods, such as Neural Marching Cubes \cite{Chen21c} and Neural Dual Contouring (NDC)~\cite{Chen22b} have been proposed. They are designed to emulate the behavior of the original methods while being differentiable. Unfortunately, due to their reliance on sign flips, these approaches cannot be directly applied to UDFs, with the exception of NDC discussed below.  

\parag{Triangulating Unsigned Fields.}

Due to increased interest in modeling open non-watertight surfaces that are best represented by UDFs, several approaches have been proposed recently. We group them into approaches that rely on primal methods such as MC and approaches that use dual methods such as DC.

{\bf Primal Methods.}
In NDF~\cite{Chibane20b}, a dense point cloud is extracted from an unsigned field and then triangulated using the Ball Pivoting algorithm~\cite{Bernardini99}, which is computationally demanding. As Marching Cubes, MeshUDF~\cite{Guillard22b} relies on a grid-based method. It assigns pseudo-signs to the corners of grid cells based on the relative orientations of the UDF gradients and relies on a surface following heuristic to enforce consistency among neighboring cells. This works well on garments but the consistency heuristic can fail on more complex shapes such as those of raw ShapeNet cars~\cite{Chang15}. It also makes the algorithm strictly sequential and thus non-parallelizable. Similarly, CAP-UDF~\cite{Zhou22} uses hand-crafted rules to determine pseudo-signs to be meshed using the MC approach. Unfortunately, this often results in unwanted artifacts and tends to underperform MeshUDF~\cite{Zhang23b}. DoubleCoverUDF (DCUDF)~\cite{Hou2023DCUDF} is a recent 3-step approach that starts by reconstructing an inflated shape using Marching Cubes on a user-defined $r$-level set. It then refines the vertex locations using a specific loss function and, finally, cuts the mesh to only keep one surface using a minimum s-t cut strategy on a graph~\cite{Boykov04}. Thanks to its loss function, it yields very smooth results but tends to produce displaced and inaccurate surfaces, unless used at very high resolutions. It is also noticeably slower than the other methods and requires manually tuning the $r$ parameter. Compared to these methods, we propose a data-driven approach that does not use hand-crafted heuristics, and can be applied to both primal and dual methods.

{\bf Dual Methods.}
Unsigned Neural Dual Contouring (UNDC)~\cite{Chen22b} and Dual-Mesh-UDF~\cite{Zhang23b} rely on Dual Contouring. UNDC is a variation of NDC that accepts an unsigned input field and uses a 3D convolutional network to predict dual vertex locations and connectivities.  It uses a learning-based way to predict which edges contain a sign flip, as we do. However, it also predicts where to place a dual vertex, using the original Dual Contouring as training data. It requires specific data augmentations---rotations, symmetries and global sign inversions---to work correctly and ignores gradient information even though it is readily available, limiting its accuracy especially on neural UDFs, as shown in~\cite{Zhang23b}. Our approach, instead, uses gradient information as well as UDF values, predicts vertex-level pseudo-signs instead of edge sign flips, allowing the use of existing SDF-based triangulation approaches, uses a simpler network architecture and we show that it outperforms UNDC by being more robust to unseen shapes and noisy settings. Our method also takes significantly less time and data to train, about 1 minute and 80 shapes, compared to UNDC which takes 12 hours and 4280 shapes.

DualMesh-UDF~\cite{Zhang23b} is a variation of the original DC algorithm. It uses a distance threshold to prune cells unlikely to contain a surface, then runs quadratic optimizations on the remaining cells for vertex localization. Resulting vertices are filtered using SVD shape analysis, and connectivity is enforced through a vicinity rule instead of the original sign-flip rule. While DualMesh-UDF approximates surfaces well, especially at corners, its vertex filtering can create holes, and the connectivity heuristic may generate spurious faces. We show that our approach can be used in conjunction with DualMesh-UDF to improve its results in noisy settings and complex shapes, and eliminating the need for manual thresholds.

\section{Method}


\definetrim{pipeline}{0cm 182cm 56cm 0cm}

\begin{figure*}[t]
    \begin{center}
    \includegraphics[width=\linewidth,pipeline]{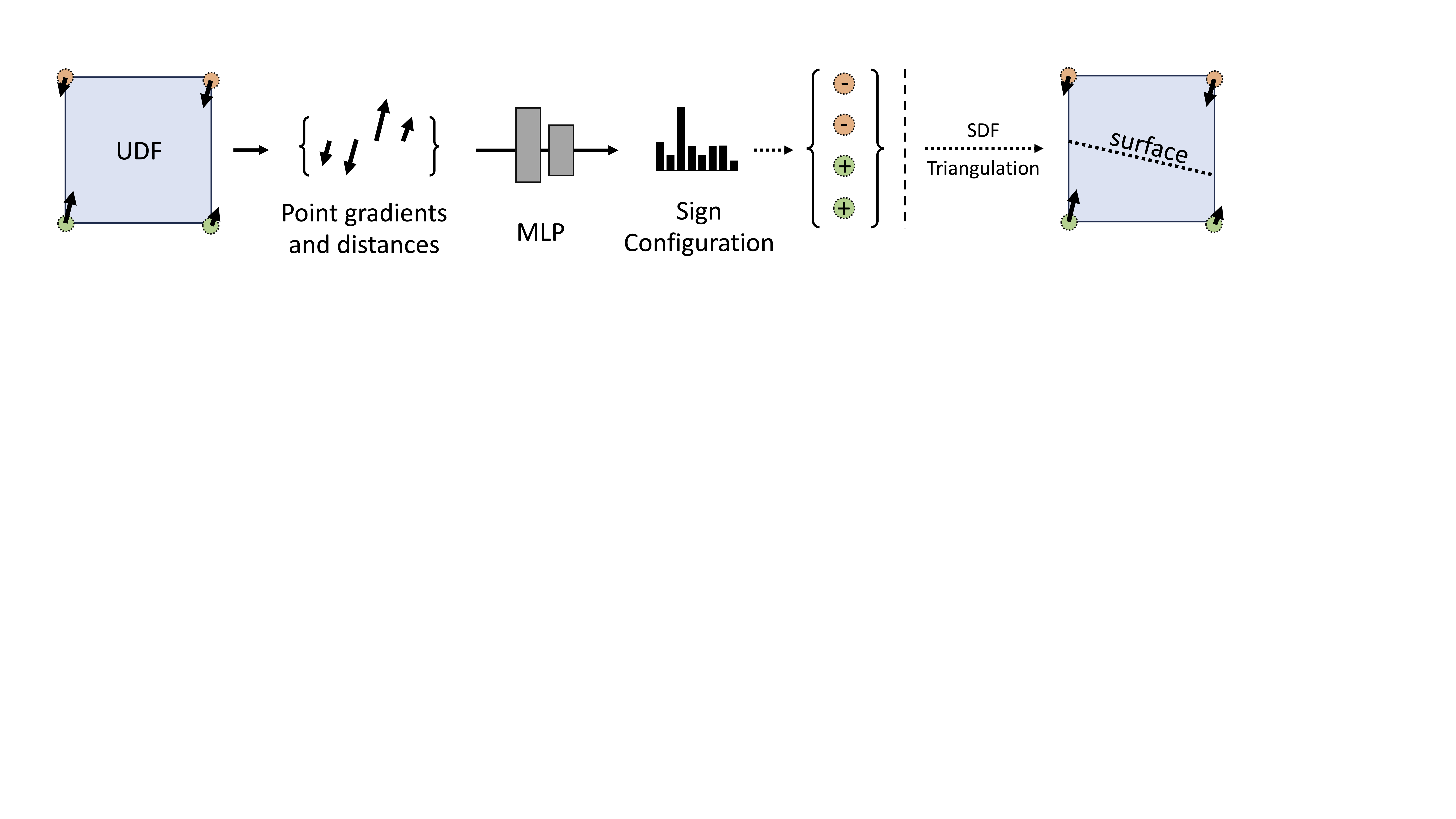}
    \end{center} 
    \caption{
    \textbf{Neural Surface Detection.}  We formulate the surface detection problem as a per-cell classification task. In each cell, we map point distances and gradients to a sign configuration of the cell vertices, which can be used to mesh the surface via Marching Cubes~\cite{Lewiner03} or Dual Contouring~\cite{Ju02}.
    }  
    \label{fig:pipeline} 
\end{figure*}

Given a surface $\mathcal{S}$ in $\mathbb{R}^3$, let the Unsigned Distance Field of $\mathcal{S}$ be
\begin{align}
    UDF_{\mathcal{S}} : \mathbb{R}^3 & \rightarrow \mathbb{R}^+ \; , \\
    \bx & \rightarrow d  \; , \nonumber
\end{align}
where $d$ is Euclidean distance from $x$ to the closest point on $\mathcal{S}$. $UDF_{\mathcal{S}}$ is zero on the surface and differentiable almost everywhere, with a gradient $\nabla UDF_{\mathcal{S}}$ pointing away from the closest point on $\mathcal{S}$ and of norm one. Non-differentiability only occurs on the surface and at points that are equidistant from two or more surface points. We now introduce our approach to locally turning $UDF_{\mathcal{S}}$ into a {\it signed} distance field, so that we can use a traditional meshing algorithm, such as Marching Cubes (MC) or Dual Contouring (DC), for triangulation purposes.

\subsection{From UDF to Pseudo-SDF}

As in standard meshing algorithms such as MC and DC, we start with a voxel grid. However, instead of having SDF values and gradients, we have values of $UDF_{\mathcal{S}}$ and their gradients. In this setup, we cannot apply MC or DC because the UDF values all are positive and, thus, there are no sign-flips. To remedy this, for each individual grid cell, we use a neural network to estimate pseudo-signs at its 8 corners, thus locally turning the UDF into an SDF that can be used to run a standard meshing algorithm.

The network we use is a simple multi-layer perceptron (MLP) that takes as input the 8 UDF values at the corners of a cell $c \in \mathcal{S}$ and their 8 respective 3D gradients, which we denote as $UDF_{\mathcal{S}}(c)$ and $\nabla UDF_{\mathcal{S}}(c)$, respectively. Our MLP has 2 hidden layers, each with 1024 dimensions and leaky ReLU activations, and a final layer with 128 dimensions,. The output layer is interpreted as a one-hot selection between $2^7=128$ choices, each corresponding to a different combination of pseudo-signs at the 8 corners, up to a complete sign symmetry inside the cell. 

We chose this approach because estimating which corners are on one side or the other of the surface is an ill-posed problem when that surface is not watertight. Thus, we do not attempt to orient the surface {\it globally}. Instead, we detect surface crossings {\it locally} and translate them into sign flips. Furthermore, we do not employ any mechanism to enforce the network to produce consistent pseudo-signs across neighboring cells, thus avoiding to tailor the method to specific assumptions, as in MeshUDF~\cite{Guillard22b}. The experiments show that the network learns to produce consistent results. However, if an inconsistency arises, this simply leads to holes in the mesh.

The use of one-hot 128 outputs instead of a 7-output multi-label classification comes from the intuition that one-hot encoding usually improves classification accuracy. We validate the approach in Sec.~\ref{sec:ablation}.

\subsection{Network Training}
\label{sec:training}

To train our MLP, we use the ABC dataset~\cite{Koch19a} of watertight meshes for which ground-truth SDF values are available. We provide the unsigned values and their gradients to the network, which is then trained to recover their sign. 

More formally, the network implements a function $\phi_\Theta : \mathbb{R}^{8 \times 4} \rightarrow \mathbb{R}^{2 ^ 7}$, where $\Theta$ represents the network parameters. We train on watertight surfaces for which an SDF is available but, as we do not need to know which point is ``outside'' or ``inside'' the surface, we leave out one vertex at a fixed position across all cells and create a ground-truth pseudo-sign configuration indicating the signs of the remaining seven vertices, hence the dimensionality of the ouput of $\phi_\Theta$. It is necessary to first ensure that the sign of the chosen anchor vertex is consistent across all cells. To do so without loss of generality, we simply flip signs of all vertices in the cell if the sign of the anchor vertex is negative.

We train $\phi_\Theta$ to perform classification by minimizing the cross entropy loss 
%
\begin{align}
    \mathcal{L}_\Theta = \sum_{\mathcal{S} \in \mathcal{D}} & \sum_{c \in \mathcal{S}}  CE( MLP_{\mathcal{S}}(c), GT_{\mathcal{S}}(c)) * w_{GT_{\mathcal{S}}(c)} \; , \nonumber \\
    MLP_{\mathcal{S}}(c) & = \phi_\theta(UDF_{\mathcal{S}}(c),\nabla UDF_{\mathcal{S}}(c)) \; , \label{eq:mlpTrain}
\end{align}
%
%
where $\mathcal{D}$ is the dataset of shapes $\mathcal{S}$, $c$ is a grid cell, $GT_{\mathcal{S}}(c)$ is the ground-truth sign configuration for cell $c$ in mesh $\mathcal{S}$, $MLP_{\mathcal{S}}(c)$ is the corresponding network prediction, and $CE$ is the cross-entropy loss. The weight vector $w \in \mathbb{R}^{2^7}$ can be used to balance the classes, i.e. the sign configurations, since they can be heavily unbalanced. In our case, these weights are set to $1$ as we show in Sec.~\ref{sec:ablation}. We use the Adam optimizer \cite{Kingma14a} with a learning rate of $5 * 10^{-3}$ for 10 epochs, and we train the network on the first 80 shapes of the ABC dataset. The voxel grids used for training are sampled at resolution $128^3$, and only cells that are close to the surface are used for training, yielding approximately 5.5 million training examples. The training process takes around 1 minute on an NVIDIA A100 GPU, whereas the data preparation takes around 20 minutes on a laptop CPU.

\subsubsection{Augmentation.}
The input data is augmented with Gaussian noise according to the following equation:
\begin{align}
    UDF_{\mathcal{S}}(c) & \leftarrow UDF_{\mathcal{S}}(c) * (1+ \mathcal{N}(0, \sigma) )\; , \nonumber \\
    \nabla UDF_{\mathcal{S}}(c) & \leftarrow \nabla UDF_{\mathcal{S}}(c) * (1+ \mathcal{N}(0, \sigma) )\; , \nonumber
\end{align}
where $\sigma$ has been empirically set to $1$. In Sec.~\ref{sec:ablation}, we show the importance of training with noise augmentation.

The input UDF values are further normalized by the size of the grid cells used for sampling, so that the network becomes resolution-invariant. In the supplementary, we validate the approach by showing that training on multiple resolutions does not improve the results.

\subsection{Surface Triangulation}
\label{sec:triangulation}

Our algorithm turns a UDF into pseudo-SDF locally. The simplest way to take advantage of this is to use the popular Marching Cube algorithm~\cite{Lorensen87}, that needs no additional information. However, the more sophisticated Dual Contouring algorithm~\cite{Ju02} is known to give more accurate results in some settings, especially in the presence of sharp edges. Unfortunately, it requires accurately sampling normals near the surface, which have been shown to be very noisy in a UDF setting~\cite{Guillard22b,Zhang23b}. To use DC nevertheless, we extend DualMeshUDF~\cite{Zhang23b}, a UDF-based implementation of DC: we relax its thresholding strategies and filter the vertices produced by the algorithm using sign-flip information from our pseudo-SDF. More specifically when DualMeshUDF fails to locate a vertex in a specific grid cell which contains a sign-flip according to our method, instead of discarding it, we place the vertex at the center of the cell. We show in the result section that this boosts the performance of DualMeshUDF by closing the holes that tend to appear when the UDF is noisy or the surface complex. Note that this is a very simple way to integrate our approach into DualMeshUDF. We show in the results section that it already gives good results. However, a more sophisticated  is approach possible and we will look into it in future work.

\section{Experiments}
\label{sec:exp}

To demonstrate the effectiveness of our approach, we ran experiments on three different datasets ABC~\cite{Koch19a}, MGN~\cite{Bhatnagar19}, and  ShapeNet-Cars~\cite{Chang15}. We considered two different scenarios: 1) triangulating the ground-truth UDFs; and 2) triangulating the less-exact but, in practice, more realistic  neural UDFs obtained as the output of two different deep UDF autodecoders, trained on 300 shapes from MGN and 20 shapes from ShapeNet-Cars, respectively. The autodecoders were trained following the approach and sampling strategy of DeepSDF~\cite{Park19c} using a 12-layer ReLU-based neural network. Technical details are provided in the supplementary. 

%
%
To quantify accuracy, we computed the \textbf{Chamfer Distance (CD)} (200.000 samples for each mesh, unless stated otherwise) and \textbf{Image Consistency (IC)}. For both metrics, we found the median values to be more stable and representative of each method compared to the mean values, that can be heavily influenced by a few mis-reconstructed examples. We organize our results into two main groups, those obtained with Marching Cubes (MC) in Section~\ref{sec:mc} and those obtained with Dual Contouring (DC) in Section~\ref{sec:dc}. Please refer to the supplementary materials for full results. 

\subsection{Neural Surface Detection and Marching Cubes}
\label{sec:mc}


\begin{table}[t]
    \renewcommand{\arraystretch}{1.0}
    \caption{\small \textbf{Marching Cubes-based triangulation.} Median L2 Chamfer Distance $\times 10^{-5}$ with 200k sample points (CD) and Image Consistency (IC) are reported at varying grid resolutions. The best results are in bold, the second-best in italics. Results meshed with Marching Cubes on the ground-truth SDF are reported for reference on ABC, where extracting an SDF is possible. *Resolution is doubled for experiments with ShapeNet-Cars due to the higher complexity of the shapes.
    }
    \label{tab:mc}
    \begin{scriptsize}
        \begin{center}
            \setlength{\tabcolsep}{3pt}
            \begin{tabular}{cc|cc|cc|cc|cc|cc} 
                \multicolumn{2}{c}{} & \multicolumn{2}{c}{ABC~\cite{Koch19a}} & \multicolumn{2}{c}{MGN~\cite{Bhatnagar19}} & \multicolumn{2}{c}{Cars~\cite{Chang15}}  & \multicolumn{2}{c}{MGN autodec.} & \multicolumn{2}{c}{Cars autodec.} \\ 
                Res. & Method & CD $\downarrow$ & IC $\uparrow$ & CD $\downarrow$ & IC $\uparrow$ & CD $\downarrow$ & IC $\uparrow$ & CD $\downarrow$ & IC $\uparrow$ & CD $\downarrow$ & IC $\uparrow$ \\ 
                \midrule
                \multirow{4}{*}{32*} 
                    & CAP-UDF~\cite{Zhou22} & 1070 & 51.0                         & 138 & 64.9                        & 54.1 & 79.3                       & 219 & 57.2                        & 191 & 71.9 \\
                    & MeshUDF~\cite{Guillard22b} & \textit{48.8} & \textit{89.4}       & \textit{18.1} & \textit{91.0}     & \textit{20.9} & \textit{84.4}     & \textit{18.4} & \textit{90.9}     & \textit{24.1} & \textit{86.4} \\
                    & DCUDF~\cite{Hou2023DCUDF} & 1200 & 78.0                           & 603 & 76.0                        & 559 & 69.7                        & 503 & 77.0                        & 318 & 83.7 \\
                    & Ours + MC~\cite{Lewiner03} & \textbf{19.0} & \textbf{91.8}     & \textbf{8.09} & \textbf{92.0}     & \textbf{9.89} & \textbf{87.4}     & \textbf{8.37} & \textbf{91.9}     & \textbf{13.2} & \textbf{87.3} \\
                    \midrule 
                    & GT + MC & 17.6 & 91.8 & - & - & - & - & - & - & - & - \\ \toprule
                \multirow{4}{*}{64*} 
                    & CAP-UDF~\cite{Zhou22} & 239 & 73.4                          & 12.0 & 85.4                       & 14.3 & 87.9                       & 18.8 & 79.4                       & 58.1 & 83.2 \\
                    & MeshUDF~\cite{Guillard22b} & \textit{6.76} & \textit{94.7}       & \textit{3.55} & \textit{94.7}     & \textit{7.51} & \textit{89.8}     & \textit{3.89} & \textit{94.1}     & \textit{14.9} & \textbf{88.6} \\
                    & DCUDF~\cite{Hou2023DCUDF} & 291 & 86.5                            & 155 & 85.6                        & 169 & 81.3                        & 92.4 & 87.3                       & 55.0 & 87.3 \\
                    & Ours + MC~\cite{Lewiner03} & \textbf{4.46} & \textbf{95.4}     & \textbf{2.10} & \textbf{95.5}     & \textbf{4.47} & \textbf{91.2}     & \textbf{2.64} & \textbf{94.7}     & \textbf{10.0} & \textit{88.5} \\
                    \midrule 
                    & GT + MC & 4.52 & 95.7 & - & - & - & - & - & - & - & - \\ 
                    \toprule
                \multirow{4}{*}{128*} 
                    & CAP-UDF~\cite{Zhou22} & 21.3 & 94.1                         & 1.99 & 95.7                       & 4.19 & \textit{93.0}              & 3.13 & 91.8                       & 39.9 & 87.6 \\
                    & MeshUDF~\cite{Guillard22b} & \textit{2.97} & \textit{97.1}       & \textit{1.59} & \textit{96.7}     & \textit{3.61} & 92.8              & \textit{2.42} & 95.0              & \textit{16.7} & \textbf{88.6} \\
                    & DCUDF~\cite{Hou2023DCUDF} & 44.6 & 92.5                           & 26.1 & 92.3                       & 113 & 87.7                        & 6.34 & \textbf{95.4}              & 355 & 78.2 \\
                    & Ours + MC~\cite{Lewiner03} & \textbf{2.52} & \textbf{97.4}     & \textbf{1.40} & \textbf{97.1}     & \textbf{3.13} & \textbf{94.2}     & \textbf{2.06} & \textit{95.2}     & \textbf{14.5} & \textit{87.9} \\
                    \midrule 
                    & GT + MC & 2.74 & 97.4 & - & - & - & - & - & - & - & - \\ 
                    \toprule
                \multirow{4}{*}{256*} 
                    & CAP-UDF~\cite{Zhou22} & 2.53 & 97.8                         & 1.40 & \textit{97.7}              & 3.41 & \textit{95.0}              & 2.15 & 94.6                       & \textbf{37.8} & \textbf{87.5} \\
                    & MeshUDF~\cite{Guillard22b} & \textit{2.42} & \textit{98.0}       & \textit{1.36} & \textit{97.7}     & \textit{3.10} & \textit{95.0}     & \textit{2.05} & \textbf{94.8}     & 92.2 & 81.7 \\
                    & DCUDF~\cite{Hou2023DCUDF} & 11.3 & 95.0                           & 8.09 & 94.8                       & 104 & 89.5                        & 3.40 & 91.1                       & 1930 & 36.0 \\
                    & Ours + MC~\cite{Lewiner03} & \textbf{2.33} & \textbf{98.2}     & \textbf{1.33} & \textbf{97.9}     & \textbf{3.03} & \textbf{95.8}     & \textbf{2.02} & \textbf{94.8}     & \textit{62.2} & \textit{83.8} \\ 
                    \midrule
                    & GT + MC & 2.55 & 98.2 & - & - & - & - & - & - & - & - \\ 
                    \toprule
            \end{tabular}
        \end{center}
    \end{scriptsize}
\end{table}


\newlength{\mcfigwidth}
\setlength{\mcfigwidth}{0.18\linewidth}
\definetrim{trimabc}{2cm 1cm 1cm 1cm}
\definetrim{trimmgn}{5cm 2cm 5cm 1cm}
\definetrim{trimsn}{0cm 12cm 9cm 12cm}
\definetrim{trimmgnad}{4cm 30cm 4cm 0cm}
\definetrim{trimsnad}{0cm 15cm 0cm 20cm}
\setlength\mytabcolsep{\tabcolsep}
\setlength\tabcolsep{1pt}

\begin{figure}[ht!]
	\begin{center}
	{\scriptsize
		\begin{tabular}{llccccc}
			\rotatebox[origin=c]{90}{ABC~\cite{Koch19a}} & \rotatebox[origin=c]{90}{64} &
			\includegraphics[valign=m,width=\mcfigwidth,trimabc]{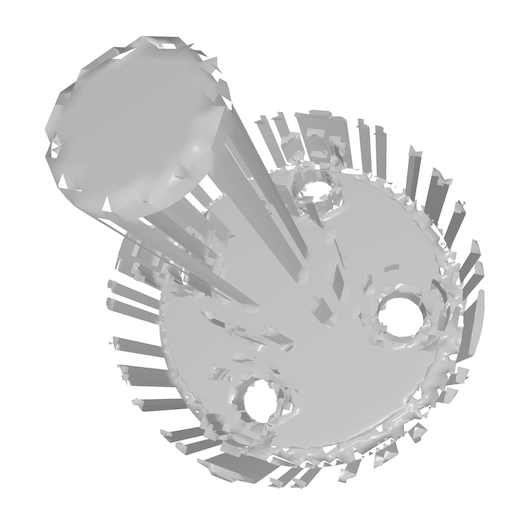}
			& \includegraphics[valign=m,width=\mcfigwidth,trimabc]{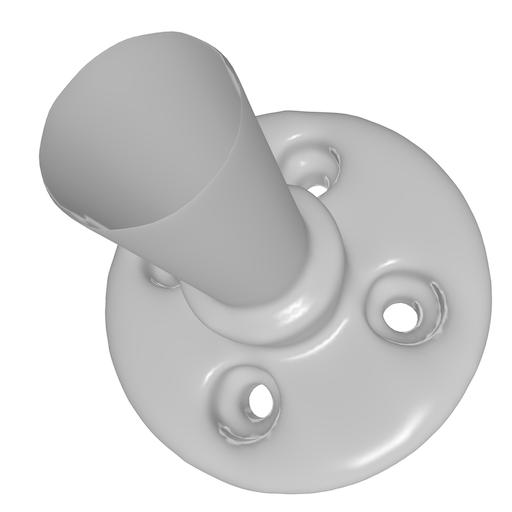}
			& \includegraphics[valign=m,width=\mcfigwidth,trimabc]{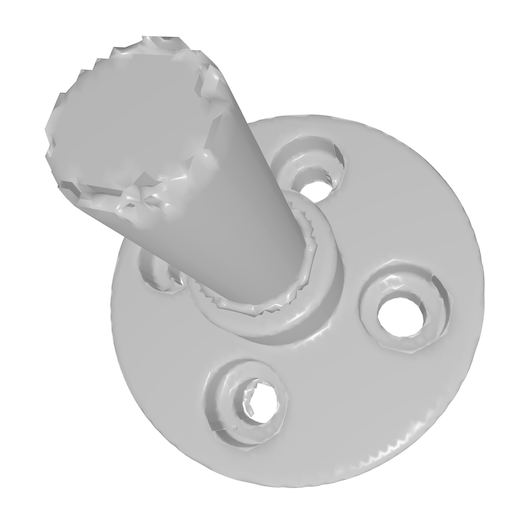}
			& \includegraphics[valign=m,width=\mcfigwidth,trimabc]{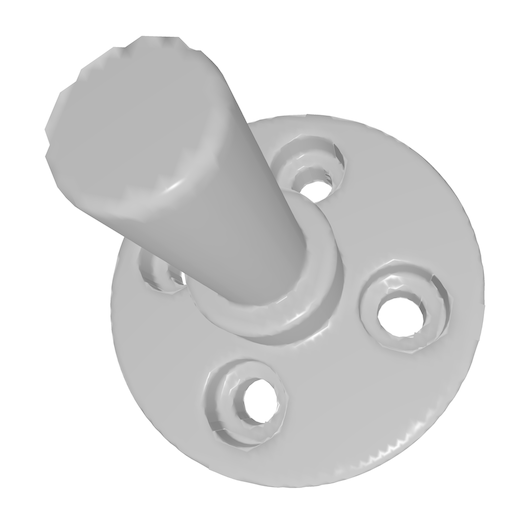} 
			& \includegraphics[valign=m,width=\mcfigwidth,trimabc]{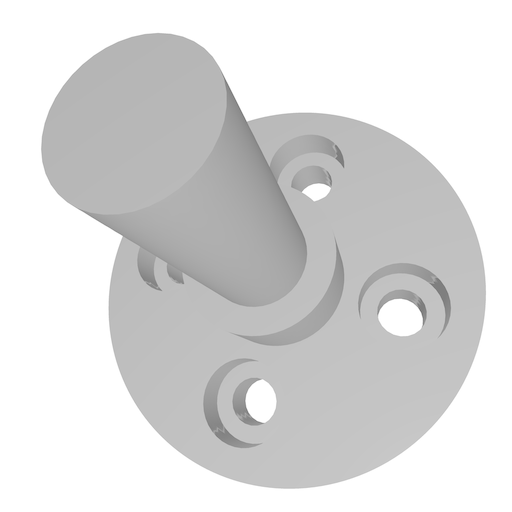}\\ \addlinespace[2\tabcolsep]
			\rotatebox[origin=c]{90}{MGN~\cite{Bhatnagar19}} & \rotatebox[origin=c]{90}{128} &
			\includegraphics[valign=m,width=\mcfigwidth,trimmgn]{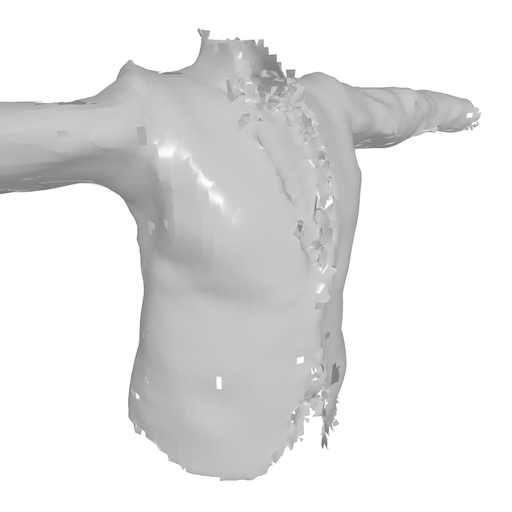}
			& \includegraphics[valign=m,width=\mcfigwidth,trimmgn]{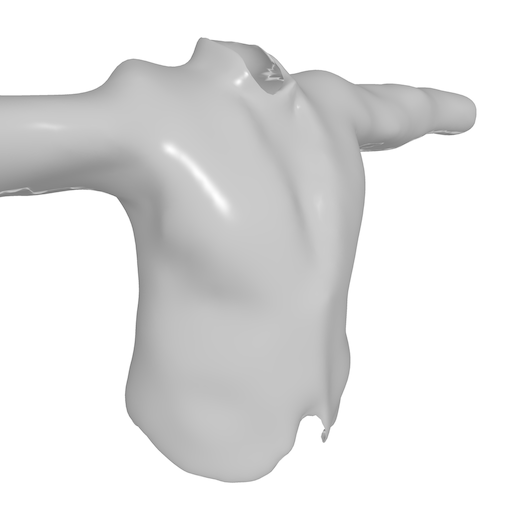}
			& \includegraphics[valign=m,width=\mcfigwidth,trimmgn]{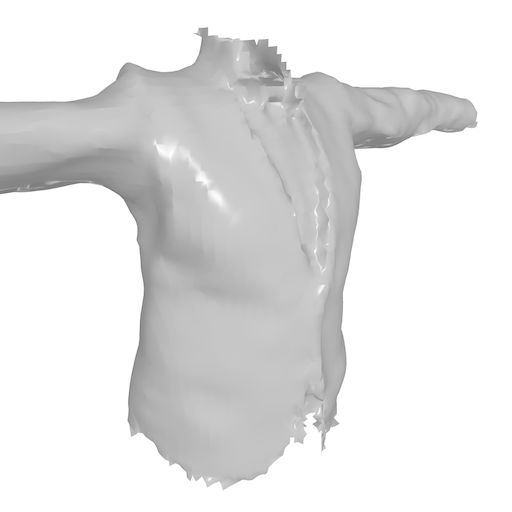}
			& \includegraphics[valign=m,width=\mcfigwidth,trimmgn]{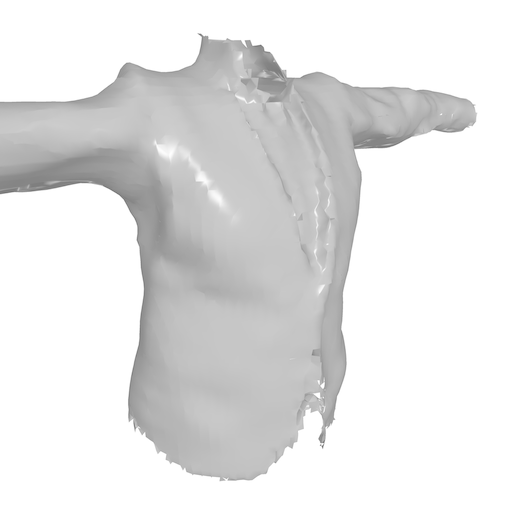}
			& \includegraphics[valign=m,width=\mcfigwidth,trimmgn]{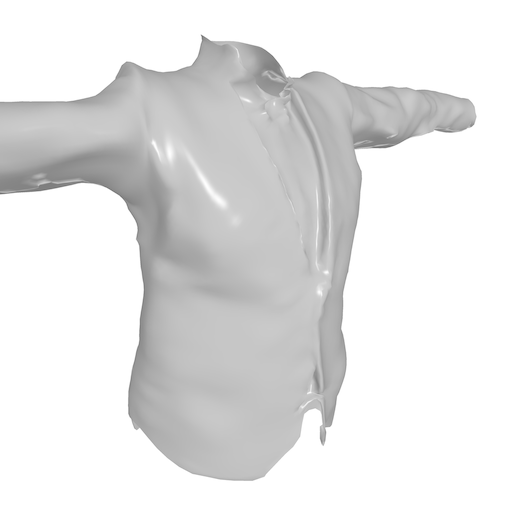}\\ \addlinespace[2\tabcolsep]
			\rotatebox[origin=c]{90}{Cars~\cite{Chang15}} & \rotatebox[origin=c]{90}{256} &
			\includegraphics[valign=m,width=\mcfigwidth,trimsn]{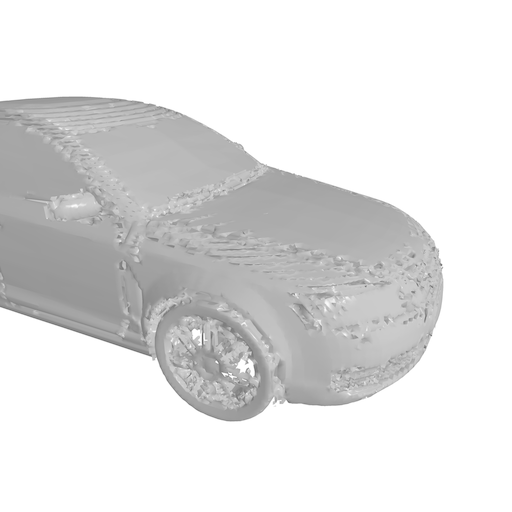}
			& \includegraphics[valign=m,width=\mcfigwidth,trimsn]{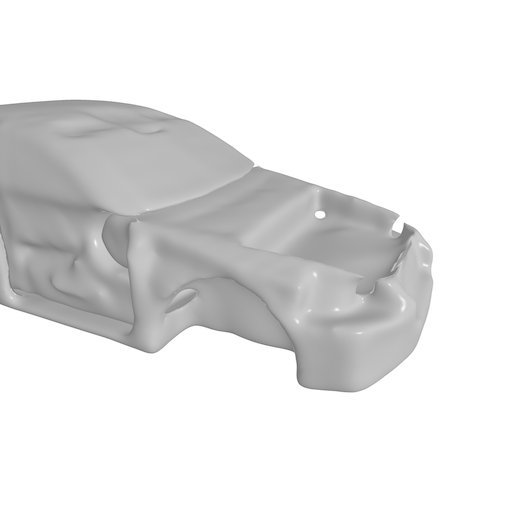}
			& \includegraphics[valign=m,width=\mcfigwidth,trimsn]{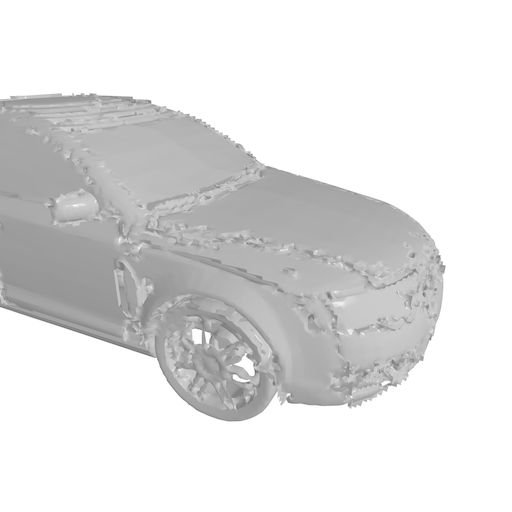}
			& \includegraphics[valign=m,width=\mcfigwidth,trimsn]{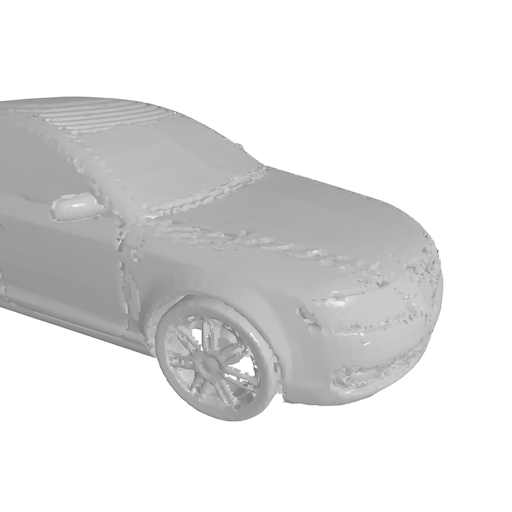}
			& \includegraphics[valign=m,width=\mcfigwidth,trimsn]{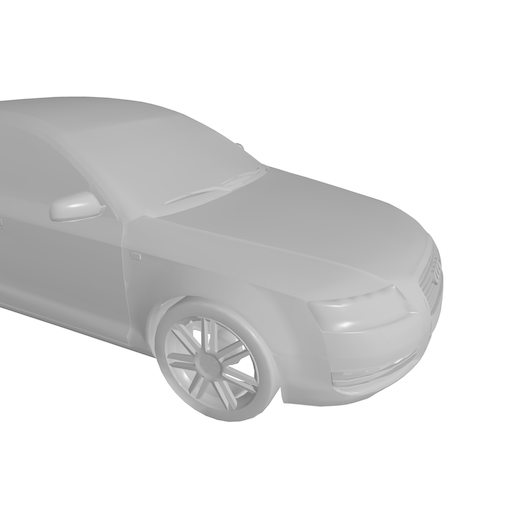}\\ \addlinespace[2\tabcolsep] \hline \addlinespace[2\tabcolsep]
			\multirow{2}{*}{\rotatebox[origin=c]{90}{MGN autodec.\phantom{AA}}} & \rotatebox[origin=c]{90}{128} &
			\includegraphics[valign=m,width=\mcfigwidth,trimmgnad]{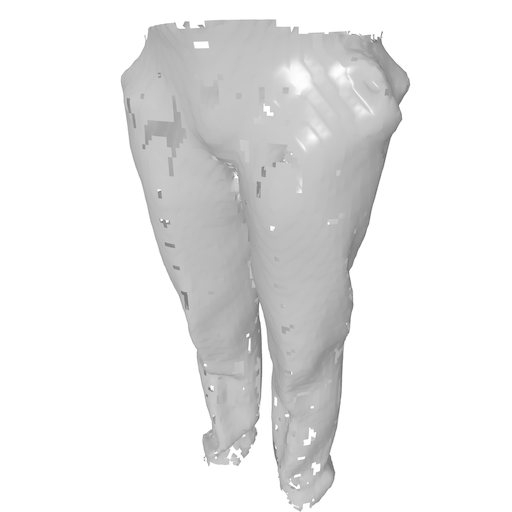}
			& \includegraphics[valign=m,width=\mcfigwidth,trimmgnad]{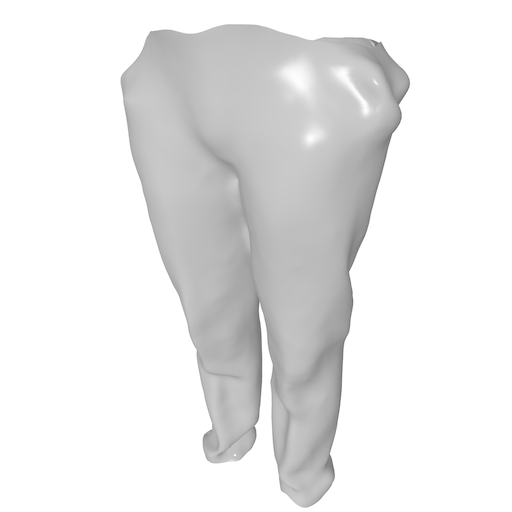}
			& \includegraphics[valign=m,width=\mcfigwidth,trimmgnad]{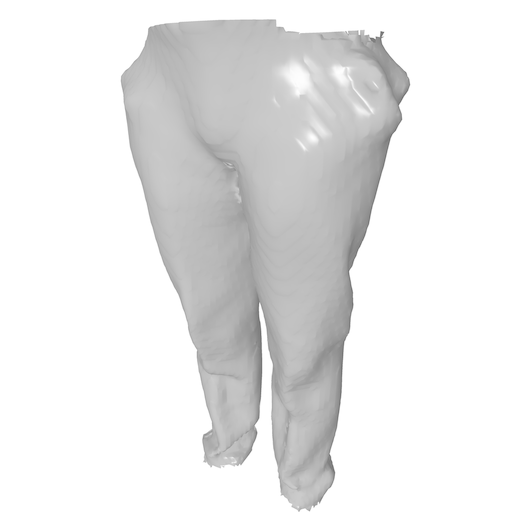}
			& \includegraphics[valign=m,width=\mcfigwidth,trimmgnad]{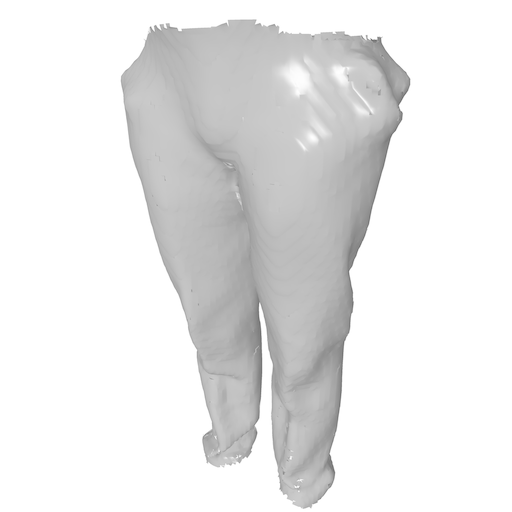}
			& \includegraphics[valign=m,width=\mcfigwidth,trimmgnad]{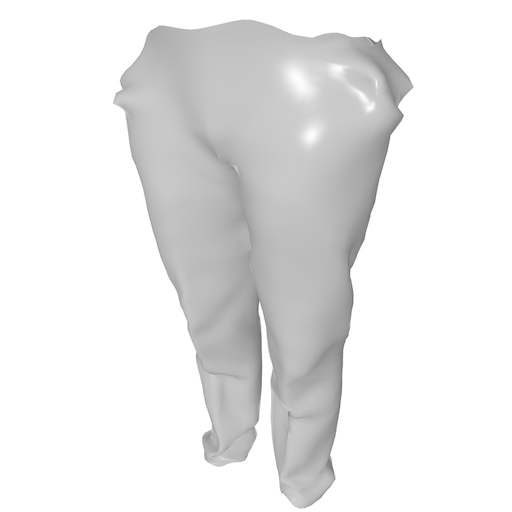}\\ \addlinespace[2\tabcolsep]
			& \rotatebox[origin=c]{90}{256} &
			\includegraphics[valign=m,width=\mcfigwidth,trimmgnad]{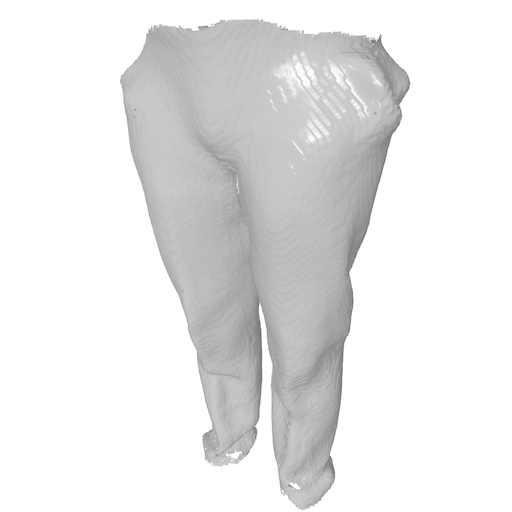}
			& \includegraphics[valign=m,width=\mcfigwidth,trimmgnad]{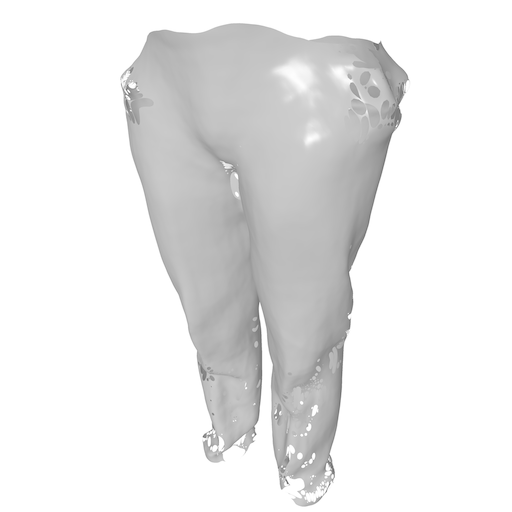}
			& \includegraphics[valign=m,width=\mcfigwidth,trimmgnad]{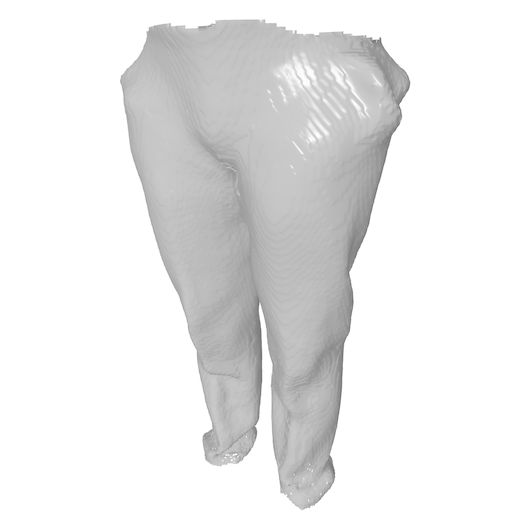}
			& \includegraphics[valign=m,width=\mcfigwidth,trimmgnad]{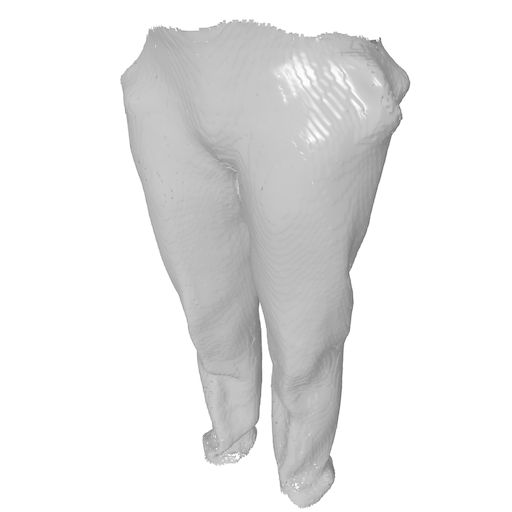}
			& \includegraphics[valign=m,width=\mcfigwidth,trimmgnad]{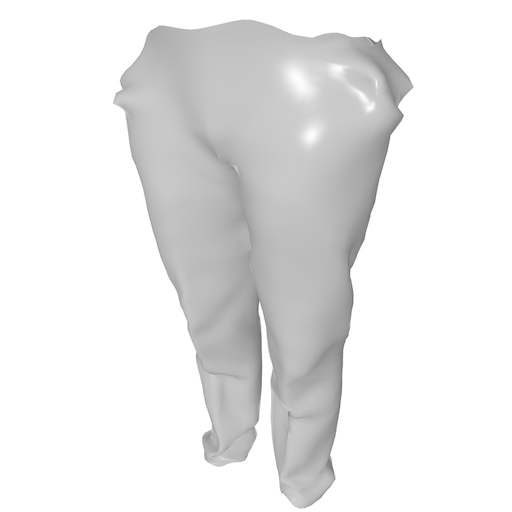}\\ \addlinespace[2\tabcolsep]
			\multirow{2}{*}{\rotatebox[origin=c]{90}{Cars autodec.}} & \rotatebox[origin=c]{90}{256} &
			\includegraphics[valign=m,width=\mcfigwidth,trimsnad]{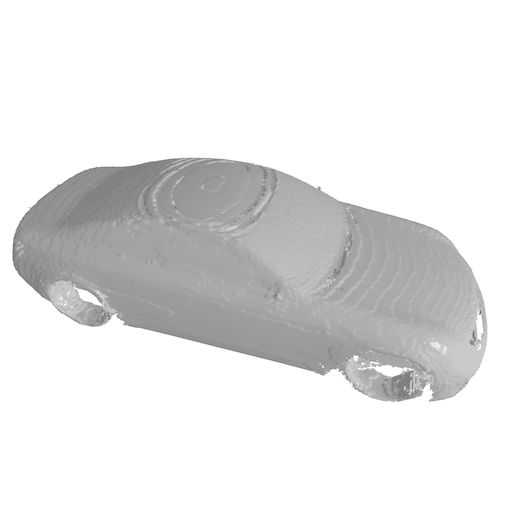}
			& \includegraphics[valign=m,width=\mcfigwidth,trimsnad]{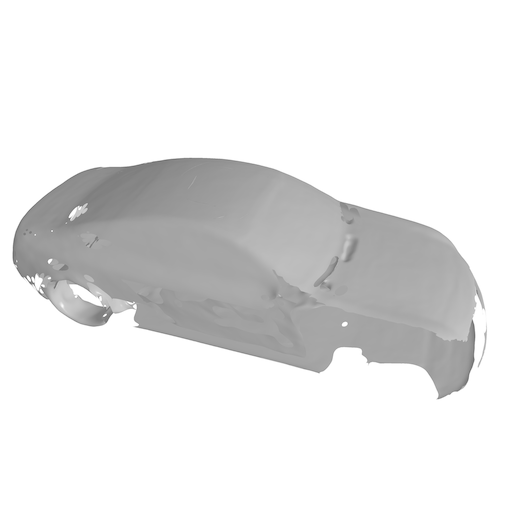}
			& \includegraphics[valign=m,width=\mcfigwidth,trimsnad]{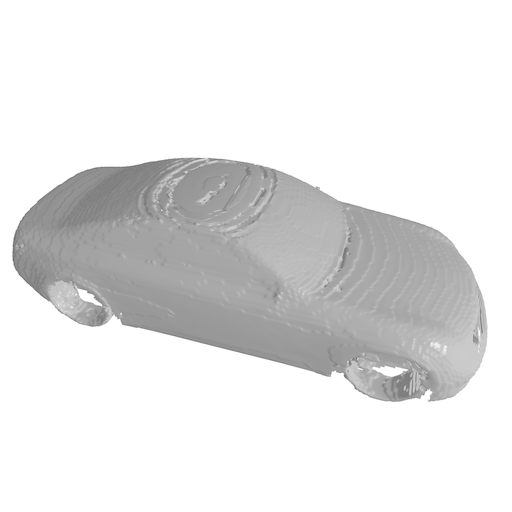}
			& \includegraphics[valign=m,width=\mcfigwidth,trimsnad]{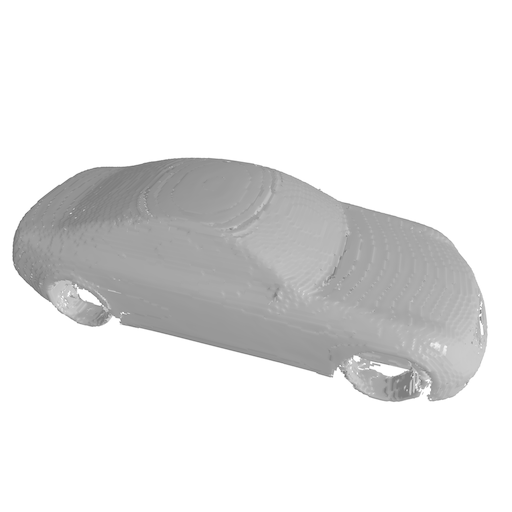}
			& \includegraphics[valign=m,width=\mcfigwidth,trimsnad]{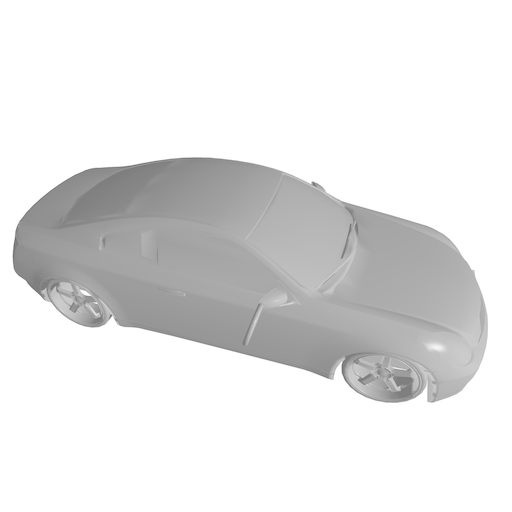}\\ \addlinespace[2\tabcolsep]
			& \rotatebox[origin=c]{90}{512} &
			\includegraphics[valign=m,width=\mcfigwidth,trimsnad]{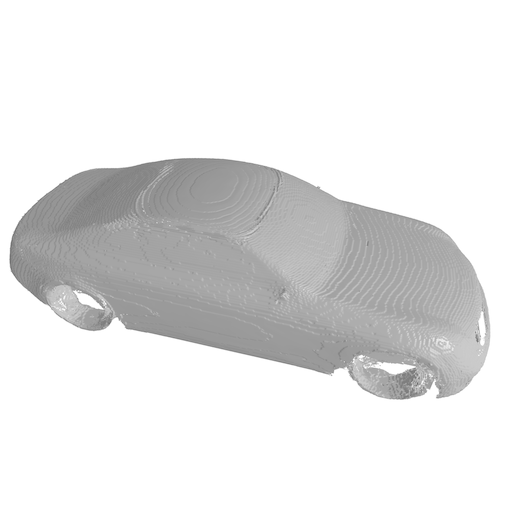}
			& \includegraphics[valign=m,width=\mcfigwidth,trimsnad]{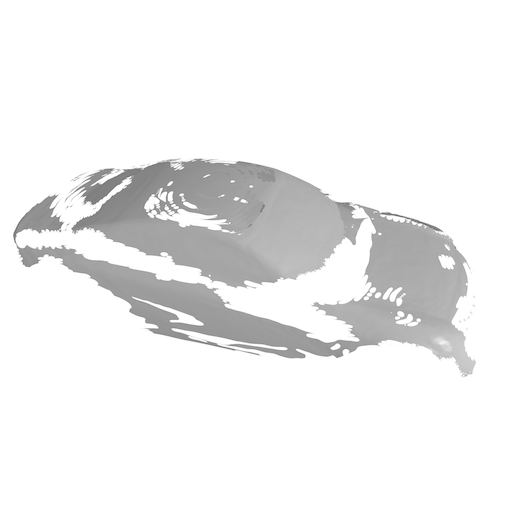}
			& \includegraphics[valign=m,width=\mcfigwidth,trimsnad]{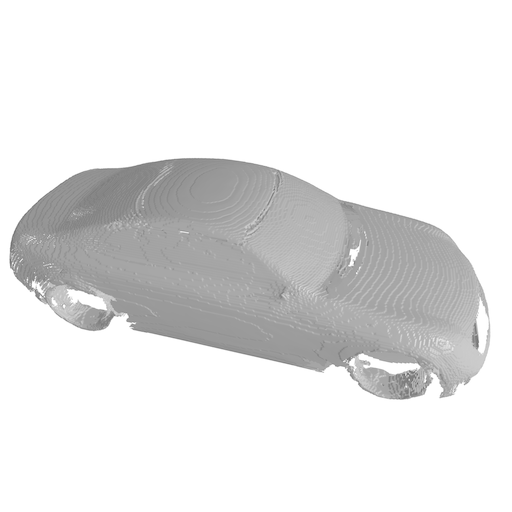}
			& \includegraphics[valign=m,width=\mcfigwidth,trimsnad]{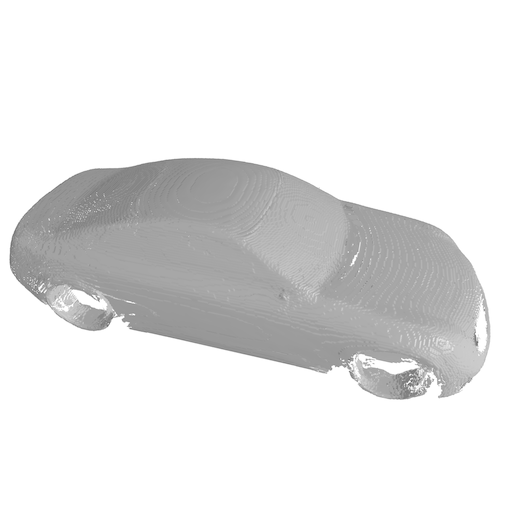}
			& \includegraphics[valign=m,width=\mcfigwidth,trimsnad]{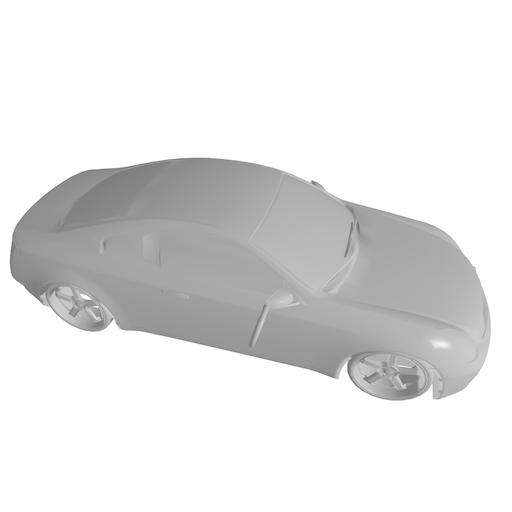}\\ \addlinespace[2\tabcolsep]
			&& CAP-UDF~\cite{Zhou22} & DCUDF~\cite{Hou2023DCUDF} & MeshUDF~\cite{Guillard22b} & Ours & GT
		\end{tabular}}
		\caption{\textbf{Reconstruction methods based on Marching Cubes.} We use our method to transform UDFs into pseudo-SDFs, which we mesh using MC. We compare our results to MC-based baselines on the ABC, MGN, and ShapeNet Cars datasets at various resolutions. More reconstructions are provided in the supplementary material.}
		\label{fig:mc}
	\end{center}
\end{figure}

\setlength{\tabcolsep}{\mytabcolsep}

As discussed in Section~\ref{sec:related}, some of the most notable UDF meshing methods that rely on Marching Cubes are CAP-UDF~\cite{Zhou22}, MeshUDF~\cite{Guillard22b} and DCUDF~\cite{Hou2023DCUDF}. For fairness, all methods have been run without any post-processing or smoothing step because it can be done independently from the meshing procedure. CAP-UDF and MeshUDF have been run with default paramenters. DCUDF, which relies on a threshold for the double covering step, has been run with 3 different thresholds for every experiment and every resolution: in each scenario we select the one that performs the best. We now compare a MC-based triangulation of our pseudo-SDF against these baselines on ground-truth and neural UDFs.

\subsubsection{Meshing Ground-Truth UDFs.}

The ABC dataset is partitioned in chunks of 10000 shapes and we used the last 300 from the first one, which we did not use for training. To triangulate them, we used voxel grids of size $n \times n \times n$, with $n$ being  32, 64, 128, or 256. In other words, the greater $n$, the finer the resolution.  We show qualitative results in  Fig.~\ref{fig:mc} and report comparative results against our baselines in Tab.~\ref{tab:mc}. We consistently outperform them across all resolutions.

We even approach the results obtained by Marching Cubes on the ground-truth SDF, as opposed to UDF, which is available for this dataset because the objects are watertight. On some very thin regions, we even outperform it because MC fails to reconstruct a surface at all from the SDF, whereas our approach does. See Fig.~\ref{fig:mc_fail} for an example. This particularly affects the mean CD values, which are 331 for MC and 35.3 for our method at resolution 32, for example, but it does not mean that in general our method can reconstruct better meshes than MC with a ground-truth SDF, as testified by the median scores in Tab.~\ref{tab:mc}. Noise augmentation during training plays an important role in the reconstruction of thin surfaces, as discussed in our ablation study below.

Among the baselines, CAP-UDF produces fairly accurate results, but also tends to produce unwanted holes. MeshUDF does not have this problem, thanks to its surface-following heuristic, but it tends to produce artifacts such as spurious faces near borders and sharp angles, which are better reconstructed by our method, as can be seen in the first three rows of Fig.~\ref{fig:mc}. At higher resolutions, such artifacts are less visible, but still present. DCUDF~\cite{Hou2023DCUDF} reconstructs very smooth meshes, thanks to its optimization procedure, however they can be too smooth and the surfaces tend to be reconstructed with an offset from the ground truth. DCUDF also tends to be far more computationally demanding than the other methods. 

On the MGN dataset we use the training split, consisting of 300 garments, at the four resolutions $n$ 32, 64, 128, or 256. For Shapenet-Cars, due to the greater complexity of the models, we use 64, 128, 256, and 512. We provide qualitative results in Fig.~\ref{fig:mc}, and report quantitative ones in Tab.~\ref{tab:mc}. They are consistent with those observed on ABC. Our method, which is only trained on 80 shapes from ABC, generalizes well to very different geometric structures. The advantage of our method is especially evident at lower resolutions because, at higher resolutions, the surfaces become easier to mesh. Nevertheless, as can be seen in the middle rows of Fig.~\ref{fig:mc}, our method produces cleaner meshes than the baselines on ShapeNet-Cars even at high resolution, and produces more precise borders on MGN, which can be smoothed with a postprocessing step. The difference in reconstruction accuracy is particularly visible in detailed areas of cars, such as tires, rooftops and front hoods. The only exception is DCUDF, which produces clean and smooth meshes but fails to reconstruct parts of the shapes, likely due to the cutting algorithm. See supplementary for details.

For all methods, however, there remain holes and artifacts in the reconstructed meshes from Shapenet-Cars, mainly due to the many details and the presence of T-junctions, and uneven borders in MGN, with the exception of DCUDF thanks to its integrated smoothing optimization. The problem is less severe at higher resolutions but cannot be handled by Marching Cubes during triangulation, due to its inability to place vertices inside the cells. In such cases, switching from MC to DC is particularly beneficial, as shown in the following section.

\subsubsection{Meshing Neural UDFs.}

To test the behavior of our approach on the kind of unsigned distance functions one is likely to encounter when using implicit representations for object reconstruction, we train two deep UDF auto-decoders on MGN and ShapeNet-Cars, and we sample the voxel grids of UDFs and gradients from these auto-decoders without access to ground-truth data. The gradients are computed by backpropagation in the autodecoder, with respect to the input coordinates. In such a scenario, robustness to noise in the distance field and in the gradient field becomes crucial, as well as the ability to reconstruct usable meshes at low resolutions. In fact the representation capacity of an autodecoder is limited, and increasing the resolution also increases the amount of noise in the UDF and gradient samples, which can complicate the surface detection task for all methods.

We report the results in Tab.~\ref{tab:mc}. Again our method outperforms the baselines in most cases, especially at lower resolutions. MeshUDF, thanks to its surface following heuristic, produces accurate meshes on MGN at $n=256$ and delivers an accuracy similar to ours. However, it can be seen in the bottom rows of Fig~\ref{fig:mc} that meshes from this MGN autodecoder are smoother at resolution 128. Higher-resolution results are impacted by the limited network capacity and exhibit blockiness, indicating that the UDF values are too imprecise for a correct interpolation by MC. DCUDF always produces smooth surfaces, which is particularly useful in this noisy scenario. However, it tends to be less accurate than the other methods, even though it has been shown to be accurate at very high resolutions~\cite{Hou2023DCUDF}. Again, this accuracy does not materialize because of the limited representational capacity of autodecoders.

Similar conclusions can be drawn from the ShapeNet-Cars experiment, where our method also outperforms the baselines at lower resolutions, and achieves results similar to MeshUDF at higher resolutions (better CD but worse IC). While CAP-UDF has the theoretically best metrics at the highest resolution (512), all methods tend to miss parts of the surface, suggesting that a lower resolution is better suited for this UDF autodecoder. CAP-UDF still reconstructs most of the surfaces, but they do not appear very smooth.

%
%

\subsection{Neural Surface Detection and Dual Contouring}
\label{sec:dc}


\begin{table}[t]
    \renewcommand{\arraystretch}{1.0}
    \caption{\small \textbf{Dual Contouring-based triangulation.} Median L2 Chamfer Distance $\times 10^{-5}$ with 2M sample points (CD) and Image Consistency (IC) are reported at varying grid resolutions. The best results are in bold, the second-best in italics. *Resolution is doubled for ShapeNet-Cars due to the higher complexity of the shapes.
    }
    \label{tab:dc}
    \begin{scriptsize}
        \begin{center}
            \setlength{\tabcolsep}{3pt}
            \resizebox{1.0\linewidth}{!}{
            \begin{tabular}{cc|cc|cc|cc|cc|cc} 
                \multicolumn{2}{c}{} & \multicolumn{2}{c}{ABC~\cite{Koch19a}} & \multicolumn{2}{c}{MGN~\cite{Bhatnagar19}} & \multicolumn{2}{c}{Cars~\cite{Chang15}}  & \multicolumn{2}{c}{MGN autodec.} & \multicolumn{2}{c}{Cars autodec.} \\ 
                Res. & Method & CD $\downarrow$ & IC $\uparrow$ & CD $\downarrow$ & IC $\uparrow$ & CD $\downarrow$ & IC $\uparrow$ & CD $\downarrow$ & IC $\uparrow$ & CD $\downarrow$ & IC $\uparrow$ \\  
                \midrule
                \multirow{4}{*}{32*} 
                    & UNDC~\cite{Chen22b}          & 11.8 & 92.4                       & 6.81 & 89.7                       & 7.60 & 88.2                       & 10.8 & 86.2                       & 14.9 & 86.4 \\
                    & DMUDF~\cite{Zhang23b}         & \textit{9.83} & \textbf{97.0}     & \textbf{2.55} & \textbf{94.3}     & \textit{4.15} & \textit{92.2}     & 339 & 69.3                        & 805 & 45.3 \\
                    & DMUDF-T       & - & -                             & - & -                             & - & -                             & \textbf{2.84} & \textbf{94.0}     & \textit{6.96} & \textbf{89.8} \\
                    & Ours+DMUDF    & \textbf{9.66} & \textbf{97.0}     & \textit{2.74} & \textit{94.2}     & \textbf{3.92} & \textbf{92.3}     & \textit{3.00} & \textbf{94.0}     & \textbf{6.73} & \textbf{89.8} \\
                    \midrule 
                \multirow{4}{*}{64*} 
                    & UNDC~\cite{Chen22b}          & 0.783 & 95.8                      & 0.926 & 92.9                      & 1.35 & 91.0                       & 1.83 & 90.7                       & 16.8 & 85.9 \\
                    & DMUDF~\cite{Zhang23b}         & \textit{0.579} & \textbf{98.6}    & \textit{0.195} & \textbf{96.9}    & \textit{0.813} & \textit{94.1}    & 216 & 68.4                        & 954 & 45.5 \\
                    & DMUDF-T       & - & -                             & - & -                             & - & -                             & \textit{0.805} & \textbf{95.4}    & \textit{5.52} & \textit{89.5} \\
                    & Ours+DMUDF    & \textbf{0.574} & \textbf{98.6}    & \textbf{0.194} & \textbf{96.9}    & \textbf{0.787} & \textbf{94.2}    & \textbf{0.803} & \textbf{95.4}    & \textbf{5.43} & \textbf{89.8} \\
                    \midrule 
                \multirow{4}{*}{128*} 
                    & UNDC~\cite{Chen22b}          & 0.0877 & 97.2                     & 0.140 & 94.7                      & 0.195 & 94.0                      & 1.06 & 88.7                       & 57.7 & 74.8 \\
                    & DMUDF~\cite{Zhang23b}         & \textit{0.00450} & \textbf{99.0}  & \textit{0.0194} & \textbf{98.0}   & \textit{0.173} & \textbf{95.7}    & 176 & 66.4                        & 846 & 45.1 \\
                    & DMUDF-T       & - & -                             & - & -                             & - & -                             & \textit{0.722} & \textbf{95.1}    & \textit{10.6} & \textit{87.0} \\
                    & Ours+DMUDF    & \textbf{0.00492} & \textbf{99.0}  & \textbf{0.0185} & \textbf{98.0}   & \textbf{0.169} & \textbf{95.7}    & \textbf{0.713} & \textbf{95.1}    & \textbf{10.0} & \textbf{88.2} \\
                    \midrule 
                \multirow{4}{*}{256*} 
                    & UNDC~\cite{Chen22b}          & 0.0146 & 98.0                     & 0.0251 & 96.5                     & - & -                             & 1.68 & 82.3                       & - & - \\
                    & DMUDF~\cite{Zhang23b}         & \textbf{0.000143} & \textbf{99.1} & \textit{0.00200} & \textbf{98.4}  & \textit{0.0731} & \textit{96.3}   & 167 & 63.9                        & 871 & 43.0 \\
                    & DMUDF-T       & - & -                             & - & -                             & - & -                             & \textit{0.834} & \textit{93.0}    & \textit{37.9} & \textit{79.2} \\
                    & Ours+DMUDF    & \textit{0.000166} & \textbf{99.1} & \textbf{0.00199} & \textbf{98.4}  & \textbf{0.0362} & \textbf{96.6}   & \textbf{0.804} & \textbf{93.6}    & \textbf{37.7} & \textbf{82.8} \\ 
                    \midrule
            \end{tabular}
            }
        \end{center}
    \end{scriptsize}
\end{table}


\newlength{\dcfigwidth}
\setlength{\dcfigwidth}{0.18\linewidth}
\definetrim{trimabc2}{2cm 1cm 1cm 1cm}
\definetrim{trimmgn2}{10cm 2cm 10cm 1cm}
\definetrim{trimsn2}{0cm 12cm 9cm 12cm}
\definetrim{trimmgnad2}{2cm 7cm 2cm 10cm}
\definetrim{trimsnad2}{0cm 12cm 0cm 25cm}
\setlength\mytabcolsep{\tabcolsep}
\setlength\tabcolsep{1pt}

\begin{figure}[ht!]
	\begin{center}
	{\scriptsize
		\begin{tabular}{llccccc}
			\rotatebox[origin=c]{90}{ABC~\cite{Koch19a}} & \rotatebox[origin=c]{90}{64} &
			\includegraphics[valign=m,width=\dcfigwidth,trimabc2]{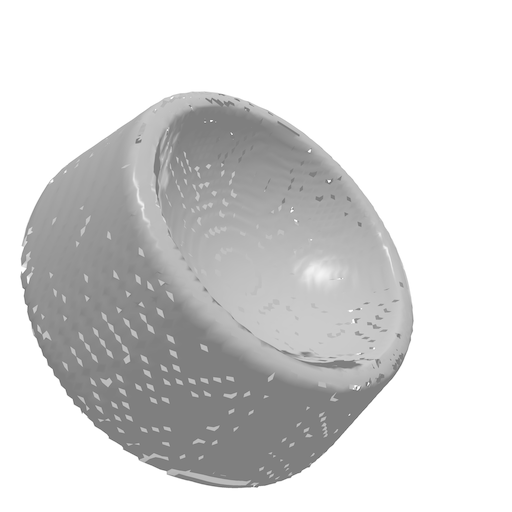}
			& \includegraphics[valign=m,width=\dcfigwidth,trimabc2]{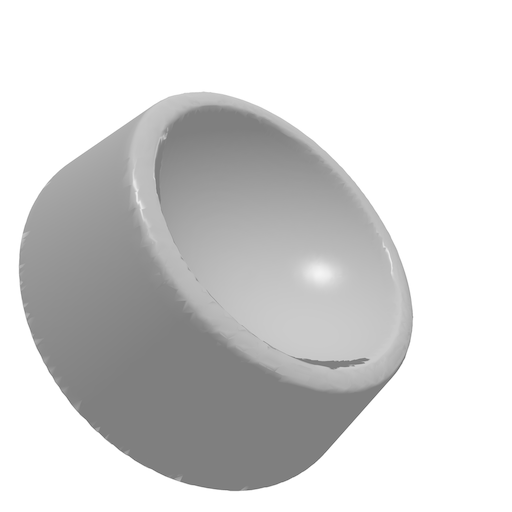}
			& 
			& \includegraphics[valign=m,width=\dcfigwidth,trimabc2]{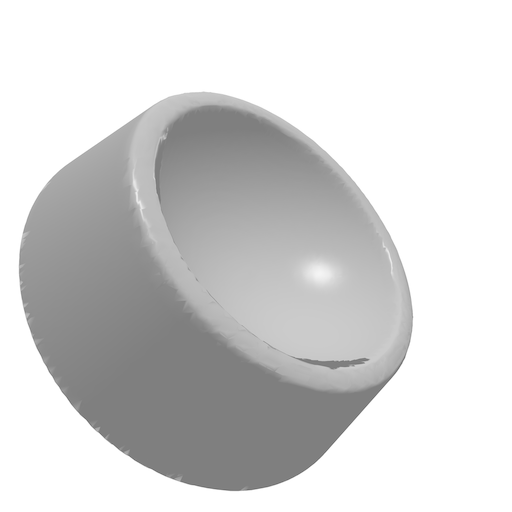} 
			& \includegraphics[valign=m,width=\dcfigwidth,trimabc2]{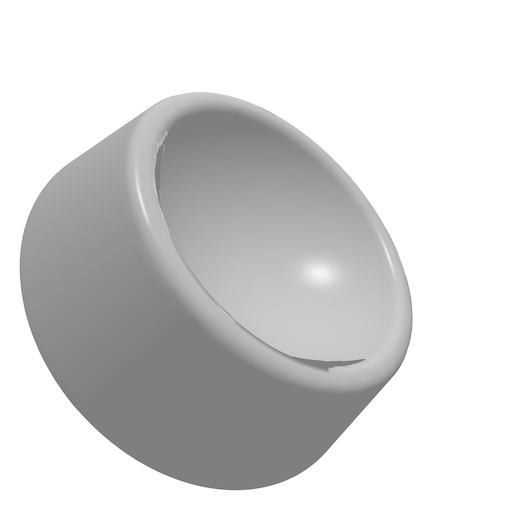}\\ \addlinespace[2\tabcolsep]
			\rotatebox[origin=c]{90}{MGN~\cite{Bhatnagar19}} & \rotatebox[origin=c]{90}{128} &
			\includegraphics[valign=m,width=\dcfigwidth,trimmgn2]{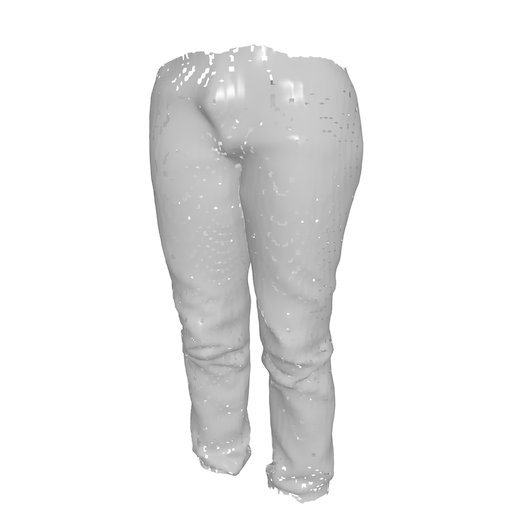}
			& \includegraphics[valign=m,width=\dcfigwidth,trimmgn2]{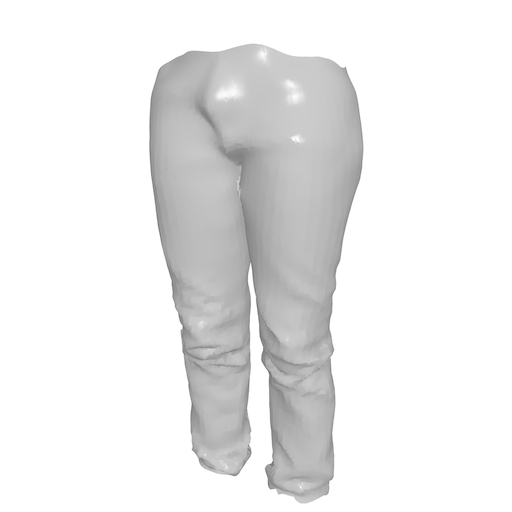}
			& 
			& \includegraphics[valign=m,width=\dcfigwidth,trimmgn2]{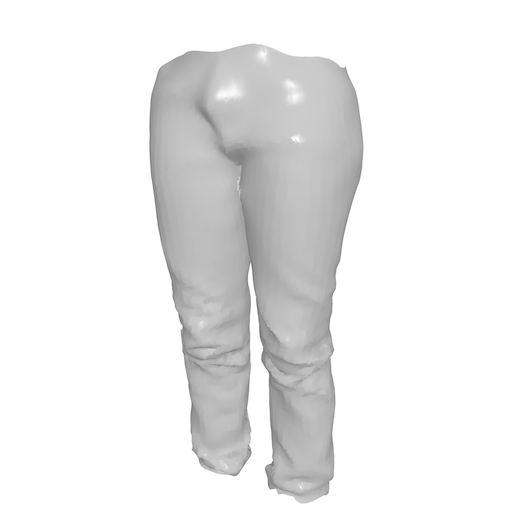}
			& \includegraphics[valign=m,width=\dcfigwidth,trimmgn2]{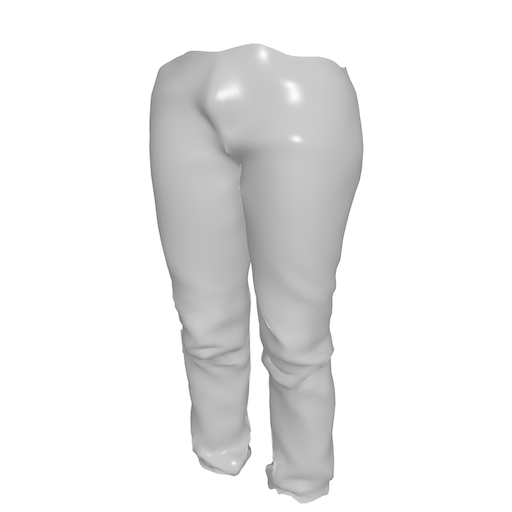}\\ \addlinespace[2\tabcolsep]
			\rotatebox[origin=c]{90}{Cars~\cite{Chang15}} & \rotatebox[origin=c]{90}{512} &
			\includegraphics[valign=m,width=\dcfigwidth,trimsn2]{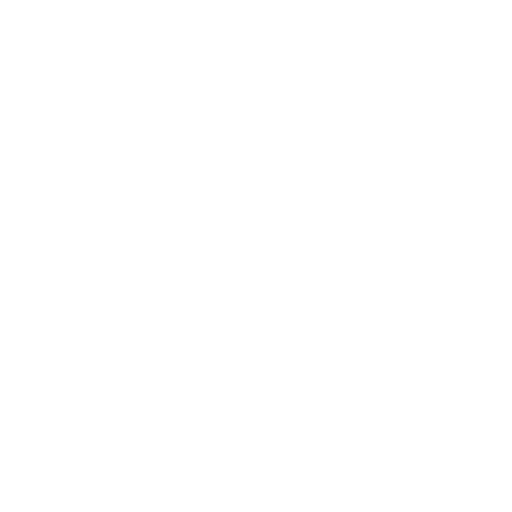}
			& \includegraphics[valign=m,width=\dcfigwidth,trimsn2]{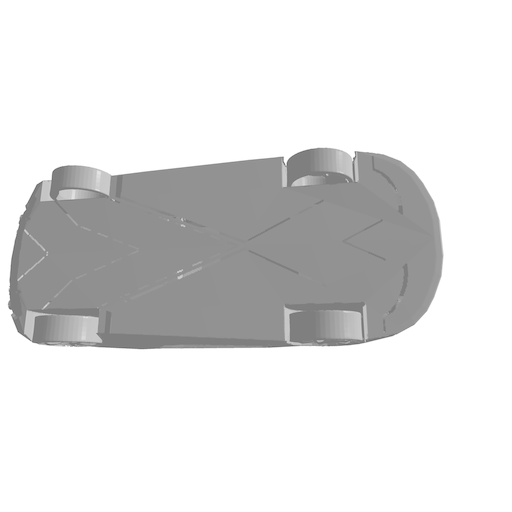}
			& 
			& \includegraphics[valign=m,width=\dcfigwidth,trimsn2]{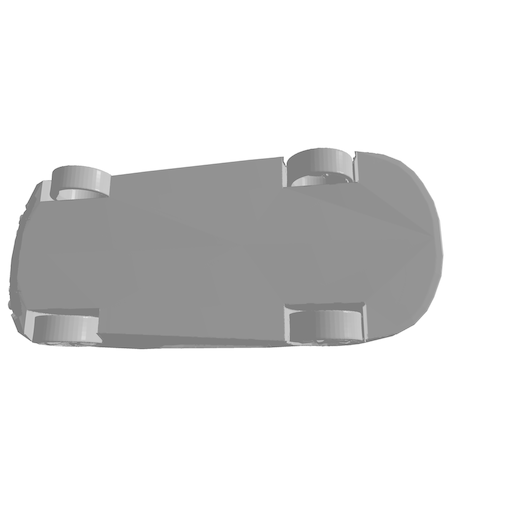}
			& \includegraphics[valign=m,width=\dcfigwidth,trimsn2]{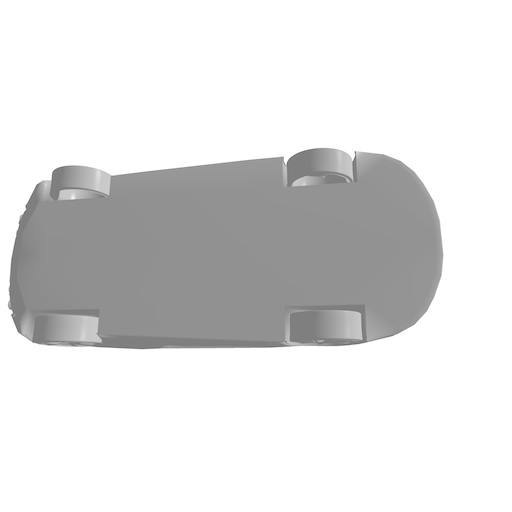}\\ \addlinespace[2\tabcolsep] \hline \addlinespace[2\tabcolsep]
			\rotatebox[origin=c]{90}{MGN autodec.} & \rotatebox[origin=c]{90}{128} &
			\includegraphics[valign=m,width=\dcfigwidth,trimmgnad2]{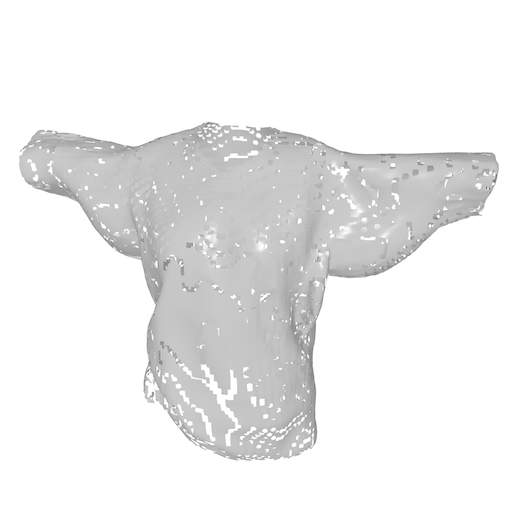}
			& \includegraphics[valign=m,width=\dcfigwidth,trimmgnad2]{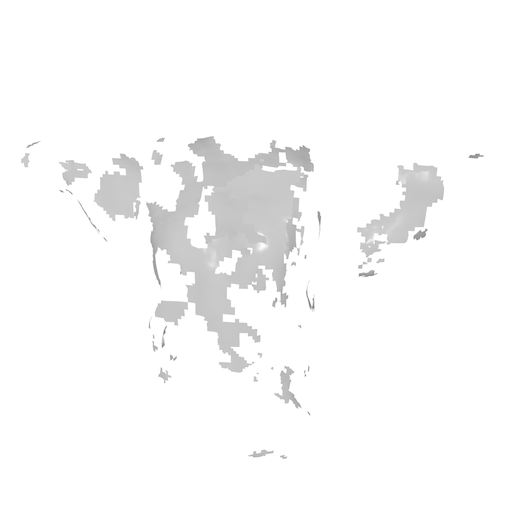}
			& \includegraphics[valign=m,width=\dcfigwidth,trimmgnad2]{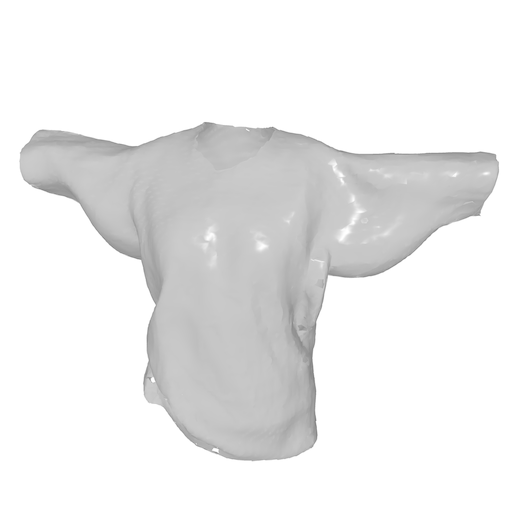}
			& \includegraphics[valign=m,width=\dcfigwidth,trimmgnad2]{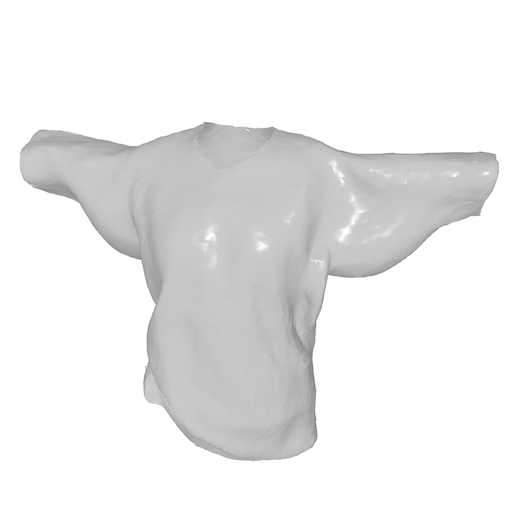}
			& \includegraphics[valign=m,width=\dcfigwidth,trimmgnad2]{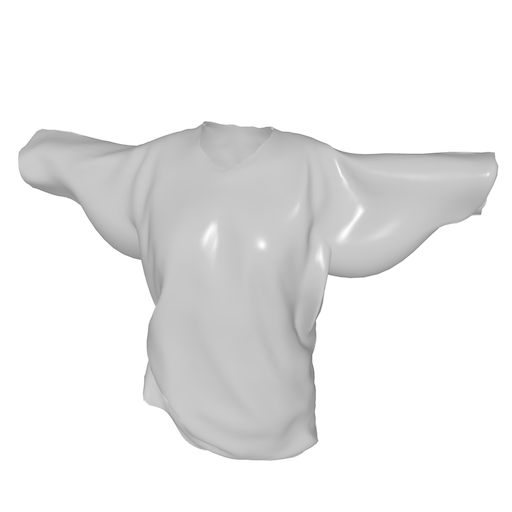}\\ \addlinespace[2\tabcolsep]
			\rotatebox[origin=c]{90}{Cars autodec.} & \rotatebox[origin=c]{90}{256} &
			\includegraphics[valign=m,width=\dcfigwidth,trimsnad2]{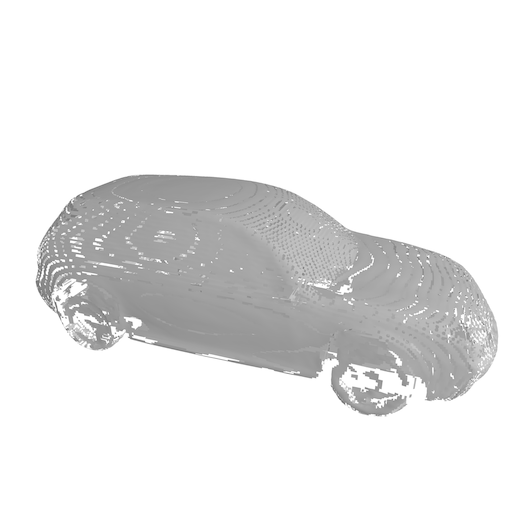}
			& \includegraphics[valign=m,width=\dcfigwidth,trimsnad2]{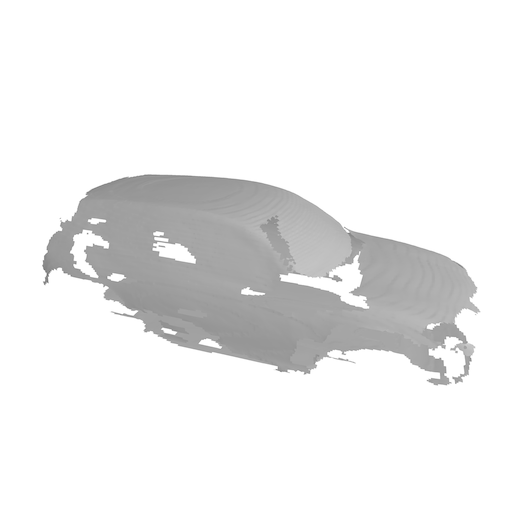}
			& \includegraphics[valign=m,width=\dcfigwidth,trimsnad2]{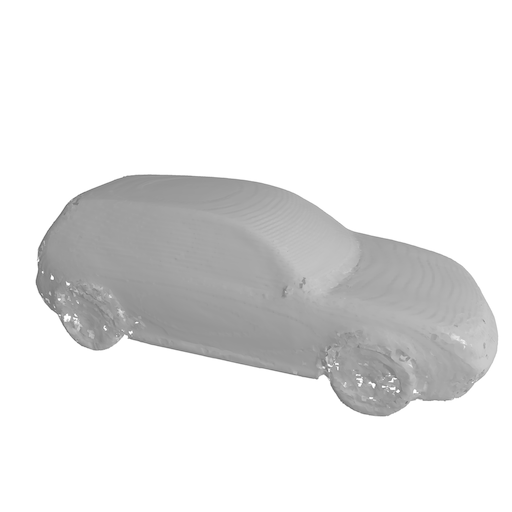}
			& \includegraphics[valign=m,width=\dcfigwidth,trimsnad2]{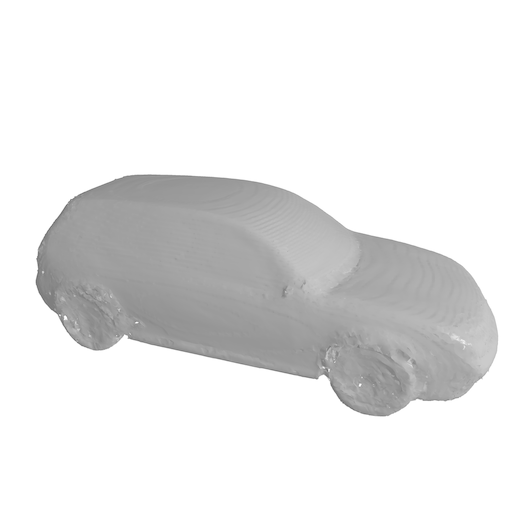}
			& \includegraphics[valign=m,width=\dcfigwidth,trimsnad2]{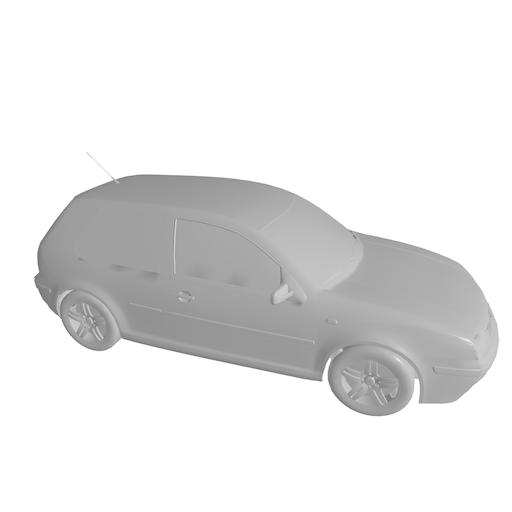}\\ \addlinespace[2\tabcolsep]
			&& UNDC~\cite{Chen22b} & DMUDF~\cite{Zhang23b} & DMUDF-T & Ours+DMUDF & GT
		\end{tabular}}
		\caption{\textbf{Reconstruction methods based on Dual Contouring.} We use our method in conjunction with DualMeshUDF. We compare our results to Unsigned Neural Dual Contouring, DualMeshUDF with default parameters, and DualMeshUDF with tuned parameters.}
		\label{fig:dc}
	\end{center}
\end{figure}

\setlength{\tabcolsep}{\mytabcolsep}

We discussed in Sec.~\ref{sec:related} dual methods for meshing unsigned fields, including DualMesh-UDF~\cite{Zhang23b} (DMUDF) and Unsigned Neural Dual Contouring~\cite{Chen22b} (UNDC). For fairness, all methods are run without any post-processing. UNDC proposes such a step, but it does not alter the results greatly: they are reported for completeness in the supplementary. Additionally, UNDC fails to reconstruct meshes at resolution 512 due to GPU memory requirements.


All methods discussed in this section tend to reconstruct surfaces accurately but with holes. To correctly quantify the difference between the reconstructed meshes, we use 2 million sample points for the Chamfer Distance, instead of 200.000. As a result, CD values are not comparable between the previous section and this section. IC values, however, remain comparable.

\subsubsection{Meshing Ground-Truth UDFs.}

DualMesh-UDF proves to be a very accurate method in this scenario, outperforming UNDC in all cases and not exhibiting particular artifacts, as shown in Tab.~\ref{tab:dc} and Fig.~\ref{fig:dc}. Our simple integration with DualMesh-UDF, aimed primarily at closing holes, does not significantly improve the results, with the two methods trading blows on ABC and MGN across different resolutions. However, when more complex shapes are involved, such as in ShapeNet-Cars, our method can bring measurable improvements. This is clearly visible in the third row of Fig.~\ref{fig:dc}, where DualMesh-UDF fails to reconstruct the surface at the bottom of the car, whereas our method correctly detects it and closes the holes.

\subsubsection{Meshing Neural UDFs.}

DualMesh-UDF (DMUDF) requires setting multiple parameters to work correctly and comes with a default setting, which we used to generate the above results. However, these do not work well on autodecoder generated surfaces: the algorithm filters out too many vertices as erroneous. We therefore tuned these parameters manually for optimal performance on neural UDFs and used that version of the algorithm, which we refer to as DMUDF-T, for our comparative tests.

In contrast, our approach needs no tuning and delivers comparable or better results than DMUDF-T and strictly better results than the untuned DMUDF, especially on ShapeNet-Cars, as can be seen in Fig.~\ref{fig:dc}, closing most of the holes without generating unwanted surfaces, and without requiring any parameter tuning from the user.


\subsection{Ablation studies}
\label{sec:ablation}


\begin{table}[t]
    \renewcommand{\arraystretch}{1.0}
    \caption{\small \textbf{Ablation study.} L2 Chamfer Distance $\times 10^{-5}$ (CD) and Image Consistency (IC) are reported at varying grid resolutions. The best results are in bold. Triangulation is performed using Marching Cubes. \textit{Ours} refers to our method as presented in the previous sections (i.e. with noise injection, unbalanced CE loss, training on 80 shapes). Median values are reported except for column \textit{ABC 64 (Mean)}.
    }
    \vspace{-5mm}
    \label{tab:ablation}
    \begin{scriptsize}
        \begin{center}
            \setlength{\tabcolsep}{3pt}
            \begin{tabular}{c|cc|cc|cc|cc} 
                 & \multicolumn{2}{c}{ABC 64} & \multicolumn{2}{c}{ABC 64 (Mean)} & \multicolumn{2}{c}{MGN 128} & \multicolumn{2}{c}{MGN autodec. 256} \\ 
                 Method & CD $\downarrow$ & IC $\uparrow$ & CD $\downarrow$ & IC $\uparrow$ & CD $\downarrow$ & IC $\uparrow$ & CD $\downarrow$ & IC $\uparrow$ \\ 
                \midrule
                    Balanced CE         & 4.55 & 95.3                           & 35.8 & 92.8                           & 1.44 & 96.9                           & \textbf{2.01} & \textbf{94.8}\\
                    No noise            & \textbf{4.36} & \textbf{95.5}         & 58.1 & 92.4                           & \textbf{1.40} & \textbf{97.1}         & 2.04 & 94.5\\
                    1 shape, no noise   & 4.83 & 95.2                           & 40.0 & 92.7                           & 1.43 & \textbf{97.1}                  & 2.04 & \textbf{94.8}\\
                    7 outputs           & 5.23 & 94.8                           & 9.97 & 92.8                           & 1.66 & 94.2                           & 2.05 & 93.5\\
                    Ours                & 4.46 & 95.4                           & \textbf{9.62} & \textbf{93.2}         & \textbf{1.40} & \textbf{97.1}         & 2.02 & \textbf{94.8}\\
                    \midrule 
            \end{tabular}
            \vspace{-8mm}
        \end{center}
    \end{scriptsize}
\end{table}


\newlength{\mcfailfigwidth}
\setlength{\mcfailfigwidth}{0.24\linewidth}
\definetrim{mcfail}{10cm 2cm 10cm 35cm}
\setlength\mytabcolsep{\tabcolsep}
\setlength\tabcolsep{1pt}

\begin{figure}[t]
	\begin{center}
	{\scriptsize
		\begin{tabular}{llcccc}
		    \rotatebox[origin=c]{90}{} & \rotatebox[origin=c]{90}{} &
			\includegraphics[valign=m,width=\mcfailfigwidth,mcfail]{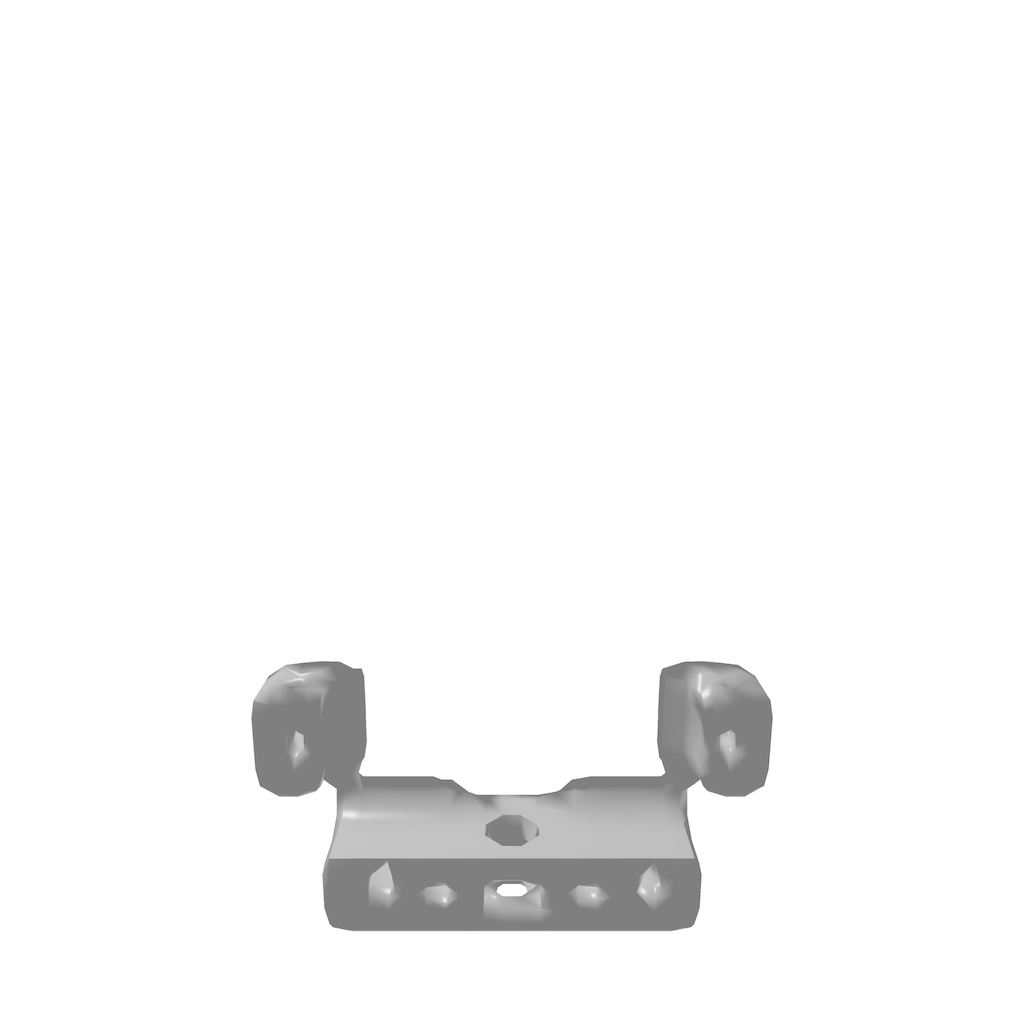}
			& \includegraphics[valign=m,width=\mcfailfigwidth,mcfail]{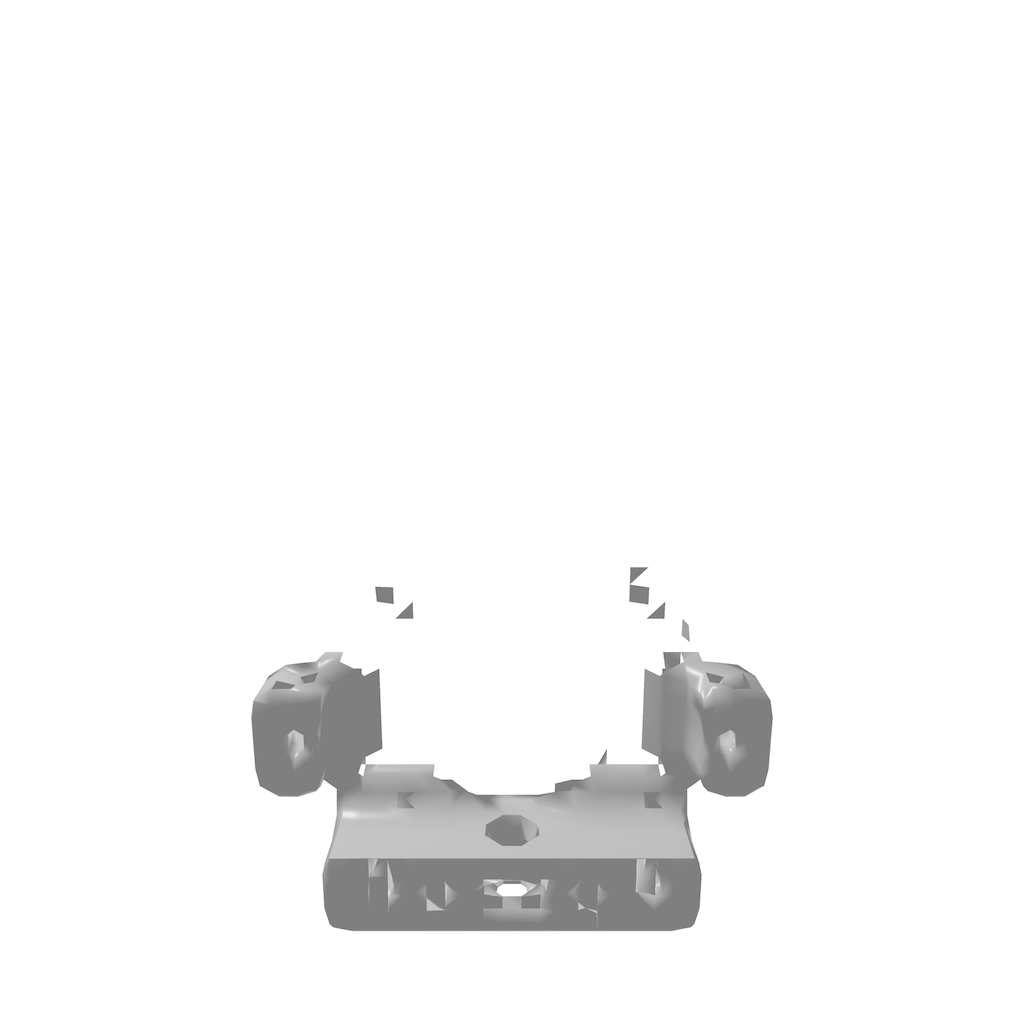}
			& \includegraphics[valign=m,width=\mcfailfigwidth,mcfail]{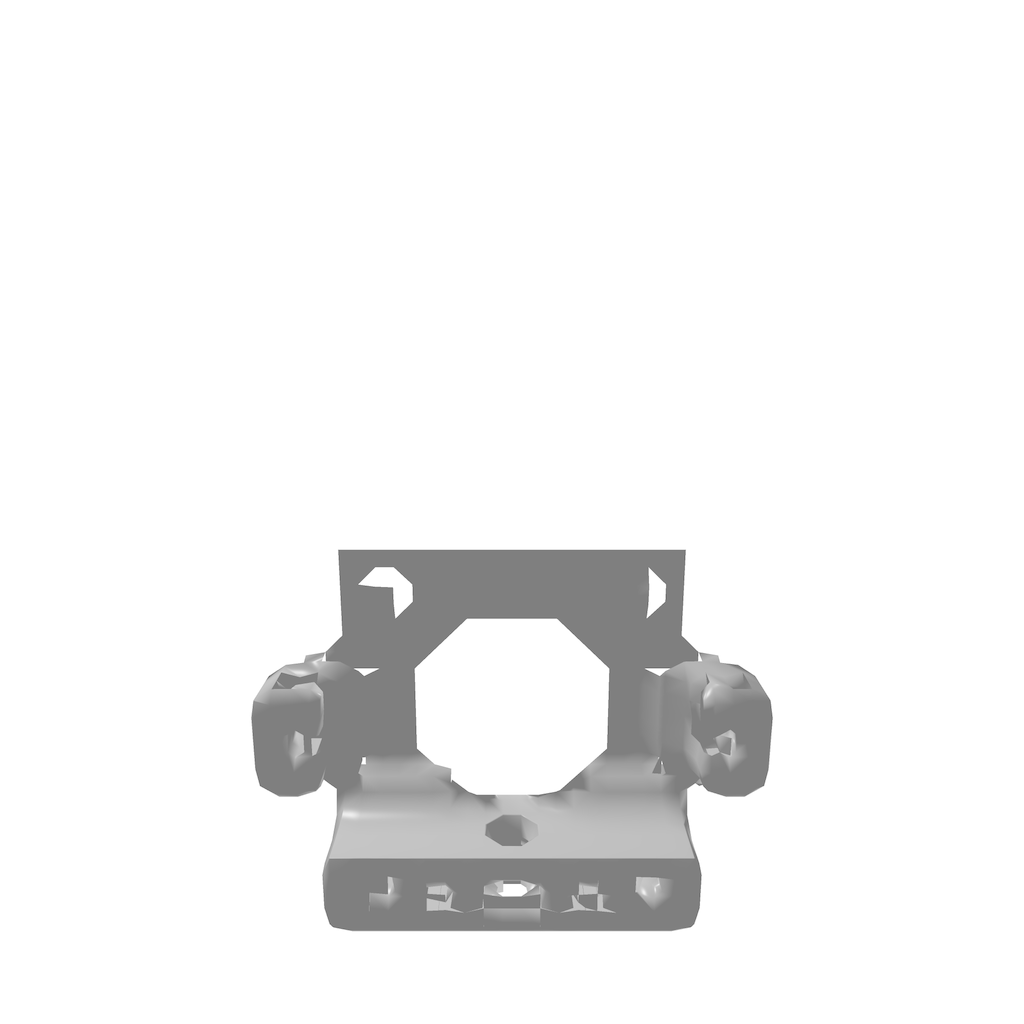}
			& \includegraphics[valign=m,width=\mcfailfigwidth,mcfail]{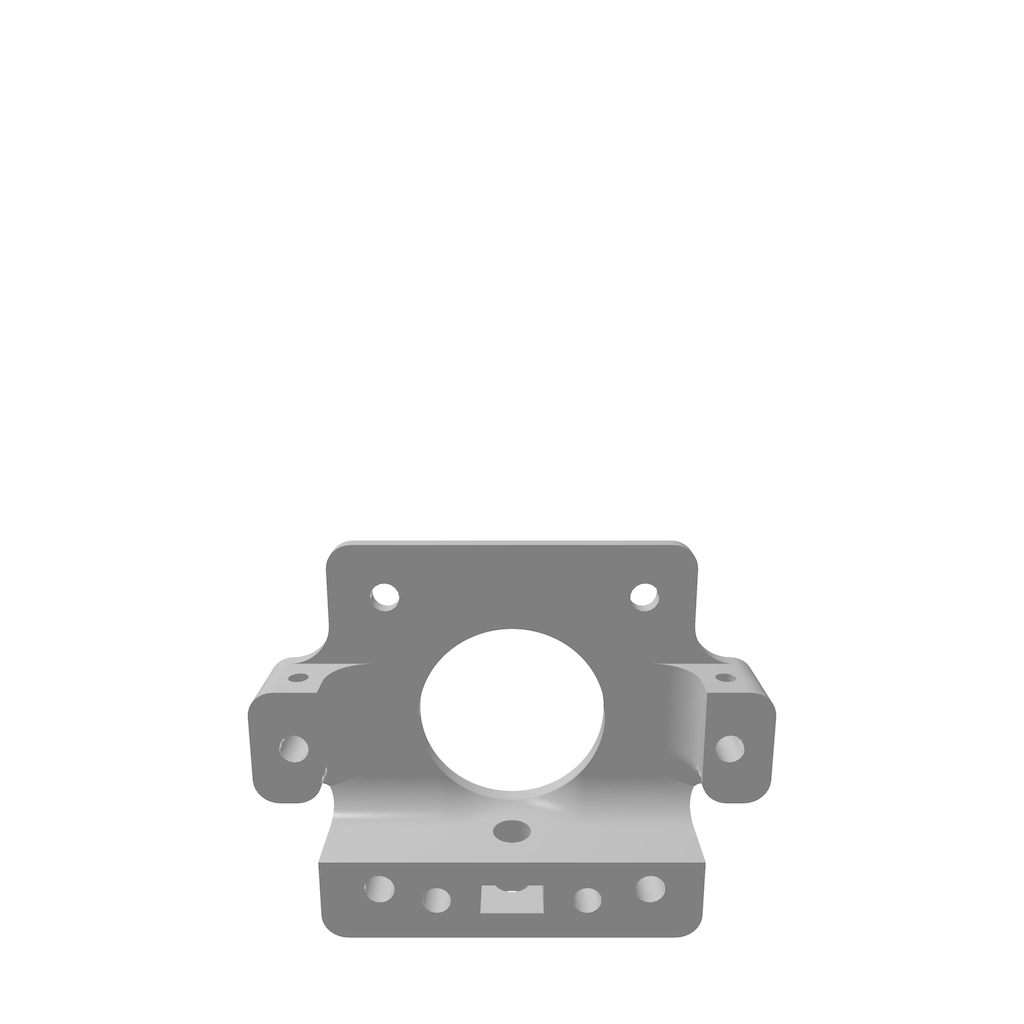}\\ \addlinespace[2\tabcolsep]
			&& MC on SDF & Ours (noiseless training) & Ours & GT
		\end{tabular}}
		\caption{\textbf{Reconstructing thin surfaces at very low resolutions.} When trained with noise augmentation, our approach can reconstruct thin surfaces, here at resolution 32, which are missed by Marching Cubes.}
		\label{fig:mc_fail}
	\end{center}
\end{figure}

\setlength{\tabcolsep}{\mytabcolsep}

We perform several ablation studies to confirm the effectiveness of our design choices. In each row of Tab.~\ref{tab:ablation}, we only modify the specified aspect of the method, keeping the rest as described. The metrics are obtained after triangulating with Marching Cubes. While balancing the Cross-Entropy weights $w$ might seem a good idea, it is detrimental to accuracy in most scenarios, likely because it is more important to reconstruct correctly the most common cases than those appearing seldom, which is achieved by not balancing the loss function. The noise augmentation, on the other hand, is crucial to the performance of our method. Unsusprisingly, training without noise increases the performance on ground-truth evaluations, but it decreases it in the most important scenarios, which are noisy autodecoders. Training with noise also makes our method more robust to thin volumes, as shown by the mean values of the ABC experiment. In the noiseless scenario, our network overfits the behavior of MC and similarly fails to reconstruct thin volumes, see Fig.~\ref{fig:mc_fail} for an example at resolution 32. In the noisy scenario, with thin volumes being few in the training set, the network learns to ignore them, and the reconstruction becomes more accurate. We experimented with multiple kinds of noise augmentations, and a simple Gaussian scaling noise on the network inputs proved to deliver the best results. We refer to the supplementary for details and ablations on other noise types.

Another interesting result relates to the training set size. A training set comprising a single ABC shape, which amounts to around 90k training examples (cells), is already enough to achieve a good performance, especially on a ground-truth UDF scenario, and only requires a few seconds to train on a laptop CPU for 50 epochs. Training on more shapes slightly improves the metrics on ground-truth UDFs, and allows for a more accurate reconstruction on neural UDFs, while still requiring only about one minute to train on an NVIDIA A100 GPU.

Using a 7-output multi-label classification instead of a one-hot 128 output results in lower accuracy, confirming our intuition.


\section{Conclusion}

We have presented a deep-learning approach to locally turning a UDF into an SDF, up to a cell-wise sign flip, making it possible to directly triangulate it using a standard algorithm such as Marching Cubes. We showed that we can consistently outperform, or at worst tie,  state-of-the-art algorithms that also rely on Marching Cubes. Our approach can also be used in conjunction with a method such as DualMesh-UDF, which relies on Dual Contouring instead of Marching Cubes, and improve its performance as well. 

The heuristic we used to integrate our approach into DualMesh-UDF is quite simple and only involves filtering. While it delivers good results, it could be improved. Thus, in future work, we will also reformulate the quadratic optimization problem at the heart of DualMesh-UDF for tighter integration and better performance. Another limitation of the proposed method lies in the use of SDF as a ground truth for training, which prevents the method from accurately reconstructing borders when coupled with MC, resulting in jagged edges (see Fig.~\ref{fig:mc}). This is a common issue in MC-based methods and is usually addressed with a post processing step, but it is a limitation nonetheless. Moreover, compared to methods such as MeshUDF that use a complex and sequential heuristic to follow the surface, our method is more suitable for non-garment surfaces but it tends to produce a higher number of small cracks, which can be unwanted in some applications.
We will also look into another limitation that is inherent to all UDF-based methods, specifically the noisiness of gradients around the surface, which makes surface detection more difficult, even when augmenting the training data to account for it. 

\section*{Acknowledgements}
This work was funded in part by the Swiss National Science Foundation.

\fi

%
%
\bibliographystyle{splncs04}
\bibliography{main}

\renewcommand\thefigure{A.\arabic{figure}} 
\renewcommand\thetable{A.\arabic{table}} 
\renewcommand\thesection{A.\arabic{section}}
\setcounter{figure}{0}
\setcounter{table}{0}
\setcounter{section}{0}

\newpage

\end{document}